%% file: main.tex
\documentclass{article}

\usepackage{graphicx} 
\usepackage{multirow}
\usepackage{amsmath,amssymb,amsfonts}
\usepackage{tikz}
\usepackage{pgfplots}
\usepgfplotslibrary{groupplots}
\pgfplotsset{compat=1.18}
\usepackage[mode=image|tex]{standalone}
\usepackage{stmaryrd}
\usepackage{tcolorbox}
\usetikzlibrary{arrows.meta, positioning, calc,shapes.geometric}
\usepackage{capt-of}
\usepackage{url}
\usepackage[utf8]{inputenc} 
\usepackage[T1]{fontenc}  
\usepackage{booktabs}
\usepackage{enumerate}

\usepackage{arxiv}
\usepackage{amsthm}
\newtheorem{lemma}{Lemma}\newtheorem{proposition}{Proposition}

\theoremstyle{definition}

\theoremstyle{remark}\newtheorem{remark}{Remark}
\usepackage{algorithm}
\usepackage{algpseudocode}
\usepackage{mathtools}
\usepackage{hyperref}
\usepackage{cleveref}
\usepackage[noblocks]{authblk}

\mathtoolsset{showonlyrefs=true}
\usepackage[backend=biber, style=alphabetic]{biblatex}
\usepackage{macros}
\title{Kernel-based Interpolatory Methods for Context Compression of
Attention Mechanisms}
\author[1,*]{Michel Fabrice Serret}
\author[1,2]{Laura Grigori}

\affil[1]{Center for Scientific Computing, Theory and Data, Paul Scherrer Institute, Switzerland.}
\affil[2]{Institute of Mathematics, EPFL, Switzerland}
\affil[*]{Corresponding Author (\href{mailto:michel.serret@psi.ch}{michel.serret@psi.ch})}

\date{\today}

\begin{document}
\maketitle

\begin{abstract}
The bottleneck of attention mechanisms, which are the fundamental building block of modern language models, is their quadratic complexity in terms of context size. There has thus been an increasing interest in the compression of the context, also called KV-cache compression. 
In this article, we consider the problem of compressing the context of attention mechanisms by first formulating it in the Hilbert space associated with its kernel. In this formalism, attention can be represented as the ratio of two linear maps whose interpolant is also the ratio of interpolants, this allows us to consider the interpolation of the linear maps directly. We thus introduce two novel KV-cache compression methods which we call GAIM and Q-DAIM, based on the Generalized Empirical Interpolation Method (GEIM) and the Q-Discrete Empirical Interpolation Method (Q-DEIM), respectively. As the cost of these methods is quadratic  in terms of context length, we propose surrogate models based on the Taylor series, the Performer and the Random Fourier Features method which allow the compression to be done in linear time. Finally, we extend this result to the streaming attention setting, providing new methods which allow one to penalize the selection of interpolation nodes based on the previously interpolated context and to update the previous interpolation based on new information. We then provide  numerical results on the convergence of the residuals and the output error in future queries and preliminary results on the LongBench-E benchmark showing performance comparable to the state-of-the-art methods. These experiments suggest that the range revealed by our Q-DAIM method could help identify where information is concentrated in the KV cache, thus enabling an efficient compression.

\end{abstract}

\section{Introduction}
\label{sec:intro}
The attention mechanism, one of the main building blocks of the
Transformer architecture~\cite{vaswaniAttentionAllYou2017}, is based on functions that involve pairwise interactions of its inputs. Indeed, attention mechanisms are used to construct functions that can change depending on the context\footnote{For a more detailed description of attention in the context of LLMs, see~\cite{serret2026understandingtransformersattentionmechanisms}.}, given by their key $K\in\RR^{N\times d}$ and value matrices $V\in\RR^{N\times d}$ with $N\in\NN^*$. They are applied to an input known as a query vector $q\in\RR^{d}$. In the context of Large Language Models, this context corresponds to a set of $N$ tokens, sets of characters,  and their associated query, key, and value vectors. 
The attention mechanism allows the models to encode the semantic information contained in the relationships of the input token with the context tokens.
To do so, the rows of the key matrices, $(K_i)_{i\in[N]}$, are used to calculate a similarity between the input token's query vector and each contextual key, via a kernel function $\kappa_{\exp}:x,y\mapsto \exp\big(\langle x,y\rangle/\sqrt d\big)$.
This distribution of similarities over the keys is then normalised resulting in the normalised attention scores given by the softmax function \[ \operatorname{softmax}(q,K)=\left(\frac{\exp(\langle q,K_i\rangle/\sqrt{d})}{\sum_{j\in [N]}\exp(\langle q,K_j\rangle/\sqrt{d})}\right)_{i\in[N]}\in\RR^{N}.\] For each context token $i\in[N]$, the associated value row $V_i$ is weighted by the attention score associated with its key. The attention on a single query $q\in\RR^{d}$ is then given by
\begin{equation}\label{eq:att_defn}
\Att(q,K,V)=\operatorname{softmax}(q,K)^\T V=\sum_{i\in[N]}
\frac{\kappa_{\exp}(q,K_i) V_i}{\sum_{j\in[N]}\kappa_{\exp}(q,K_j)}\in\RR^{d},
\end{equation}
where $\kappa_{\exp}$ is the exponential kernel of the scaled dot-product.

However, computing the attention has a quadratic cost in the sequence length $N$ and, at inference time, requires caching the key and value embeddings of all past
tokens. Attention is hence the main computational and memory bottleneck of modern Large Language Models (LLMs). Implementations such as FlashAttention~\cite{daoFlashAttentionFastMemoryEfficient2022,daoFlashAttention2FasterAttention2023}
compute attention exactly while reducing memory traffic; however, it still has a quadratic scaling.
Ever since the rise of LLMs in the last few years, there has been heightened interest in the development of sub-quadratic approximations of the attention mechanism.
The methods proposed come from several fields of research, ranging from clustering and
sparsity~\cite{kitaevReformerEfficientTransformer2020,zandiehKDEformerAcceleratingTransformers2023,hanHyperAttentionLongcontextAttention2023,hooperMultipoleAttentionEfficient2025} to low-rank and subspace
projection-based methods~\cite{wangLinformerSelfAttentionLinear2020,singhaniaLokiLowrankKeys2024}. Recent works also used sketching and tensor decompositions~\cite{phamTensorSketchFast2025} as well as
kernel-based approximations, which replace the exponential kernel of attention with finite-dimensional
features~\cite{choromanskiRethinkingAttentionPerformers2022,kachamPolySketchFormerFastTransformers2024}
or by Nystr\"om approximations~\cite{chenSkyformerRemodelSelfAttention2021,leeProxyformerNystromBasedLinear2024,schroderWildCatNearLinearAttention2026} of the attention scores. A classification
of these methods through the lens of numerical analysis can be found in~\cite{serretAttentionMechanismsThrough2026}. Other methods attempt to compress the KV
cache of a pretrained model directly, removing tokens from the KV cache depending on their associated 
attention-score statistics, discrepancy measures, or following heuristics based on the norm of the key and value vectors~\cite{zhangH2OHeavyHitterOracle2023,liSnapKVLLMKnows2024,xiaoEfficientStreamingLanguage2024,yangPyramidInferPyramidKV2024,hanStreamingAttentionApproximation2025,chenTightBoundsStreaming2026}.
The observation at the basis of our method is shared with the WildCat method~\cite{schroderWildCatNearLinearAttention2026}, attention can be written as the ratio of two linear maps, and the orthogonal projections of these maps onto a well-chosen subspace provide an approximation of the attention mechanism.
Indeed, formulated in the Reproducing Kernel Hilbert Space (RKHS), $H$, of the exponential kernel $\kappa_{\exp}$ underlying scaled dot-product attention, attention becomes a ratio of linear maps in the query feature $\phi [q]\in H$,
\[\Att(q,K,V)=\frac{\phi [q]^{\T}K_\phi^{\T}V}{\phi [q]^{\T}K_\phi^{\T}\mathbf 1_N},\]
where $K_\phi:H\to\RR^{N}$ denotes the linear map whose rows are the features $\phi[K_i]$, $i\in[N]$.
The numerator is given by the feature-value operator $K_\phi^{\T}V$ and the denominator by the feature-normalization operator $K_\phi^{\T}\mathbf 1_N$.
As they share their left factor $K_\phi^\T$, both are generated by the same key features and have ranges in the same $N$-dimensional subspace $\Span(K_\phi)$ of $H$. They can thus be approximated jointly through the augmented operator $M:=K_\phi^\T\Vaug$, with $\Vaug:=[V\mid\mathbf 1_N]\in\RR^{N\times (d+1)}$ the augmented value matrix.

Given a matrix $L\in\RR^{n\times d}$ of $n<N$ keys, let $\Pi_{L_\phi}$ be the orthogonal projector onto $\Span(L_\phi):=\big\{\sum_j\alpha_j \phi [L_j]\in H \,\big|\, \alpha_j\in\RR\big\}$ and let $\Waug:=[W\mid W_{\mathrm{Norm}}]\in\RR^{n\times (d+1)}$ be the weights such that $L_\phi^\T \Waug = \Pi_{L_\phi} M$. Projecting both linear maps yields the approximation
\[
    \widetilde\Att_L(q,K,V)
=\frac{\phi [q]^{\T} \Pi_{L_\phi} K_\phi^{\T}V}
      {\phi [q]^{\T} \Pi_{L_\phi} K_\phi^{\T}\mathbf 1_N}
=\frac{\phi [q]^{\T}L_\phi^{\T}W}{\phi [q]^{\T}L_\phi^{\T}W_{\mathrm{Norm}}},
\]
which WildCat~\cite{schroderWildCatNearLinearAttention2026} interprets as a Nystr\"om-based approximation of the attention mechanism.

In this article, we adopt instead the point of view of interpolation. The rows of $L$ are the \emph{interpolation nodes}, which play the
role of the magic points of the generalized empirical interpolation method (GEIM)~\cite{madayGeneralizedEmpiricalInterpolation2013}, $\mathcal B:=\Span(L_\phi)$ is the basis space onto which we interpolate, and $\Waug$ is the projection of $M$ onto $\mathcal B$, expressed in the node features. The pair $(L,\Waug)$ then replaces $(K,\Vaug)$ as the compressed KV cache.
Since $K_\phi^\T V$ and $K_\phi^\T \mathbf1_N$ are interpolated on the same basis space, the ratio of their interpolants is the interpolant of their ratio: it reproduces attention exactly whenever the query is one of the interpolation nodes.
The compression of attention thus divides into two tasks: the \emph{selection} of nodes $L$, which determines $\mathcal B$, and the \emph{projection} of $M$ onto $\mathcal B$, which determines $\Waug$. The selection phase, described in \S\ref{sec:selection} and \S\ref{sec:sel_qdeim}, uses the information contained in the keys and the values, through $M$, to choose the interpolation nodes $L$ that minimize a given residual norm.
The natural residual is the worst-case attention error in the $p$-norm, $\mathcal R_{p}(L)=\max_{q\in\RR^{d}}\|\Att(q,K,V)-\widetilde\Att_L(q,K,V)\|_{p}$. However, this residual is nonlinear and very sensitive to small errors in the denominator. We therefore consider five variants of the attention residual, based on different approximations of $\mathcal R_p$.
The projection phase, described in \S\ref{sec:proj}, is formalized as a Petrov-Galerkin problem: it computes the weights $\Waug$ for which $L_\phi^\T\Waug$ best approximates $M$ on a given test space $\mathcal T$, the orthogonal projection above being the Galerkin case $\mathcal T=\mathcal B$.

However, classical empirical interpolation methods such as GEIM~\cite{madayGeneralizedEmpiricalInterpolation2013} or Q-DEIM~\cite{drmacNewSelectionOperator2016,drmacDiscreteEmpiricalInterpolation2018} cannot be applied directly to this problem for three reasons.
First, due to the infinite-dimensional nature of the RKHS, operators such as $M$ and features such as the rows of $K_\phi$ cannot be constructed explicitly, and hence the usual pivot-based matrix factorizations cannot, in general, be applied. We have to resort instead to methods relying on kernel evaluations rather than on the elements of $H$.
Second, the kernel norms of the keys, $\|\phi[K_i]\|_H=\exp\big(\|K_i\|^2/(2\sqrt d)\big)$, span several orders of magnitude, so that the exponential kernel Gram matrices involved in the projection are ill-conditioned.
Third, evaluating the exact residual and its updates through kernel evaluations has a cost quadratic in the number of keys, which defeats the purpose of compression.

Our main contribution is to introduce two KV-cache compression methods for attention interpolation that overcome these obstacles, being numerically stable and scaling linearly with the context size $N$:
\begin{itemize}
  \item GAIM, a greedy selection in the style of GEIM~\cite{madayGeneralizedEmpiricalInterpolation2013,madayGeneralMultipurposeInterpolation2007}, which greedily picks the interpolation nodes minimizing the $L^\infty$-norm of the chosen residual, by selecting at each step the node with the largest current residual.
  It evaluates the augmented residual operator $M_\Delta:=M-\Pi_{L_\phi}M=\Pi_{L_\phi}^\perp M$ on the key features, $K_\phi M_\Delta$ and selects the row with the largest $L^\infty$-norm as the next interpolation node and updates the residual through a rank-one update.
  \item Q-DAIM, a Q-DEIM-based selection~\cite{drmacNewSelectionOperator2016,drmacDiscreteEmpiricalInterpolation2018}, which minimizes the $L^2$-norm of the residual in the metric induced by the map using the orthogonal basis obtained through a CholeskyQR / Cholesky whitening of the augmented operator.
  It relies on a QR-like factorization $M=UR$, where $U:\RR^{r}\to H$ is an orthonormal basis in $H$ of the range of $M$, of rank $r\le d+1$, and $R\in\RR^{r\times(d+1)}$.
  As $U$ cannot be constructed explicitly, we instead compute a Cholesky factorization of its Gram matrix, $G:=M^\T M=R^\T R\in\RR^{(d+1)\times(d+1)}$ which only requires kernel evaluations, to obtain $R$. The orthogonal operator onto the range of $M$ is then given by $U=MR^{\dagger}=K_\phi^\T\Vaug R^{\dagger}$.
  By evaluating $U$ on the keys, $F:=K_\phi U\in\RR^{N\times r}$, we obtain a matrix whose rows are the coordinates of the key features in the revealed range, then we select the nodes to be the pivots of a column-pivoted QR factorization of $F^\T$, which greedily maximizes the $L^2$ norm of the reconstructed basis of the revealed range. Additionally, the Q-DEIM interpolation operator then provides us with an alternative Petrov-Galerkin projection whose test space is given by the revealed range of $M$.
\end{itemize}
In practice, both selections are carried out in the Gaussian reformulation described below.
Both methods are designed for streaming attention, where the interpolation is applied on successive segments of the context as they come in, and where each interpolation can be improved by using information from previously interpolated segments.
In \S\ref{sec:streaming}, we decompose the interpolation over a stream into a four-stage process. An orthogonalization of the residual against a basis constructed from previously selected interpolation nodes, which we call deflation for brevity, the selection of new interpolation nodes, the projection of the values, and an optional update of the previously interpolated weights.

The contributions that make these methods numerically stable and computationally feasible are threefold.
\begin{enumerate}[(i.)]
\item \textbf{A Gaussian-kernel and scaled key norm reformulation.} \S\ref{sec:rkhs} develops attention
interpolation in the RKHS of the exponential kernel.
To address the ill-conditioning of this formulation, we provide an exact Gaussian reformulation of the attention mechanism and highlight several of its
invariants, namely query norm scaling, global key norm scaling, and global key translation.
Combined, they separate the RKHS interpolation problem into two parts, a well-conditioned linear problem in translation-invariant Gaussian features and a kernel norm rescaling, which stabilizes our methods.

\item \textbf{Kernel-based selection and Petrov-Galerkin projections.}
Splitting the compression into selection and projection allows both phases to be carried out through kernel evaluations only.
For the projection, we provide several alternative Petrov-Galerkin methods, which differ by their test space $\mathcal T$, and hence by the metric, based on the interpolation nodes $L$, in which $L_\phi^\T\Waug$ approximates $M$.

\item \textbf{Finite-dimensional surrogate residuals.}
The exact residual and its update are costly, $O(\Nc ^2d)$, to evaluate directly, where $\Nc$ denotes the size of a segment. We therefore introduce finite-dimensional
\emph{surrogate} residuals based on finite-dimensional approximations of the exponential kernel's feature map. We namely consider feature maps constructed using the first-order Taylor approximation,  Random Fourier Features \cite{rahimiRandomFeaturesLargescale2007} or from the Performer FAVOR+ features~\cite{choromanskiRethinkingAttentionPerformers2022}. Due to the finite-dimensional nature of these maps, key features can be represented as matrices and this allows for a change in the order of operations that reduces the cost of selection and residual updates to $O(\Nc d^2)$.
These surrogate residuals are used as alternative selection criterion for our magic points. 
However, to limit the approximation error, the projection step is still obtained through an exact Petrov-Galerkin projection of the operator on the chosen nodes, which costs on the order of $O(\Nc d^2 + d^3)$ for a single segment of size $\Nc $ with a number of interpolation nodes proportional to $d$, $\nc=O(d)$.
\end{enumerate}

We provide a complexity analysis for segments of size $\Nc\in\NN$ with $\nc$ interpolation nodes per segment of our streaming interpolation methods in \S\ref{sec:complexity}.
We also provide a numerical study of our proposed interpolation methods in \S\ref{sec:results} on Qwen2.5-7B-Instruct~\cite{qwenQwen25TechnicalReport2025}, in terms of the convergence of the residuals, of the attention error on future queries of the same text, and of the performance on the LongBench-E benchmark~\cite{baiLongBenchBilingualMultitask2024}.
Overall, in a preliminary LongBench-E evaluation at $25\%$ context compression, Q-DAIM obtains the best average score of all the methods we evaluated, retaining $87.1\%$ of the score of the uncompressed model and slightly outperforming both GAIM and the RP-Cholesky selection of WildCat~\cite{schroderWildCatNearLinearAttention2026} implemented in the same framework. For this benchmark, we computed several variants of the model to isolate the impact of the different parameters of the methods on the result.
We observe that a lower attention error does not necessarily translate into better benchmark scores, the selection of the interpolation nodes having a more decisive impact on performance than the projection.
Interestingly, it appears that the revealed range of Q-DAIM provides both better selection and better projection performance than the GAIM or RP-Cholesky methods.
Indeed, by looking at the variants where no projection is used, i.e. $\Waug=\Vaug$, Q-DAIM selection appears to outperform all other methods, and on the other hand the best performing variant on the most tasks for Q-DAIM is the one using the revealed-range projection, i.e. the Q-DEIM interpolation operator.
This suggests that the range of the augmented operator revealed by Q-DAIM is a better indicator of the information retained by a method than the residuals and that the revealed range provides additional information which seems to not be accessible to the other methods. As the revealed range is only approximately reconstructed in this setting, being limited to $d$ vectors for its reconstruction but requiring the complete set of keys for a perfect reconstruction, this motivates future research for the direct reconstruction of $U$.

The present work belongs to the kernel-based family of attention approximation methods and is most closely related to the WildCat method~\cite{schroderWildCatNearLinearAttention2026}, which also approximates the augmented linear map similarly but whose selection uses a Randomly Pivoted Cholesky (RP-Cholesky) on the Gram matrix of the keys for a rescaled kernel. WildCat was also recently combined
with the Thinformer approximation inside the causal Express wrapper
of~\cite{gongExpressLanguageModeling2026}.
Compared to the WildCat~\cite{schroderWildCatNearLinearAttention2026} method, whose
random selection is optimal to reconstruct the span of the keys in the Hilbert space $H$, our method takes into account the value matrix and focuses on minimizing the residual of attention.
\section{Attention in the Hilbert space of the scaled dot-product exponential kernel}
\label{sec:rkhs}
In this section, we first define the attention mechanism and highlight its formulation in the Reproducing Kernel Hilbert Space setting in \S\ref{sec:feature_space}, and proceed to define its interpolation on a set of keys in \S\ref{sec:interp}.
Then we consider the Gaussian reformulation of the linear maps in \S\ref{sec:gauss_main} and describe the invariance properties of the attention mechanism in \S\ref{sec:invariances} which are the basis of our numerical implementations. 
Indeed, they allow us to improve the conditioning of the interpolation problem and provide better numerical stability.

\subsubsection*{Notation}
In the paper, we use the following notation. We write $\NN$ for the set of non-negative integers, $\NN^*=\NN\setminus\{0\}$, and $[n]:=\{1,\dots,n\}$ for $n\in\NN^*$. We denote arbitrary vectors using $x\in\RR^d$ and $y\in\RR^d$.  We set $\langle\cdot,\cdot\rangle$ and $\|\cdot\|$ to denote the $L^2$ inner product and norm, and $\mathbf 1_n\in\RR^{n}$ to be the vector of ones for any $n\in\NN^*$. 
For any matrix $A\in\RR^{m\times n}$, we denote by $A^{\T}$ its transpose and by $A_i\in\RR^{n}$ the vector corresponding to its $i$-th row, and we also set $A^\dagger$ to be the pseudo-inverse. $\|A\|_F$ denotes the Frobenius norm and $\Tr(A)$ the trace of the matrix $A$. 
Given a Hilbert space $H$, we denote by $\langle\cdot,\cdot\rangle_H$ and $\|\cdot\|_H$ its associated inner product and norm. We denote using the notation $\Pi_{\mathcal S}$ the orthogonal projector onto a closed subspace $\mathcal S\subseteq H$, with $\Pi_{\mathcal S}^{\perp}:=\ID-\Pi_{\mathcal S}$. We also use by abuse of notation $\Pi_{\Psi}$ for a linear map $\Psi: H\to\RR^{n}$ to denote the orthogonal projector onto its row-space $\Span(\Psi)$. For any feature map $\phi:\RR^d\to H$, we denote for any $x\in\RR^d$ the mapped feature as $\phi[x]\in H$.
Occasionally, for clarity, we use Python notation to identify the matrix rows, columns and elements: $A[i,j]$ is used to denote the entry in row $i$ and column $j$, $A[i,:]$ denotes the $i$-th row and $A[:,j]$ the $j$-th column. More generally, given index sets $I\subseteq[m]$ and $J\subseteq[n]$, we set $A[I,J]$ to be the submatrix of $A$ composed of the rows indexed by $I$ and columns indexed by $J$. We also set, when used in python notation, $i:j=\{i,\dots,j\}$, for any $i,j\in\NN^*$ with $i\le j$.
For any $x\in\RR^{d}$, we set $\Diag(x)\in\RR^{d\times d}$ to be  the diagonal matrix whose diagonal is given by the elements of $x$. On the other hand, when used with a matrix $A\in\RR^{n\times n}$, it corresponds to the diagonal of the matrix in $\Diag(A)=(A[i,i])_{i\in[n]}\in\RR^{n}$. The Hadamard, or elementwise,  product is denoted by $\odot$. The $\exp$ and $\log$ functions are considered to be applied elementwise to vectors and matrices. 

\subsection{Attention as a ratio of two linear maps}
\label{sec:feature_space}

As described in the context of the Nyström approximation used in WildCat method~\cite{schroderWildCatNearLinearAttention2026}, the above formulation of attention~\eqref{eq:att_defn} can be rewritten as the ratio of two linear maps through the kernel function $\kappa_{\exp}$.  In our case, it is more useful to describe attention as a ratio of linear maps in the Reproducing Kernel Hilbert Space, $H$, of $\kappa_{\exp}$ by its explicit feature map, $\phi:\RR^{d}\to H$.
Indeed, let $H_1:=\RR^d$, the RKHS associated to $\kappa_{\exp}$ is given by $H=\bigoplus_{r\in\NN}\mathrm {Sym}(H_1^{\otimes r})$, the direct sum of the Hilbert spaces of symmetric tensors~\cite{comonSymmetricTensorsSymmetric2008} of powers of $H_1$\footnote{With the convention that $H_1^{\otimes0}:=\RR$ and $x^{\otimes0}=1$}. Its associated feature map, $\phi:\RR^{d}\to H$, relies on the Taylor expansion, given, for any $x,y\in\RR^d$, as
\begin{equation}\label{eq:kernel_tensor_prod}
\langle\phi [x],\phi [y]\rangle_H=\sum_{r\in\NN} \frac{\langle x,y\rangle^r}{r!\sqrt{d^r}}=\sum_{r\in\NN}\frac{\langle x^{\otimes r},y^{\otimes r}\rangle}
{r!\sqrt{d^{r}}}.\end{equation}
For any $r\in\NN$ and any $x\in\RR^d$, we set $\phi_r[x]:=\frac{x^{\otimes r}}{\sqrt{r! d^{r/2}}}$.
Then the feature map satisfying \eqref{eq:kernel_tensor_prod}, is $\phi [x]=(\phi_{r}[x])_{r\in\NN}.$

We remark here that any sum of features generated by $\phi$ is a direct sum of symmetric
tensors of degree $r$, one for each $r\in\NN$. 
We denote by $K_\phi:H\to\RR^{N}$ the linear map whose rows are the vectors
$\phi [K_j]\in H$, $j\in[N]$. Being bounded and of finite rank, $K_\phi$
is a Hilbert-Schmidt operator, which allows us to manipulate it as we would with a matrix. Attention can then be expressed as
\begin{equation}
\Att(q,K,V)
=\frac{\phi [q]^{\T}K_\phi^{\T}V}{\phi [q]^{\T}K_\phi^{\T}\mathbf 1_N}
=\frac{\phi [q]^{\T} \Pi_{K_\phi} K_\phi^{\T}V}
      {\phi [q]^{\T} \Pi_{K_\phi} K_\phi^{\T}\mathbf 1_N},
\qquad
\Pi_{K_\phi}=K_\phi^{\T}\big(K_\phi K_\phi^{\T}\big)^{\dagger}K_\phi ,
\label{eq:att_compact}
\end{equation}
where $\Pi_{K_\phi}$ is the orthogonal projector onto $\Span(K_\phi)$ in
$H$. 
As already mentioned in the introduction, the attention is thus a ratio of two linear maps of $\phi [q]$, both of
which act on the same $N$-dimensional subspace of $H$, $\Span(K_\phi)$ determined by the key vectors. Consider the two following sets of elements of $H$ constructed from $K$, the elements of the first consists in the feature map of a sum of vectors while those of the second consists of a sum of feature maps.
\begin{align}
\phi\left[\Span(K)\right]&:=\phi\left[\left\{\sum_j\alpha_jK_j\in\RR^{d}
 \Big|\, \alpha_j\in\RR\right\}\right],
&
\Span(K_\phi)&:=\left\{\sum_j\alpha_j \phi [K_j]\in H
 \Big|\, \alpha_j\in\RR\right\}.
\end{align}
The first set is a submanifold while the second is a vector subspace, and they only match on the feature maps of the keys themselves.
 Indeed, $\phi:\RR^{d}\to H$ is differentiable, injective, and a homeomorphism over $\phi[\RR^{d}]$ using the restriction of the feature map to $H^{\otimes 1}$, $\phi_1:\RR^{d}\to \RR^{d}$.
\begin{proposition}\label{prop:intersection}
Let $K\in \RR^{N\times d}$, with distinct and nonzero rows. Then the intersection of the two sets is given by
\[ 
\Span(K_{\phi})\cap\phi[\Span(K)]
=\Span(K_{\phi})\cap\phi[\RR^{d}]=\{\phi [K_j]:j\in[N]\}.\]
\end{proposition}
The proof, given in Appendix~\ref{app:proofs}, relies on the following lemma.

\begin{lemma}
\label{lem:unicity_key_fm}
For any $u\in\RR^d\setminus\{0\}$ and any $\alpha\in\RR\setminus\{0,1\}$,
$\alpha \phi[u]\notin\phi[\RR^d]$. Furthermore, let $u \in \RR^d\setminus\{0\}$,
$k \in \NN^*$, and $u_i = \alpha_i u$ for $i \in [k]$, with pairwise distinct
$\alpha_1,\dots,\alpha_k \in \RR\setminus\{0\}$. Then, for every $R \in \NN$, the truncated
features $\big(\phi_r[u_i]\big)_{r=R}^{R+k-1}$, $i \in [k]$, are linearly independent and thus
$\dim \Span\{\phi[u_1],\dots,\phi[u_k]\} = k$.
\end{lemma}
The proof of \Cref{lem:unicity_key_fm} is also given in Appendix~\ref{app:proofs}.

The consequence we shall use is that the only elements of $\Span(K_\phi)$ which coincide with the actual feature map of a vector of $\RR^d$ are in the keys themselves.

\subsection{Interpolating attention}
\label{sec:interp}
We are interested in interpolating attention on a given reduced set of key vectors.
Let $L\in\RR^{n\times d}$ with $n\ll N$ be a matrix formed by a set of key vectors,
whose rows we call the \emph{interpolation nodes}, or magic points,
onto which we wish to interpolate the map
$\Att(\cdot,K,V):\RR^{d}\to\RR^{d}$, and let $L_\phi:H\to\RR^{n}$ be the
operator with rows $\phi [L_i]$. The reduced set of key features $\{\phi [L_i]\}_{i\in[n]}$ then span the basis space $\mathcal B$ onto which we interpolate, i.e. 
\begin{equation}
\mathcal B:=\Span(L_\phi)\subset \Span(K_{\phi})\subset H.
\label{eq:trial_def}
\end{equation} 
We now discuss an approach for interpolating the attention on this reduced set of keys. For any $q$ with $\phi [q]\in\mathcal B$, we can insert $\Pi_{L_\phi}$, the orthogonal projector onto $\mathcal B$, into
\eqref{eq:att_compact} without changing its value,
\[
\Att(q,K,V)
=\frac{\phi [q]^{\T} \Pi_{L_\phi} \Pi_{K_{\phi}} K_\phi^{\T}V}
      {\phi [q]^{\T} \Pi_{L_\phi} \Pi_{K_{\phi}} K_\phi^{\T}\mathbf 1_N},
\]
and by \Cref{prop:intersection} the nodes $q\in\{L_1,\dots,L_n\}$ are the only queries for which $\phi[q]\in\mathcal B$. The interpolation equality thus holds exactly at the nodes, although it may also hold at other queries, as attention only depends on $\phi[q]$ through its projection $\Pi_{K_\phi}\phi[q]$ onto $\Span(K_\phi)$, which may lie in $\mathcal B$ even when $\phi[q]$ does not. For any two matrices $A\in\RR^{a\times d}$ and $B\in\RR^{b\times d}$, we define
\[ 
  E_{AB}[i,j]:=\langle\phi [A_i],\phi [B_j]\rangle_H=\exp(\langle A_i,B_j\rangle/\sqrt d),\quad i\in[a],j\in[b],\qquad\text{i.e.}\quad E_{AB}=\exp(AB^{\T}/\sqrt d),
\]
where the element of the matrix in the \(i\)-th row and the \(j\) -th
column is denoted using square brackets and where the exponential is applied entrywise, we
obtain, 
\begin{equation*}
\Pi_{L_\phi}K_\phi^{\T}=L_\phi^{\T}(L_\phi L_\phi^{\T})^{\dagger}L_\phi   K_\phi^{\T}=L_\phi^{\T} E_{LL}^{\dagger} E_{LK}.
\end{equation*}
Hence, for a given \(L\), we have (using a similar method to the one used in
\cite{schroderWildCatNearLinearAttention2026}) that the associated interpolated
value vector \(W\in\RR^{n\times d}\) and the normalization weight
\(W_{\mathrm{Norm}}\) are the solutions to the following equations
\begin{equation}
  E_{LL}W = E_{LK} V\quad \text{ and }\quad
  E_{LL}W_{\mathrm{Norm}} = E_{LK} \mathbf 1_N.
  \label{eq:waug_solve}
\end{equation}
Setting \(\Waug\in\RR^{n\times (d+1)}\), and assuming \(E_{LL}\) is invertible, we can write, for simplicity, the solution to be
\begin{equation}
\Waug:=E_{LL}^{\dagger}E_{LK}\Vaug=[ W  \mid W_{\mathrm{Norm}} ],
\qquad \Vaug:=[ V  \mid \mathbf 1_N ].
\label{eq:waug}
\end{equation}
Indeed, as the numerator and denominator share their left factor \(K_{\phi}^\T\), 
we use a single basis space for the interpolation of both the numerator and denominator by considering the joint 
\(\RR^{d+1}\) augmented interpolation problem given by \begin{equation}\label{eq:lin_sys}E_{LL}\Waug=E_{LK}\Vaug.\end{equation}
The interpolant of attention on $\mathcal B$ is then given by
\begin{equation}
\widetilde\Att_L(q,K,V)
=\frac{\phi [q]^{\T} \Pi_{L_\phi} K_\phi^{\T}V}
      {\phi [q]^{\T} \Pi_{L_\phi} K_\phi^{\T}\mathbf 1_N}
=\frac{\phi [q]^{\T}L_\phi^{\T}W}{\phi [q]^{\T}L_\phi^{\T}W_{\mathrm{Norm}}} .
\label{eq:interpolant}
\end{equation}
Hence, \emph{the interpolant of
the attention mechanism is the ratio of the interpolants of the two
linear maps}. 

The residual we ideally want to minimise in the context of interpolation is the maximum error in $p\in\NN^*$ norm over all possible queries $q\in\RR^{d}$,
\[ \mathcal R_p(L)=\max_{q\in\RR^{d}}\lVert\Att(q,K,V)-\widetilde\Att_L(q,K,V)\rVert_p.\]
We note that, contrary to the exact denominator, the interpolated denominator $\phi [q]^{\T}L_\phi^{\T}W_{\mathrm{Norm}}$ is not guaranteed to be positive and may be equal to zero for some queries. In that case $\widetilde\Att_L$ has a pole and $\mathcal R_p(L)$ is infinite, we introduce a mean vector approximation in \S~\ref{sec:rc} to reduce the impact of this problem.

\begin{remark} To identify the sources of the interpolation error between $\phi[q]\Pi_{K_\phi}$ and $\phi[q]\Pi_{L_\phi}$, we first decompose $\phi[q]$ onto the $\Span(K_\phi)$ and its orthogonal complement 
by writing $\phi [q]^\T=\phi[q]^\T\Pi_{K_{\phi}}+\phi[q]^\T\Pi_{K_{\phi}}^\perp$.
Next, we look at the projection onto $\Span(L_\phi)$ and its complement $\Span(L_\phi)^\perp$ given by
\begin{align}
\phi [q]^{\T}\Pi_{L_\phi}\Pi_{K_\phi}
&=\phi [q]^{\T}\Pi_{K_\phi}\Pi_{L_\phi}\Pi_{K_\phi}
 +\phi [q]^{\T}\Pi^{\perp}_{K_\phi}\Pi_{L_\phi}\Pi_{K_\phi},
\label{eq:K_projections}\\
\phi [q]^{\T}\Pi^\perp_{L_\phi}\Pi_{K_\phi}
&=\phi [q]^{\T}\Pi_{K_\phi}\Pi^{\perp}_{L_\phi}\Pi_{K_\phi}
 +\phi [q]^{\T}\Pi^{\perp}_{K_\phi}\Pi^{\perp}_{L_\phi}\Pi_{K_\phi}.
\label{eq:K_projections_perp}
\end{align}
The first terms of both equations, \eqref{eq:K_projections} and \eqref{eq:K_projections_perp}, correspond to the part of the vector in $\Span(K_\phi)$ that is correctly reconstructed for the respective projections onto $\Span(L_\phi)$ and its complement, and their sum is equal to $\Pi_{K_\phi}$.
On the other hand, the second terms correspond to the part of the vector in $\Span(K_\phi)^\perp$ which is sent onto $\Span(K_\phi)$ by the projections and their sum equals zero, i.e. one is the correction to the other.
As \eqref{eq:K_projections_perp} corresponds to $\phi[q]^\T\Pi_{K_\phi}-\phi[q]^{\T}\Pi_{L_\phi}\Pi_{K_\phi}$, its terms represent how aligned or misaligned the two subspaces are. When $L_\phi$ is formed from the rows of $K_\phi$, we have that $\Span(L_\phi)\subseteq\Span(K_\phi)$. In that case, we have the inclusion $\Span(L_\phi)^{\perp}\supseteq\Span(K_\phi)^{\perp}$ and the misalignment terms vanish leaving the remaining first terms in \eqref{eq:K_projections} and respectively \eqref{eq:K_projections_perp} to represent the projectors onto the intersection $\Span(L_\phi)\cap \Span(K_\phi)$ and the orthogonal difference
$\Span(K_\phi)\ominus\Span(L_\phi)
=\Span(K_\phi)\cap\Span(L_\phi)^{\perp}$.
We are thus left with
\begin{equation}
\phi [q]^{\T}\Pi_{K_\phi}
=\phi [q]^{\T}\Pi_{L_\phi}+\phi [q]^{\T}\Pi_{K_\phi\ominus L_\phi},
\qquad
\Pi_{K_\phi\ominus L_\phi}:=\Pi_{K_\phi}\Pi_{L_\phi}^{\perp}\Pi_{K_\phi}.
\label{eq:ominus}
\end{equation}
In the context of this article, we restrict the selection to vectors based on the rows of $K_\phi$ for the interpolation nodes.
\end{remark}

\subsection{Invariants of attention}
\label{sec:invariances}
However, the linear problem \eqref{eq:waug_solve} is extremely ill-conditioned and computing its numerical solution is therefore very unstable. Indeed, the RKHS norms of the keys vary by orders of magnitude, resulting in ill-conditioned $E_{LK}$ and $E_{KK}$ matrices. However, using the invariants of the attention mechanism, we can solve an equivalent but better-conditioned problem through a Gaussian kernel formulation.

In this subsection, we consider three invariants of the attention mechanism and describe a Gaussian kernel reformulation of the attention mechanism (\S\ref{sec:gauss_main}). 
We then show that recentering the keys~\eqref{eq:gram_translation}, also used in WildCat~\cite{schroderWildCatNearLinearAttention2026}, gives us an additional degree of freedom and finally combine these results in \S\ref{sec:stab} to improve the conditioning of the interpolation procedure.

\begin{proposition}[Invariants of attention]
\label{prop:att_inv}
Consider \(q\in\RR^{d}\), \(K\in\RR^{N\times d}\) and \(V\in\RR^{N\times d}\). 
\begin{enumerate}[(i.)]
\item Query kernel-norm scaling:
Given $\lambda \in\RR\setminus\{0\}$, the attention as defined in~\eqref{eq:att_compact} is invariant under the transform $\phi[q]\to \lambda \phi[q]$.
Attention thus depends only on the direction of \(\phi [q]\) in \(H\), not on its magnitude.
\item Global key mass scaling:
 Given $\lambda \in\RR\setminus\{0\}$, the attention output is invariant under the rescaling of every key feature by the same $\lambda$, i.e. \(\phi
[K_i]\to\lambda \phi [K_i]\) for all \(i\in[N]\). 
\item Global key translation: 
For any \(\delta\in\RR^{d}\), the attention output is invariant under the translation $K_{i}\to K_{i}+\delta$ for all \(i\in[N]\), i.e. \(\Att(q,K+\delta,V)=\Att(q,K,V)\).
\end{enumerate}
\end{proposition}
The proof is given in Appendix~\ref{app:proofs}.
Each of these properties of attention allows us to consider for our interpolation a chosen representative of the `equivalence' class they create. 
\begin{enumerate}[I.]
\item Kernel-normalized queries.
As we approximate the ratio of linear maps by a ratio of two interpolated linear maps, we can restrict ourselves to kernel-normalized query vectors to reproduce the effect of attention\footnote{Note however that this does not imply any restriction on the norm of \(q\). Indeed, the only kernel-normalized vector of \(\phi[\RR^d]\) is \(\phi[0]\).  In the Gaussian rewriting \S\ref{sec:gauss_main}, this corresponds to restricting ourselves to the directional vectors given by the Gaussian feature map \(\breve\phi\)}, simplifying our Gaussian kernel reformulation in \S\ref{sec:gauss_main}.
\item Key normalization gauge. As attention is invariant by a global rescaling of the kernel norms of the keys, $\|\phi[K_i]\|_H$, only their ratios as defined later in equation~\eqref{eq:centered_etas} carry information. In \S\ref{sec:gauss_main} we use this to define the max-normalised gauge which allows us to restrict the numerical range of all kernel norms $\left(\|\phi [K_i]\|_{H}\right)_{i\in[N]}$ to \((0,1]\), and hence their logarithms, the shifted scaled squared $L^2$ norms $(\|K_i\|^2-\max_{j\in[N]}\|K_j\|^{2})/(2\sqrt d)$,  to \([-\max_{j\in[N]}\|K_j\|^{2}/(2\sqrt d), 0]\).
\item Key centering gauge. As attention is invariant to the translation of the keys by a same vector, the keys can be recentered. In~\eqref{eq:mean_centering}, we consider the mean of the key vectors as the origin, allowing us to reduce the spread of the mass ratios.
\end{enumerate}
Query translation, on the other hand, does change the output of attention (otherwise, it would be constant).
\begin{proposition}[Query translation]
\label{prop:query_translation_inv}Given $q, \delta\in\RR^d$, the effect of a translation of $\delta$ on the output of attention is given by
\[
\Att(q+\delta,K,V)=\frac{\left[\exp\left(q^\T K^\T/\sqrt{d}\right)\odot\exp\left(\delta^\T K^\T/\sqrt{d}\right)\right] V} {\left[\exp\left(q^\T K^\T/\sqrt{d}\right)\odot\exp\left(\delta^\T K^\T/\sqrt{d}\right)\right]\mathbf 1_N}=\frac{\phi [q]^{\T}K_\phi^{\T}\Xi_{K,\delta}V}{\phi [q]^{\T}K_\phi^{\T}\Xi_{K,\delta}\mathbf 1_N}, \]
where we set $\Xi_{K,\delta}=\Diag\left((\langle \phi [\delta],\phi[K_i]\rangle)_{i\in[N]}\right)$.
\end{proposition}
\subsection{Gaussian formulation and centering of the keys}
\label{sec:gauss_main}
The main tool that allows us to make use of these invariants is the reformulation of attention using the Gaussian kernel. We denote the Gaussian kernel (or Radial Basis Function kernel) by $\breve\kappa(x,y)=\exp\left({-\| x-y\|^2/2\sqrt d}\right)=\|\phi [x-y]\|_{H}^{-1}$; we denote with the breve symbol $\breve\cdot$ the objects attached to this Gaussian kernel. For any $x,y\in\RR^d$, the exponential kernel can be rewritten as
\begin{equation}\label{eq:gauss_exp}
\kappa_{\exp}(x,y)=\|\phi[x]\|_H \breve\kappa(x,y) \|\phi [y]\|_H.
\end{equation}
This can equivalently be expressed by defining the Gaussian feature map $\breve \phi[x]:=\frac{\phi[x]}{\|\phi[x]\|_{H}}$, such that, for any $x,y\in\RR^{d}$, we have $\langle\breve \phi [x],\breve\phi[y]\rangle_{H}=\breve\kappa(x,y)$. As discussed later, a useful property of this reformulation is that it separates the norm $\|\phi[x]\|_{H}$, and direction $\breve \phi[x]$, of the exponential kernel features $\phi[x]$. Then, for any Gaussian kernel evaluations on the matrices $A\in\RR^{a\times d}$ and $B\in\RR^{b\times d}$, we set
\begin{equation}
\breve E_{AB}[i,j]=\frac{\langle\phi [A_i],\phi [B_j]\rangle_H}
      {\lVert\phi [A_i]\rVert_H \lVert\phi [B_j]\rVert_H} = \langle\breve\phi[A_i],\breve\phi[B_j]\rangle_H,\quad i\in[a],j\in[b],\label{eq:gauss_def}      
\end{equation}
where we denote the element of the matrix in the $i$-th row and $j$-th column using square brackets.
The Gaussian gram matrix can hence be related to the exponential kernel one as
\begin{equation}
\label{eq:gauss_gram_rewrite}
E_{LL}= D_{L_\phi}^{1/2}\breve E_{LL}D_{L_\phi}^{1/2},\quad E_{KK}= D_{K_\phi}^{1/2}\breve E_{KK}D_{K_\phi}^{1/2}
\end{equation}
 where we set, for any $B\in\RR^{ b\times d}$, $D_{B_\phi}=\Diag\left(\exp({\| B_i\|^2_2/\sqrt d})\right)_{i\in[b]}=\Diag(\|\phi [B_i]\|_{H}^2)_{i\in[b]}$, abbreviated $D_B$ when no ambiguity arises. With
$\breve B_\phi:=D_B^{-1/2}B_\phi$ the operator with rows
$\breve\phi[B_i]$ for any $B\in\RR^{b\times d}$, we obtain
\begin{equation}\label{eq:gauss_attn}
    \Att(q,K,V)=\frac{\breve \phi[q]^{\T}\breve K_\phi ^\T D_{K_\phi}^{1/2} V}{\breve\phi[q]^{\T}\breve K_\phi ^\T D_{K_\phi}^{1/2} \mathbf 1_N}=\frac{\breve E_{qK}D_{K_\phi}^{1/2}V}{\breve E_{qK}D_{K_\phi}^{1/2}\mathbf 1_N}.
\end{equation} 
We also obtain for the interpolation,
\begin{equation}\label{eq:gauss_interp}
\widetilde {\Att}_L(q,K,V)=\frac{\breve \phi[q]^{\T}\breve L_\phi ^\T D_{L_\phi}^{1/2} W}{\breve\phi[q]^{\T}\breve L_\phi ^\T D_{L_\phi}^{1/2} W_{\mathrm{Norm}}}=\frac{\breve E_{qL}D_{L_\phi}^{1/2}W}{\breve E_{qL}D_{L_\phi}^{1/2}W_{\mathrm{Norm}}}
\end{equation}
Hence, the interpolated weights \eqref{eq:waug} are given as 
\begin{equation}
W=E_{LL}^{\dagger}E_{LK}V
=D_L^{-1/2} \breve E_{LL}^{\dagger} \breve E_{LK} D_K^{1/2} V ,\quad W_\mathrm{Norm}=E_{LL}^{\dagger}E_{LK}\mathbf 1_N
=D_L^{-1/2} \breve E_{LL}^{\dagger} \breve E_{LK} D_K^{1/2} \mathbf1_N .
\label{eq:Gauss_galerkin}
\end{equation}

Furthermore, due to invariant \emph{(ii.)} and Gaussian formulation, we can replace \(D_{K_\phi}\) in Equation~\eqref{eq:gauss_attn} by its
\(L^\infty\)-normalisation, which we denote as
\begin{equation}
  \eta_i := 
  \frac{\lVert\phi [K_i]\rVert_H}{\max_{j}\lVert\phi [K_j]\rVert_H}
   = \exp\!\Big(\frac{\lVert K_i\rVert^2-\max_j\lVert K_j\rVert^2}
  {2\sqrt d}\Big)
   = \frac{D_{K,ii}^{1/2}}{\lVert D_K^{1/2}\rVert_\infty} \in (0,1].
  \label{eq:kernel_norm_weight}
\end{equation}

The global key translation invariance \emph{(iii)} of \Cref{prop:att_inv} represents an additional degree of freedom for interpolation. As we have seen before,  the Gaussian formulation allows us to separate $\phi[x]$ into its norm from the direction given by the feature map $\breve \phi$, i.e. the normalised vector in the RKHS. Due to the translation invariance of the Gaussian kernel, only the norm of the keys depends on the recentering.

\begin{proposition}[Gram matrices for translated keys]
  \label{prop:gauss_inv}
  For any \(A\in\RR^{a\times d}\), \(B\in\RR^{b\times d}\) and any \(\delta\in\RR^{d}\), we have $\breve E_{A-\delta,B-\delta}=\breve E_{AB}$ and hence
  \[
    E_{A-\delta,B-\delta}=D_{(A-\delta)_\phi}^{1/2} \breve E_{AB} D_{(B-\delta)_\phi}^{1/2}=\Xi^{-1}_{A,\delta} E_{AB}\Xi_{B,\delta}^{-1}\cdot\exp\left(\frac{\|\delta\|^2}{\sqrt{d}}\right).
  \]
  where $\Xi$ is defined as in Proposition~\ref{prop:query_translation_inv} and we set
\[
  D_{(A-\delta)_\phi}^{1/2}[i,i]=\|\phi [A_i-\delta]\|_H = e^{\|A_i\|^2/2\sqrt{d}}e^{-\langle A_i,\delta\rangle/\sqrt{d}}e^{\|\delta\|^2/2\sqrt{d}}.
\]
\end{proposition}
The proof is given in Appendix~\ref{app:proofs}.
In our case, this leads to the following,
\begin{equation}\label{eq:gram_translation}
  E_{L-\delta,K-\delta}=e^{\|\delta\|^2/\sqrt{d}}\cdot \Xi_{L,\delta}^{-1} E_{LK}\Xi_{K,\delta}^{-1}\quad \text{ and } \quad E_{K-\delta,K-\delta}=e^{\|\delta\|^2/\sqrt{d}}\cdot \Xi_{K,\delta}^{-1} E_{KK}\Xi_{K,\delta}^{-1}.
\end{equation}

Comparing \(D_{K_\phi}\) and \(D_{(K-\delta)_\phi}\), we see that the effect of recentering on the ratio of norms is given by
\[
  \frac{\|\phi [K_i-\delta]\|_H}{\|\phi [K_j-\delta]\|_H}=e^{\frac{\|K_i\|^2-\|K_j\|^2-2\langle K_i-K_j,\delta\rangle}{2\sqrt{d}}}=\frac{\|\phi [K_i]\|_H}{\|\phi [K_j]\|_H}\cdot e^{-\frac{\langle K_i-K_j,\delta\rangle}{\sqrt{d}}}.
  \]
We note that instead of being exact at the chosen keys, the interpolation is exact at the translated nodes $L_i-\delta$.

\begin{proposition}[Interpolation with translated keys]
  \label{prop:translated_interp}
  Given $\delta\in\RR^d$, the interpolation of attention on $\delta$-translated keys gives
  \[
  \widetilde\Att_{L-\delta}(q,K-\delta,V)
  =\frac{E_{qL}\Xi_{L,\delta}E_{LL}^{\dagger}E_{LK}\Xi^{-1}_{K,\delta}V}
        {E_{qL}\Xi_{L,\delta}E_{LL}^{\dagger}E_{LK}\Xi^{-1}_{K,\delta}\mathbf 1_N}.
\]
The interpolation is hence exact at the translated nodes $L_i-\delta$, $i\in[n]$, whose features $\{\phi[L_i-\delta], i\in[n]\}$ are the only points of $\phi[\RR^d]$ in the interpolation basis,
and the interpolation basis is
\[
  \mathcal B_\delta:=\Span\big((L-\delta)_\phi\big) \neq \mathcal B.
\]
Translation invariance is in particular recovered when $\Pi_{L_\phi}K^\T_{\phi}=K^{\T}_{\phi}$.
\end{proposition}
The proof is given in Appendix~\ref{app:proofs}.
Let us now consider which point of origin we want to choose. As we have seen above, the selection of the centering point $\delta$ only affects the distribution of the scaled squared $L^2$ norms, $\|K_i-\delta\|^2/(2\sqrt{d})=\log\|\phi[K_i-\delta]\|_H$ for $i\in[N]$. Therefore, for stability, our choice is to minimize the sum of the scaled squared $L^2$ norms, in general improving the conditioning of the problem while making the norms of the vectors more comparable by concentrating the distribution of the norm of the keys. Indeed, as the Gaussian Gram matrix is constructed from normalised vectors and the Gaussian scalar product decreases exponentially fast with distance, the conditioning of the problem depends mainly on the $D_{K_\phi}$ matrices.
Similarly to  \cite{schroderWildCatNearLinearAttention2026}, we choose to recenter the key vectors on the mean vector
\begin{equation}\label{eq:mean_centering}
\mu=\frac{1}{N}\sum_{j=1}^{N}K_j.
\end{equation}
This choice of the center $\mu$ is motivated by the fact that, as the mean of the key distribution, it minimizes the sum of squared distances $\mu=\underset{\delta\in\RR^{d}}{\arg \min}\sum_{i\in[N]}\|K_{i}-\delta\|^2$, and thus the sum of the scaled squared $L^2$ norms, $\sum_{i\in[N]}\|K_i-\delta\|^2/(2\sqrt{d})$. We note however that this does not in general minimize the spread of these norms, which may increase after centering for particular distributions of the keys.
\subsubsection{Stabilizing interpolation through the Gaussian formulation and centering of the keys}
\label{sec:stab}

The Gaussian reformulation described in  \S\ref{sec:gauss_main} and the centering of the key vectors with respect to the mean described in equation \eqref{eq:mean_centering} now allow us to construct a better conditioned linear system equivalent to equation~\eqref{eq:waug_solve}.
As discussed above, the Gaussian kernel matrices $\breve E_{LL}$ and $\breve E_{LK}$ are in general much better conditioned than their exponential counterparts (as shown in Appendix~\ref{app:conditioning}). This means that the ill-conditioning is almost entirely due to the matrices $D_{L_\phi}$ and $D_{K_\phi}$, whose conditioning depends only on the spread of the scaled squared $L^2$ norms of the keys, $\|K_i\|^2/(2\sqrt{d})$ for $i\in[N]$.
 We now look at how the invariants allow us to stabilize the problem.
 Let \(\eta^{(c)}_{L}\in(0,1]^{n}\) and
\(\eta^{(c)}_{K}\in(0,1]^{N}\) be the gauged norms, which we will refer to as masses, of the centered interpolation nodes and
keys,
\begin{equation}
\label{eq:centered_etas}
  \eta^{(c)}_{L,i}=\frac{\|\phi [L_i-\mu]\|_H}{\max_{j}\|\phi [K_j-\mu]\|_H},
  \qquad
  \eta^{(c)}_{K,i}=\frac{\|\phi [K_i-\mu]\|_H}{\max_{j}\|\phi [K_j-\mu]\|_H},
\end{equation} 
both evaluated in log space.
Then, the augmented Galerkin weights \eqref{eq:Gauss_galerkin} can be obtained by the two-step solve and scale method
\begin{equation}
  \breve E_{LL} X=\breve E_{LK} \Diag\big(\eta^{(c)}_{K}\big) \Vaug,
  \qquad
  {\Waug}^{(c)}=\Diag\big(\eta^{(c)}_{L}\big)^{-1}X.
  \label{eq:stable_galerkin}
\end{equation}
Then, the interpolated attention on $\mathcal B_{\mu}$ is given by
\begin{equation}
  \widetilde\Att_{L^{(c)}}(q,K,V)=\frac{\breve E_{q,L-\mu} \Diag(\eta^{(c)}_{L}) W^{(c)}}
  {\breve E_{q,L-\mu} \Diag(\eta^{(c)}_{L}) W^{(c)}_{\mathrm{Norm}}} =\frac{E_{qL}W^{(c)}}{E_{qL}W_{\mathrm{Norm}}^{(c)}},
  \label{eq:stable_readout}
\end{equation}
where we note that the second equality uses $L$ and not $L-\mu$.
The masses of the nodes are kept bounded in \((0,1]\) thanks to the $\eta$ formulation, the Gaussian kernel
matrix has in general a moderate condition number (see Appendix~\ref{app:conditioning}) and the right hand side's rescaling is controlled thanks to the mean recentering.
\emph{From now on, unless otherwise specified, we will consider the problems in the $\mu$-centered context and drop the $(c)$.}
Note also that $\Waug$ is remultiplied by \(\Diag(\eta^{(c)}_L)^{-1}\) after the solve, rescaling it to be in the same gauge as $\Vaug$.
Indeed, as the two terms in Eq.\eqref{eq:centered_etas} share their denominators, the gauge is cancelled by construction.
While having different denominators would have no impact on the interpolated attention output,
it would affect the scaling of the resulting interpolated value matrix. We choose to keep it
  comparable to $V$, notably for range correction which is described in \S\ref{sec:rc}.
\begin{remark}[Clustering the data]
  When the data is distributed in clusters, even for example as a mixture of Gaussians,
  the conditioning of the problem can then be heavily impacted through the renormalised masses $\eta$.
  Partitioning the nodes into clusters \(C_1,\dots,C_m\) and anchoring each at its own mean \(\mu_{C_{i}}\)
  combined with a separate solve on each cluster would provide a way to take this issue into account.
  As we shall see in \S\ref{sec:streaming}, the recombination is similar to what we already do in the context of streaming attention for
  each successive segment, the content-based clustering is left as future work.
\end{remark}

Hence, while the key centering can be seen as a hyperparameter which allows us to select our exactness set through $\mathcal B_\delta$,
it also affects our capacity to solve the interpolation problem and we will restrict ourselves to the $\mu$-centering in the rest of this article.
Indeed, while equivalent in exact arithmetic, the origin centering would require too much precision due to the redundant linear term while the mean centering keeps
every direction resolvable~\eqref{eq:mean_centering}. Furthermore, the mean centering allows  to have an interpolation basis which may be more representative of the future queries as they tend to not align exactly with the keys.

\section{Empirical interpolation in infinite-dimensional spaces}
\label{sec:methods}
As discussed in the previous section, interpolation can be divided into two distinct but connected tasks, the \emph{Selection} of the basis keys, i.e.
 constructing a basis space $\mathcal B$, and the \emph{Projection} of the augmented linear map onto it, i.e.  approximating $K_\phi^\T V$ through a Petrov-Galerkin \emph{Projection} by minimising the residual on a test space $\mathcal T$ that is potentially different from $\mathcal B$.
\subsection{Basis selection}
\label{sec:sel}
The selection chooses the basis space $\mathcal B$ of the RKHS, as previously described in \S\ref{sec:interp}.
Projection determines the linear operator that maps the input (projected onto the basis), and thus the output
space.
In this section, we first define the residual operators and error terms (\S\ref{sec:err}), and present two selectors:
\begin{itemize} 
  \item The GAIM selector in  \S\ref{sec:selection} is based on greedily selecting one after the other the interpolation nodes with the highest residuals in the $L^\infty$ norm, iteratively removing the components with the highest error in any output dimension, following the GEIM method~\cite{madayGeneralizedEmpiricalInterpolation2013}. We also provide an $L^2$ norm alternative that chooses the directions that minimize the sum of squared errors.
  \item The Q-DAIM selector in \S\ref{sec:sel_qdeim} is based on the
  Q-DEIM~\cite{drmacDiscreteEmpiricalInterpolation2018,drmacNewSelectionOperator2016}
  method which reframes the \(L^2\) formulation of the residual minimization problem in the metric induced by the linear map. This is done by constructing an orthogonal basis for the linear map through either a CholeskyQR \cite{stathopoulosBlockOrthogonalizationProcedure2002} or through the Singular Value Decomposition, a method known in the data-science literature as Cholesky or PCA whitening respectively \cite{Kessy02102018}. The algorithm then selects nodes to maximise the volume of the approximation in the obtained basis. 
  \end{itemize} 
We also present an approach for taking into account multiple query heads in $L^2$ norm using a multi-head scoring metric in
\S\ref{sec:multihead}. Finally, we present approximate finite-dimensional surrogates for the computation of the residuals 
that allow a linear complexity in terms of context size in \S\ref{sec:surrogate} and thus allow for the compression of longer contexts.
Let us now take a look at the residuals we want to minimize.

\subsubsection{Approximation of the residual error}
\label{sec:err}

Consider the numerator and denominator of the attention output that we denote $g(q)=\phi [q] ^\T M^{g}\in\RR^{d}$ and $h(q)=\phi [q] ^\T M^{h}\in\RR^+$, where we set $M^{g}:=K_\phi^{\T}V$ and  $M^{h}=K_\phi^{\T}\mathbf 1_N$. The ratio $g(q)/h(q)=\frac{\phi [q] ^\T M^{g}}{\phi [q] ^\T M^{h}}$ is therefore determined by applying the combined \emph{augmented} linear operator to $\phi[q]$, this operator is defined as
\begin{equation}\label{eq:M_op}
M:=K_\phi^{\T}\Vaug=\big[ M^{g} \mid M^{h} \big].
\end{equation}
For interpolated attention, we denote $\tilde g=\phi [q]^{\T}M_L^{g}$ and $\tilde h=\phi [q]^{\T}M_L^{h}$ similarly, where $M_L:=\Pi_{L_\phi}M=L_\phi^{\T}\Waug$.
We can then write that \[
\Att(q,K,V)=g(q)/h(q)\quad\text{ and }\quad\widetilde\Att_L(q,K,V)=\tilde g(q)/\tilde h(q).\] 
Using $\Pi_{L_\phi}K_\phi^{\T}=L_\phi^{\T}E_{LL}^{\dagger}E_{LK}$ and the weights
\eqref{eq:waug}, the decomposition \eqref{eq:ominus} splits $M$ into
its component in the basis space $\mathcal B$ and its orthogonal complement in $\Span(K_\phi)$,
\begin{equation}
M=M_L+M_\Delta,
\qquad\text{ with }
\qquad
M_\Delta:=\Pi_{K_\phi\ominus L_\phi}M=M-L_\phi^{\T}\Waug.
\label{eq:resop}
\end{equation}
We call $M_\Delta$ the \emph{augmented residual operator}. Indeed, the residual error of $g$
and $h$ due to the interpolation method are given by the first $d$ columns and the last column of
$M_\Delta=\big[ M^{g}_{\Delta} \mid M^{h}_{\Delta} \big]$ respectively. The residual error is then given by
\[
\Delta g=g-\tilde g=\phi [q]^{\T}M^{g}_{\Delta},
\quad\text{ and }
\Delta h=h-\tilde h=\phi [q]^{\T}M^{h}_{\Delta}.
\]
The residual error on the attention output is then obtained as
\[
\Att-\widetilde\Att_L
=\frac{g}{h}-\frac{\tilde g}{\tilde h}=\Big(\frac{g}{h}-\frac{g}{\tilde h}\Big)
+\Big(\frac{g}{\tilde h}-\frac{\tilde g}{\tilde h}\Big)
=-\frac{g}{h \tilde h} \Delta h+\frac{\Delta g}{\tilde h}
=\frac{1}{\tilde h}\Big(\Delta g-\frac{g}{h} \Delta h\Big).
\]
This admits the closed form
\begin{equation}
\Att-\widetilde\Att_L
=\frac{1}{\underbrace{\phi [q]^{\T}L_\phi^{\T}W_{\mathrm{Norm}}}_{T_3}}
\Big(\underbrace{\phi [q]^{\T}M^{g}_{\Delta}}_{T_1}
 - \underbrace{\frac{\phi [q]^{\T}M^{g}}
{\phi [q]^{\T}M^{h}} \phi [q]^{\T}M^{h}_{\Delta}}_{T_2}\Big).
\label{eq:residual_exact}
\end{equation}

This formulation allows the usage of the residual given in \eqref{eq:residual_exact} for interpolation.
Indeed, each of the three terms depends only on the query through the prefactor $\phi [q]^{\T}$ and is linear in $M_\Delta$, $M_{\Delta}^h$ and $M_{L}^{h}$, respectively.
Hence, for any new node $x\in\RR^d$, we can update the residual to the new set $L'=  L\cup\{x\}$ using rank-$1$ updates of $M_\Delta$ and $M_{L}$.
Indeed, let $\rho[x]:=\Pi_{K_{\phi}\ominus L_{\phi}}\phi [x]$ be the projection of the new node on $\Span(K_\phi)\ominus \Span(L_\phi)$ and set $s_{L}(a,b):=\langle a,\Pi_{K_{\phi}\ominus L_{\phi}}b\rangle_H=E_{ab}-E_{aL}E_{LL}^{\dagger}E_{Lb}$, the scalar product in $H$ restricted to $\Span(K_\phi)\ominus \Span(L_\phi)$. The norm of $\rho[x]$ is given by the norm of $s_{L}(x,x)=E_{xx}-E_{xL}E_{LL}^{\dagger}E_{Lx}$. Then the projection onto $\Span(L'_{\phi})$ is given by $\Pi_{L'_\phi}=\Pi_{L_\phi}+\rho [x]\rho [x]^{\T}/s_{L}(x,x)$ and we have
\[
M_{L'}=M_{L}+\frac{\rho [x] r_x}{s_{L}(x,x)},
\qquad
M_{\Delta}'=M_{\Delta}-\frac{\rho [x] r_x}{s_{L}(x,x)},
\qquad
r_x:=\phi [x]^\T \Pi_{K_{\phi}\ominus L_{\phi}} M=\phi [x]^\T M_{\Delta}
=\begin{bmatrix} r^{g}_{x} &  r^{h}_{x} \end{bmatrix}.
\]
At a query $q$, the updates to the three blocks are then obtained through the
scalar $s_L(q,x):=\phi[q]^{\T}\Pi_{L_\phi}^{\perp}\phi[x]=\phi [q]^{\T}\rho [x]$:
\begin{equation}\label{eq:rank1_T123}
T_1'=T_1-\frac{s_L(q,x)}{s_L(x,x)} r^{g}_{x},
\qquad
T_2'=T_2-\frac{g}{h} \frac{s_L(q,x)}{s_L(x,x)} r^{h}_{x},
\qquad
T_3'=T_3+\frac{s_L(q,x)}{s_L(x,x)} r^{h}_{x},
\end{equation}
where we note that the values that need to be computed are the evaluations of 
\[ r_{x}= \left(E_{xK}-E_{xL}E_{LL}^{\dagger}E_{LK}\right)\Vaug,\quad s_{L}(x,x)=E_{xx}-E_{xL}E_{LL}^{\dagger}E_{Lx}\quad \text{ and }\quad s_L(q,x) =E_{qx}-E_{qL}E_{LL}^{\dagger}E_{Lx}.\]
Hence, $T_1$, $T_2$ and $T_3$ can be
updated \emph{separately} for each rank-one extension of $L$, which is what makes our GEIM-based greedy selection computationally feasible.
We note that the above quantities cannot be computed through $\rho [x]$, which lives
in $H$. They are thus obtained through kernel evaluations and projected
 through either the inverse of the node or using a solve from its Cholesky factorization. 
Thus, we keep in memory either the inverse or a Cholesky factorization of the Gram matrix for the projection step.
With the addition of a new node $x\in \RR^d$, we therefore use a rank-$1$ update for the addition of the new border
\[
  E_{L'L'}=\begin{pmatrix}
    E_{LL} & E_{Lx}\\
    E_{xL} & E_{xx}
  \end{pmatrix}.
\]
Setting $s_L(x,x)=E_{xx}-E_{xL}E_{LL}^{\dagger}E_{Lx}$,
$u=E_{LL}^{\dagger}E_{Lx}$ and $\Gamma_{LL}\Gamma_{LL}^{\T}=E_{LL}$ the
Cholesky factorization, we can write
\begin{equation}\label{eq:rank1_inverse}
E_{L'L'}^{\dagger}
=\begin{pmatrix}
E_{LL}^{\dagger}+\dfrac{u u^{\T}}{s_L(x,x)} & -\dfrac{u}{s_L(x,x)}\\[6pt]
-\dfrac{u^{\T}}{s_L(x,x)} & \dfrac{1}{s_L(x,x)}
\end{pmatrix},
\qquad
\Gamma_{L'L'}
=\begin{pmatrix}
\Gamma_{LL} & 0\\
\gamma^{\T} & \sqrt{s_L(x,x)}
\end{pmatrix},
\quad \gamma:=\Gamma_{LL}^{\dagger}E_{Lx},
\end{equation}
both at $O(n^{2})$ cost in memory per node.
We note that in practice, the computations are done using the  Gaussian formulation from equation~\eqref{eq:gauss_gram_rewrite}, which gives the following. \[E_{LL}^{\dagger}=D_{L_\phi}^{-1/2}\breve E_{LL}^{\dagger}D_{L_\phi}^{-1/2}.\]
Thus, we can compute using the same formulae as the evaluations replaced by their Gaussian counterparts $\breve s_L(a,b)=\frac{s_{L}(a,b)}{\|\phi [a]\|_{H}\|\phi[b]\|_H}$ and $\breve \rho[x]=\rho[x]/\|\phi[x]\|_{H}$.
A useful corollary is that the term $\breve s_L(x,x):=1-\breve E_{xL} \breve E_{LL}^{\dagger} \breve E_{Lx}$ can then be used to determine the collinearity of a new node in the selection loop.

The residual is composed of three terms, $T_1=\Delta g$ associated with the linear residual of the numerator $M^g_{\Delta}$, $T_2=\frac{g}{h} \cdot \Delta h$ corresponding to the linear effect of the residual of the denominator $M_{\Delta}^{h}$ on the error and
the prefactor $\frac{1}{T_3}$ that corresponds to the multiplicative effect related to the interpolated denominator. For our method, we consider five variants of the residual combining these terms:
\begin{itemize}
  \item[\emph{(v1.)}] The value residual $T_1$,
  \item[\emph{(v2.)}] the unnormalised error residual $T_1-T_2$,
  \item[\emph{(v3.)}] the full residual  $\frac{T_1-T_2}{T_3}$,
  \item[\emph{(v4.)}] the $h$-normalised full residual $\frac{T_1-T_2}{h}$, replacing the denominator by its exact value,
  \item[\emph{(v5.)}] and the augmented-moment residual $\phi [q]^{\T}M_\Delta$, which corresponds to the residual of the interpolation map itself.
\end{itemize}
We also consider two possible norms, $L^\infty$ and $L^2$, the first one more suited to the GEIM-inspired method and the second one to the Q-DEIM method of the next subsection.

\subsubsection{Greedy selection of the magic points by GEIM}
\label{sec:selection}
Following the generalized empirical interpolation method \cite{madayGeneralizedEmpiricalInterpolation2013} (and the original empirical
interpolation method),
we construct $L$ by greedily selecting the nodes from a dictionary of vectors that iteratively best
minimize one of the residual variants v1-v5,
evaluated at the Gaussian directions of the keys, $\breve\phi[K_i]$.
We name this the \emph{greedy attention
interpolation method} (GAIM).
 In this paper, we restrict the dictionary to the rows of $K$, which allows the error to be restricted to $\Span(K_\phi)\ominus\Span(L_\phi)$,  as mentioned in \S~\ref{sec:interp}.

In practice, while we choose our dictionary of possible directions as the set of keys itself, it should be noted
that any arbitrary set of vectors of $\RR^d$ may be chosen for this method. Furthermore, one might ideally want to 
obtain the vector in $H$ maximizing the residual. However, constructing such a vector often requires 
access to all the keys as we can only reconstruct it through the feature map $\phi$. 

Let $\Vaug_\eta:=\Diag(\eta) \Vaug=[V_\eta\mid \eta]$, where $\eta$ are the gauged masses defined in 
\eqref{eq:kernel_norm_weight}.
 We consider the $\eta$-gauged augmented operator and residual operator equivalent to \eqref{eq:M_op},
\[
M_\eta:=\breve K_\phi^{\T}\Vaug_\eta
=\frac{M}{\lVert D_K^{1/2}\rVert_\infty},
\qquad
M_{\Delta,\eta}:=\Pi_{K_\phi\ominus L_\phi} M_\eta
=\frac{M_\Delta}{\lVert D_K^{1/2}\rVert_\infty} .
\]
Thus, in the following proposition, every residual score is expressed in the $L^\infty$ gauge.

\begin{proposition}[Residual scores]
\label{prop:discrete_scores}
Let
$\breve S_L:=\breve E_{KK}-\breve E_{KL} \breve E_{LL}^{\dagger} \breve E_{LK}$
be the projection onto the orthogonal complement of $\Span(L_\phi)$ of the  Gaussian Gram matrix $\breve E_{KK}$. Then
$\breve S_L=\breve K_\phi \Pi_{K_\phi\ominus L_\phi} \breve K_\phi^{\T}\succeq0$
and for every $i\in[N]$, the residual of the linear map $M$ evaluated on the keys $K$ is given by
\[
R_\Delta[i,:]:=\Big\langle\breve \phi [K_i], 
M_{\Delta,\eta}\Big\rangle_H
=\big[ \breve S_L \Vaug_\eta \big]_{i,:} \in \RR^{d+1}.
\]
\end{proposition}
The proof is given in Appendix~\ref{app:proofs}.
Then every residual variant can be assembled from $R_\Delta$,  and from $R^{0}:=\breve E_{KK} \Vaug_\eta$, the evaluation of the augmented operator on the keys. Indeed, the row $i$ of these matrices gives us
\[
 g_i=R^{0}[i,:d],\quad h_i=R^{0}[i,d+1]\quad \text { and }\quad \Delta g_i=R_{\Delta}[i,:d],\quad \Delta h_i=R_\Delta[i,d+1].
\]
Hence, we can update $T_1(i)=\Delta g_i$, $T_2(i)=(g_i/h_i) \Delta h_i$ and $T_3(i)=h_i-\Delta h_i$, at a cost
$O(N(d+1))$ per pick.

Let $R_{\mathrm{sel}}$ be the matrix of the chosen residual
variant and the norm $p\in\{\infty,2\}$, the next node is obtained using
\begin{equation}
i^\star=\arg\max_{i\notin L} 
\big\lVert R_{\mathrm{sel}}[i,:] \big\rVert_p,
\qquad L\leftarrow L\cup\{K_{i^\star}\},
\label{eq:greedy_argmax}
\end{equation}
where for simplicity, we denote, for any $v\in\RR^{d}$ and $A\in\RR^{a\times d}$, by $A\cup v$ the concatenation of the vector $v$ at the end of $A$.
We note that, similarly to GEIM, this is a greedy method which selects the node with the largest current residual, which is not guaranteed to be the optimal selection to reduce the residual. 
Indeed, for the Galerkin residual, adding the key $K_i$ decreases $\|M_{\Delta,\eta}\|_{\mathrm{HS}}^2$ by $\|R_\Delta[i,:]\|_2^2/\breve S_L[i,i]$.

In practice, we note that for the pairing (v5.,\(L^\infty\)) the addition of the denominator column does not seem to work well with the infinity norm.

The exact-kernel selection loop is summarized in Algorithm~\ref{alg:gaim_exact}.
The rank-1 residual updates, computed for $q\in(K_{i})_{i\in [N]}$ using \eqref{eq:rank1_T123}, have only a linear cost in $N$. The overall complexity is therefore of order $O(N^2d+Nn(d+n))$.
Indeed, we only maintain the residual rows and $\breve S_L$, although at the cost of keeping the $n\times n$ matrix $E_{LL}^{\dagger}$ in memory. The quadratic cost of the method comes from the initialization of $R^{0}$ which requires computing $E_{KK}$ and hence the scalar product between each pair of keys for a total of $N^{2}$ kernel evaluations.

\begin{algorithm}[t]
\caption{Exact-kernel greedy GAIM selection}
\label{alg:gaim_exact}
\begin{algorithmic}[1]
\Require keys $K\in\RR^{N\times d}$, values $V\in\RR^{N\times d}$,
budget $n$, residual variant $R_{\mathrm{sel}}$, norm $p$, collinearity floor $\tau_{\sigma}$
\State $\Vaug\gets[ V \mid \mathbf 1_N ]$;\quad
$\eta\gets$ gauged masses of \eqref{eq:kernel_norm_weight}, in log space;\quad
$\mathfrak L\gets\varnothing$;\quad $\breve E_{LL}^{\dagger}\gets[ ]$
\State $\breve E_{KK}\gets\big[\exp(-\|K_i-K_j\|^2/2\sqrt{d})\big]_{i,j\in[N]}$;\quad
$R^{0}\gets\breve E_{KK} \Diag(\eta) \Vaug$;\quad
$R_{\Delta}\gets R^{0}$
\Comment{Gram, evaluations and residual rows at $L=\varnothing$}
\For{$t=1,\dots,n$}
  \State Update $R_{\mathrm{sel}}$ from
  $(R_{\Delta},R^{0})$
  \Comment{chosen residual variant, \S\ref{sec:selection}}
  \State $i^\star\gets\arg\max_{i\notin \mathfrak L}
  \lVert[R_{\mathrm{sel}}]_{i,:}\rVert_p$
  \Comment{selection in chosen norm}
  \State $u_{i^\star}\gets[\breve E_{KK}]_{:,i^\star} - \breve E_{KL}\breve E_{LL}^{\dagger}[\breve E_{LK}]_{:,i^\star}$;\quad
  $\sigma\gets [u_{i^\star}]_{i^\star}$;\quad check $\sigma\ge\tau_{\sigma}$
  \Comment{Schur complement column, collinearity check}
  
  \State $R_{\Delta}\gets R_{\Delta}-u_{i^\star}[R_{\Delta}]_{i^\star,:}/\sigma$
  \Comment{Orthogonalisation}
  \State Grow $\breve E_{LL}^{\dagger}$ by the border of $i^\star$ with pivot
  $\sigma$;\quad $\mathfrak L\gets \mathfrak L\cup\{i^\star\}$
  \Comment{rank-$1$ bordered update, \S\ref{sec:err}}
\EndFor
\State \Return nodes $L=K[\mathfrak L,:]$, inverse $\breve E_{LL}^{\dagger}$
\end{algorithmic}
\end{algorithm}

\subsubsection{Greedy selection by rank-revealed Q-DEIM}
\label{sec:sel_qdeim}
While the GAIM method greedily minimizes the residual for any given norm, the \emph{Q-DEIM-based Attention Interpolation Method} (Q-DAIM) only targets the residual in the norm $L^2$, similar to Q-DEIM \cite{drmacNewSelectionOperator2016}. To do so, instead of considering the residual in its original space, we consider the Cholesky-whitened~\cite{Kessy02102018} counterpart of the augmented operator, i.e. the orthogonal projector onto the range of $M_\eta$ obtained through a CholeskyQR.
Then we follow the Q-DEIM methodology and iteratively choose our nodes greedily to maximize the volume spanned in the  orthogonal basis, i.e. we choose the keys having the highest overlap with the basis vectors via the pivots of a greedily pivoted QR (or for more stability a rank-revealing pivoted QR with theoretical guarantees).  Furthermore, as we have the projection of the keys in an orthogonal basis, compared to the GAIM method, there is no need to compute $\Pi_L$ for the orthogonalisation procedure of the operator and all nodes can be selected sequentially without having to update a factorisation of the Gram matrix $E_{LL}$.
However, this comes at the cost of only being able to consider linear residuals, thus being restricted to v1 and v5.
Note that we use this method on the Gaussian formulation of the problem from \eqref{eq:gauss_attn}, allowing it to take into account the metric induced by both $\Vaug$ and by the mass ratios $\eta$, resulting in a better conditioned problem.

Let us now look at how we implement this in the context of attention. 
As mentioned before, we wish to find the orthogonal projector onto the range of $M_{\eta}$.
As the augmented operator $M_\eta$ has rank $r$ bounded by $\min(N,d+1)$, the range of $M_\eta$
is a subspace of $H$ of dimension $r$ and there exists a decomposition
\[
  M_\eta =\breve K^\T_{\phi} \Vaug_\eta = U R
\]
where $U:\RR^{r}\to H$ is an isometry whose columns span the range of $M_\eta$, and $R\in \RR^{r\times (d+1)}$.
Ideally, we would obtain this decomposition from a rank-revealing QR decomposition of $M_\eta$.
However, due to the infinite-dimensional nature of the problem, we will need to resort to 
alternative formulations to use $U$.

To obtain the factorisation, we consider a method known as whitening~\cite{Kessy02102018}, which groups CholeskyQR-like~\cite{highamAnalysisCholesky1990} methods where, given the Gram matrix \begin{equation}
G=M_{\eta}^{\T}M_{\eta}={\Vaug_\eta}^{\T} \breve E_{KK} \Vaug_\eta  = R^\T R
 \in \RR^{(d+1)\times(d+1)},
\label{eq:qdeim_gram}
\end{equation} we compute a factorization with permutations $P_R^\T G P_R=R^\T R$ and apply the (pseudo-)inverse of $R$ to $M_\eta$ to obtain $U$.
Indeed, $G$ is symmetric and positive semi-definite as the Gram of the column `vectors' of $M_\eta$, furthermore its rank is upper bounded by $\min(N,d+1)$. To obtain $R\in\RR^{r \times (d+1)}$,  we thus consider a \emph{rank-revealing} diagonally pivoted Cholesky~\cite{highamAnalysisCholesky1990} where we pivot on the largest remaining element on the diagonal of the residual, and stop when it falls below $\tau_{\mathrm{chol}}$ times the largest diagonal entry of $G$.
Then, the whitened coefficient matrix is given by 
\begin{equation}
C=\Vaug_\eta P_RR^\dagger \in\RR^{N\times  r},
\quad\text{ where }\quad  C^{\T}\breve E_{KK}C=I_{ r}.
\end{equation}
This then gives us the orthogonal operator 
\begin{equation} 
U=\breve K_\phi^{\T}C,\quad\text{ with }\quad U^{\T}U=I_{ r}.
\label{eq:qdeim_whiten}
\end{equation}
We note that the identities $C^{\T}\breve E_{KK}C=I_r$ and $U^{\T}U=I_r$ hold when the Cholesky factorization reveals the exact rank of $G$. 
When the factorization is stopped earlier, we instead have $P_R^\T G P_R=R^\T R+E$ with $E\succeq0$, such that $U^{\T}U=I_r+{R^{\dagger}}^{\T}ER^{\dagger}$. In this case $U$ is only approximately orthonormal. 
An exactly orthonormal basis can then be recovered from the pivoted columns only, $U=M_\eta P_R[:,1:r]R_{11}^{-1}$, where $R_{11}$ denotes the leading $r\times r$ block of $R$.
An error is then introduced due to the truncation procedure, it corresponds to $\|\Pi_{U}^{\perp} M_\eta \|^2=\|(I-UU^{\T})M_\eta\|^2$. 

In practice, since $G$ is of size $(d+1)\times(d+1)$, the rank revealed by the Cholesky factorization always satisfies $r\le d+1$.
When the budget is small and we have $2n+p\le d+1$, for a chosen oversampling $p\ge0$, we stop the Cholesky factorization at $r=2n+p$, even if the threshold has not yet been reached, in which case $U$ is only approximately orthonormal, as noted above.
Furthermore, while it might seem reasonable to take small thresholds, this can sometimes amplify noise as each vector in the revealed range provides an equal weight in the selection criterion.
We also note that any alternative square-root of $G$ may be used for the whitening, the most relevant alternative consisting in directly computing the counterpart to $R^\dagger$ obtained from the eigendecomposition of $G$.

Compared to the unwhitened selection of $L^2$ GAIM. Q-DAIM appears to be less sensitive to the keys with the largest kernel norms which tend to monopolize the pivots in GAIM. 
In the same manner, applying the method with the augmented value matrix (i.e. v5.) is also impactful, as it allows the map of the denominator to be taken into account in the metric.

For the selection procedure, we consider the projection of the key vectors onto the range of $M_\eta$,
\begin{equation}
F= \breve K_{\phi}U=\breve E_{KK} C\in\RR^{N\times  r},
\qquad
F[i,:]=\big\langle\breve\phi[K_i], U\big\rangle ,
\label{eq:qdeim_F}
\end{equation}
and a column-pivoted QR (Businger-Golub~\cite{busingerLinearLeast1965})
is applied to $F^{\T}$. We set $P=(p_i)_{i\in[n]}\subset[N]$ to be the indices of the $n$ selected pivots such that $K[P,:]=L$.
However, when the node budget $n\ge r$, the column-pivoted QR described above only gives us at most $r$ informative pivots, the remaining selections being chosen on numerical noise. 
Hence, for budgets $n>r$, we consider optional restart procedures, as described in Appendix~\ref{app:qdaim_restart}.

An alternative to the fixed number of selected nodes for interpolation
is to instead consider keeping pivots as long as the addition of a new pivot improves the reconstruction above a certain threshold, i.e. keep pivots only while
$\nu_\ell^2/\nu_1^2\ge\tau$, where $\nu_\ell$ denotes the remaining norm of the $\ell$-th pivot of the QR-CP of $F^\T$, or alternatively while the relative operator energy missed by the nodes remains above a threshold.

Finally, our Q-DAIM interpolant\footnote{
Note that, writing \(\mathcal I_{L}^* = U F_{P,:}^{\dagger}\breve L_\phi=U(\breve L_{\phi}U)^{\dagger}\breve L_{\phi}\) for the usual DEIM operator on \(H\), the operator we use for
the interpolation is thus the adjoint of the usual DEIM operator, i.e. $\mathcal I_L= (\mathcal I_L^* )^{*}$ .
Indeed, this arises because in the classical DEIM method the convention is for multiplication to occur on the right, i.e. given $\mathrm{M}=\mathrm V\mathrm S\mathrm U^\T\in\RR^{d\times N}$, the transposed equivalent to our operator $M_\eta$, $\mathrm x\in\RR^{N}$ and a pivoting matrix $\mathrm P$, Q-DEIM approximates as $\mathrm M \mathrm x\approx \mathrm M \mathrm U{(\mathrm P^\T \mathrm U)}^{\dagger}\mathrm P^\T \mathrm x$. 
Whereas we apply multiplication on the left, $\phi[q]^\T \breve K_\phi^\T \Vaug_\eta \approx \phi[q]^\T (\breve K_{\phi}[P,:])^\T (U^\T \breve K_\phi[P,:]^\T )^{\dagger}U^\T \breve K_\phi^\T \Vaug_\eta$,
but the method is the same in both cases, we project the input, $\phi[q]$ in our case, onto the reduced basis determined by the pivots.} is then given by the oblique projection
\begin{equation}
  \mathcal I_LM_\eta
  =\breve L_{\phi}^{\T}F_{P,:}^{\dagger\T}F^{\T}\Vaug_\eta
  =\breve L_{\phi}^{\T}F_{P,:}^{\dagger\T}{{R^\dagger}^\T P_R^\T}G=\breve L_{\phi}^{\T}F_{P,:}^{\dagger\T}R P_R^\T,\qquad
  \mathcal I_L=\breve L_{\phi}^\T( U^\T\breve L_\phi^\T)^{\dagger}U^{\T},
\end{equation}
where we use $\dagger$ for the pseudo-inverse.
We note that $\mathcal I_L$ has rank at most $r$. Hence, when $n>r$, it does not project onto the whole basis space $\mathcal B$ but only onto a subspace of $\mathcal B$ of dimension at most $r$, the pseudo-inverse selecting the minimum-norm coefficients. Furthermore, as its test space $\Pi_U\mathcal B$ differs from $\mathcal B$, this projection is in general not exact at the interpolation nodes, but instead at the projections of their features onto $\Span(U)$, i.e. $(\Pi_U\breve\phi[L_i])^\T\mathcal I_L M_\eta=(\Pi_U\breve\phi[L_i])^\T M_\eta$ for all $i\in[n]$.

We note that an improvement to our implementation would be the use of randomisation-based QR methods suited to the case $N\gg d\ge r$. 
Such methods include randomised methods preserving rank-revealing properties such as \cite{fakihEfficientQRbasedColumn2026} and recent methods specifically suitable for the Q-DEIM method such as \cite{cortinovisAdaptiveRandomizedPivoting2026}.

\begin{algorithm}[t]
\caption{Batch Q-DAIM selection}
\label{alg:qdaim}
\begin{algorithmic}[1]
\Require keys $K\in\RR^{N\times d}$ (centered), values $V\in\RR^{N\times d}$, budget $n$,
oversampling $p$, Cholesky tolerance $\tau_{\mathrm{chol}}$,
linear residual variant (v1. or v5. of \S\ref{sec:selection})
\State $\Vaug\gets[ V \mid \mathbf 1_N ]$;\quad
$\eta\gets$ gauged masses of \eqref{eq:kernel_norm_weight}
\State $V_{\mathrm{sel}}\gets\Diag(\eta)V$ (v1.), or  
$\Diag(\eta)\Vaug$ (v5.)
\State $G\gets{V_{\mathrm{sel}}}^{\T} \breve E_{KK} V_{\mathrm{sel}}$
\Comment{operator Gram \eqref{eq:qdeim_gram}}
\State $(P_{R},R)\gets$ diagonally pivoted Cholesky of $G$
\Comment{rank-revealing factorisation}
\State $C\gets V_{\mathrm{sel}} P_{R}{R^{\dagger}}$
\Comment{whitened coefficients \eqref{eq:qdeim_whiten}}
\State $F\gets\breve E_{KK} C$
\Comment{evaluation matrix \eqref{eq:qdeim_F}}
\State $\mathfrak L\gets$ first $n$ pivots of the QR-CP of $F^{\T}$
\Comment{or volume-thresholded stop, see the remark above}
\State \Return $L=K[\mathfrak L,:]$
\Comment{weights: any projection of \S\ref{sec:proj}}
\end{algorithmic}
\end{algorithm}

The full Q-DAIM batch selector is summarised in Algorithm~\ref{alg:qdaim}. 
Similarly to the GAIM method, the Q-DAIM selector also has a complexity in $N^2$ due to the construction of the Gram matrix $\breve E_{KK}$.

\subsubsection{Multi-headed $L^2$ weighted residuals}
\label{sec:multihead}
One issue that arises in the context of multi-headed attention is that the attention residual for a key head does not
take into account the attention residual summed over the query heads.
Given $\Nh$ query heads, we set $O_h=\Att(Q_{h},K_{h},V_{h})$ to be the output of query head $h$ and the associated output projection matrix $\WO_h\in\RR^{d\times d}$ \footnote{We use $d$ for both dimensions for simplicity, see \cite{serretAttentionMechanismsThrough2026} for more details on multi-headed attention}. 
The output of a single layer of attention is given by $O=\sum_{h\in[\Nh]}O_h\WO_h$. Under grouped-query
attention (GQA)~\cite{ainslieGQATrainingGeneralized2023} the $\Nh$ query
heads share $\Ng<\Nh$ key-value heads through a surjection
$h\mapsto \mathfrak g_h\in[ \Ng]$ such that
\[
  O_{h}=\Att(Q_{h},K_{\mathfrak{g}_h},V_{\mathfrak{g}_h}),\quad \sum_{h\in[\Nh]}O_h\WO_h=\sum_{k\in[\Ng]}\sum_{h:\mathfrak g_{h}=k} \Att(Q_{h},K_{\mathfrak{g}_h},V_{\mathfrak{g}_h})W^{O}_h
\] 

In practice, as we shall see in \S~\ref{sec:streaming}, we interpolate on different nodes for each \emph{KV head}, hence the output error of $k$-th KV head is given
by the sum of query heads of its group, i.e. heads such that $\mathfrak g_{h}=k$.

For an $L^2$ norm residual, we can take this into account by weighting the output dimensions of $V$ through the Cholesky factor 
\begin{equation}
\Omega_k = \sum_{h : \mathfrak g_h=k}\WO_h {\WO_h}^{\T}
 \in \RR^{d\times d},
\qquad\text{for which}\qquad
\sum_{h : \mathfrak g_h=k}\big\lVert e \WO_h\big\rVert_2^2
=e \Omega_k e^{\T}
=\big\lVert e \Omega_k^{1/2}\big\rVert_2^2
\label{eq:output_gram}
\end{equation}
given an error term $e\in\RR^{d}$ shared by the group. We note that, as the query heads of a group have different errors $e_h$ and the output error $\sum_{h}e_h\WO_h$ contains cross terms between heads, this weighting is a proxy for the error of the layer output rather than its exact value.
We thus reweight the value columns of the residual by $\Omega_k^{1/2}$,  scoring
$\lVert[ \breve S_L\Diag(\eta)\Vaug ]_{i,1:d} \Omega_k^{1/2}\rVert_2$ in
place of \eqref{eq:greedy_argmax}. 
Two things to note, first is that this method can only be applied to the $L^2$ norm, and that as it does not change the range, it is only applicable to GAIM.
Second, each $\Omega_k$ only needs to be computed once at initialisation as it only depends on the head and the model being used.

\subsubsection{The finite-dimensional surrogate}
\label{sec:surrogate}

The main bottleneck of both the GAIM and Q-DAIM methods resides in the quadratic cost of the computation of the Gram matrix $\breve E_{KK}$.
As we multiply it by $V_\eta$ afterwards, if the attention was linear, a simple matrix product reordering would allow us to go from $O(N^2d) \to O(N d^2)$.
To enable this re-ordering, we resort to approximating $\breve \phi$ using finite feature maps $\psi:\RR^{d}\to \RR^{D}$, with $D\in\NN^*$. 
Indeed, as we know that $\eta$ carries some important information for the selection of key nodes, we choose to preserve it.
Set  $K_\psi=(\psi [K_j])_{j\in[N]}$ and $L_\psi=(\psi [L_k])_{k\in[n]}$ the feature matrices and 
$D_{K_\psi}=\Diag(K_\psi K_\psi^{\T})$, and $\breve K_\psi:=D_{K_\psi}^{-1/2}K_\psi$ then, our method consists in replacing
\begin{equation}
\breve K_\phi^{\T} \eta_K \longrightarrow 
K_\psi^{\T} D_{K_\psi}^{-1/2} \eta_K,
\qquad
\breve L_\phi^{\T} \eta_L \longrightarrow 
L_\psi^{\T} D_{L_\psi}^{-1/2} \eta_L,
\label{eq:surrogate_sub}
\end{equation} and equivalently for their transposes.
In practice we consider three main methods to obtain such a $\psi$:
\begin{itemize}
  \item The linear Taylor map $\psi(x)=(1, x/d^{1/4})\in\RR^{d+1}$, corresponding to the first order Taylor approximation of $\kappa_{\exp}$,
  \item the Performer FAVOR$+$ randomised feature map~\cite{choromanskiRethinkingAttentionPerformers2022} approximating $\kappa_{\exp}$,
  \item and the Random Fourier Features~\cite{rahimiRandomFeaturesLargescale2007} which approximates the Gaussian kernel directly.
\end{itemize}

For the GAIM method, we consider the linear-Taylor map and the Performer method, whereas for the Q-DAIM method, we use the random Fourier features for
the Gaussian matrix which approximate the Gaussian gram directly.
For clarity, we denote by $\breve\psi$, these features, and the approximation thus reads $\breve E_{KK}\approx \breve K_{\breve\psi} \breve K_{\breve\psi}^{\T}$.

\paragraph{GAIM linear-complexity surrogates.}
For each residual operator $M_{*}$ described in \S\ref{sec:err}, we write $M_{*,\psi}$
for their \emph{finite-dimensional surrogate operator}, obtained
through substitution as in \eqref{eq:surrogate_sub}, thus we obtain
\begin{equation}
    M_{\psi}:=\breve K_\psi^{\T}\Vaug_\eta=\big[ M^{g}_{\psi} \mid M^{h}_{\psi} \big]
\end{equation}
We introduce a first variant GAIM-LIN-O, where we compute the residual operators using the $L_\phi$-orthogonal projection of the map, and substitute $K_\phi$ in the same way as above,
\begin{equation}
    \hat M_{\Delta,\psi}:=
    M_\psi-\breve L_\psi^\T \Waug_\eta =:\big[ \hat M^{g}_{\Delta,\psi} \mid \hat M^{h}_{\Delta,\psi} \big],
\end{equation} where $\Waug_\eta:=\breve E_{LL}^{\dagger}\breve E_{LK}\Vaug_\eta$ is the gauged Galerkin solution $X$ of \eqref{eq:stable_galerkin}.
However, this $\phi$-orthogonal surrogate requires the computation of the cross-Gram and Gram matrices, $E_{LK}$ and $E_{LL}$, as well as the inverse of the latter. To avoid this, we define a second variant whose residual operators are given as
\begin{align}
M_{\Delta,\psi} := M_{\psi}-\Pi_{L_\psi}M_{\psi} =:\big[  M^{g}_{\Delta,\psi} \mid  M^{h}_{\Delta,\psi} \big],
\end{align} where $\Pi_{L_\psi}$ is the orthogonal projector onto $\Span(L_\psi)\subset\RR^{D}$.
To approximate this, we consider a Matching Pursuit approach similar to \cite{pati1993orthogonal,davis1997greedy,tropp2006algorithms}, which we call GAIM-LIN, where after each greedy selection indexed by $i^*\in[N]$ for the chosen residual, we set instead \[ M_{\Delta,\psi}^{(t)}=M^{(t-1)}_{\Delta,\psi}-\left(\breve K_{\psi}[i^*,:]^\T\breve K_{\psi}[i^*,:]\right) M^{(t-1)}_{\Delta,\psi},\] orthogonalizing the residual only with respect to the last selection.

\begin{algorithm}[t]
\caption{Surrogate greedy GAIM selection : GAIM-LIN}
\label{alg:selection}
\begin{algorithmic}[1]
\Require keys $K\in\RR^{N\times d}$, values $V\in\RR^{N\times d}$,
budget $n$, feature map $\psi$, residual variant $R_{\mathrm{sel}}$, norm $p$
\State $\Vaug\gets[ V \mid \mathbf 1_N ]$;\quad
$\eta\gets$ gauged masses of \eqref{eq:kernel_norm_weight};\quad
$\mathfrak L\gets\varnothing$
\State $\breve K_\psi\gets D_{K_\psi}^{-1/2}K_\psi$;\quad
$M_\psi\gets \breve K_\psi^{\T}\Diag(\eta)\Vaug\in\RR^{D\times (d+1)}$
\Comment{normalised features and moment}
\State $R^{0}\gets\breve K_\psi M_\psi$;\quad $R_{\Delta}\gets R^{0}$
\Comment{evaluations and residual rows at $L=\varnothing$}
\For{$t=1,\dots,n$}
  \State Update $R_{\mathrm{sel}}$ from $(R_{\Delta},R^{0})$;\quad
  $i^\star\gets\arg\max_{i\notin \mathfrak L}\lVert[R_{\mathrm{sel}}]_{i,:}\rVert_p$
  \Comment{chosen variant, \S\ref{sec:selection}}
  \State $R_{\Delta}\gets R_\Delta - \breve K_\psi \breve K_{\psi,i^\star}^{\T}
  [R_{\Delta}]_{i^*,:}$;\quad $\mathfrak L\gets \mathfrak L\cup\{i^\star\}$
  \Comment{deflation of the moment, updated residual rows}
\EndFor
\State \Return $L=K[\mathfrak L,:]$
\Comment{weights: any projection of \S\ref{sec:proj}}
\end{algorithmic}
\end{algorithm}

\paragraph{Q-DAIM Random Fourier Features linear complexity surrogate.}
We tackle the quadratic cost of Q-DAIM by building a surrogate Gram matrix $G_{\breve \psi}$ which allows changing the order of multiplication and computing $M_{\breve\psi}=\breve K_{\breve \psi} ^\T \Vaug_{\eta}\in\RR^{D\times (d+1)}\approx M_\eta$ first. We use the notation $\breve \psi$ for the Random Fourier Feature map to highlight the fact that the feature map approximates the Gaussian features directly and thus does not require renormalization through $D_{K_\psi}$. This then gives us 
\[G_{\breve \psi}:=M_{\breve\psi}^\T M_{\breve\psi}={\Vaug_{\eta}}^\T \breve K_{\breve\psi} \breve K_{\breve\psi}^\T \Vaug_{\eta}\in\RR^{(d+1)\times (d+1)}.\]
 The surrogate Gram $G_{\breve\psi}$ is still positive semi-definite regardless of $\breve\psi$, and the remaining Q-DAIM procedure, whitening followed by selection of nodes based on QRCP pivots, remains unchanged.
In particular, the rank-revealing pivoted Cholesky factorization used for whitening~\eqref{eq:qdeim_whiten} is now applied to $G_{\breve\psi}$ instead of $G$. The evaluation matrix \eqref{eq:qdeim_F} is formed similarly through $M_{\breve \psi}$ as
\[
F_{\breve \psi} = \breve K_{\breve\psi} \big(\breve K_{\breve\psi}^{\T}C_{\breve \psi}\big),
\]
with $C_{\breve \psi}=\Vaug_\eta P_{R_{\breve \psi}}R_{\breve \psi}^{\dagger}$ and $P_{R_{\breve \psi}}^\T G_{\breve \psi}P_{R_{\breve \psi}}=R_{\breve \psi}^\T R_{\breve \psi}$.
Hence, the overall complexity is reduced from $O(Nd(N+d))$ to $O(D(d+1)(N+d+1))$, the second term coming from the surrogate Gram $G_{\breve\psi}$.

\subsection{Alternative projections}
\label{sec:proj}
The methods we have developed above all possess a natural associated projection, i.e., a way to construct \(\Waug\), used to actually select the nodes. 
As we shall see, each of these methods can be classified as a specific case of a Petrov-Galerkin projection.
Indeed, Petrov-Galerkin methods approximate a continuous operator $M$ in a space $H$ by an operator $M_L$ defined in a finite-dimensional subspace, $\mathcal B\subset H$, of functions called the trial space, which corresponds to our basis space $\mathcal B$.
The approximated operator is then obtained by choosing a set of functions $\mathcal T$, called the test space, and its associated norm in which the residual is minimized. 

In the case of direct interpolation/GAIM, the approximated linear map is obtained by setting the
residual to zero at the interpolation nodes. This corresponds to a
Bubnov-Galerkin method, as we test the function on the basis space \(\mathcal T = \mathcal B\) and consider the orthogonal projection
of our linear maps onto the basis space. For Q-DAIM, the test space corresponds to the projection of $\mathcal B$ on the revealed range of the operator, i.e., $\mathcal T= \Pi_{U}\mathcal B$, which coincides with $\mathcal B$ only if $\mathcal B\subseteq \Span( U)$, which requires $n\le r$.

The above methods can equivalently be seen as orthogonality constraints. On the other hand, the least-square approximation $\mathcal T=\Span(K_\phi)$ consists in a case where the test space contains directions orthogonal to the trial space, i.e. $\mathcal T \cap \mathcal B^\perp \neq \{0\}$, and where the dimensions of $\mathcal T$ and $\mathcal B$ may not match.
In our case, it is overdetermined. 
To handle this as an orthogonality constraint, we consider the least-square Petrov-Galerkin consisting in solving the orthogonality constraints arising from the normal equations instead.

Finally, we also consider the projection obtained by a direct minimization of the overdetermined system's residual. We note that, while this projection and the least-square Petrov-Galerkin coincide in exact arithmetic, their numerical solutions need not be the same, the normal equations squaring the condition number of the system.
In the following we describe in more detail these different types of projection. We now formalise how these methods  can be framed as Galerkin approximations of $M_\eta=\breve K_\phi^{\T}\Vaug_\eta\in H^{d+1}$, the gauged augmented operator, on the trial space
$\mathcal B=\Span\{\breve\phi[\ell_1],\dots,\breve\phi[\ell_n]\}\subset H$.

Let $\mathcal W$ be a symmetric positive semi-definite operator acting on $H$.
The Petrov-Galerkin projection finds $X\in\RR^{n\times(d+1)}$ such that the
residual $\breve L_\phi^{\T}X-M_\eta$ is $\mathcal W$-orthogonal to
$\mathcal B$, i.e.
\begin{equation}
  \big\langle\breve\phi[\ell_i]^\T,
  (\breve L_\phi^{\T}X-M_\eta)[:,j]\big\rangle_{\mathcal W}=0
  \quad\text{for all }i\in[n] \text{ and } j\in[d+1],
  \label{eq:galerkin_var}
\end{equation}
where we set $\langle f,g\rangle_{\mathcal W}:=\langle f,\mathcal Wg\rangle_H$.  The solution is obtained by solving the $n\times n$ linear system,
\begin{equation}
  A_{\mathcal W} X=B_{\mathcal W},
  \qquad
  A_{\mathcal W}:=\breve L_\phi \mathcal W \breve L_\phi^{\T},
  \qquad
  B_{\mathcal W}:=\breve L_\phi \mathcal W M_\eta.
  \label{eq:galerkin_sys}
\end{equation}
The projected weights are then obtained by
\begin{equation}
  \Waug=\Diag(\eta_L)^{-1} X.
  \label{eq:galerkin_ungauge}
\end{equation}

The projection methods correspond to the following
choices of $\mathcal W$, where we set $\Sigma_K=\breve K_\phi^{\T}\breve K_\phi$ and $\Pi_{U}=UU^{\T}$ the projector
onto $\mathcal U$.
We provide the equivalent linear systems for each projection method in Table~\ref{table:projections}, as well as the formulation of the WildCat~\cite{schroderWildCatNearLinearAttention2026} temperature-scaled projection in the context of Petrov-Galerkin, more details can be found in Appendix~\ref{app:wilcat_proj}.
We note that, for the Q-DAIM and WildCat rows of the table, the $\mathcal W$ column gives the associated oblique projectors $\mathcal I_L$ and $P_\tau$, which are not symmetric. These projections correspond to Petrov-Galerkin methods with distinct trial and test spaces rather than to a symmetric positive semi-definite $\mathcal W$. Furthermore, the solution of \eqref{eq:galerkin_sys} is unique only when $A_{\mathcal W}$ is nonsingular.
\begin{table}
\begin{center}
\begin{tabular}{llll}
\hline
Projection method & $\mathcal W$ & $A_{\mathcal W}$ & $B_{\mathcal W}$ \\
\hline
GAIM    & $\Pi_{L_{\phi}}$
  & $\breve E_{LL}$
  & $\breve E_{LK}\Vaug_\eta$ \\
PG least squares    & $\Sigma_K$
  & $\breve E_{LK}\breve E_{KL}$
  & $\breve E_{LK}\breve E_{KK}\Vaug_\eta$ \\
Q-DAIM revealed-range & $\mathcal I_{L}/\Pi_{U}$
  & $F_{P,:}F_{P,:}^{\T}$
  & $F_{P,:}F^{\T}\Vaug_{\eta}$\\
Least squares & - & $\breve E_{KL}$&$\breve E_{KK}\Vaug_{\eta}$\\
WildCat $\tau$ scaling & $P_{\tau}=L_\phi^\T E_{\tau^{-2}L,L}^{-1}L^{\tau^{-2}}_\phi$ & $\breve E_{\tau^{-1} L,\tau^{-1}L}$&$\breve E_{\tau^{-1}L,\tau^{-1}K}\Vaug_{\eta,\tau}$\\
\hline
\end{tabular}
\end{center}
\caption{Matrices and projectors associated with the different projection methods.}
\label{table:projections}
\end{table}

\subsection{Positioning with respect to kernel-based selection methods}
We now discuss how our two selectors relate to WildCat~\cite{schroderWildCatNearLinearAttention2026} and, more generally, to greedy kernel interpolation methods~\cite{wenzelNovelClassStabilized2021,wenzelAnalysisTargetDatadependent2023}.
  WildCat selects its nodes through a randomly pivoted Cholesky~\cite{epperlyEmbraceRejectionKernel2025} of the key Gram matrix $E_{KK}$ alone, i.e. the next node is sampled with probability $i\sim\|\Pi^\perp_{L_\phi}\phi[K_i]\|_H^2$.
 Hence, it aims at reconstructing the span of $K_\phi$ and does not take into account the effect of $V$ on the interpolated output.
  In the context of kernel interpolation methods~\cite{demarchiNearoptimalDataindependentPoint2005,santinConvergenceRateDataindependent2017}, 
  this corresponds to a randomized version of a $P$-greedy method, setting the power function $P_L(x)=\lVert\Pi_{L_\phi}^{\perp}\phi[x]\rVert_H$. Similarly, Q-DAIM can be seen as a $P$-greedy method for the revealed-range projection norm $P_{L}(x)=\|\Pi_{\Pi_U L_\phi}^{\perp}\Pi_U \breve \phi[x]\|_H$, 
  which to our knowledge has not been studied in the context of kernel interpolation.
 GAIM, on the other hand, corresponds to what is referred to as an $f$-greedy method in the case of linear variants based on $g$ and $h$.
 Note, however, that the convergence theory of the greedy kernel methods concerns the interpolation of functions of the native space, here $g$ and $h$, whereas the error on attention itself is only related to them through \eqref{eq:residual_exact}. As our method consists in considering the ratio $g/h$ of two such interpolants, we have the following rudimentary bound
\begin{equation}\label{eq:att_error_bound}
  \left\|\Att(q,K,V)-\widetilde\Att_L(q,K,V)\right\|_2
  \le\frac{\|M^{g}_{\Delta}\|_{\mathrm{HS}}+\|\Att(q,K,V)\|_2\|M^{h}_{\Delta}\|_{H}}
          {\left|\breve\phi[q]^{\T}M^{h}_{L}\right|},
\end{equation}
where we set $\|M^{g}_{\Delta}\|_{\mathrm{HS}}=\sqrt{\Tr({M^{g}_{\Delta}}^\T M^{g}_{\Delta})}$ to be the Hilbert-Schmidt norm of the numerator residual operator. We leave a more detailed analysis of the  error bounds for future work.

\section{Interpolation for streaming attention}
\label{sec:streaming}

The interpolation methods we have defined above consider a fixed number $N$ of tokens and associated $K$ and $V\in\RR^{N\times d}$ matrices 
to be compressed.
In modern attention-based language models, each layer of the model contains a
multi-headed attention mechanism. For each of these heads of attention, the key
and value vectors for the text that has been seen until a given moment in time
are stored in memory to allow faster computation once a new sentence is
provided to, or generated by, the model. This is known as KV caching.
Once the input context is read, the key and value vectors of a given token impact the construction of the query, key and values of subsequent tokens in the same head. Additionally, as the layers are sequential, a given token's key and value vector also affects all the layers that follow.

We therefore choose to read the context in segments of \(\Nc\) tokens and alternate two steps.
First, the model is computed on the new segment using the compressed cache. Then, we compress the keys and
values produced by the model using one of our methods. Beyond the first layer, the keys and values of a segment thus depend on the compressed attention of the previous layers  and 
the errors due to the interpolation of the previous segments propagate both  to the generation and compression of the vectors of the layers that follow. 
Only the first segment's key and value matrices are generated using exact attention. 
For each layer and head, the cache thus holds, after $N_{\mathrm{tot}}$ tokens have been read and compressed:
\begin{itemize}
  \item $\tsink\in\NN$ fixed sink tokens at the beginning of the context, kept uncompressed;
  \item $\trec\in\NN$ recent tokens, corresponding to the last $\trec$ tokens read, kept uncompressed;
  \item  the key nodes $L\in\RR^{n\times d}$ and associated weights $\Waug\in\RR^{n\times (d+1)}$ corresponding to the compressed $N=N_{\mathrm{tot}}-\tsink-\trec$ tokens in between.
\end{itemize}  
After a new sentence's key and values vectors are computed, its keys and values are appended to the cache. The last $\trec$ of these new vectors form the new recent window and the other remaining vectors correspond to the segment $(K_{\mathrm{seg}},V_{\mathrm{seg}})$ which, once it reaches a given size $\Nc$, is then compressed into $\nc$ nodes, for every layer and head at once.
Once the context has been read, the generated tokens are appended raw to $K_{\mathrm{raw}},V_{\mathrm{raw}}$ and are not compressed.

Although the compression of the first segment is done using the exact same process as before, once a set $L\in\RR^{n\times d}$ and $\Waug\in\RR^{n\times (d+1)}$ have been instantiated,
we slightly modify the interpolation process to take into account the compressed data.
In order to preserve the coherence of the interpolation between the different segments, i.e. the impact of the previous segments on the compression of a new one, we consider two additional steps to the interpolation's selection and projection, an orthogonalisation step based on the compressed cache which we call the deflation step, and an associated weight update step on previous nodes after projection which we call the refit step.
Additionally, the centering of the keys needs to be refined, since the mean of the keys changes as new segments are added.

\subsection{Bridging the gap between global and local interpolation}
Let $K_{\mathrm{seg}},V_{\mathrm{seg}}\in\RR^{\Nc \times d}$ be the key and value matrices of a given segment that we will compress into $\nc \in\NN^*$ nodes $L_{\mathrm{seg}}$ and associated weights $\Waug_{\mathrm{seg}}$ and let $L\in\RR^{n\times d}$, with $n\in\NN$, be the previously compressed nodes  and weights $\Waug\in\RR^{n\times (d+1)}$.
Then, we can interpolate the segment using the selection and projection techniques defined in the fixed context case, with some optional modifications, based on the  previously selected nodes, which allow to reduce the gap between the global and segment interpolation.
The first consideration to take into account is the centering of the keys. Indeed, the mean key vector $\mu$ is not fixed in the context of the streaming attention, and we thus define two alternatives for the  interpolation centering:
\begin{itemize}
  \item Global centering:  Set $\mu_{g}=\frac{1}{N}\sum_{i=1}^{N}K_{\tsink + i}$ the  mean of the current global, sinkless, context.
  \item Segment centering: Set $\mu_{\mathrm{seg}}=\frac{1}{\Nc }\sum_{i=1}^{\Nc }K_{\mathrm{seg},i}$ the mean of the current segment.
\end{itemize}
Once this is fixed, along with the choice of selection and projection method, we modify the selection method to remove the redundancy with the historical nodes through a deflation method.
In the same way as the basis deflation that is used after the selection of a node, this consists in considering only the part of the residual orthogonal to a given span in $H$.
We consider either the revealed span, through $\Pi_U$, or the selected nodes, via $\Pi_L$, subtracting the part of the residual explained by the historical nodes/linear map.
Note that, in the rest of this section, every object (masses, values and weights) is handled in the $\eta$ gauge of \eqref{eq:kernel_norm_weight}; we drop the $\eta$ subscripts for simplicity.
This results in two types of deflation which can be obtained as
\begin{itemize}
  \item Basis deflation  $\Pi_{L_{\phi}}^\perp$ : For any selection method, we consider  the projection onto the orthogonal complement of the basis space $\mathcal B = \Span(L_\phi)$, obtained as
  \[
  \Pi_{L_{\phi}}^\perp\breve K_{\mathrm{seg},\phi}^\T=\breve K_{\mathrm{seg},\phi}^\T - \breve L_\phi^\T \breve E_{LL}^{\dagger}\breve E_{LK_{\mathrm{seg}}}.
  \]
   The  Gram matrix $\breve E_{KK}$ is then replaced by its orthogonal projection, the deflated Gram matrix given by
  \[ 
    \breve S_{L}=\breve K_{\mathrm{seg},\phi}\Pi^\perp_{L_{\phi}}{\breve K_{\mathrm{seg},\phi}}^\T=\breve E_{K_{\mathrm{seg}}K_{\mathrm{seg}}} - \breve E_{K_{\mathrm{seg}}L}\breve E_{LL}^{\dagger}\breve E_{LK_{\mathrm{seg}}}.
  \]
  \item Range deflation $\Pi_{U}^\perp$: For any selection method, we consider its projection onto the orthogonal complement of $\Span(U)$, obtained as
  \[ \Pi_{U}^\perp\breve K_{\mathrm{seg},\phi}^\T=\breve K_{\mathrm{seg},\phi}^\T - \breve L_\phi^\T \Waug G_{L}^{\dagger}{\Waug}^\T \breve E_{LK_{\mathrm{seg}}},\] where $G_{L}={\Waug}^\T \breve E_{LL}\Waug\in\RR^{(d+1)\times (d+1)}$.
  This gives us the  deflated Gram matrix,
  \[
    \breve S_{U}=\breve K_{\mathrm{seg},\phi}\Pi^\perp_{U}\breve K_{\mathrm{seg},\phi}^\T=\breve E_{K_{\mathrm{seg}}K_{\mathrm{seg}}} - \breve E_{K_{\mathrm{seg}}L}\Waug G_{L}^{\dagger} {\Waug}^\T \breve E_{LK_{\mathrm{seg}}},
  \] 
\end{itemize}

In practice, the range deflation may also be computed using the Cholesky/eigendecomposition of $G_L$ in the same way as the Q-DAIM deflation, i.e. through $F_{U,ch}=\breve K_{\mathrm{seg},\phi}\breve L_\phi^\T \Waug P_{R_L} R_L^{\dagger}$ with $P_{R_{L}}^\T G_LP_{R_{L}}=R_{L}^\T R_{L}$.
Furthermore, instead of being recomputed from $L$ and $\Waug$, the matrix $G_L$ can be updated for each new segment as,
\[
G_{[L\cup L_{\mathrm{seg}}]}=G_L+G_{L_{\mathrm{seg}}}
+{\Waug}^\T\breve E_{LL_{\mathrm{seg}}}\Waug_{\mathrm{seg}}
+{\Waug_{\mathrm{seg}}}^\T\breve E_{L_{\mathrm{seg}}L}\Waug,
\]
which requires computing only the matrix $\breve E_{LL_{\mathrm{seg}}}$ coupling the  previously and newly selected nodes, at a cost of $O(n\nc (d+1))$, the products with the weights adding $O(\nc (d+1)(\nc +d+1))$.
We note that this update assumes that the historical weights $\Waug$ are unchanged. When a refit replaces them by $\Waug+\Waug_{\mathrm{refit}}$, the first term becomes $(\Waug+\Waug_{\mathrm{refit}})^\T\breve E_{LL}(\Waug+\Waug_{\mathrm{refit}})$ and the cross terms must be computed with the refitted weights.

The modified operators for each deflation method are described in table~\ref{tab:deflation}.
\begin{table}
\caption{Modified matrices for each deflation type.}
\label{tab:deflation}
\begin{center}
\begin{tabular}{lcccc}
\hline
Deflation & $M$ & $G$ & $\breve S$ & $R$ \\
\hline
None & $\breve K_{\mathrm{seg},\phi}^{\T}\Vaug_{\mathrm{seg}}$ & ${\Vaug_{\mathrm{seg}}}^{\T}\breve E_{K_{\mathrm{seg}}K_{\mathrm{seg}}}\Vaug_{\mathrm{seg}}$ & $\breve E_{K_{\mathrm{seg}}K_{\mathrm{seg}}}$ & $\breve E_{K_{\mathrm{seg}}K_{\mathrm{seg}}}\Vaug_{\mathrm{seg}}$ \\
Basis & $\Pi^{\perp}_{L_{\phi}}\breve K_{\mathrm{seg},\phi}^{\T}\Vaug_{\mathrm{seg}}$ & ${\Vaug_{\mathrm{seg}}}^{\T}\breve S_{L}\Vaug_{\mathrm{seg}}$ & $\breve S_{L}$ & $\breve S_{L}\Vaug_{\mathrm{seg}}$ \\
Range & $\Pi^{\perp}_{U}\breve K_{\mathrm{seg},\phi}^{\T}\Vaug_{\mathrm{seg}}$ & ${\Vaug_{\mathrm{seg}}}^{\T}\breve S_{U}\Vaug_{\mathrm{seg}}$ & $\breve S_{U}$ & $\breve S_{U}\Vaug_{\mathrm{seg}}$ \\
\hline
\end{tabular}
\end{center}
\end{table}

We also note that for the finite-dimensional surrogate methods, we replace $\Pi_{L_\phi}^\perp \breve K_{\mathrm{seg},\phi}^\T$ by  \[\Pi_{L_{\phi},\psi}^\perp\breve K_{\mathrm{seg},\psi}^\T=\breve K_{\mathrm{seg},\psi}^\T - \breve L_\psi^\T \breve E_{LL}^{\dagger}\breve E_{LK_{\mathrm{seg}}},\] and $\Pi_U^\perp \breve K_{\mathrm{seg},\phi}^\T$ by
\[\Pi_{U,\psi}^\perp\breve K_{\mathrm{seg},\psi}^\T:=\breve K_{\mathrm{seg},\psi}^\T - \breve L_\psi^\T \Waug G_{L}^{\dagger}{\Waug}^\T \breve E_{LK_{\mathrm{seg}}}.\]

After the selection, the projection is done on the undeflated operator, as in fixed context interpolation, and the resulting residual operator $M_{\Delta}=K_{\mathrm{seg},\phi}^\T\Vaug_{\mathrm{seg}}-L_{\mathrm{seg},\phi}^\T\Waug_{\mathrm{seg}}$ is used to update the weights of the historical nodes.
To do so, although any projection method may be used for the refit, the machinery used for the deflation allows a refit counterpart consisting in updating $\Waug$ with the weights associated to the orthogonal projection used for the deflation, making it less costly.
Thus, denoting $\Waug_\mathrm{refit}$ the update to the weights of the historical nodes, we obtain the updates 
\[
L\gets \begin{bmatrix}L\\ L_{\mathrm{seg}}\end{bmatrix},
\qquad
\Waug \gets \begin{bmatrix}\Waug\\ 0\end{bmatrix}
+ \begin{bmatrix}\Waug_\mathrm{refit}\\ \Waug_{\mathrm{seg}}\end{bmatrix}.
\]
Thanks to the linearity of the attention's numerator and denominator, we can directly append the new segments nodes and weights, $L_{\mathrm{seg}}$ and $\Waug_{\mathrm{seg}}$, to $L$ and $\Waug$.
Indeed, once all the segments have been compressed into $L\in\RR^{n\times d}$ and $\Waug\in\RR^{n\times (d+1)}$,  we can approximate, denoting by $K_{\mathrm{raw}}$ and $V_{\mathrm{raw}}$ the key and values associated with the sink and recent tokens, the attention's output as\[ 
\widetilde \Att_{L}(Q,K,V)=\frac{
    E_{qK_{\mathrm{raw}}}V_{\mathrm{raw}}+E_{qL}W}{
      E_{qK_{\mathrm{raw}}}\mathbf 1_{N_{\mathrm{raw}}}+E_{qL}W_{\mathrm{Norm}}}.
\]

In practice, an issue that may arise due to the interpolation is the ill-conditioning of the ratio due to the interpolated denominator.
To mitigate this, we introduce a mean vector approximation of the remaining keys and values which allows us to better approximate the range of the operator, we hence call this method range correction.
\subsubsection{Mean vector approximation and range correction}
\label{sec:rc}
Given $K,V\in\RR^{N\times d}$, its associated compressed set of nodes $L\in\RR^{n\times d}$, and their weights $\Waug=[W \mid W_{\mathrm{Norm}}]\in\RR^{n\times (d+1)}$.
Range correction then consists in completing the 'missing weights' at origin associated to $\Span(K_\phi)\ominus\Span(L_\phi)$ using a `mean-node' approximation of the remaining nodes.
To do so, we consider the average of the remaining nodes and weights, given by
\begin{equation*}
\bar L:=\frac{\sum_{i=1}^{N}K_i-\sum_{j=1}^{n}L_j}{N-n},\quad
\bar W:=\frac{\sum_{i=1}^{N} V_i-\sum_{j=1}^{n} W_j}{N-n},\quad
\text{ and } \quad\bar W_{\mathrm{Norm}}:=\frac{N-\sum_{j=1}^{n}W_{\mathrm{Norm},j}}{N-n}.
\end{equation*}
Given the $N_\mathrm{raw}:=\tsink+\trec$ raw tokens $K_{\mathrm{raw}}, V_{\mathrm{raw}}\in\RR^{N_{\mathrm{raw}}\times d}$ corresponding to the sink and recent windows. 
The range correction procedure consists in computing 
\[
\widetilde \Att_{RC,L}(Q,K,V)=\frac{
    E_{qK_{\mathrm{raw}}}V_{\mathrm{raw}}+E_{qL}W+\bar N E_{q\bar L}\bar W}{
      E_{qK_{\mathrm{raw}}}\mathbf 1_{N_{\mathrm{raw}}}+E_{qL}W_{\mathrm{Norm}}+\bar N E_{q\bar L}\bar W_{\mathrm{Norm}}},\]
where we set $\bar N :=N-n$.
Using this, we can approximate each unselected node by the average of the unselected nodes, giving us at the origin
\[ 
\widetilde \Att_{RC,L}(0,K,V)=\frac{\mathbf 1_{N_{\mathrm{raw}}}^\T V_{\mathrm{raw}} +\mathbf 1_{n}^\T W+\bar N \bar W}{N_{\mathrm{raw}} + \mathbf 1_n^\T W_{\mathrm{Norm}}+\bar N \bar W_{\mathrm{Norm}}}=\Att(0,K,V).
\]
\subsubsection{Positional encoding removal}
Additionally, for sufficiently distant tokens, an issue that arises is that it appears that the RoPE positional encoding~\cite{suRoFormerEnhancedTransformer2023} of tokens adversely affects the selection of interpolation nodes.
To reduce this effect, we also consider a 'RoPE Stripping' method, wherein we consider the RoPE-less key vectors for selection only. Alternatively, we also consider stripping only the rotation matrices with frequencies such that their period is higher than the size of the segment.

\subsection{Streaming interpolation methods}
Our streaming compression techniques are hence completely defined by the
choice of the key centers and of the method used at each of the four
stages, summarised in table~\ref{tab:str_att_params}.
\begin{table}
\caption{Parameters for interpolation of streaming attention.}\label{tab:str_att_params}
\begin{center}
\begin{tabular}{lll}
\hline
Choice & Options & \\
\hline
Key centering & Segment (c) / Global (g) & \S\ref{sec:streaming} \\
\emph{(1)} Deflation & None / Basis ($\breve S_L$) / Range ($\breve S_U$)
  & \S\ref{sec:streaming} \\
\emph{(2)} Selection & GAIM / Q-DAIM
  & \S\ref{sec:selection}\\
\quad features & Exact / Linear-Taylor / Performer / RFF
  & \S\ref{sec:surrogate} \\
\quad residual & v1($T_1$) / v2($T_1{-}T_2$) / v3($\tfrac{T_1-T_2}{T_3}$) / v4($\tfrac{T_1-T_2}{h}$) / v5($\phi [q]^{\T}M_{\Delta}$)
  & \S\ref{sec:selection} \\
\emph{(3)} Projection & Basis (Galerkin) / Range / PG least squares / least squares
  & \S\ref{sec:proj} \\
\emph{(4)} Refit & None / any projection method on $L$
  & \S\ref{sec:streaming} \\
  \\
  \hline
  \multicolumn{3}{c}{Additional parameters}
  \\
  \hline
 Range correction & None / Mean vector (\S\ref{sec:rc})
  & \S\ref{sec:streaming} \\
RoPE Stripping & All frequencies/ frequency truncation
  & \S\ref{sec:streaming} \\
\hline\\
\end{tabular}
\end{center}
\end{table}
One thing to note is that deflation and refit both require keeping, or recomputing, information
on the previously compressed nodes. Indeed, the basis variants require the cached inverse, or Cholesky factorisation, of $\breve E_{LL}$ at a memory cost of $O(n^2)$ or a recomputation cost in $O(n^3)$.
The range variants on the other hand require the Gram $G_L$ and its Cholesky factorisation or inverse, at a memory cost of $O((d+1)^2)$, or a recomputation cost in $O(n(d+1)(n+d+1))$.
Furthermore, not all combinations of choices are possible, Q-DAIM only admits the linear residuals v1 and v5.
We thus define our two main strategies:
\begin{itemize}
  \item Streaming interpolation: The sequential application of the above methods to successive segments, the deflation and refit of each segment being based on the nodes and weights accumulated so far.
  \item Block interpolation: A fully parallelizable strategy in which, given a block consisting of multiple segments, each segment is interpolated in parallel, the deflation and refit being based on the $L$ and $\Waug$ of the previous blocks only, and not on the segments of the current block.
\end{itemize}

\begin{algorithm}[!ht]
\small
\caption{Compression of one segment, for one layer and one KV head.
The four stages are given as input functions, chosen among the combinations of Table~\ref{tab:str_att_params};
the range correction (\S\ref{sec:rc}) is omitted.}
\label{alg:streaming}
\begin{algorithmic}[1]
\Require segment $(K_{\mathrm{seg}},V_{\mathrm{seg}})$; nodes and weights $(L,\Waug)$; budget $\nc $;
stage functions $\textsc{Deflate}$, $\textsc{Select}$, $\textsc{Project}$, $\textsc{Refit}$
\State $(\breve S,R,\mathrm{cache})\gets\textsc{Deflate}(L,\Waug,K_{\mathrm{seg}},\Vaug_{\mathrm{seg}})$
\Comment{\emph{(1)}: deflated Gram and residual, cached factorisations}
\State $L_{\mathrm{seg}}\gets\textsc{Select}(\breve S,R,\nc )$
\Comment{\emph{(2)}: at most $\nc $ nodes of $K_{\mathrm{seg}}$}
\State $\Waug_{\mathrm{seg}}\gets\textsc{Project}(L_{\mathrm{seg}},K_{\mathrm{seg}},\Vaug_{\mathrm{seg}})$
\Comment{\emph{(3)}: on the undeflated operator}
\State $\Waug_{\mathrm{refit}}\gets\textsc{Refit}(\mathrm{cache},L,L_{\mathrm{seg}},\Waug_{\mathrm{seg}},K_{\mathrm{seg}},\Vaug_{\mathrm{seg}})$;\quad
$\Waug\gets\Waug+\Waug_{\mathrm{refit}}$
\Comment{\emph{(4)}: refit on the residual of the projected segment}
\State $(L,\Waug)\gets(L\cup L_{\mathrm{seg}},[\Waug;\Waug_{\mathrm{seg}}])$
\Comment{append the new nodes and weights}
\State \Return $(L,\Waug)$
\end{algorithmic}
\end{algorithm}

\subsection{Complexity and memory}
\label{sec:complexity}
Let us now discuss the complexity and memory requirements of the different interpolation strategies, to do so, we first recall the following notation:
\begin{center}
\begin{tabular}{llll}
  \multicolumn{4}{c}{Notation reminder}\\
\hline
$N$ & Pre-compression context size & $\Nc $ & segment size\\
$n$ & Historical/interpolation nodes & $\nc $ & Per-segment node budget\\
$d$ & Head dimension & $D$ & Surrogate feature dimension\\
$\tsink$, $\trec$ & sink and recent window sizes & $\theta= n/N$ & compression ratio\\
$r\le d+1$ & revealed rank (\S\ref{sec:sel_qdeim}) & & \\
\hline
\end{tabular}
\end{center}

In Tables~\ref{tab:mem} and~\ref{tab:cost} we list the memory and computational complexity associated to each step of the streaming interpolation methods defined above.
Here, the advantage of the surrogate methods is clear, while the cost of exact GAIM and Q-DAIM selection are dominated by the $O(\Nc ^2d)$ construction of the Gram matrix, 
the surrogate methods are able to reduce it to $O(\Nc  d D)$, making them usable even when segment sizes become prohibitive for the exact methods.
We recall that, as stated in the introduction, we consider a number of interpolation nodes per segment proportional to the head dimension, $\nc=O(d)$, so that these costs are linear in $\Nc$. For a fixed compression ratio, i.e. $\nc=\theta\Nc$, the terms in $\Nc\nc$ would instead be quadratic in the segment size.
Furthermore, a new issue arises with basis deflation and refit, indeed the requirement to keep in memory the $n^2$ matrix associated to the Gram's Cholesky factorisation (or its inverse) can quickly become prohibitive and grow beyond the initial KV cache's footprint.
The computation of the historical Gram itself is also a bottleneck as it costs $O(n^2\nc )$ per segment for a total $O(n^3)$ cost.
These increased costs make the basis deflation and refit unusable in practice.
Their Range-revealed counterparts however, represent a less costly and more memory-preserving alternative. Their computational cost and memory requirements are more manageable, scaling linearly in the number of historical nodes.
We now look at some numerical results which will enable us to shed some light on the performance of our interpolation methods, the impact of deflation and refit and a look into how the variants compare with each other.

\begin{table}[h!]
\centering\footnotesize
\vspace{2pt}
\begin{tabular}{llccl}
\hline
 & Matrix & Size\\
\hline
\multicolumn{3}{l}{\emph{Post-Compression}}\\
Nodes  & $L$ & $n\times d$\\
Weights & $\Waug$ & $n\times(d+1)$ \\
Running mean & $\mu_g$ & $d$ \\
Range-correction weights & $\bar W,\bar W_{\mathrm{Norm}}$ & $d+1$\\
Sink & $(K,V)_{\mathrm{sink}}$ & $2\tsink d$ \\
Recent  & $(K,V)_{\mathrm{rec}}$ & $2\trec d$\\

\hline
\multicolumn{3}{l}{\emph{Deflation/Refit}}\\
Basis projection & $\breve E_{LL}^{\dagger}$ or $\Gamma_{LL}$ & $n^2$ \\
Range projection  & $G_L$, $R_{L}$ & $(d+1)^2$ \\
\hline
\multicolumn{3}{l}{\emph{Post-Compression: $2(n+1+\tsink+\trec)d+n+1$}}\\
\multicolumn{3}{l}{\emph{Compression Ratio: $\theta\approx n/N$}}\\
\hline
\end{tabular}
\caption{Data in memory per head and per layer kept throughout the context compression.}
\label{tab:mem}
\end{table}

\begin{table}[h!]
\centering\footnotesize
\caption{Computational cost of each step, for a single head, per segment. The selection and residual update costs are summed over the $\nc $ picks, and the surrogate costs assume $D\ge d$.}
\label{tab:cost}
\vspace{2pt}
{\setlength{\tabcolsep}{4pt}%
\begin{tabular*}{\linewidth}{@{\extracolsep{\fill}}lcccc@{}}
\toprule
Selection & \begin{tabular}{@{}c@{}}Residual\\instantiation\end{tabular} & Selection & \begin{tabular}{@{}c@{}}Residual\\update\end{tabular} & Total\\
\midrule
\multicolumn{5}{@{}l}{\emph{Exact kernel}}\\
GAIM
  & $O(\Nc ^2 d)$ & $O(\Nc \nc  d)$ & $O(\Nc \nc (\nc +d))$ & $O(\Nc (\Nc  d+\nc ^2))$\\
Q-DAIM$^{\dagger}$
  & $O(\Nc  d(\Nc +d))$ & $O(\Nc (\nc +r))$ & $O(\Nc  r\min(\nc ,r))$ & $O(\Nc  d(\Nc +d))$\\
\addlinespace
\multicolumn{5}{@{}l}{\emph{Surrogate}}\\
GAIM-LIN
  & $O(\Nc  Dd)$ & $O(\Nc \nc  d)$ & $O(\Nc \nc D)$ & $O(\Nc D(d+\nc ))$\\
GAIM-LIN-O$^{*}$
  & $O(\Nc  Dd)$ & $O(\Nc \nc  d)$ & $O(\Nc \nc D+\nc ^3)$ & $O(\Nc D(d+\nc )+\nc ^3)$\\
Q-DAIM RFF$^{\dagger}$
  & $O(Dd(\Nc +d))$ & $O(\Nc (\nc +r))$ & $O(\Nc  r\min(\nc ,r))$ & $O(Dd(\Nc +d))$\\
\bottomrule
\end{tabular*}}

\vspace{8pt}
\begin{tabular*}{\linewidth}{@{\extracolsep{\fill}}lccc@{}}
\toprule
Projection & Matrix construction & Solve & Total\\
\midrule
Galerkin
  & $O(\Nc \nc  d)$ & $O(\nc ^2(\nc +d))^{\ddagger}$ & $O(\Nc \nc  d+\nc ^3)$\\
Revealed-range
  & $O(\Nc  rd)$ & $O(\nc  rd)$ & $O(\Nc  rd)$\\
Least squares
  & $O(\Nc ^2 d)^{\S}$ & $O(\Nc \nc (\nc +d))$ & $O(\Nc ^2 d+\Nc \nc ^2)$\\
\bottomrule
\end{tabular*}

\vspace{8pt}
{\setlength{\tabcolsep}{4pt}%
\begin{tabular*}{\linewidth}{@{\extracolsep{\fill}}lccc@{}}
\toprule
Deflation and refit & From scratch & Cached factorisations & Surrogate\\
\midrule
Basis deflation
  & $O(n(\Nc +n)(n+d))$ & $O(n(\Nc (n+d)+nd))$ & $O(nd(\Nc +n+D))$\\
Range deflation, $G_L$ update
  & $O(d(\Nc +n)(n+d)+d^3)$ & $O(\Nc  d(n+d)+d^3)$ & $O(d(\Nc +D)(n+d))$\\
Bordered update of $\breve E_{LL}^{\dagger}$
  & --- & $O(\nc  n(n+d)+\nc ^3)$ & ---\\
\addlinespace
Basis refit
  & $O(n((\Nc +n)d+n^2))$ & $O(nd(n+\nc ))$ & ---\\
Range refit
  & $O(nd(\Nc +n+d)+d^3)$ & $O(d^2(n+\nc +d))$ & ---\\
Least-squares refit
  & \multicolumn{2}{c}{$O(\Nc  n(\min(n,\Nc )+d))^{\S}$} & ---\\
\bottomrule
\end{tabular*}}

\vspace{4pt}
\begin{minipage}{\linewidth}\scriptsize
$^{*}$Galerkin coefficients updated by bordering at each pick, so that each residual update is rank-1.\\
$^{\dagger}$Single pass of Algorithm~\ref{alg:qdaim}, $r\le\min(\Nc ,d+1)$. The restarts of Appendix~\ref{app:qdaim_restart} ($\nc >r$) add $O(\Nc \nc (\nc +r))$ to the exact selector.\\
$^{\ddagger}$Absorbed when $\breve E_{LL}^{\dagger}$ is maintained by bordered updates (exact GAIM, GAIM-LIN-O, streaming basis variants).\\
$^{\S}$$R^0=\breve E_{KK}\Vaug_\eta$ costs $O(\Nc ^2 d)$ unless reused from an exact selection, as assumed for the refit.
\end{minipage}
\end{table}
\pagebreak
\section{Numerical results}
\label{sec:results}
In this section, we discuss the efficiency of our method by considering two different aspects.
First, we focus on the convergence of the interpolation method in terms of the norm of the different residuals considered in this work and in terms of the downstream exact attention error for queries that appear after the context is compressed. Second, we focus on the score of the methods on a set of benchmarks, LongBench-E~\cite{baiLongBenchBilingualMultitask2024}, used in the literature to determine the performance of compression methods.
Our method uses the following setting:
\begin{itemize}
  \item Model: Qwen2.5-7B-Instruct~\cite{qwenQwen25TechnicalReport2025}: $28$ layers, $\Nh=28$ query heads, $\Nkv=4$ KV heads, $d=128$ with bf16 weights.
  \item Library: Huggingface Transformers Library~\cite{wolf-etal-2020-transformers} with PyTorch 2.12 backend.
  \item Backend: Model forward: bf16 on Metal(MPS), Compression: float64 on CPU.
  \item Hardware: Apple M4 Pro with 48GB of Unified Memory.
\end{itemize}

\subsection{Interpolation Residuals}\label{sec:resid_setup}
To study the convergence of the GAIM and Q-DAIM approaches described in \S\ref{sec:selection} and \S\ref{sec:sel_qdeim}, we use a single long document from the Wiki-40b~\cite{guo-etal-2020-wiki} dataset: the English Wikipedia article on the American Civil War (Wikidata Q8676, index 449 of the English-language train split).
After the $\tsink=32$  sink tokens, as introduced in \S~\ref{sec:streaming} this corresponds to the first tokens of the context, we compress the key and value vectors of the next $N=4096$ tokens and evaluate the query vectors of the following $4096$ tokens, which we call \emph{future queries}.
Unless stated otherwise, every metric is the median over all  KV heads\footnote{In certain types of attention called Grouped Query Attention (GQA), multiple heads of attention share the same set of key and value vectors but possess different query vectors, this shared head of attention is what we refer to here as a KV head. In MHA every head of attention is a KV head.} of the layers $\{0,7,14,21\}$.
Recall that Q-DAIM only admits the linear residuals v1 and v5, which explains the missing Q-DAIM curves for v2, v3 and v4 in the following figures.

\subsubsection{Direct Interpolation residuals}
As a first experiment, we look at the interpolation of the complete context ($N=4096$) using our different methods and their variants.
We look at three metrics which depend only on the keys and values, all expressed with the basis deflated Gram matrix $\breve S_L$ of \S\ref{sec:streaming} and the Hilbert--Schmidt norm:
\begin{itemize}
  \item relative norm of the augmented residual operator: $\frac{\|M_{\Delta,\eta}\|_{2}^2}{\|M_{\eta}\|_{2}^2}=\frac{\Tr({\Vaug_\eta}^\T\breve S_{L}\Vaug_\eta)}{\Tr({\Vaug_\eta}^\T \breve E_{KK}\Vaug_\eta)}$.
  \item relative norm of the numerator residual operator: $\frac{\|M^{g}_{\Delta,\eta}\|_{2}^2}{\|M^g_{\eta}\|_{2}^2}=\frac{\Tr(V_\eta^\T\breve S_{L} V_\eta)}{\Tr(V_\eta^\T \breve E_{KK}V_\eta)}$.
  \item relative norm of the denominator residual operator: $\frac{\|M^{h}_{\Delta,\eta}\|_{2}^2}{\|M^h_{\eta}\|_{2}^2}=\frac{\Tr(\eta^\T\breve S_{L} \eta)}{\Tr(\eta^\T \breve E_{KK}\eta)}$.
\end{itemize}
Figure~\ref{fig:resid_complete_context} displays the median squared norms of the augmented and denominator operators, the numerator being indistinguishable from the augmented one. We observe that, as expected, variants v1 and v5, which directly optimize these residuals, perform better than the other variants in these metrics. A surprising result is the fact that the residual norm of $M^{h}_{\Delta,\eta}$ is also minimised by the selection of the un-augmented variant v1.

\begin{figure}[!htbp]
  \centering
  \includestandalone{figures/tikz/fig_resid_complete_context}
  \caption{Residual operator convergence for the different methods.
  Relative squared norm of the augmented residual operator (top) and of the denominator residual operator (bottom) against the number of nodes $n$, median over the $16$ KV heads.
  Each panel shows the variants of one method in colour and the curves of the other methods in grey.
  The norm of the numerator residual operator $M^{g}_{\Delta,\eta}$ stays within about $3\%$ of that of the augmented one and is not shown.}
  \label{fig:resid_complete_context}
\end{figure}
Figures~\ref{fig:resid_exact_gh} and~\ref{fig:resid_exact} display the median error for the numerator, denominator, and exact attention on future queries. To make heads comparable, we divide the errors by the mean output norm over the 10\% of future queries with the largest attention on the same head, which illustrates how the error compares with the heavy hitters of that head, i.e. the set of queries with high attention output norms relative to the rest of the queries.
We define this as the median heavy-hitter relative error, given the future queries matrix $Q\in\RR^{N_Q\times d}$,
\begin{equation}\label{eq:rel_att_err}
\varepsilon_g^{10\%}=\frac{\mathrm{med}_{q\in Q}\,\|\Delta g(q)\|_{2}}{\mathrm{mean}_{q\in Q^{10\%}_{g}}\,\|g(q)\|_{2}},\quad
\varepsilon_h^{10\%}=\frac{\mathrm{med}_{q\in Q}\,|\Delta h(q)|}{\mathrm{mean}_{q\in Q^{10\%}_{h}}\,|h(q)|},\quad
\varepsilon_{\Att}^{10\%}=\frac{\mathrm{med}_{q\in Q}\,\|\Att(q,K,V)-\widetilde \Att_L(q,K,V)\|_{2}}{\mathrm{mean}_{q\in Q^{10\%}_{\Att}}\,\|\Att(q,K,V)\|_{2}},
\end{equation}
where $Q^{10\%}_{f}\subset Q$ denotes the $10\%$ of queries on which $\|f(q)\|_2$ is largest.

The dash-dotted line indicates the \emph{mean-vector approximation}, i.e. the range-corrected estimate (described in \S\ref{sec:rc}) with $n=0$ nodes, and the dotted line indicates the median output norm, i.e. the zero-prediction error.
As expected, the v2, v3 and v4 variants outperform v1 and v5 in terms of attention error for all GAIM-based methods while, by design, the opposite is true for the residual norms, as can be expected given that v1 and v5 directly minimise the residual norms.

\begin{figure}[!htbp]
    \centering
  \includestandalone{figures/tikz/fig_resid_exact_gh}
  \caption{Heavy hitter relative error~\eqref{eq:rel_att_err} in numerator and denominator with $\tsink=32$ and without range correction.
  Top: $\varepsilon_g^{10\%}$; bottom: $\varepsilon_h^{10\%}$; both on the future queries.
  Each curve is the median of this ratio over the $112$ query heads; each panel shows the variants of one method in colour and the curves of the other methods in grey. The interquartile range over the heads is shown in Figure~\ref{fig:resid_exact_gh_iqr} of Appendix~\ref{app:additional_figures}.}
  \label{fig:resid_exact_gh}
\end{figure}
\begin{figure}[!htbp]
  \centering
  \includestandalone{figures/tikz/convergence_exact_linf_sink32}
  \caption{Attention error in $L^2$ norm on future queries with $\tsink=32$ and $\trec=0$.
  Heavy hitter relative attention error $\varepsilon^{10\%}_{\Att}$ without (top) and with (bottom) range correction.
  Each curve is the median of this ratio over the $112$ query heads; each panel shows the variants of one method in colour and the curves of the other methods in grey. For both rows, the zero-readout errors are at least 3 times higher and have thus not been drawn in the figures. The interquartile range over the heads is shown in Figure~\ref{fig:resid_exact_iqr} of Appendix~\ref{app:additional_figures}.}
  \label{fig:resid_exact}
\end{figure}

\subsubsection{Impact of range-correction on a long context example}
 \input{tables/rc_head_linf.tex}

Table~\ref{tab:rc_heads_improved_linf_layers} shows the number of heads of each layer that were improved by range correction in terms of $L^\infty$ attention error in future queries. 
The decisive advantage of the range-corrected estimators on the median error observed in Figure~\ref{fig:resid_exact} is less clear once we look at specific heads. Indeed, as can be seen from Table~\ref{tab:rc_heads_improved_linf_layers}, while most heads are improved, few of them are improved by more than $10\%$. To illustrate this, Figure~\ref{fig:rc_worst_per_q} and Figure~\ref{fig:rc_best_per_q} show two failure modes for the interpolation, one for the range-corrected case and one for the raw interpolation.
Each figure plots the norm of the attention error for each future query, ordered according to the norm of the attention of the query; the solid black line indicates the norm of the attention output vector while the dotted line corresponds to the mean vector approximation.
As can be seen from Figure~\ref{fig:rc_worst_per_q}, for some heads range correction adversely affects the performance of the interpolation.
Indeed, as the mean vector approximation has a high error, range-corrected interpolation performs poorly until a sufficient number of interpolation nodes are chosen, compensating the mean value approximation, whereas the uncorrected interpolation performs well even with few nodes.
It is thus not a panacea but a parameter which may improve performance in certain cases, and for certain tasks as we shall see on the benchmarks.
On the other hand, we can see in Figure~\ref{fig:rc_best_per_q} a case where, when the chosen interpolation nodes are not able to represent attention at the queries correctly, range correction allows to mitigate the failure of interpolation and, given a sufficient number of nodes, to extract some additional performance from the interpolation.

\begin{figure}[!htbp]
  \centerline{\includestandalone{figures/tikz/perquery_smoothnorms_L0q4_worst_errinf}}
  {\centering
  \caption{Worst-case error for range-corrected GAIM interpolation in $L^\infty$ norm. Layer 0, query head 4.
  GAIM exact with the nodes selected in the $L^\infty$ score norm; each curve is a sliding mean over $205$ consecutive queries.}
    \label{fig:rc_worst_per_q}
  }
\end{figure}

\begin{figure}[!htbp]
  \centerline{\includestandalone{figures/tikz/perquery_smoothnorms_L1q16_best_errinf}}
  {\centering
  \caption{Worst-case GAIM interpolation error and its improvement with range-correction in $L^\infty$ norm. Layer 1, query head 16; this layer lies outside the layers $\{0,7,14,21\}$ used for the median curves.
  GAIM exact with the nodes selected in the $L^\infty$ score norm; each curve is a sliding mean over $205$ consecutive queries. Range correction enables some of the variants to improve their error levels, especially on the tokens with higher attention scores.}
  \label{fig:rc_best_per_q}
  }
\end{figure}

\subsection{Alternative projection comparison}
Figures~\ref{fig:proj_gaim_exact} and \ref{fig:proj_qdaim_exact} illustrate the impact of alternative projections described in \S\ref{sec:proj} on the residual error of the GAIM v4 and Q-DAIM v5 selection methods described above. These show that on future queries, the oblique projections of Q-DAIM (in the case of Q-DAIM selection) and of WildCat provide a definite improvement in terms of residual attention error.
As we will see in the LongBench-E benchmarks, this reduction in attention error with respect to the uncompressed attention does not correlate one-to-one with better performance on tasks in practice. Although it is worth noting that very poor attention reconstruction error correlates with poor task results, generally because they result in garbled nonsensical text. Table~\ref{tab:temperature_heads_improved_linf} displays the number of heads for which WildCat temperature scaling improves the attention error on future queries.
\input{tables/temp_comp}
For the rest of the residual-based experiments, we will consider the Galerkin, i.e. the orthogonal projection, onto the interpolation nodes.

\begin{figure}[!htbp]
  \centering
  \includestandalone{figures/tikz/projections_gaim_exact_v4_linf_sink32}
  \caption{Heavy-hitter relative attention error $\varepsilon^{10\%}_{\Att}$~\eqref{eq:rel_att_err} using GAIM exact v4 interpolation nodes for the different projection methods described in Table~\ref{table:projections}, and without projection, i.e. keeping the values of the selected keys, $W=V[\mathfrak L,:]$. Median over the $112$ query heads; the interquartile range over the heads is shown in Figure~\ref{fig:proj_gaim_exact_iqr} of Appendix~\ref{app:additional_figures}. The methodology used is the same as the experiments in \S\ref{sec:resid_setup}.}  \label{fig:proj_gaim_exact}
\end{figure}
\begin{figure}[!htbp]
  \centering
  \includestandalone{figures/tikz/projections_qdaim_exact_v5_linf_sink32}
 \caption{Heavy-hitter relative attention error $\varepsilon^{10\%}_{\Att}$~\eqref{eq:rel_att_err} using Q-DAIM exact v5 interpolation nodes for the different projection methods described in Table~\ref{table:projections}, and without projection, i.e. keeping the values of the selected keys, $W=V[\mathfrak L,:]$. Median over the $112$ query heads; the interquartile range over the heads is shown in Figure~\ref{fig:proj_qdaim_exact_iqr} of Appendix~\ref{app:additional_figures}. The methodology used is the same as the experiments in \S\ref{sec:resid_setup}.}\label{fig:proj_qdaim_exact}
\end{figure}
\subsubsection{Block solve with global projection}
We focus now on the same document on the effect of a projection restricted to a segment versus a global projection on a complete interpolation basis.
To do so, we select the nodes with the best-performing methods on the above metrics, in a per-segment interpolation without deflation or refit.
For each of these methods, we compare the projection restricted to each segment with the projection against all interpolation nodes.
In Figure~\ref{fig:chunk_vs_global_solve}, we plot, for each method, the ratio of the error obtained with the global projection to the error obtained with the segment-local one, computed head by head.
Interestingly, in most cases the residual error of the segment local interpolation is within $5\%$ of the global interpolation.
\begin{figure}[!htbp]
  \centering
  \includestandalone{figures/tikz/global_solve_ratio_cs256_p25_sink32}
    \caption{Comparison of segment-local projection versus global interpolation node projection on attention residual.
    Ratio, head by head, of the error obtained with the global projection to the error obtained with the segment-local one, where the error is the median attention error $\mathrm{med}_q\,\|\Att(q)-\widetilde\Att_L(q)\|_2$ over the future queries (left, centre) or the fit residual $\|\breve E_{KK}\Vaug_\eta-\breve E_{KL}\,\eta_L\Waug\|_F$ at the keys (right).
    The dot is the median over the heads, the thick bar spans the quartiles and the thin line the 10th to 90th percentiles; a ratio below $1$ favours the global projection, and ``worse in'' gives the share of heads on which it has the larger error.
    $112$ query heads for the future queries and $16$ KV heads for the fit at the keys, with $\Nc =256$, $\nc =64$ and $\tsink=32$.}
  \label{fig:chunk_vs_global_solve}
\end{figure}
\subsubsection{Comparison between streaming and block methods}
Finally, we compare the streaming method described in  \S\ref{sec:streaming} with range deflation and refit to block methods without deflation or refit.
In order for the comparison to reflect the interpolation methods as used in practice, we proceed as follows.
First, we keep the raw keys and values of the sink tokens $\tsink=32$, then we compress the key and value vectors of the subsequent tokens $N=4096$ and keep the key and value vectors of the next tokens $\trec=32$.
Then, for each query after the $\tsink+N+\trec$ tokens, we compute the attention with the additional raw keys and values of the tokens up to the query token, as would occur in practice if one were to compress only the $N$ tokens.
\begin{figure}[!htbp]
  \centering
  \includestandalone{figures/tikz/graph3_heatmap_aug_sink32_rc}
      \caption{Norm of the augmented residual operator for segmented interpolation.
      Top: relative squared norm $\|M_{\Delta,\eta}\|_2^2/\|M_{\eta}\|_2^2$ of the block method, in percent and shaded on a logarithmic scale (median over the heads).
      Bottom: difference of this norm between the streaming (with range deflation and refit) and the block method, streaming minus block, in percentage points (blue: lower with streaming, red: higher).
      Range correction on, $\tsink=32$.}
\label{fig:aug_heatmap}
\end{figure}
\begin{figure}[!htbp]
    \centering
  \includestandalone{figures/tikz/graph3_realq_heatmap_aug_sink32_rc_recent32}
    \caption{Impact of segment-interpolation parameters on best-performing GAIM and Q-DAIM variants.
    Top: relative attention error $\mathrm{med}_q\,\|\Att(q)-\widetilde\Att_L(q)\|_2\,/\,\mathrm{mean}_{q\in\mathrm{top}\,10\%}\,\|\Att(q)\|_2$ on the future queries for the block method, in percent (median over the heads).
    Bottom: difference of this error between the streaming (with range deflation and refit) and the block method, streaming minus block, in percentage points (blue: lower with streaming, red: higher).
    Range correction on, $\tsink=32$, $\trec=32$.}
  \label{fig:res_heatmap}
\end{figure}
Some observations are in order. First, the impact of the streaming configuration on the norm of the augmented residuals in Figure~\ref{fig:aug_heatmap} is small in most cells (for the variants v1 or v5 that converge on the operator norms). For future queries attention residual, it appears to be a small but net positive for the smaller segment-size and smaller compression ratio cells of the GAIM-exact method, while its effect on the other methods is unclear. All in all, it appears that while the impact of range deflation and refit is small, it is nonetheless there. And as we shall see in the next experiments, the relationship between the performance in language processing tasks and the remaining residual norm/future query attention error is unclear. Indeed, while a high residual norm or future query error implies a poor performance, the lowest residual does not imply the best performance, which indicates that further investigations into a better metric to optimise might be necessary.

\subsection{LongBench-E experiments}
Indeed, while the residual error allows us to compare the output of the interpolated attention with the exact attention, this does not necessarily correlate with actual performance on natural language processing tasks.
In order to illustrate the impact of the compression on the performance of the LLM on given tasks, we provide in Table~\ref{tab:longbench_e_ledger} a preliminary result on the LongBench-E~\cite{baiLongBenchBilingualMultitask2024} benchmark.
For a few variants of our GAIM and Q-DAIM methods, we report, for each LongBench-E task, the mean score over $25$ samples per task.
The bold scores correspond to the best-performing compression method, and their distribution illustrates that the different parameters introduced in the article favour different types of tasks.
While the exact GAIM and Q-DAIM methods were restricted to segment sizes on the order of $d$ due to their quadratic scaling, we observe that the surrogate methods perform on par with the exact ones even though they scale linearly with $\Nc $. For comparison, we include the performance of CompressKV, the WildCat-based compression technique reported in \cite{schroderWildCatNearLinearAttention2026} on the complete benchmark and for one-to-one comparison we also show \emph{RP-Cholesky}, a Randomly-Pivoted Cholesky selection in the WildCat $\tau$-scaled kernel, similar to WildCat but using our segment-based numerical framework. Additionally, we also computed GAIM-LIN and Q-DAIM exact in the WildCat $\tau$-scaled kernel formulation, for both projection and selection, which provided a considerable improvement in residual error experiments. In these benchmarks, it appears to mostly reduce the performance of GAIM while slightly improving Q-DAIM on some specific tasks. 
Most notably, some question answering tasks, MFQA and HotpotQA, scores seem to be positively impacted by the temperature scaling both in terms of selection and projection.

\input{tables/longbench_e_ledger.tex}

\subsection{Observations}

As the experiments above show, the attention residual does not seem to be correlated to the benchmark scores directly.
Indeed, as we have seen in Figure~\ref{fig:proj_gaim_exact} and Figure~\ref{fig:proj_qdaim_exact} the temperature-scaled projection lowers the attention error, yet loses 12 to 14 points on qasper for the GAIM-LIN selection and 3 points for the Q-DAIM selection. The selection metric however does seem to determine which type of task performs better.
At the same budget, Q-DAIM and GAIM-LIN reach nearly the same LongBench-E mean, albeit with different profiles. Comparing their Galerkin variants with range correction, Q-DAIM performs better on summarisation (GovRep, MultiNews), on HotpotQA and on code completion in LCC, while GAIM-LIN leads on single-document question answering (Qasper, MFQA) and, by a smaller margin, on 2Wiki and RepoB-P.
This suggests that the selection criterion impacts which information the interpolation preserves, and that the range of the augmented operator, revealed by Q-DAIM, helps identify the features that contain it. It is also interesting to note that the selection of interpolation nodes has the most decisive impact on output performance. While projection does help in some respects, simply using the values associated with the nodes without projection, also works well.
For instance, the projection-free Q-DAIM variant is the best performer on average ($50.40$, $87.1\%$ of Dense), although it wins mainly by a large margin on a single task, 2Wiki, where it reaches $60.5$ (almost on par with the uncompressed model) against $35.0$ for the Galerkin variant with range correction ($45.9$ without), while performing less well than the former on eight of the twelve other tasks.
Without projection or range correction, each node keeps its own key and value, so the score depends only on which tokens are kept; the fact that Q-DAIM still outperforms GAIM-LIN in this setting therefore suggests that its selection chooses better nodes.

Finally, Q-DAIM's revealed-range projection restricted to the revealed rank (range$_r$) performs very well. Obtaining $50.02$ on average ($86.5\%$ of Dense), with the best scores on TriviaQA and SAMsum and the shared best on PRetr.
The restriction to the revealed rank, which is not necessary for the other projections, is due to the fact that on the span revealed by $U$ of dimension $r$, the first $r$ pivots reduce the residual on this subspace of $H$ to zero and the projected values on the remaining pivots are numerical noise.
Indeed, under range deflation, the operator in a later segment can have fewer revealed directions $r$ than the node budget $\nc$, down to one or two for $\nc=16$ in some heads of the middle layers. Unrestricted range projection then reduces the performance dramatically due to the induced noise ($32.1$ on average, with garbled outputs on several prompts).
Restricting the projection to the first $r$ pivots, removes this failure but leaves part of the budget unused in those segments.
Our restart method, consisting in projecting onto the orthogonal complement of the chosen nodes, allows us to find new directions that improve the reconstruction of $U$ on $H$, but with diminishing returns as all the keys are required to construct $U$ completely.
Finding a method that would allow one to represent the revealed range $U$ more fully, could use this budget to further improve the method (\S\ref{sec:discussion}).

We also note that removing the RoPE positional embedding on the key vectors has a noticeable impact on the performance of our methods, illustrated by the best-performing PCount score.
This provides some weight to the argument that better compression may require taking into account positional information separately from semantic information. 

We highlight again the two cases illustrated in Figures~\ref{fig:rc_best_per_q} and~\ref{fig:rc_worst_per_q}, namely the fact that several regimes appear to occur in different heads, corresponding to the ability of either mean vector approximation or kernel based interpolation to provide an effective compression of the context and the interaction between the two methods.
 These cases provide some evidence to support the idea that several different structures emerge in different attention heads and that the best methods of approximation for each structure differ.  It seems reasonable to expect that such structures not only appear in the studied model but that similar structures occur in most learned models of attention and that the correct way to compress the information contained in the data must take these structures into account.

\section{Discussion and Future work}
\label{sec:discussion}
The experimental results we obtained are promising, showing a convergence in terms of residual norm and a reasonable LongBench-E performance, comparable to state of the art methods (see~\cite{schroderWildCatNearLinearAttention2026}), which may, given more experimentation on the choice of method and associated best set of parameters, allow for further improvements in performance. Furthermore, the interpolation methods we developed do not rely on the textual nature of the context, but only on the kernel underlying scaled dot-product attention.
They are thus also suited to other types of attention mechanisms, such as the self-attention and cross-attention layers found in models aimed at vision, such as Vision Transformer~\cite{dosovitskiy2021imageworth16x16words} based models~\cite{simeoni2025dinov3,teamGemma3Technical2025} or graph attention models~\cite{Sun_2023}.
In these non-causal settings, the complete context is usually available at once while the context size is then set by a fixed resolution input. Furthermore, given the fact that the nature of the output is different, the performance of the interpolated context might be more directly related to its residual operator norm or attention error.
The interpolatory framework developed in this article motivates several extensions, which will be the subject of future work.
First, since the revealed range of Q-DAIM $\mathcal U=\Span(U)$ exists in the infinite dimensional RKHS, we never manipulate it directly but only through its evaluations of key features $K_\phi$, requiring the storage of the complete set of $N$ keys. The natural extension is thus to try to represent $U$ directly in order to reduce this costly dependence on the keys. An avenue to achieve this might be through the use of a random range finder for operators in Hilbert spaces~\cite{kressnerRandomizedSVDInfinite2025}.
Second, our experiments provide support for the fact that different heads of attention possess different structures for the information they contain. These structures respond differently to interpolation and range correction, and thus motivate the construction of an efficiently computable metric to identify the methods suitable for the approximation.
One such candidate is clustering, as it has been studied extensively for attention~\cite{zandiehKDEformerAcceleratingTransformers2023,hanHyperAttentionLongcontextAttention2023,hooperMultipoleAttentionEfficient2025} but not for kernel interpolation, and a direct extension of our methods would be to interpolate each cluster separately along with its own mean vector range correction.
A related finding which appears to affect the approximability of these structures is positional encoding, and applying our methods to Latent Attention~\cite{deepseek-aiDeepSeekV2StrongEconomical2024} might shed some light on its impact due to its separation of positional from semantic information. Furthermore, applying our methods to it through conversion methods such as TransMLA~\cite{mengTransMLAMultiHeadLatent2025} would let us compare how the same data interpolate in latent and regular attention.

Finally, the methods developed in this article reduce the context size $N$, and can thus be complementary to quantization-based methods such as TurboQuant~\cite{zandieh2025turboquantonlinevectorquantization} or PolarQuant~\cite{han2025polarquantquantizingkvcaches}. Further improvements may also be obtained by using quantization to develop a more efficient way of selecting keys and projecting values by concentrating the information contained in the key and value vectors.

\section{Acknowledgments}
The first author (M.F.S.) would like to thank Alexandre Pasco for the lively discussions that provided additional insights on the attention mechanism, Johannes Laute for the many conversations on the numerical implementation of our methods, and Benjamin Carrel for the numerous insightful discussions concerning interpolation.    

\subsection*{Code and Data}
The code used for the numerical experiments in Section~\ref{sec:results}
was written with the assistance of Claude Opus 4.8 and Fable 5.1 (Anthropic), accessed
through Claude Code in 2026. The authors designed the experiments, reviewed
and tested all generated code, and take full responsibility for the results.
The experimental code is based upon the Huggingface transformers~\cite{wolf-etal-2020-transformers} python library and all experiments were run on Qwen2.5-7B-Instruct~\cite{qwenQwen25TechnicalReport2025}. The data used for the experiments comes from the Wiki-40b~\cite{guo-etal-2020-wiki} dataset and the LongBench benchmark~\cite{baiLongBenchBilingualMultitask2024} dataset.
\subsection*{A.I. Usage for typesetting and literature review}
In addition to the use of AI for coding, AI assistance using the Claude Code Harness (Opus 4.8, 5.5 and Fable 5.1) was used to search the literature, to generate diagrams, and to format and typeset parts of the text and ensure coherence of notation, as well as provide feedback on the text.
\printbibliography
\appendix

\section{Proofs}
\label{app:proofs}

\begin{proof}[Proof of \Cref{prop:intersection}]
We first prove the result when the rows of $K$ are, in addition, pairwise non-collinear, i.e. such that, for all distinct $i,j\in[N]$, there exists no $\alpha$ for which $K_i=\alpha K_j$. Let $\Psi\in\Span(K_{\phi})\cap\phi\left[\Span(K)\right]$, we can then set, dropping the prefactor  \[\Psi=\sum_{i=1}^{N}\alpha_i\phi[K_i]=\left((d^{r/4}\sqrt{r!})^{-1}\sum_{i=1}^{N}\alpha_iK_i^{\otimes r}\right)_{r\in\NN},\] where, for all $i\in[N]$, $\alpha_i\in\RR$. Since $\Psi\in\phi[\Span(K)]$ there also exists a $v\in\Span(K)$ such that $\Psi=\phi [v]$, so the degree-$r$ component is
$\Psi_r=(d^{r/4}\sqrt{r!})^{-1}v^{\otimes r}$.
For the case $v=0$, we have $\phi_0[v]=1=\sum_{i=1}^{N}\alpha_i$ and for any $r\ge1$, $\phi_r[v]=\sum_{i=1}^{N}\alpha_i\phi_r[K_i]=0$, which thanks to the linear independence we shall prove below implies that, for all $i\in[N]$, $\alpha_i=0$, and hence $1=\sum_{i=1}^{N}\alpha_i=0$. This contradiction then requires $v\neq 0$.
For $v\neq0$, we have for every $r\ge1$, that the matricization of $\Psi_{2r}$ has rank
$1$.  
Let us show that this leads to a unique non-zero coefficient $\alpha_i=1$ for some $i\in[N]$.
Indeed, for sufficiently large
$r$, the rank of the matricization of
$\Psi_{2r}=\sum_{i=1}^{N}\alpha_iK_i^{\otimes r}\otimes K_i^{\otimes r}$ is equal to the number of non-zero coefficients. Indeed, as the rows of $K$ are pairwise linearly independent vectors, we have that, the $r$-th tensor power of the keys $K_i^{\otimes r}$, for all $i\in[N]$, are linearly independent~\cite[Corollary 4.4]{comonSymmetricTensorsSymmetric2008}.
Hence, the rank of $\Psi_{2r}$ is equal to the number of non-zero coefficients. 
Thus, as the rank needs to be equal to 1, we have that there exists a single non-zero coefficient $\alpha_j\in\RR^{*}$ such that $\Psi=\alpha_j \phi [K_j]$ for
some $j$. The first result of \Cref{lem:unicity_key_fm} then
gives us that $\alpha_j=1$.
Finally, thanks to the second result of Lemma~\ref{lem:unicity_key_fm}, we can extend the result to $K$ with distinct, non-zero rows.
Indeed, consider a key matrix whose rows form $p\in[N]$ equivalence classes $C_1,\dots,C_p$ for the pairwise collinearity, with representatives $u_1,\dots,u_p$, and set $K_i=c_iu_k$ for all $i\in C_k$, where the $c_i$ are non-zero and distinct inside each equivalence class. Let $m\in[N]$ be the cardinality of the biggest collinearity class, $R:=\max(1,p-1)$, $\mathfrak{R}:=\{R,\dots,R+m-1\}$ and, for any $x\in\RR^d$, let $\phi_{\mathfrak{R}}[x]:=(\phi_r[x])_{r\in\mathfrak{R}}$ denote the truncated features.
First, the truncated features $\phi_{\mathfrak{R}}[K_i]$, $i\in[N]$, are linearly independent. Indeed, let $\gamma\in\RR^N$ be such that $\sum_{i=1}^{N}\gamma_i\phi_{\mathfrak{R}}[K_i]=0$, i.e. such that, for all $r\in\mathfrak{R}$,
\[
\sum_{k=1}^{p}\Big(\sum_{i\in C_k}\gamma_ic_i^{r}\Big)\phi_r[u_k]=0.
\]
As $r\geq p-1$, the $r$-th tensor powers of the representatives $u_k^{\otimes r}$ are linearly independent~\cite[Corollary 4.4]{comonSymmetricTensorsSymmetric2008}, so that $\sum_{i\in C_k}\gamma_ic_i^{r}=0$ for all $k\in[p]$ and $r\in\mathfrak{R}$, and the second result of Lemma~\ref{lem:unicity_key_fm}, applied inside each equivalence class, gives us that $\gamma=0$. In particular, the case $v=0$ is excluded in the same way as above, as it would give $\sum_{i=1}^{N}\alpha_i\phi_{\mathfrak{R}}[K_i]=0$.
Then, in the same way as above, consider the matricization of $\sum_{i=1}^{N}\alpha_i\phi_{\mathfrak{R}}[K_i]\otimes\phi_{\mathfrak{R}}[K_i]$. Since, for any $x\in\RR^d$ and $r,s\in\mathfrak{R}$, $\phi_r[x]\otimes\phi_s[x]=\binom{r+s}{r}^{1/2}\phi_{r+s}[x]$, its $(r,s)$ block is given by
\[
\sum_{i=1}^{N}\alpha_i\phi_r[K_i]\otimes\phi_s[K_i]=\binom{r+s}{r}^{1/2}\Psi_{r+s}=\binom{r+s}{r}^{1/2}\phi_{r+s}[v]=\phi_r[v]\otimes\phi_s[v],
\]
so that $\sum_{i=1}^{N}\alpha_i\phi_{\mathfrak{R}}[K_i]\otimes\phi_{\mathfrak{R}}[K_i]=\phi_{\mathfrak{R}}[v]\otimes\phi_{\mathfrak{R}}[v]$, whose matricization has rank $1$ as $v\neq0$. Now, as before, since the truncated feature vectors are all linearly independent, we have the same rank argument as in the non-collinear case and thus the result, there is a unique $j\in[N]$ such that $\phi[v]=\phi[K_j]$.
\end{proof}

\begin{proof}[Proof of \Cref{lem:unicity_key_fm}]
Suppose $\phi[v]=\alpha\phi[u]$ for some $v\in\RR^d$ and $\alpha\notin\{0,1\}$. The equality of
the degree-$1$ components gives $v=\alpha u$. The equality of the degree-$2$ components then
requires $\alpha \phi_2[u]=\phi_2[\alpha u]=\alpha^2\phi_2[u]$ with $\phi_2[u]\neq0$, so
$\alpha\in\{0,1\}$, contradicting the supposition.

For the second claim, let $\beta\in\RR^{k}$ be such that, for all $r\in\{R,\dots,R+k-1\}$,
\[
 \sum_{i=1}^{k}\beta_i\phi_r[u_i]=\Big(\sum_{i=1}^{k}\beta_i\alpha_i^r\Big)\phi_r[u]=0 .
\]
Since $u\neq0$, $\phi_r[u]\neq0$, this is equivalent to
\begin{equation}\label{eq:collinear_moments}
  \sum_{i=1}^{k}\beta_i\alpha_i^r=0,\qquad r\in\{R,\dots,R+k-1\}.
\end{equation}
We show by induction on $k$ that this implies $\beta=0$, for any $R\in\NN$,
any pairwise distinct $\alpha_1,\dots,\alpha_k\in\RR\setminus\{0\}$ and any $\beta\in\RR^k$.
For $k=1$, \eqref{eq:collinear_moments} implies $\beta_1\alpha_1^R=0$, hence $\beta_1=0$ as
$\alpha_1\neq0$. For $k\geq2$, assume the claim holds for the $k-1$ vectors.
Let us consider $\beta$ such that for all $r\in[R,R+k-1]$,
\[
  \sum_{i=1}^{k}\beta_i\alpha_i^{r}
  =\sum_{i=1}^{k-1}\beta_i\alpha_i^{r} + \beta_k\alpha_k^{r}=0 .
\]
For each
$r\in\{R,\dots,R+k-2\}$, subtracting $\alpha_k$ times the equation of degree $r$ from that of degree $r+1$ eliminates the last term, giving us
\[
  \sum_{i=1}^{k}\beta_i\alpha_i^{r+1}-\alpha_k\sum_{i=1}^{k}\beta_i\alpha_i^{r}
  =\sum_{i=1}^{k-1}\beta_i(\alpha_i-\alpha_k)\alpha_i^{r}=0 .
\]
We identify this as the property \eqref{eq:collinear_moments} for $\alpha_1,\dots,\alpha_{k-1}$ on the $k-1$
consecutive degrees $R,\dots,R+k-2$, with coefficients $\beta_i(\alpha_i-\alpha_k)$. Hence, by the
induction hypothesis, $\beta_i(\alpha_i-\alpha_k)=0$, and since the $\alpha_i$ are pairwise
distinct, $\beta_i=0$ for all $i\in[k-1]$. The equation of degree $R$ then reduces to
$\beta_k\alpha_k^R=0$, so $\beta_k=0$. This proves the claim at rank $k$, and thus for all
$k\in\NN^*$ by induction. The truncated features are thus linearly independent, hence so are the
$\phi[u_i]$, and $\dim\Span\{\phi[u_1],\dots,\phi[u_k]\}=k$.
\end{proof}

\begin{proof}[Proof of \Cref{prop:att_inv}]
  \begin{enumerate}[(i.)]
  \item The numerator and denominator are linear in $\phi[q]$, so any scalar cancels the ratio.
  \item The numerator and denominator are also linear  in the key features $(\phi [K_i])_{i\in[N]}$, so rescaling all of them by $\lambda$ multiplies both by $\lambda$.
  \item Translating all keys by $\delta$ multiplies each evaluation of the kernel by the single factor per-query $e^{\langle q,\delta\rangle/\sqrt d}$, which cancels in the ratio,
  \begin{equation}
  \Att(q,K+\delta,V)
  =\frac{e^{\langle q,\delta\rangle/\sqrt d}\exp\left(q^\T K^\T/\sqrt{d}\right) V} {e^{\langle q,\delta\rangle/\sqrt d}\exp\left(q^\T K^\T/\sqrt{d}\right)\mathbf 1_N}=\frac{\exp\left(q^\T K^\T/\sqrt{d}\right) V} {\exp\left(q^\T K^\T/\sqrt{d}\right)\mathbf 1_N}=\Att(q,K,V).
  \end{equation}
  \end{enumerate}
\end{proof}

\begin{proof}[Proof of \Cref{prop:gauss_inv}]
  \(\breve\kappa(x-\delta,y-\delta)=\exp(-\|x-y\|^2/2\sqrt d)=\breve\kappa(x,y)\), and the second identity corresponds to applying equation \eqref{eq:gauss_gram_rewrite} to the translated matrices.
\end{proof}

\begin{proof}[Proof of \Cref{prop:translated_interp}]
  The interpolated attention can be written as
  \[
  \widetilde\Att_{L-\delta}(q,K-\delta,V)
  =\frac{E_{q,L-\delta}E^{\dagger}_{L-\delta,L-\delta}E_{L-\delta,K-\delta}V}
        {E_{q,L-\delta}E^{\dagger}_{L-\delta,L-\delta}E_{L-\delta,K-\delta}\mathbf 1_N}.
\]
Using \Cref{prop:gauss_inv} and equation~\eqref{eq:gram_translation}, we obtain
  \[
  \widetilde\Att_{L-\delta}(q,K-\delta,V)
  =\frac{\Xi_{q,\delta}^{-1}}{\Xi_{q,\delta}^{-1}}
  \frac{E_{q,L}\Xi_{L,\delta}E^{\dagger}_{L,L}\Xi_{L,\delta}\Xi_{L,\delta}^{-1}E_{L,K}\Xi^{-1}_{K,\delta}V}
       {E_{q,L}\Xi_{L,\delta}E^{\dagger}_{L,L}\Xi_{L,\delta}\Xi_{L,\delta}^{-1}E_{L,K}\Xi_{K,\delta}^{-1}\mathbf 1_N}.
\]
Simplifying the $\Xi$ terms then gives us
\[
  \widetilde\Att_{L-\delta}(q,K-\delta,V)
  =\frac{E_{q,L}\Xi_{L,\delta}E^{\dagger}_{L,L}E_{L,K}\Xi_{K,\delta}^{-1}V}
        {E_{q,L}\Xi_{L,\delta}E^{\dagger}_{L,L}E_{L,K}\Xi_{K,\delta}^{-1}\mathbf 1_N}
  =\frac{E_{q+\delta,L}E^{\dagger}_{L,L}E_{L-\delta,K}V}
        {E_{q+\delta,L}E^{\dagger}_{L,L}E_{L-\delta,K}\mathbf 1_N},
\]
where the first equality gives us the identity of the proposition.
For the $i$-th row $\ell_\delta$ of $L-\delta$, $E_{\ell_\delta+\delta,L}E^{\dagger}_{L,L}=e_i^\T$ and the equation simplifies to
\[
\widetilde\Att_{L-\delta}(\ell-\delta,K-\delta,V)
  =\frac{E_{\ell-\delta,K}V}
        {E_{\ell-\delta,K}\mathbf 1_N}=\Att(\ell-\delta,K,V),
\]
which shows that the interpolation is exact at the translated nodes, whose features are, by \Cref{prop:intersection}, the only points of $\phi[\RR^d]$ in $\mathcal B_\delta$.
\end{proof}

\begin{proof}[Proof of \Cref{prop:discrete_scores}]
Using the identities of \eqref{eq:gauss_gram_rewrite}, we have $E_{LL}^{\dagger}=D_{L_{\phi}}^{-1/2}\breve E_{LL}^{\dagger}D_{L_{\phi}}^{-1/2}$ and $E_{LK}=D_{L_{\phi}}^{1/2}\breve E_{LK}D_{K_{\phi}}^{1/2}=E_{KL}^\T$.
Applying this to each block of the Schur complement
$S_L:=E_{KK}-E_{KL}E_{LL}^{\dagger}E_{LK}$, the $D_{L_\phi}$ terms cancel within $E_{KL}E_{LL}^{\dagger}E_{LK}$ and thus $S_L=D_{K_\phi}^{1/2} \breve S_L D_{K_\phi}^{1/2}$. Hence, using $E_{KL}\Waug=E_{KL}E_{LL}^{\dagger}E_{LK}\Vaug$ and
$\lVert\phi [K_i]\rVert_H=\kappa_{\exp}(K_i,K_i)^{1/2}=D_{K_\phi}^{1/2}[i,i]$, we can write
\[
\breve K_\phi M_\Delta
=D_K^{-1/2}\big(E_{KK}\Vaug-E_{KL}\Waug\big)
=D_K^{-1/2}S_L\Vaug
=\breve S_L D_K^{1/2} \Vaug,
\]
dividing both sides by $\lVert D_K^{1/2}\rVert_\infty$, gives us the formula.
The identity $\breve S_L=\breve K_\phi\Pi_{K_\phi\ominus L_\phi}\breve K_\phi^{\T}$
follows from $\Pi_{K_\phi\ominus L_\phi}K_\phi^{\T}
=K_\phi^{\T}-L_\phi^{\T}E_{LL}^{\dagger}E_{LK}$, and the positive
semi-definiteness from $\Pi_{K_\phi\ominus L_\phi}$ being an orthogonal
projector.
\end{proof}

\section{WildCat~\cite{schroderWildCatNearLinearAttention2026} temperature scaling as an oblique projection} \label{app:wilcat_proj}

WildCat introduces an alternative rescaled scalar product in the RKHS given by $\kappa_\tau(x,y)=\exp\left(\langle\frac{x}{\tau},\frac{y}{\tau}\rangle/\sqrt{d}\right)$ for a temperature $\tau>0$. The selection and projection steps, of $L$ and $W$ in our formulation, of WildCat, then consist in considering the orthogonal projection of the linear map in this metric. In this subsection, we provide an equivalent formulation in terms of an oblique projection in the original kernel, allowing us to represent the WildCat temperature method as an oblique Petrov-Galerkin projection.
Let $\alpha\in\RR\setminus\{0\}$ and for any $A\in\RR^{a\times d}$, set $A_\phi^{\alpha}=\left(\phi[\alpha A_i]\right)_{i\in[a]}$ and $\mathfrak D_\alpha=(\alpha^{r}I_{r})_{r\in\NN}$.
We then have, for any $x\in\RR^{d}$, $\phi[\alpha x]=\mathfrak D_\alpha\phi[x]$, hence $A_\phi^{\alpha}=A_\phi \mathfrak D_\alpha$. We note that $\mathfrak D_\alpha$ is unbounded on $H$ for $|\alpha|>1$, so that, for $\tau\neq1$, either $\mathfrak D_\tau$ or its inverse is unbounded. The identities below are thus to be understood on the finite spans of feature vectors, the final expression of $P_\tau$ being bounded and used as its definition. Let
$$\Pi_{L_\phi^\alpha}={L_\phi^\alpha}^{\T}E_{\alpha L,\alpha L}^{-1}L_\phi^\alpha$$
be the orthogonal projector onto $\Span(L_\phi^\alpha)$.
Then the WildCat $\tau$ projector is given by the oblique projector:
$$P_\tau:=\mathfrak D_{\tau}\Pi_{L_{\phi}^{\tau^{-1}}}\mathfrak D_{\tau}^{-1}=L_\phi^\T E_{\tau^{-2}L,L}^{-1}L^{\tau^{-2}}_\phi,$$
and the set of exact interpolation nodes is given by $\mathcal B_{L_\phi^{\tau^{-2}}}$.
In our Petrov-Galerkin formulation, this amounts to a trial space given by $\Span(L_\phi)$ and a test space given by $\Span(L_\phi^{\tau^{-2}})$. Thanks to the obliqueness of the projector, it extrapolates differently than the orthogonal projector and as shown in \cite{schroderWildCatNearLinearAttention2026} its error can be bounded under reasonable assumptions.
For the weights, this gives us,
$$W=E_{L,\tau^{-2}L}^{-1}E_{L,\tau^{-2}K}V.$$
In the Gaussian reformulation, we have, for any $A\in\RR^{a\times d}$, $B\in\RR^{b\times d}$ and any $\alpha,\beta\in\RR$,
$$E_{\alpha A,\beta B}=D^{\alpha^2/2}_A\breve E_{\alpha A,\beta B}D_{B}^{\beta^2/2}.$$
Using this, the fact that  $E_{\tau^{-2}L,L}=E_{\tau^{-1}L,\tau^{-1}L}$ and that, for any $\alpha\in\RR$, $\breve E_{\alpha A,\alpha B}=(\breve E_{AB}[i,j]^{\alpha^2})_{i\in [a],j\in [b]}$, we obtain that
\begin{equation}W=D_L^{-\tau^{-2}/2}\breve E_{\tau^{-1}L,\tau^{-1}L}^{-1}\breve E_{\tau^{-1}L,\tau^{-1}K}\,D_K^{\tau^{-2}/2}\,V
\end{equation}
In the $\eta$ gauge with centering,  the two-step solve \eqref{eq:stable_galerkin} then becomes
\begin{equation}
\breve E_{\tau^{-1}L,\tau ^{-1}L}X=\breve E_{\tau^{-1}L,\tau ^{-1}K}\underbrace{\Diag\big(\eta^{(c)}_{K}\big)^{\tau^{-2}}\Vaug}_{=:\Vaug_{\eta,\tau}},
\qquad
{\Waug}^{(c)}=\Diag\big(\eta^{(c)}_{L}\big)^{-\tau^{-2}}X,
\end{equation}
while the readout \eqref{eq:stable_readout} is unchanged.

\section{Additional figures}
\label{app:additional_figures}
Figures~\ref{fig:resid_exact_gh_iqr}--\ref{fig:proj_qdaim_exact_iqr} repeat Figures~\ref{fig:resid_exact_gh}, \ref{fig:resid_exact}, \ref{fig:proj_gaim_exact} and~\ref{fig:proj_qdaim_exact} with, for every curve, the interquartile range over the $112$ query heads in dashed lines of the same colour. 
For each head, the error is the median over the future queries of~\eqref{eq:rel_att_err}, as in the main text. 
Reference levels lying far above the curves are not drawn and their values are given in the legend.

\begin{figure}[!htbp]
  \centering
  \includestandalone{figures/tikz/fig_resid_exact_gh_iqr}
  \caption{Heavy hitter relative error~\eqref{eq:rel_att_err} in numerator (top, $\varepsilon_g^{10\%}$) and denominator (bottom, $\varepsilon_h^{10\%}$) on the future queries, with $\tsink=32$ and without range correction, as in Figure~\ref{fig:resid_exact_gh}.
  Solid: median over the $112$ query heads; dashed: interquartile range over the heads.}
  \label{fig:resid_exact_gh_iqr}
\end{figure}

\begin{figure}[!htbp]
  \centering
  \includestandalone{figures/tikz/convergence_exact_linf_sink32_iqr}
  \caption{Heavy hitter relative attention error $\varepsilon^{10\%}_{\Att}$ on the future queries without (top) and with (bottom) range correction, with $\tsink=32$ and $\trec=0$, as in Figure~\ref{fig:resid_exact}.
  Solid: median over the $112$ query heads; dashed: interquartile range over the heads.}
  \label{fig:resid_exact_iqr}
\end{figure}

\begin{figure}[!htbp]
  \centering
  \includestandalone{figures/tikz/projections_gaim_exact_v4_linf_sink32_iqr}
  \caption{Heavy hitter relative attention error $\varepsilon^{10\%}_{\Att}$ using GAIM exact v4 interpolation nodes for the projection methods of Table~\ref{table:projections} and without projection ($W=V[\mathfrak L,:]$), as in Figure~\ref{fig:proj_gaim_exact}.
  Solid: median over the $112$ query heads; dashed: interquartile range over the heads.}
  \label{fig:proj_gaim_exact_iqr}
\end{figure}

\begin{figure}[!htbp]
  \centering
  \includestandalone{figures/tikz/projections_qdaim_exact_v5_linf_sink32_iqr}
  \caption{Heavy hitter relative attention error $\varepsilon^{10\%}_{\Att}$ using Q-DAIM exact v5 interpolation nodes for the projection methods of Table~\ref{table:projections} and without projection ($W=V[\mathfrak L,:]$), as in Figure~\ref{fig:proj_qdaim_exact}.
  Solid: median over the $112$ query heads; dashed: interquartile range over the heads.}
  \label{fig:proj_qdaim_exact_iqr}
\end{figure}

Figures~\ref{fig:proj_operator_gaim} and~\ref{fig:proj_operator_qdaim} give, for the same projections, the residual operator they leave on the keys, i.e. the relative squared norm $\|M_\eta-\breve L_\phi^\T X\|_H^2/\|M_\eta\|_H^2$ of the augmented operator and of its denominator part, computed from $\breve E_{KK}$, $\breve E_{LK}$ and $\breve E_{LL}$ on the context keys, the sink excluded. The GAIM exact nodes are those of the v1 variant: with the v4 nodes of Figure~\ref{fig:proj_gaim_exact}, every projection leaves the whole operator, about $0.9999$ of its norm at $n=512$. The Galerkin projection, the orthogonal projection in $H$, gives the smallest residual by construction, and the least-squares projection and the values of the selected keys, $W=V[\mathfrak L,:]$, are within $0.1\%$ of it. The revealed-range projection is larger and peaks around $n=129$, where the number of nodes reaches the revealed rank. The WildCat temperature projection leaves the largest residual, above the norm of the denominator operator itself for small $n$, although it gives the lowest attention error at the future queries in Figures~\ref{fig:proj_gaim_exact} and~\ref{fig:proj_qdaim_exact}.

\begin{figure}[!htbp]
  \centering
  \includestandalone{figures/tikz/projections_operator_gaim_exact_v1_linf_sink32}
  \caption{Relative squared norm of the residual operator on the keys left by each projection of Table~\ref{table:projections}, and without projection ($W=V[\mathfrak L,:]$), on GAIM exact v1 interpolation nodes ($L^\infty$ score), $\tsink=32$. Left: augmented operator; right: denominator. Median over the $16$ KV heads.}
  \label{fig:proj_operator_gaim}
\end{figure}

\begin{figure}[!htbp]
  \centering
  \includestandalone{figures/tikz/projections_operator_qdaim_exact_v5_linf_sink32}
  \caption{Relative squared norm of the residual operator on the keys left by each projection of Table~\ref{table:projections}, and without projection ($W=V[\mathfrak L,:]$), on the Q-DAIM exact v5 interpolation nodes of Figure~\ref{fig:proj_qdaim_exact}, $\tsink=32$. Left: augmented operator; right: denominator. Median over the $16$ KV heads.}
  \label{fig:proj_operator_qdaim}
\end{figure}

\section{Conditioning of the key Gram matrices}
\label{app:conditioning}
In this appendix, we illustrate the conditioning argument of \S\ref{sec:stab} on the document used in \S\ref{sec:resid_setup}. Figure~\ref{fig:gram_conditioning} shows, for each layer of the model and each of its four KV heads, the condition number $\kappa_2$ of the Gram matrix of the context keys, $E_{KK}$, and of the Gram matrix of the $n=512$ interpolation nodes, $E_{LL}$. In both cases the $\tsink=32$ sink keys are excluded, and the context consists of the first $4096$ tokens of the document, i.e. $N=4064$ keys once the sinks are removed, rather than the $4096$ tokens following the sinks used in \S\ref{sec:resid_setup}. The interpolation nodes are selected either by GAIM exact, using the v2 residual in $L^2$ norm, or by Q-DAIM exact, where the nodes past the revealed rank are chosen as described in Appendix~\ref{app:qdaim_restart}.

For each of these sets of keys, we compare three Gram matrices. The first one is the Gaussian Gram matrix $\breve E$ which, by Proposition~\ref{prop:gauss_inv}, does not depend on the centering of the keys. The two others correspond to the exponential Gram matrix $E=D^{1/2}\breve E D^{1/2}$ of \eqref{eq:gauss_gram_rewrite}, computed either with the keys taken at the origin or with the keys centered on their mean. The condition numbers are obtained from the eigenvalues computed in double precision, which only allows us to resolve $\kappa_2$ up to $1/(n\epsilon)$. Beyond this level, we instead give the lower bound
\[
\kappa(D)=\frac{\max_i E_{ii}}{\min_i E_{ii}}=\exp\Big(\frac{\max_i\|K_i\|^2-\min_i\|K_i\|^2}{\sqrt d}\Big),
\]
which can be computed exactly as it only involves the diagonal of $E$, together with the upper bound $\kappa(D)\,\kappa(\breve E)$.

Over the $112$ KV heads of the model, the Gaussian Gram matrix of the context keys remains well conditioned, with a median $\log_{10}\kappa_2$ of $0.41$ and a maximum of $4.1$, reached in layer $0$. The Gaussian Gram matrices of the $512$ interpolation nodes all stay below $10^{2.4}$, with medians of $10^{0.14}$ for Q-DAIM and $10^{0.33}$ for GAIM exact. On the other hand, the exponential Gram matrix of the keys taken at the origin is beyond the double-precision resolution on $105$ of the $112$ heads, its diagonal bound having a median of $10^{16.8}$ and reaching $10^{354}$ in layer $27$. Centering the keys on their mean lowers these values to a median of about $10^{11.5}$ and to at most $10^{34}$, although $47$ heads still remain beyond the resolution. For the $512$ interpolation nodes, the centered exponential Gram matrix has a median condition number of $10^{5.6}$ for Q-DAIM and $10^{11.0}$ for GAIM exact. As discussed in \S\ref{sec:stab}, the ill-conditioning of the interpolation problem is thus carried by the norm terms $D$, and the solves in the Gaussian formulation remain well conditioned in every head.

\begin{figure}[!htbp]
  \centering
  \includestandalone{figures/tikz/gram_conditioning_per_layer}
  \caption{Condition number $\kappa_2$ per layer of the Gram matrices of the context keys, $E_{KK}$ ($N=4064$, left), and of the $n=512$ interpolation nodes selected by GAIM exact (centre) and Q-DAIM exact (right), $E_{LL}$; $\tsink=32$ sink keys excluded. Blue: Gaussian Gram $\breve E$ (any centering); red: exponential Gram $E$ with the keys at the origin; orange: exponential Gram with the keys centred on their mean. Lines: median over the four KV heads of the layer; small marks: one KV head. Past the double-precision resolution $1/(n\epsilon)$ (dotted), hollow marks give the lower bound $\kappa(D)$ and the bars extend to $\kappa(D)\,\kappa(\breve E)$; triangles mark the heads beyond $10^{40}$, with the largest exponent of the layer.}
  \label{fig:gram_conditioning}
\end{figure}

\section{Q-DAIM past the revealed rank}
\label{app:qdaim_restart}
The operator Gram matrix $G$ of \eqref{eq:qdeim_gram} is of size $(d+1)\times(d+1)$. Hence, the revealed range $U$ has rank $r\le d+1$ and the evaluation matrix $F$ of \eqref{eq:qdeim_F} has $r$ columns. As the column-pivoted QR of Algorithm~\ref{alg:qdaim} deflates the rows of $F$ in these $r$ coordinates, once $r$ pivots have been selected, the remaining rows then give zero up to numerical error and any further pivot is thus chosen on numerical noise.
For the document of \S\ref{sec:resid_setup}, we observe that the norm of the pivot drops by about fifteen orders of magnitude after the $129$-th selected node, and even earlier in heads of lower numerical rank, for example after $77$ nodes in a head of layer $0$ and after $10$ nodes in a head of layer $14$.
However, the nodes beyond $r$ remain useful for the interpolation. Indeed, each direction of the revealed range is in general a combination of the features of many keys. We thus select these additional nodes by restarting the selection, in one of the two following ways.

\paragraph{Q-DAIM exact.} In this case, the range $U=\breve K_\phi^\T C$ is revealed only once and the selection proceeds in blocks. Within each block, the nodes are given by the pivots of the column-pivoted QR of $F^\T$, so that the first block coincides with Algorithm~\ref{alg:qdaim}. Once the residual of the block collapses, i.e. once the largest remaining squared row norm falls below $10^{-12}$ times that of the first pivot of the block, we replace $F$ by
\[
F_L=\breve K_\phi\,\Pi^\perp_{L_\phi}\,U=\breve S_L\,C,
\]
where $\breve S_L$ is the projection of the Gram matrix $E_{KK}$ onto the orthogonal complement of $\Span(L_\phi)$ of Proposition~\ref{prop:discrete_scores}. The rows of $F_L$ are the coordinates, in the revealed range, of the key features orthogonalized in $H$ against all the selected nodes, and a new block is started by applying the column-pivoted QR to $F_L^\T$. Furthermore, $F_L$ can be updated at a cost $O(Nr)$ per selected node. Indeed, setting $\sigma=\breve S_L[i,i]$ and $g=\breve S_L[:,i]/\sqrt\sigma$ for the selected node $i$, we have $F_L\leftarrow F_L-g\,F_L[i,:]/\sqrt{\sigma}$, where the columns $g$ form the partial Cholesky factor of $\breve E_{KK}$, which allows us to obtain the next column of $\breve S_L$ at a cost $O(Nn)$.

\paragraph{Q-DAIM RFF.} For the surrogate selection of \S\ref{sec:surrogate}, we instead proceed in rounds of $d$ nodes. After each round, the coefficients of the selected nodes are computed using the exact kernel, $X=\breve E_{LL}^{-1}\breve E_{LK}\Vaug_\eta$. The next round then reveals the range of, and selects the pivots on, the surrogate residual operator $M_{\breve\psi}-\breve L_{\breve\psi}^\T X$ instead of $M_{\breve\psi}$, so that each round has up to $d+1$ new directions to reveal.

Figure~\ref{fig:qdaim_restart} compares both restart procedures with the single pass of Algorithm~\ref{alg:qdaim}, for the v1 and v5 variants. The curves coincide up to about $129$ nodes, after which the restarts lower the norm of the residual operator on the keys. At $n=512$, this norm decreases from $8.9\times10^{-3}$ (v1) and $4.3\times10^{-3}$ (v5) to $3.8\times10^{-3}$ for Q-DAIM exact, and from $1.05\times10^{-2}$ (v1) and $1.47\times10^{-2}$ (v5) to $3.9\times10^{-3}$ for Q-DAIM RFF, which then matches the exact selector.
The attention error on the future queries, on the other hand, changes little. With range correction, the restarted selections and the single pass of Q-DAIM exact reach $0.147$--$0.148$, compared to $0.144$ for the single pass of Q-DAIM RFF. Without range correction, the restarted selections reach $0.184$--$0.187$, compared to $0.178$--$0.188$ for the single pass of Q-DAIM exact and $0.176$--$0.177$ for that of Q-DAIM RFF. As observed in \S\ref{sec:results}, a lower residual on the keys therefore does not necessarily translate into a lower error on the future queries.

\begin{figure}[!htbp]
  \centering
  \includestandalone{figures/tikz/qdaim_restart_comparison}
  \caption{Q-DAIM exact (left) and Q-DAIM RFF (right) past the revealed rank, single pass of Algorithm~\ref{alg:qdaim} (dashed) against the restarted selections of Appendix~\ref{app:qdaim_restart} (solid), for the v1 and v5 targets. Top: relative squared norm of the augmented residual operator on the keys, without sink, as in Figure~\ref{fig:resid_complete_context}. Middle and bottom: heavy hitter relative attention error $\varepsilon^{10\%}_{\Att}$ on the future queries with $\tsink=32$, without and with range correction, as in Figure~\ref{fig:resid_exact}. Each curve is the median over the heads. Dotted vertical lines: restarts, every $d=128$ nodes for Q-DAIM RFF; for Q-DAIM exact, where at least half of the KV heads reset $F$, i.e. after $128$, $256$ and $384$ nodes for the v1 target and one node later per block for v5 (on $14$ of the $16$ KV heads; the other two, of lower numerical rank, reset more often).}
  \label{fig:qdaim_restart}
\end{figure}

\end{document}

%% file: figures/tikz/fig_resid_complete_context.tex
\begin{tikzpicture}
  \pgfplotsset{every axis/.append style={
      scale only axis, /tikz/font=\footnotesize, label style={align=center}, title style={align=center},
      legend cell align=left,
      legend style={font=\footnotesize, /tikz/every even column/.append style={column sep=7pt}}}}
  \pgfplotsset{
    v1/.style={blue, mark=*},
    v2/.style={red, mark=square*},
    v3/.style={cyan!60!black, mark=triangle*},
    v4/.style={orange!50!black, mark=diamond*},
    v5/.style={magenta!70!black, mark=pentagon*},
  }
  \pgfplotsset{
    context/.style={gray!40, thin, mark=none},
    mean value/.style={black, thin, dashdotted, mark=none},
    zero readout/.style={black, thin, densely dotted, mark=none},
    every axis plot/.append style={very thick, mark size=1.8pt, mark options={solid}, mark repeat=8, forget plot},
    legend image code/.code={\draw[very thick, mark size=1.8pt, mark options={solid}, mark repeat=2, mark phase=2, #1]
      plot coordinates {(0pt,0pt) (10pt,0pt) (20pt,0pt)};},
  }
  \pgfplotstableread[row sep=\\]{
    n gaim_exact_v1 gaim_exact_v2 gaim_exact_v3 gaim_exact_v4 gaim_exact_v5 gaim_mp_v1 gaim_mp_v2 gaim_mp_v3 gaim_mp_v4 gaim_mp_v5 gaim_omp_v1 gaim_omp_v2 gaim_omp_v3 gaim_omp_v4 gaim_omp_v5 qdaim_exact_v1 qdaim_exact_v5 qdaim_rff_v1 qdaim_rff_v5\\
    1 0.47703 0.93601 0.93601 1 0.47703 0.78622 1 1 1 0.82835 0.78622 1 1 1 0.82835 0.61627 0.61627 0.97563 0.97563\\
    17 0.05876 0.86605 0.90199 1 0.05876 0.33825 1 1 1 0.34189 0.12018 1 1 1 0.12242 0.083694 0.083694 0.084589 0.084589\\
    33 0.032796 0.84622 0.87429 1 0.032829 0.26976 0.99999 0.99999 0.99999 0.28109 0.092493 0.99999 1 0.99999 0.10233 0.041902 0.041902 0.041124 0.041124\\
    49 0.023115 0.84246 0.87038 1 0.023063 0.26951 0.99997 0.99997 0.99997 0.22232 0.079187 0.99998 1 0.99998 0.068547 0.03561 0.030647 0.035452 0.035566\\
    65 0.017984 0.84058 0.86278 1 0.017984 0.1966 0.99995 0.99995 0.99995 0.22229 0.056999 0.99996 0.99997 0.99996 0.061985 0.025152 0.025289 0.033553 0.033591\\
    81 0.014601 0.8402 0.86059 0.99999 0.014594 0.19069 0.99995 0.99995 0.99995 0.21638 0.042065 0.99994 0.99995 0.99994 0.048871 0.02112 0.02112 0.0211 0.03118\\
    97 0.012116 0.83671 0.81855 0.99999 0.012036 0.17779 0.9999 0.9999 0.9999 0.21623 0.039557 0.99994 0.99994 0.99994 0.041894 0.01999 0.01999 0.02003 0.025492\\
    113 0.009993 0.81372 0.81126 0.99999 0.0099232 0.1775 0.99989 0.99989 0.99989 0.20282 0.037991 0.99993 0.99994 0.99993 0.040571 0.018887 0.018934 0.019021 0.024489\\
    129 0.0087396 0.81364 0.80062 0.99999 0.0084969 0.17594 0.99988 0.99988 0.99988 0.18916 0.036953 0.99991 0.99994 0.99991 0.039676 0.018019 0.017734 0.01807 0.023674\\
    145 0.007666 0.81288 0.79591 0.99999 0.0073949 0.1759 0.99975 0.99975 0.99985 0.18912 0.035961 0.99989 0.99993 0.9999 0.03939 0.017226 0.015881 0.017197 0.023175\\
    161 0.0066349 0.77788 0.79499 0.99999 0.0064441 0.17589 0.99973 0.99973 0.99981 0.1885 0.035808 0.99987 0.99993 0.99989 0.038909 0.016426 0.015364 0.016519 0.021149\\
    177 0.0058078 0.77692 0.79309 0.99999 0.0056427 0.1662 0.99972 0.99972 0.9998 0.16556 0.03488 0.99987 0.99993 0.99989 0.037589 0.014725 0.014758 0.01593 0.019889\\
    193 0.0050858 0.77375 0.79039 0.99999 0.0050305 0.16345 0.99971 0.99971 0.99972 0.15748 0.034139 0.99986 0.99993 0.99987 0.037215 0.014228 0.013889 0.015471 0.019214\\
    209 0.0046862 0.77335 0.78756 0.99999 0.0045741 0.14919 0.9997 0.9997 0.9997 0.13129 0.03362 0.99985 0.99993 0.99986 0.036549 0.013912 0.013466 0.013964 0.018687\\
    225 0.0043277 0.7727 0.78172 0.99999 0.0042474 0.14682 0.99969 0.99969 0.99969 0.12573 0.031104 0.99985 0.99992 0.99986 0.035505 0.013527 0.013068 0.013621 0.018353\\
    241 0.0039182 0.59561 0.77732 0.99999 0.0038988 0.14555 0.99967 0.99967 0.99967 0.12548 0.030994 0.99984 0.99992 0.99984 0.034925 0.013218 0.012745 0.013388 0.018036\\
    257 0.0036181 0.59278 0.77653 0.99998 0.0035705 0.14436 0.99967 0.99965 0.99967 0.12019 0.029725 0.99984 0.99989 0.99984 0.030643 0.012871 0.012476 0.013121 0.0178\\
    273 0.0033447 0.56775 0.77264 0.99998 0.0033102 0.12821 0.99965 0.99965 0.99965 0.11558 0.029383 0.99983 0.99986 0.99983 0.03036 0.012606 0.012196 0.012857 0.017639\\
    289 0.0031346 0.5543 0.76974 0.99998 0.0030785 0.12151 0.99962 0.99964 0.99962 0.1142 0.029014 0.99983 0.99986 0.99983 0.029967 0.012349 0.011993 0.012569 0.017408\\
    305 0.0029402 0.55413 0.76282 0.99998 0.0028721 0.11218 0.99962 0.99962 0.99962 0.11398 0.028908 0.99982 0.99986 0.99982 0.029738 0.012159 0.0070398 0.012387 0.017187\\
    321 0.0027154 0.55371 0.76255 0.99998 0.0026861 0.11213 0.9996 0.99958 0.9996 0.11312 0.028765 0.9998 0.99986 0.9998 0.029537 0.011898 0.0068253 0.012186 0.017075\\
    337 0.0025232 0.5529 0.76188 0.99998 0.0025198 0.11212 0.9996 0.99956 0.9996 0.11303 0.028316 0.99979 0.99986 0.99979 0.029449 0.011689 0.0066441 0.011998 0.016911\\
    353 0.0023419 0.43064 0.73819 0.99998 0.0023458 0.11156 0.99956 0.99955 0.99956 0.11299 0.027321 0.99979 0.99977 0.99979 0.027004 0.011535 0.0064708 0.011826 0.016745\\
    369 0.0021815 0.42464 0.73814 0.99998 0.0021782 0.11153 0.99954 0.99953 0.99954 0.11278 0.027079 0.99978 0.99967 0.99978 0.026695 0.011348 0.0062755 0.011678 0.016587\\
    385 0.0020343 0.40276 0.73657 0.99997 0.0020322 0.11118 0.99953 0.99952 0.99953 0.11271 0.026132 0.99978 0.99965 0.99978 0.025341 0.010513 0.0060779 0.011524 0.01643\\
    401 0.0018954 0.38762 0.73648 0.99997 0.0018954 0.11117 0.99952 0.99951 0.99952 0.10995 0.025154 0.99977 0.99964 0.99977 0.025155 0.010295 0.0057673 0.011354 0.016255\\
    417 0.0017815 0.38754 0.73623 0.99997 0.0017794 0.10899 0.99948 0.9995 0.99948 0.10951 0.024391 0.9997 0.99964 0.9997 0.023494 0.010153 0.0055893 0.011185 0.016153\\
    433 0.001668 0.38034 0.73611 0.99997 0.0016686 0.10898 0.99948 0.9995 0.99948 0.10946 0.023183 0.9997 0.9996 0.9997 0.02231 0.0096187 0.0054392 0.011024 0.016015\\
    449 0.0015671 0.37885 0.73083 0.99997 0.0015697 0.10894 0.99944 0.99946 0.99944 0.10894 0.022426 0.99967 0.99959 0.99967 0.021564 0.009445 0.0052633 0.010858 0.015709\\
    465 0.0014762 0.37882 0.73066 0.99997 0.0014787 0.10542 0.99943 0.99946 0.99943 0.10892 0.022035 0.99967 0.99958 0.99967 0.021228 0.0093117 0.0047193 0.010774 0.015568\\
    481 0.0013929 0.37693 0.72109 0.99997 0.0013975 0.10539 0.99942 0.99944 0.99942 0.089927 0.021034 0.99965 0.99956 0.99965 0.020809 0.0091814 0.0045852 0.01068 0.015417\\
    497 0.0013111 0.37535 0.71524 0.99996 0.0013144 0.10536 0.99942 0.99944 0.99942 0.089848 0.017923 0.99965 0.99951 0.99965 0.017784 0.0090361 0.0044365 0.010581 0.015036\\
    512 0.0012407 0.37504 0.71472 0.99996 0.0012424 0.1053 0.99941 0.99943 0.99941 0.088981 0.017187 0.99957 0.9995 0.99957 0.017512 0.00891 0.0043187 0.010518 0.014733\\
  }\curvesa
  \pgfplotstableread[row sep=\\]{
    n gaim_exact_v1 gaim_exact_v2 gaim_exact_v3 gaim_exact_v4 gaim_exact_v5 gaim_mp_v1 gaim_mp_v2 gaim_mp_v3 gaim_mp_v4 gaim_mp_v5 gaim_omp_v1 gaim_omp_v2 gaim_omp_v3 gaim_omp_v4 gaim_omp_v5 qdaim_exact_v1 qdaim_exact_v5 qdaim_rff_v1 qdaim_rff_v5\\
    1 0.57711 0.9326 0.9326 1 0.57711 0.80221 1 1 1 0.80306 0.80221 1 1 1 0.80306 0.63792 0.63792 0.9758 0.9758\\
    17 0.074342 0.88972 0.91001 1 0.074342 0.38153 1 1 1 0.38389 0.12981 1 0.99999 1 0.13021 0.094057 0.094057 0.094845 0.094845\\
    33 0.033252 0.85172 0.87914 1 0.031486 0.32284 0.99999 0.99999 0.99999 0.33578 0.098924 0.99999 0.99999 0.99999 0.097758 0.058829 0.058829 0.058312 0.058312\\
    49 0.019754 0.84846 0.87181 1 0.019675 0.30048 0.99998 0.99998 0.99998 0.24763 0.085723 0.99999 0.99999 0.99999 0.075977 0.049642 0.045756 0.049271 0.049558\\
    65 0.01564 0.84561 0.86495 0.99999 0.01564 0.2318 0.99997 0.99997 0.99997 0.24759 0.069607 0.99998 0.99998 0.99998 0.067659 0.039104 0.039243 0.046304 0.046354\\
    81 0.012606 0.84527 0.86411 0.99999 0.012306 0.22687 0.99997 0.99997 0.99997 0.24265 0.057432 0.99997 0.99997 0.99997 0.060717 0.032846 0.032493 0.032695 0.042585\\
    97 0.010065 0.83266 0.82133 0.99999 0.010065 0.21956 0.99994 0.99994 0.99994 0.24254 0.054955 0.99997 0.99997 0.99997 0.053452 0.030504 0.030504 0.030525 0.035985\\
    113 0.0086317 0.80932 0.81898 0.99999 0.0085842 0.21915 0.99994 0.99994 0.99994 0.2073 0.053369 0.99996 0.99996 0.99996 0.053043 0.028784 0.028839 0.028932 0.034396\\
    129 0.0076284 0.8084 0.80718 0.99999 0.0075727 0.21707 0.99993 0.99993 0.99993 0.19886 0.052269 0.99995 0.99995 0.99995 0.051817 0.027515 0.027007 0.027317 0.033155\\
    145 0.0066922 0.80415 0.80514 0.99999 0.0066045 0.21703 0.99981 0.99981 0.99981 0.19881 0.051264 0.99994 0.99995 0.99994 0.049504 0.026143 0.023722 0.025937 0.032271\\
    161 0.0060352 0.76485 0.79932 0.99999 0.0059428 0.21702 0.9998 0.9998 0.9998 0.1924 0.05107 0.99993 0.99995 0.99993 0.049049 0.024867 0.022887 0.024873 0.02885\\
    177 0.0052728 0.76315 0.79586 0.99999 0.0052464 0.19828 0.99979 0.99979 0.99979 0.18099 0.050042 0.99993 0.99995 0.99993 0.047894 0.021953 0.0218 0.023981 0.027358\\
    193 0.0046951 0.76237 0.79126 0.99999 0.0046523 0.19488 0.99978 0.99979 0.99978 0.17015 0.049038 0.99992 0.99995 0.99992 0.047223 0.020963 0.020651 0.023196 0.025986\\
    209 0.0042405 0.76176 0.78898 0.99999 0.004167 0.18134 0.99976 0.99978 0.99976 0.14668 0.048453 0.99992 0.99995 0.99992 0.046664 0.020387 0.019937 0.020599 0.025093\\
    225 0.0038641 0.76146 0.78293 0.99999 0.0037969 0.17781 0.99976 0.99976 0.99976 0.1396 0.045765 0.99992 0.99995 0.99992 0.045246 0.019763 0.019224 0.019993 0.024512\\
    241 0.0034882 0.70065 0.77933 0.99999 0.0034436 0.17665 0.99974 0.99974 0.99974 0.13939 0.045615 0.99991 0.99995 0.99991 0.044217 0.019234 0.018682 0.019607 0.024014\\
    257 0.0031957 0.69698 0.77876 0.99999 0.003139 0.17557 0.99974 0.99972 0.99974 0.1368 0.043714 0.99991 0.99989 0.99991 0.04332 0.018648 0.018212 0.01918 0.023582\\
    273 0.0029191 0.69415 0.77578 0.99999 0.0028828 0.13694 0.99972 0.99971 0.99972 0.13541 0.043268 0.99991 0.99989 0.99991 0.042579 0.018141 0.017702 0.018643 0.023303\\
    289 0.0026395 0.67988 0.77294 0.99999 0.0026164 0.12777 0.99969 0.99971 0.99969 0.13422 0.042871 0.9999 0.99989 0.9999 0.042098 0.017725 0.017365 0.017974 0.022853\\
    305 0.0024469 0.67936 0.76808 0.99999 0.002429 0.12695 0.99969 0.9997 0.99969 0.13405 0.042729 0.9999 0.99989 0.9999 0.041855 0.017364 0.012186 0.017661 0.022481\\
    321 0.0022746 0.67879 0.76781 0.99998 0.0022585 0.12683 0.99968 0.99967 0.99968 0.13349 0.042536 0.99983 0.99988 0.99983 0.041224 0.016913 0.011789 0.017316 0.022273\\
    337 0.0021362 0.67765 0.76703 0.99998 0.0021142 0.12683 0.99968 0.99967 0.99968 0.13346 0.032825 0.99983 0.99988 0.99983 0.029971 0.016522 0.011475 0.016977 0.021988\\
    353 0.0019877 0.53616 0.70232 0.99998 0.0019787 0.1264 0.99967 0.99967 0.99967 0.13338 0.026344 0.99983 0.99988 0.99983 0.023612 0.016254 0.011169 0.016665 0.021694\\
    369 0.0018575 0.53126 0.70121 0.99998 0.0018447 0.12639 0.99965 0.99966 0.99965 0.13317 0.026111 0.99982 0.99984 0.99982 0.023268 0.015949 0.010785 0.016413 0.021453\\
    385 0.0017369 0.41802 0.69781 0.99998 0.001728 0.12618 0.99965 0.99965 0.99965 0.13307 0.024874 0.99982 0.99971 0.99982 0.022079 0.014906 0.010463 0.016124 0.021142\\
    401 0.0016312 0.40658 0.69708 0.99998 0.0016312 0.12609 0.99965 0.99965 0.99965 0.13103 0.024724 0.99982 0.99971 0.99982 0.021931 0.014439 0.010027 0.015804 0.02076\\
    417 0.0015512 0.40649 0.69562 0.99998 0.0015373 0.12598 0.99965 0.99965 0.99965 0.13071 0.023541 0.99977 0.9997 0.99977 0.020704 0.014189 0.0096484 0.015511 0.020572\\
    433 0.0014647 0.40049 0.69471 0.99998 0.0014562 0.12596 0.99965 0.99965 0.99965 0.13066 0.022419 0.99977 0.99968 0.99977 0.019602 0.013531 0.0093887 0.015178 0.020354\\
    449 0.0013869 0.39899 0.69258 0.99998 0.0013794 0.12596 0.99955 0.99958 0.99955 0.13013 0.021903 0.99974 0.99968 0.99974 0.019113 0.013225 0.0091214 0.014879 0.019898\\
    465 0.0013189 0.39897 0.69216 0.99998 0.0013069 0.12202 0.99954 0.99957 0.99954 0.13012 0.021649 0.99974 0.99967 0.99974 0.0189 0.012988 0.0080961 0.01475 0.019637\\
    481 0.0012568 0.39705 0.68948 0.99997 0.0012411 0.12196 0.99953 0.99956 0.99953 0.1123 0.020391 0.99973 0.99965 0.99973 0.018619 0.012776 0.0080652 0.01458 0.01943\\
    497 0.0011908 0.39524 0.66936 0.99997 0.0011747 0.12189 0.99953 0.99956 0.99953 0.11226 0.017283 0.99973 0.99961 0.99973 0.01563 0.012522 0.0078675 0.014389 0.019203\\
    512 0.0011399 0.39498 0.66869 0.99997 0.0011136 0.12178 0.99952 0.99955 0.99952 0.11119 0.017125 0.99968 0.9996 0.99968 0.015455 0.012294 0.0076655 0.01428 0.018993\\
  }\curvesb
  \begin{groupplot}[
      group style={group size=5 by 2, horizontal sep=6pt, vertical sep=9pt,
        xticklabels at=edge bottom, yticklabels at=edge left},
      width=77pt, height=92pt, grid=major, tick label style={font=\scriptsize},
      xmin=-15, xmax=527, xtick={0,250,500}, ymode=log]
    \nextgroupplot[ymin=1e-3, ymax=1.4, title={GAIM exact}, ylabel={$\|M_{\Delta,\eta}\|_2^2\,/\,\|M_{\eta}\|_2^2$}, legend to name=legend-fig-resid-complete-context-curves, legend columns=-1]
      \addplot[context] table[x=n, y=gaim_mp_v1] {\curvesa};
      \addplot[context] table[x=n, y=gaim_mp_v2] {\curvesa};
      \addplot[context] table[x=n, y=gaim_mp_v3] {\curvesa};
      \addplot[context] table[x=n, y=gaim_mp_v4] {\curvesa};
      \addplot[context] table[x=n, y=gaim_mp_v5] {\curvesa};
      \addplot[context] table[x=n, y=gaim_omp_v1] {\curvesa};
      \addplot[context] table[x=n, y=gaim_omp_v2] {\curvesa};
      \addplot[context] table[x=n, y=gaim_omp_v3] {\curvesa};
      \addplot[context] table[x=n, y=gaim_omp_v4] {\curvesa};
      \addplot[context] table[x=n, y=gaim_omp_v5] {\curvesa};
      \addplot[context] table[x=n, y=qdaim_exact_v1] {\curvesa};
      \addplot[context] table[x=n, y=qdaim_exact_v5] {\curvesa};
      \addplot[context] table[x=n, y=qdaim_rff_v1] {\curvesa};
      \addplot[context] table[x=n, y=qdaim_rff_v5] {\curvesa};
      \addplot[v1, mark phase=2] table[x=n, y=gaim_exact_v1] {\curvesa};
      \addplot[v2, mark phase=4] table[x=n, y=gaim_exact_v2] {\curvesa};
      \addplot[v3, mark phase=6] table[x=n, y=gaim_exact_v3] {\curvesa};
      \addplot[v4, mark phase=8] table[x=n, y=gaim_exact_v4] {\curvesa};
      \addplot[v5, mark phase=3] table[x=n, y=gaim_exact_v5] {\curvesa};
      \addlegendimage{v1}\addlegendentry{v1}
      \addlegendimage{v2}\addlegendentry{v2}
      \addlegendimage{v3}\addlegendentry{v3}
      \addlegendimage{v4}\addlegendentry{v4}
      \addlegendimage{v5}\addlegendentry{v5}
      \addlegendimage{context}\addlegendentry{the other selectors}
    \nextgroupplot[ymin=1e-3, ymax=1.4, title={GAIM-LIN}]
      \addplot[context] table[x=n, y=gaim_exact_v1] {\curvesa};
      \addplot[context] table[x=n, y=gaim_exact_v2] {\curvesa};
      \addplot[context] table[x=n, y=gaim_exact_v3] {\curvesa};
      \addplot[context] table[x=n, y=gaim_exact_v4] {\curvesa};
      \addplot[context] table[x=n, y=gaim_exact_v5] {\curvesa};
      \addplot[context] table[x=n, y=gaim_omp_v1] {\curvesa};
      \addplot[context] table[x=n, y=gaim_omp_v2] {\curvesa};
      \addplot[context] table[x=n, y=gaim_omp_v3] {\curvesa};
      \addplot[context] table[x=n, y=gaim_omp_v4] {\curvesa};
      \addplot[context] table[x=n, y=gaim_omp_v5] {\curvesa};
      \addplot[context] table[x=n, y=qdaim_exact_v1] {\curvesa};
      \addplot[context] table[x=n, y=qdaim_exact_v5] {\curvesa};
      \addplot[context] table[x=n, y=qdaim_rff_v1] {\curvesa};
      \addplot[context] table[x=n, y=qdaim_rff_v5] {\curvesa};
      \addplot[v1, mark phase=2] table[x=n, y=gaim_mp_v1] {\curvesa};
      \addplot[v2, mark phase=4] table[x=n, y=gaim_mp_v2] {\curvesa};
      \addplot[v3, mark phase=6] table[x=n, y=gaim_mp_v3] {\curvesa};
      \addplot[v4, mark phase=8] table[x=n, y=gaim_mp_v4] {\curvesa};
      \addplot[v5, mark phase=3] table[x=n, y=gaim_mp_v5] {\curvesa};
    \nextgroupplot[ymin=1e-3, ymax=1.4, title={GAIM-LIN-O}]
      \addplot[context] table[x=n, y=gaim_exact_v1] {\curvesa};
      \addplot[context] table[x=n, y=gaim_exact_v2] {\curvesa};
      \addplot[context] table[x=n, y=gaim_exact_v3] {\curvesa};
      \addplot[context] table[x=n, y=gaim_exact_v4] {\curvesa};
      \addplot[context] table[x=n, y=gaim_exact_v5] {\curvesa};
      \addplot[context] table[x=n, y=gaim_mp_v1] {\curvesa};
      \addplot[context] table[x=n, y=gaim_mp_v2] {\curvesa};
      \addplot[context] table[x=n, y=gaim_mp_v3] {\curvesa};
      \addplot[context] table[x=n, y=gaim_mp_v4] {\curvesa};
      \addplot[context] table[x=n, y=gaim_mp_v5] {\curvesa};
      \addplot[context] table[x=n, y=qdaim_exact_v1] {\curvesa};
      \addplot[context] table[x=n, y=qdaim_exact_v5] {\curvesa};
      \addplot[context] table[x=n, y=qdaim_rff_v1] {\curvesa};
      \addplot[context] table[x=n, y=qdaim_rff_v5] {\curvesa};
      \addplot[v1, mark phase=2] table[x=n, y=gaim_omp_v1] {\curvesa};
      \addplot[v2, mark phase=4] table[x=n, y=gaim_omp_v2] {\curvesa};
      \addplot[v3, mark phase=6] table[x=n, y=gaim_omp_v3] {\curvesa};
      \addplot[v4, mark phase=8] table[x=n, y=gaim_omp_v4] {\curvesa};
      \addplot[v5, mark phase=3] table[x=n, y=gaim_omp_v5] {\curvesa};
    \nextgroupplot[ymin=1e-3, ymax=1.4, title={Q-DAIM exact}]
      \addplot[context] table[x=n, y=gaim_exact_v1] {\curvesa};
      \addplot[context] table[x=n, y=gaim_exact_v2] {\curvesa};
      \addplot[context] table[x=n, y=gaim_exact_v3] {\curvesa};
      \addplot[context] table[x=n, y=gaim_exact_v4] {\curvesa};
      \addplot[context] table[x=n, y=gaim_exact_v5] {\curvesa};
      \addplot[context] table[x=n, y=gaim_mp_v1] {\curvesa};
      \addplot[context] table[x=n, y=gaim_mp_v2] {\curvesa};
      \addplot[context] table[x=n, y=gaim_mp_v3] {\curvesa};
      \addplot[context] table[x=n, y=gaim_mp_v4] {\curvesa};
      \addplot[context] table[x=n, y=gaim_mp_v5] {\curvesa};
      \addplot[context] table[x=n, y=gaim_omp_v1] {\curvesa};
      \addplot[context] table[x=n, y=gaim_omp_v2] {\curvesa};
      \addplot[context] table[x=n, y=gaim_omp_v3] {\curvesa};
      \addplot[context] table[x=n, y=gaim_omp_v4] {\curvesa};
      \addplot[context] table[x=n, y=gaim_omp_v5] {\curvesa};
      \addplot[context] table[x=n, y=qdaim_rff_v1] {\curvesa};
      \addplot[context] table[x=n, y=qdaim_rff_v5] {\curvesa};
      \addplot[v1, mark phase=2] table[x=n, y=qdaim_exact_v1] {\curvesa};
      \addplot[v5, mark phase=3] table[x=n, y=qdaim_exact_v5] {\curvesa};
    \nextgroupplot[ymin=1e-3, ymax=1.4, title={Q-DAIM RFF}]
      \addplot[context] table[x=n, y=gaim_exact_v1] {\curvesa};
      \addplot[context] table[x=n, y=gaim_exact_v2] {\curvesa};
      \addplot[context] table[x=n, y=gaim_exact_v3] {\curvesa};
      \addplot[context] table[x=n, y=gaim_exact_v4] {\curvesa};
      \addplot[context] table[x=n, y=gaim_exact_v5] {\curvesa};
      \addplot[context] table[x=n, y=gaim_mp_v1] {\curvesa};
      \addplot[context] table[x=n, y=gaim_mp_v2] {\curvesa};
      \addplot[context] table[x=n, y=gaim_mp_v3] {\curvesa};
      \addplot[context] table[x=n, y=gaim_mp_v4] {\curvesa};
      \addplot[context] table[x=n, y=gaim_mp_v5] {\curvesa};
      \addplot[context] table[x=n, y=gaim_omp_v1] {\curvesa};
      \addplot[context] table[x=n, y=gaim_omp_v2] {\curvesa};
      \addplot[context] table[x=n, y=gaim_omp_v3] {\curvesa};
      \addplot[context] table[x=n, y=gaim_omp_v4] {\curvesa};
      \addplot[context] table[x=n, y=gaim_omp_v5] {\curvesa};
      \addplot[context] table[x=n, y=qdaim_exact_v1] {\curvesa};
      \addplot[context] table[x=n, y=qdaim_exact_v5] {\curvesa};
      \addplot[v1, mark phase=2] table[x=n, y=qdaim_rff_v1] {\curvesa};
      \addplot[v5, mark phase=3] table[x=n, y=qdaim_rff_v5] {\curvesa};
    \nextgroupplot[ymin=1e-3, ymax=1.4, ylabel={$\|M^{h}_{\Delta,\eta}\|_2^2\,/\,\|M^{h}_{\eta}\|_2^2$}]
      \addplot[context] table[x=n, y=gaim_mp_v1] {\curvesb};
      \addplot[context] table[x=n, y=gaim_mp_v2] {\curvesb};
      \addplot[context] table[x=n, y=gaim_mp_v3] {\curvesb};
      \addplot[context] table[x=n, y=gaim_mp_v4] {\curvesb};
      \addplot[context] table[x=n, y=gaim_mp_v5] {\curvesb};
      \addplot[context] table[x=n, y=gaim_omp_v1] {\curvesb};
      \addplot[context] table[x=n, y=gaim_omp_v2] {\curvesb};
      \addplot[context] table[x=n, y=gaim_omp_v3] {\curvesb};
      \addplot[context] table[x=n, y=gaim_omp_v4] {\curvesb};
      \addplot[context] table[x=n, y=gaim_omp_v5] {\curvesb};
      \addplot[context] table[x=n, y=qdaim_exact_v1] {\curvesb};
      \addplot[context] table[x=n, y=qdaim_exact_v5] {\curvesb};
      \addplot[context] table[x=n, y=qdaim_rff_v1] {\curvesb};
      \addplot[context] table[x=n, y=qdaim_rff_v5] {\curvesb};
      \addplot[v1, mark phase=2] table[x=n, y=gaim_exact_v1] {\curvesb};
      \addplot[v2, mark phase=4] table[x=n, y=gaim_exact_v2] {\curvesb};
      \addplot[v3, mark phase=6] table[x=n, y=gaim_exact_v3] {\curvesb};
      \addplot[v4, mark phase=8] table[x=n, y=gaim_exact_v4] {\curvesb};
      \addplot[v5, mark phase=3] table[x=n, y=gaim_exact_v5] {\curvesb};
    \nextgroupplot[ymin=1e-3, ymax=1.4]
      \addplot[context] table[x=n, y=gaim_exact_v1] {\curvesb};
      \addplot[context] table[x=n, y=gaim_exact_v2] {\curvesb};
      \addplot[context] table[x=n, y=gaim_exact_v3] {\curvesb};
      \addplot[context] table[x=n, y=gaim_exact_v4] {\curvesb};
      \addplot[context] table[x=n, y=gaim_exact_v5] {\curvesb};
      \addplot[context] table[x=n, y=gaim_omp_v1] {\curvesb};
      \addplot[context] table[x=n, y=gaim_omp_v2] {\curvesb};
      \addplot[context] table[x=n, y=gaim_omp_v3] {\curvesb};
      \addplot[context] table[x=n, y=gaim_omp_v4] {\curvesb};
      \addplot[context] table[x=n, y=gaim_omp_v5] {\curvesb};
      \addplot[context] table[x=n, y=qdaim_exact_v1] {\curvesb};
      \addplot[context] table[x=n, y=qdaim_exact_v5] {\curvesb};
      \addplot[context] table[x=n, y=qdaim_rff_v1] {\curvesb};
      \addplot[context] table[x=n, y=qdaim_rff_v5] {\curvesb};
      \addplot[v1, mark phase=2] table[x=n, y=gaim_mp_v1] {\curvesb};
      \addplot[v2, mark phase=4] table[x=n, y=gaim_mp_v2] {\curvesb};
      \addplot[v3, mark phase=6] table[x=n, y=gaim_mp_v3] {\curvesb};
      \addplot[v4, mark phase=8] table[x=n, y=gaim_mp_v4] {\curvesb};
      \addplot[v5, mark phase=3] table[x=n, y=gaim_mp_v5] {\curvesb};
    \nextgroupplot[ymin=1e-3, ymax=1.4, xlabel={number of nodes $n$}]
      \addplot[context] table[x=n, y=gaim_exact_v1] {\curvesb};
      \addplot[context] table[x=n, y=gaim_exact_v2] {\curvesb};
      \addplot[context] table[x=n, y=gaim_exact_v3] {\curvesb};
      \addplot[context] table[x=n, y=gaim_exact_v4] {\curvesb};
      \addplot[context] table[x=n, y=gaim_exact_v5] {\curvesb};
      \addplot[context] table[x=n, y=gaim_mp_v1] {\curvesb};
      \addplot[context] table[x=n, y=gaim_mp_v2] {\curvesb};
      \addplot[context] table[x=n, y=gaim_mp_v3] {\curvesb};
      \addplot[context] table[x=n, y=gaim_mp_v4] {\curvesb};
      \addplot[context] table[x=n, y=gaim_mp_v5] {\curvesb};
      \addplot[context] table[x=n, y=qdaim_exact_v1] {\curvesb};
      \addplot[context] table[x=n, y=qdaim_exact_v5] {\curvesb};
      \addplot[context] table[x=n, y=qdaim_rff_v1] {\curvesb};
      \addplot[context] table[x=n, y=qdaim_rff_v5] {\curvesb};
      \addplot[v1, mark phase=2] table[x=n, y=gaim_omp_v1] {\curvesb};
      \addplot[v2, mark phase=4] table[x=n, y=gaim_omp_v2] {\curvesb};
      \addplot[v3, mark phase=6] table[x=n, y=gaim_omp_v3] {\curvesb};
      \addplot[v4, mark phase=8] table[x=n, y=gaim_omp_v4] {\curvesb};
      \addplot[v5, mark phase=3] table[x=n, y=gaim_omp_v5] {\curvesb};
    \nextgroupplot[ymin=1e-3, ymax=1.4]
      \addplot[context] table[x=n, y=gaim_exact_v1] {\curvesb};
      \addplot[context] table[x=n, y=gaim_exact_v2] {\curvesb};
      \addplot[context] table[x=n, y=gaim_exact_v3] {\curvesb};
      \addplot[context] table[x=n, y=gaim_exact_v4] {\curvesb};
      \addplot[context] table[x=n, y=gaim_exact_v5] {\curvesb};
      \addplot[context] table[x=n, y=gaim_mp_v1] {\curvesb};
      \addplot[context] table[x=n, y=gaim_mp_v2] {\curvesb};
      \addplot[context] table[x=n, y=gaim_mp_v3] {\curvesb};
      \addplot[context] table[x=n, y=gaim_mp_v4] {\curvesb};
      \addplot[context] table[x=n, y=gaim_mp_v5] {\curvesb};
      \addplot[context] table[x=n, y=gaim_omp_v1] {\curvesb};
      \addplot[context] table[x=n, y=gaim_omp_v2] {\curvesb};
      \addplot[context] table[x=n, y=gaim_omp_v3] {\curvesb};
      \addplot[context] table[x=n, y=gaim_omp_v4] {\curvesb};
      \addplot[context] table[x=n, y=gaim_omp_v5] {\curvesb};
      \addplot[context] table[x=n, y=qdaim_rff_v1] {\curvesb};
      \addplot[context] table[x=n, y=qdaim_rff_v5] {\curvesb};
      \addplot[v1, mark phase=2] table[x=n, y=qdaim_exact_v1] {\curvesb};
      \addplot[v5, mark phase=3] table[x=n, y=qdaim_exact_v5] {\curvesb};
    \nextgroupplot[ymin=1e-3, ymax=1.4]
      \addplot[context] table[x=n, y=gaim_exact_v1] {\curvesb};
      \addplot[context] table[x=n, y=gaim_exact_v2] {\curvesb};
      \addplot[context] table[x=n, y=gaim_exact_v3] {\curvesb};
      \addplot[context] table[x=n, y=gaim_exact_v4] {\curvesb};
      \addplot[context] table[x=n, y=gaim_exact_v5] {\curvesb};
      \addplot[context] table[x=n, y=gaim_mp_v1] {\curvesb};
      \addplot[context] table[x=n, y=gaim_mp_v2] {\curvesb};
      \addplot[context] table[x=n, y=gaim_mp_v3] {\curvesb};
      \addplot[context] table[x=n, y=gaim_mp_v4] {\curvesb};
      \addplot[context] table[x=n, y=gaim_mp_v5] {\curvesb};
      \addplot[context] table[x=n, y=gaim_omp_v1] {\curvesb};
      \addplot[context] table[x=n, y=gaim_omp_v2] {\curvesb};
      \addplot[context] table[x=n, y=gaim_omp_v3] {\curvesb};
      \addplot[context] table[x=n, y=gaim_omp_v4] {\curvesb};
      \addplot[context] table[x=n, y=gaim_omp_v5] {\curvesb};
      \addplot[context] table[x=n, y=qdaim_exact_v1] {\curvesb};
      \addplot[context] table[x=n, y=qdaim_exact_v5] {\curvesb};
      \addplot[v1, mark phase=2] table[x=n, y=qdaim_rff_v1] {\curvesb};
      \addplot[v5, mark phase=3] table[x=n, y=qdaim_rff_v5] {\curvesb};
  \end{groupplot}
  \node[below=4pt] at (current bounding box.south) {\pgfplotslegendfromname{legend-fig-resid-complete-context-curves}};
\end{tikzpicture}

%% file: figures/tikz/fig_resid_exact_gh.tex
\begin{tikzpicture}
  \pgfplotsset{every axis/.append style={
      scale only axis, /tikz/font=\footnotesize, label style={align=center}, title style={align=center},
      legend cell align=left,
      legend style={font=\footnotesize, /tikz/every even column/.append style={column sep=7pt}}}}
  \pgfplotsset{
    v1/.style={blue, mark=*},
    v2/.style={red, mark=square*},
    v3/.style={cyan!60!black, mark=triangle*},
    v4/.style={orange!50!black, mark=diamond*},
    v5/.style={magenta!70!black, mark=pentagon*},
  }
  \pgfplotsset{
    context/.style={gray!40, thin, mark=none},
    mean value/.style={black, thin, dashdotted, mark=none},
    zero readout/.style={black, thin, densely dotted, mark=none},
    every axis plot/.append style={very thick, mark size=1.8pt, mark options={solid}, mark repeat=8, forget plot},
    legend image code/.code={\draw[very thick, mark size=1.8pt, mark options={solid}, mark repeat=2, mark phase=2, #1]
      plot coordinates {(0pt,0pt) (10pt,0pt) (20pt,0pt)};},
  }
  \pgfplotsset{
    band/.style={draw=none, fill, fill opacity=0.12, mark=none},
    range line/.style={thin, densely dashed, mark=none},
    range legend/.style={black, thin, densely dashed, mark=none},
    band legend/.style={legend image code/.code={\fill[black, fill opacity=0.12] (0pt,-3pt) rectangle (20pt,3pt);}},
  }
  \pgfplotstableread[row sep=\\]{
    n gaim_exact_v1 gaim_exact_v2 gaim_exact_v3 gaim_exact_v4 gaim_exact_v5 gaim_mp_v1 gaim_mp_v2 gaim_mp_v3 gaim_mp_v4 gaim_mp_v5 gaim_omp_v1 gaim_omp_v2 gaim_omp_v3 gaim_omp_v4 gaim_omp_v5 qdaim_exact_v1 qdaim_exact_v5 qdaim_rff_v1 qdaim_rff_v5\\
    1 0.082489 0.082485 0.082485 0.082473 0.082489 0.082493 0.082491 0.082491 0.082491 0.082493 0.082493 0.082491 0.082491 0.082491 0.082493 0.082489 0.082489 0.08249 0.08249\\
    17 0.082317 0.08234 0.082167 0.08219 0.082317 0.082252 0.081905 0.081905 0.081905 0.082153 0.082093 0.081549 0.082145 0.081549 0.082093 0.082291 0.082291 0.082284 0.082278\\
    33 0.082125 0.082226 0.081745 0.08087 0.082125 0.081579 0.08095 0.080937 0.08095 0.081681 0.081468 0.079558 0.081458 0.079558 0.081468 0.082124 0.082124 0.082117 0.082117\\
    49 0.081808 0.082136 0.08137 0.078485 0.081808 0.081158 0.079859 0.07983 0.079859 0.081158 0.080953 0.078525 0.080698 0.078525 0.080953 0.081764 0.08174 0.081842 0.081842\\
    65 0.081611 0.081806 0.081032 0.077816 0.081611 0.080602 0.078328 0.078464 0.078328 0.080628 0.080013 0.077555 0.079337 0.077555 0.080013 0.081614 0.08163 0.081608 0.081606\\
    81 0.081465 0.081665 0.080462 0.077424 0.081465 0.080177 0.077953 0.07788 0.077944 0.080298 0.079732 0.077283 0.077712 0.077265 0.079732 0.081433 0.081433 0.081421 0.081448\\
    97 0.081251 0.081554 0.08027 0.076781 0.081251 0.079768 0.076756 0.076879 0.076902 0.079953 0.079231 0.076923 0.07708 0.07695 0.079231 0.081112 0.081042 0.081101 0.081123\\
    113 0.080971 0.081397 0.080083 0.075144 0.080971 0.079223 0.076518 0.076604 0.075248 0.079504 0.078852 0.076689 0.076658 0.076693 0.078852 0.080847 0.080864 0.080842 0.080895\\
    129 0.080627 0.080608 0.079272 0.074227 0.080627 0.078786 0.076063 0.076428 0.075005 0.079217 0.078671 0.076332 0.075942 0.075173 0.078749 0.080648 0.080624 0.080573 0.080608\\
    145 0.080471 0.080386 0.078834 0.072265 0.080471 0.078441 0.074571 0.07592 0.074916 0.07877 0.07775 0.074867 0.075707 0.074931 0.078372 0.080435 0.080374 0.080263 0.080206\\
    161 0.080074 0.078933 0.078585 0.069855 0.080074 0.078099 0.074179 0.074483 0.074433 0.078278 0.077146 0.07438 0.075208 0.0747 0.078524 0.079931 0.079904 0.080017 0.07984\\
    177 0.079894 0.07907 0.078415 0.069778 0.079894 0.077774 0.073884 0.074354 0.073973 0.077913 0.076098 0.074307 0.074525 0.074291 0.078139 0.07957 0.079665 0.079473 0.07948\\
    193 0.079639 0.078105 0.077997 0.069655 0.079639 0.077557 0.073781 0.074143 0.073797 0.077618 0.07528 0.074224 0.074241 0.074101 0.077728 0.079211 0.07925 0.079268 0.07927\\
    209 0.079257 0.0768 0.077581 0.069504 0.079257 0.077134 0.073624 0.073841 0.07368 0.07721 0.075092 0.073941 0.073983 0.073917 0.077539 0.078889 0.078998 0.078909 0.078877\\
    225 0.079027 0.075819 0.077384 0.069503 0.079027 0.076823 0.073392 0.073496 0.073545 0.076937 0.074871 0.072366 0.073614 0.072513 0.077209 0.078445 0.078779 0.078661 0.078397\\
    241 0.078487 0.074545 0.076981 0.069304 0.078487 0.076432 0.073061 0.073275 0.07315 0.076535 0.074421 0.071793 0.072888 0.071821 0.077028 0.078238 0.078509 0.07836 0.078132\\
    257 0.078217 0.071628 0.076692 0.069253 0.078217 0.076174 0.072961 0.072742 0.072941 0.076298 0.07381 0.07146 0.072612 0.071493 0.076669 0.077982 0.078246 0.078133 0.077887\\
    273 0.077907 0.071498 0.076394 0.068953 0.077907 0.075944 0.07237 0.072698 0.072484 0.076013 0.073447 0.07099 0.07238 0.071169 0.076369 0.077756 0.078044 0.077875 0.0774\\
    289 0.077665 0.07106 0.076042 0.068893 0.077665 0.075703 0.07067 0.070844 0.07069 0.075703 0.072858 0.07076 0.071134 0.070397 0.076005 0.077467 0.077648 0.07762 0.077196\\
    305 0.077377 0.070798 0.075726 0.068814 0.077377 0.075236 0.070422 0.070703 0.070266 0.075132 0.072463 0.070401 0.070833 0.06967 0.075236 0.077149 0.077378 0.077282 0.076832\\
    321 0.077034 0.070544 0.075285 0.068711 0.077034 0.074831 0.070144 0.070496 0.069953 0.074771 0.072077 0.070029 0.070396 0.069377 0.074841 0.076916 0.077119 0.077026 0.076532\\
    337 0.07684 0.070373 0.074933 0.068052 0.07684 0.074649 0.06948 0.070035 0.069636 0.074534 0.071766 0.069723 0.070299 0.069047 0.074405 0.076642 0.076929 0.076692 0.076376\\
    353 0.07654 0.07033 0.074677 0.067924 0.07654 0.074324 0.069206 0.069541 0.068972 0.073946 0.071312 0.069232 0.069199 0.06879 0.074021 0.076361 0.076739 0.076322 0.07598\\
    369 0.076223 0.070324 0.074262 0.067629 0.076223 0.074089 0.068352 0.069286 0.068378 0.073699 0.071069 0.06848 0.067906 0.06839 0.073716 0.076101 0.076352 0.076051 0.075728\\
    385 0.075997 0.070177 0.073965 0.067341 0.075997 0.073747 0.067951 0.068665 0.067933 0.073347 0.070585 0.068155 0.067591 0.068095 0.073374 0.075791 0.076114 0.075797 0.075404\\
    401 0.075688 0.069853 0.07373 0.06718 0.075688 0.073421 0.067616 0.068351 0.067625 0.073077 0.070115 0.067998 0.067103 0.067754 0.073211 0.075476 0.075774 0.075392 0.074772\\
    417 0.075449 0.069272 0.073433 0.067106 0.075449 0.073163 0.066851 0.067685 0.067284 0.072781 0.069686 0.067788 0.066614 0.06741 0.07288 0.075128 0.075411 0.075063 0.07451\\
    433 0.074974 0.068608 0.073038 0.066871 0.074974 0.072909 0.066339 0.067461 0.066962 0.072568 0.069178 0.067536 0.066244 0.067055 0.072502 0.074845 0.075169 0.07482 0.074288\\
    449 0.074776 0.068134 0.072683 0.066456 0.074776 0.07274 0.066023 0.067209 0.066825 0.07221 0.068562 0.067261 0.065952 0.066607 0.072178 0.074641 0.074832 0.074638 0.073899\\
    465 0.074389 0.067968 0.072249 0.066444 0.074389 0.072471 0.065866 0.066771 0.066716 0.071923 0.067845 0.066775 0.065663 0.066095 0.071871 0.074506 0.074533 0.074381 0.073451\\
    481 0.074091 0.067584 0.072006 0.066222 0.074091 0.072117 0.065178 0.066637 0.066483 0.071607 0.067548 0.066571 0.065347 0.065954 0.071492 0.074165 0.07411 0.074103 0.073335\\
    497 0.073872 0.067418 0.07172 0.065621 0.073872 0.071766 0.064819 0.066333 0.066185 0.071276 0.066602 0.066327 0.064987 0.065681 0.07125 0.073788 0.073891 0.073799 0.073083\\
    512 0.073611 0.067137 0.07137 0.0655 0.073611 0.071504 0.064565 0.065924 0.065878 0.071099 0.065187 0.066123 0.064607 0.065546 0.071034 0.073452 0.073413 0.073566 0.072781\\
  }\curvesa
  \pgfplotstableread[row sep=\\]{
    n gaim_exact_v1 gaim_exact_v2 gaim_exact_v3 gaim_exact_v4 gaim_exact_v5 gaim_mp_v1 gaim_mp_v2 gaim_mp_v3 gaim_mp_v4 gaim_mp_v5 gaim_omp_v1 gaim_omp_v2 gaim_omp_v3 gaim_omp_v4 gaim_omp_v5 qdaim_exact_v1 qdaim_exact_v5 qdaim_rff_v1 qdaim_rff_v5\\
    1 0.049428 0.049429 0.049429 0.04942 0.049428 0.049394 0.049419 0.049419 0.049419 0.049394 0.049394 0.049419 0.049419 0.049419 0.049394 0.049426 0.049426 0.049375 0.049375\\
    17 0.049271 0.049384 0.049258 0.048753 0.049271 0.049207 0.049323 0.049295 0.049319 0.04926 0.049136 0.049334 0.04902 0.049351 0.049235 0.049272 0.049272 0.049273 0.049273\\
    33 0.049199 0.049263 0.049048 0.047049 0.049199 0.049047 0.049172 0.049155 0.049151 0.049076 0.049006 0.049066 0.048749 0.049082 0.049006 0.049161 0.049161 0.049161 0.049161\\
    49 0.04906 0.049175 0.048834 0.046726 0.04906 0.048866 0.048845 0.048825 0.048837 0.048862 0.048791 0.0488 0.048119 0.048826 0.048791 0.049028 0.049028 0.049028 0.049028\\
    65 0.048918 0.049108 0.048616 0.046179 0.048918 0.048537 0.048745 0.048723 0.048677 0.048569 0.048621 0.048709 0.047543 0.048744 0.048621 0.048903 0.048903 0.048915 0.048915\\
    81 0.048756 0.04899 0.048372 0.046085 0.048756 0.048245 0.048543 0.048544 0.048489 0.048286 0.048464 0.04851 0.04727 0.048483 0.048464 0.048675 0.048675 0.04868 0.04868\\
    97 0.04849 0.04889 0.048215 0.045968 0.04849 0.048061 0.048451 0.04843 0.048424 0.048062 0.048108 0.048408 0.047174 0.04843 0.048108 0.048513 0.048507 0.048509 0.048512\\
    113 0.048385 0.048696 0.04798 0.045776 0.048385 0.047839 0.04829 0.04812 0.048283 0.047839 0.047851 0.048315 0.046978 0.048398 0.047851 0.04837 0.048373 0.048367 0.048373\\
    129 0.048261 0.048561 0.047754 0.045668 0.048261 0.047675 0.047922 0.048019 0.047972 0.047656 0.047707 0.048121 0.046722 0.048201 0.047707 0.048264 0.048139 0.04811 0.048129\\
    145 0.048133 0.048436 0.047608 0.045565 0.048133 0.047432 0.047524 0.047759 0.04763 0.047376 0.047276 0.048064 0.046631 0.048083 0.047276 0.047982 0.047983 0.048003 0.048014\\
    161 0.047977 0.048243 0.047456 0.043526 0.047977 0.04717 0.047264 0.047619 0.047455 0.047111 0.04704 0.046286 0.046137 0.0463 0.04704 0.047855 0.047888 0.047909 0.047879\\
    177 0.047756 0.048066 0.047227 0.042936 0.047756 0.04679 0.047156 0.047446 0.047279 0.046754 0.046895 0.046176 0.046048 0.046199 0.046895 0.047653 0.047752 0.047789 0.047733\\
    193 0.047581 0.047946 0.047096 0.042816 0.047581 0.046556 0.046983 0.04717 0.047127 0.046546 0.0465 0.046084 0.046033 0.046109 0.046592 0.047597 0.047602 0.047651 0.047571\\
    209 0.047466 0.04765 0.046884 0.042624 0.047466 0.046334 0.046895 0.04702 0.047071 0.04628 0.04631 0.046012 0.045935 0.046019 0.04646 0.047464 0.047414 0.047489 0.04749\\
    225 0.047321 0.04743 0.046749 0.042431 0.047321 0.046138 0.046743 0.046968 0.046985 0.046152 0.046218 0.045868 0.044834 0.045888 0.046452 0.047365 0.047298 0.047341 0.047287\\
    241 0.047156 0.047146 0.046552 0.042305 0.047156 0.045983 0.045137 0.045359 0.045461 0.046082 0.046086 0.04581 0.044522 0.04578 0.046331 0.047223 0.047188 0.047225 0.047167\\
    257 0.047035 0.046864 0.04634 0.042217 0.047035 0.045767 0.044977 0.045168 0.045359 0.04589 0.045874 0.04577 0.04432 0.045757 0.046071 0.04713 0.047088 0.046994 0.047063\\
    273 0.046796 0.046536 0.046101 0.04211 0.046796 0.045623 0.044906 0.045121 0.045261 0.04569 0.045699 0.045509 0.044144 0.045509 0.045879 0.046929 0.047006 0.046752 0.04693\\
    289 0.046746 0.046371 0.04591 0.041827 0.046746 0.045459 0.044746 0.045052 0.045176 0.045554 0.045625 0.045375 0.043876 0.045375 0.045807 0.046819 0.046849 0.046658 0.046791\\
    305 0.046568 0.046052 0.045767 0.04175 0.046568 0.045329 0.044407 0.044987 0.044889 0.045421 0.045492 0.045068 0.043656 0.045068 0.045675 0.046604 0.046737 0.046514 0.046728\\
    321 0.046506 0.045898 0.045458 0.041665 0.046506 0.045127 0.044332 0.044739 0.044828 0.045231 0.045261 0.042542 0.043535 0.042729 0.045426 0.046502 0.046568 0.046349 0.046581\\
    337 0.046421 0.045611 0.04527 0.041552 0.046421 0.045009 0.044174 0.044581 0.044744 0.045118 0.045116 0.042275 0.043446 0.042488 0.045268 0.046433 0.046343 0.045846 0.046367\\
    353 0.046306 0.044587 0.044898 0.041332 0.046306 0.044869 0.043886 0.044476 0.044502 0.044934 0.044951 0.04169 0.043383 0.042041 0.045119 0.046237 0.046233 0.045592 0.046197\\
    369 0.04619 0.044477 0.044739 0.041058 0.04619 0.044528 0.04375 0.043871 0.043944 0.044751 0.04469 0.04138 0.042812 0.041727 0.044875 0.04614 0.04613 0.045416 0.04594\\
    385 0.046066 0.044374 0.044637 0.040877 0.046066 0.044333 0.0436 0.043658 0.043692 0.044614 0.044635 0.040924 0.042827 0.04128 0.044792 0.045984 0.046034 0.045224 0.045807\\
    401 0.045938 0.044247 0.044497 0.040638 0.045938 0.044028 0.043276 0.043421 0.043329 0.044288 0.044371 0.040456 0.042546 0.040846 0.044578 0.045893 0.045888 0.044956 0.045646\\
    417 0.045815 0.044067 0.044335 0.040441 0.045815 0.043725 0.042787 0.042787 0.042787 0.044005 0.044308 0.040216 0.04201 0.040601 0.044489 0.045804 0.045812 0.044847 0.045474\\
    433 0.045688 0.043985 0.044101 0.040234 0.045688 0.043557 0.041579 0.041545 0.041579 0.043838 0.044218 0.040029 0.041703 0.040371 0.044393 0.045675 0.045698 0.044736 0.045224\\
    449 0.045592 0.043778 0.043914 0.040049 0.045592 0.043268 0.040735 0.041269 0.04088 0.043468 0.04398 0.039876 0.041471 0.040128 0.044203 0.04559 0.045567 0.044598 0.04504\\
    465 0.04547 0.043662 0.043677 0.039844 0.04547 0.043017 0.040452 0.041002 0.040691 0.043129 0.043884 0.03964 0.041179 0.039977 0.044109 0.045516 0.045517 0.044441 0.044572\\
    481 0.045383 0.043547 0.043522 0.039667 0.045383 0.042705 0.040096 0.04067 0.040359 0.042941 0.043748 0.03947 0.040643 0.039863 0.043988 0.0454 0.045443 0.044247 0.044308\\
    497 0.045273 0.042493 0.043313 0.039517 0.045273 0.042581 0.039832 0.04032 0.040107 0.042669 0.043578 0.039268 0.0404 0.039668 0.043886 0.045297 0.045272 0.044114 0.044091\\
    512 0.045134 0.041629 0.043157 0.03937 0.045134 0.042438 0.039552 0.039796 0.039768 0.042471 0.043415 0.039121 0.039686 0.039574 0.043728 0.045214 0.045179 0.044002 0.043937\\
  }\curvesb
  \begin{groupplot}[
      group style={group size=5 by 2, horizontal sep=6pt, vertical sep=9pt,
        xticklabels at=edge bottom, yticklabels at=edge left},
      width=77pt, height=92pt, grid=major, tick label style={font=\scriptsize},
      xmin=-15, xmax=527, xtick={0,250,500}, scaled y ticks=false, yticklabel style={/pgf/number format/fixed, /pgf/number format/fixed zerofill, /pgf/number format/precision=3}]
    \nextgroupplot[ymin=0.06385, ymax=0.08321, title={GAIM exact}, ylabel={relative error $\varepsilon_g^{10\%}$}, legend to name=legend-fig-resid-exact-gh-curves, legend columns=-1, legend style={draw=none}]
      \addplot[context] table[x=n, y=gaim_mp_v1] {\curvesa};
      \addplot[context] table[x=n, y=gaim_mp_v2] {\curvesa};
      \addplot[context] table[x=n, y=gaim_mp_v3] {\curvesa};
      \addplot[context] table[x=n, y=gaim_mp_v4] {\curvesa};
      \addplot[context] table[x=n, y=gaim_mp_v5] {\curvesa};
      \addplot[context] table[x=n, y=gaim_omp_v1] {\curvesa};
      \addplot[context] table[x=n, y=gaim_omp_v2] {\curvesa};
      \addplot[context] table[x=n, y=gaim_omp_v3] {\curvesa};
      \addplot[context] table[x=n, y=gaim_omp_v4] {\curvesa};
      \addplot[context] table[x=n, y=gaim_omp_v5] {\curvesa};
      \addplot[context] table[x=n, y=qdaim_exact_v1] {\curvesa};
      \addplot[context] table[x=n, y=qdaim_exact_v5] {\curvesa};
      \addplot[context] table[x=n, y=qdaim_rff_v1] {\curvesa};
      \addplot[context] table[x=n, y=qdaim_rff_v5] {\curvesa};
      \addplot[mean value] coordinates {(-15,0.065696) (527,0.065696)};
      \addplot[v1, mark phase=2] table[x=n, y=gaim_exact_v1] {\curvesa};
      \addplot[v2, mark phase=4] table[x=n, y=gaim_exact_v2] {\curvesa};
      \addplot[v3, mark phase=6] table[x=n, y=gaim_exact_v3] {\curvesa};
      \addplot[v4, mark phase=8] table[x=n, y=gaim_exact_v4] {\curvesa};
      \addplot[v5, mark phase=3] table[x=n, y=gaim_exact_v5] {\curvesa};
      \addlegendimage{v1}\addlegendentry{v1}
      \addlegendimage{v2}\addlegendentry{v2}
      \addlegendimage{v3}\addlegendentry{v3}
      \addlegendimage{v4}\addlegendentry{v4}
      \addlegendimage{v5}\addlegendentry{v5}
      \addlegendimage{context}\addlegendentry{the other selectors}
    \nextgroupplot[ymin=0.06385, ymax=0.08321, title={GAIM-LIN}, legend to name=legend-fig-resid-exact-gh-references, legend columns=-1, legend style={draw=none}]
      \addplot[context] table[x=n, y=gaim_exact_v1] {\curvesa};
      \addplot[context] table[x=n, y=gaim_exact_v2] {\curvesa};
      \addplot[context] table[x=n, y=gaim_exact_v3] {\curvesa};
      \addplot[context] table[x=n, y=gaim_exact_v4] {\curvesa};
      \addplot[context] table[x=n, y=gaim_exact_v5] {\curvesa};
      \addplot[context] table[x=n, y=gaim_omp_v1] {\curvesa};
      \addplot[context] table[x=n, y=gaim_omp_v2] {\curvesa};
      \addplot[context] table[x=n, y=gaim_omp_v3] {\curvesa};
      \addplot[context] table[x=n, y=gaim_omp_v4] {\curvesa};
      \addplot[context] table[x=n, y=gaim_omp_v5] {\curvesa};
      \addplot[context] table[x=n, y=qdaim_exact_v1] {\curvesa};
      \addplot[context] table[x=n, y=qdaim_exact_v5] {\curvesa};
      \addplot[context] table[x=n, y=qdaim_rff_v1] {\curvesa};
      \addplot[context] table[x=n, y=qdaim_rff_v5] {\curvesa};
      \addplot[mean value] coordinates {(-15,0.065696) (527,0.065696)};
      \addplot[v1, mark phase=2] table[x=n, y=gaim_mp_v1] {\curvesa};
      \addplot[v2, mark phase=4] table[x=n, y=gaim_mp_v2] {\curvesa};
      \addplot[v3, mark phase=6] table[x=n, y=gaim_mp_v3] {\curvesa};
      \addplot[v4, mark phase=8] table[x=n, y=gaim_mp_v4] {\curvesa};
      \addplot[v5, mark phase=3] table[x=n, y=gaim_mp_v5] {\curvesa};
      \addlegendimage{mean value}\addlegendentry{mean vector approximation}
      \addlegendimage{empty legend}\addlegendentry{zero readout $\approx$ $0.43$ / $0.53$ (top / bottom)}
    \nextgroupplot[ymin=0.06385, ymax=0.08321, title={GAIM-LIN-O}]
      \addplot[context] table[x=n, y=gaim_exact_v1] {\curvesa};
      \addplot[context] table[x=n, y=gaim_exact_v2] {\curvesa};
      \addplot[context] table[x=n, y=gaim_exact_v3] {\curvesa};
      \addplot[context] table[x=n, y=gaim_exact_v4] {\curvesa};
      \addplot[context] table[x=n, y=gaim_exact_v5] {\curvesa};
      \addplot[context] table[x=n, y=gaim_mp_v1] {\curvesa};
      \addplot[context] table[x=n, y=gaim_mp_v2] {\curvesa};
      \addplot[context] table[x=n, y=gaim_mp_v3] {\curvesa};
      \addplot[context] table[x=n, y=gaim_mp_v4] {\curvesa};
      \addplot[context] table[x=n, y=gaim_mp_v5] {\curvesa};
      \addplot[context] table[x=n, y=qdaim_exact_v1] {\curvesa};
      \addplot[context] table[x=n, y=qdaim_exact_v5] {\curvesa};
      \addplot[context] table[x=n, y=qdaim_rff_v1] {\curvesa};
      \addplot[context] table[x=n, y=qdaim_rff_v5] {\curvesa};
      \addplot[mean value] coordinates {(-15,0.065696) (527,0.065696)};
      \addplot[v1, mark phase=2] table[x=n, y=gaim_omp_v1] {\curvesa};
      \addplot[v2, mark phase=4] table[x=n, y=gaim_omp_v2] {\curvesa};
      \addplot[v3, mark phase=6] table[x=n, y=gaim_omp_v3] {\curvesa};
      \addplot[v4, mark phase=8] table[x=n, y=gaim_omp_v4] {\curvesa};
      \addplot[v5, mark phase=3] table[x=n, y=gaim_omp_v5] {\curvesa};
    \nextgroupplot[ymin=0.06385, ymax=0.08321, title={Q-DAIM exact}]
      \addplot[context] table[x=n, y=gaim_exact_v1] {\curvesa};
      \addplot[context] table[x=n, y=gaim_exact_v2] {\curvesa};
      \addplot[context] table[x=n, y=gaim_exact_v3] {\curvesa};
      \addplot[context] table[x=n, y=gaim_exact_v4] {\curvesa};
      \addplot[context] table[x=n, y=gaim_exact_v5] {\curvesa};
      \addplot[context] table[x=n, y=gaim_mp_v1] {\curvesa};
      \addplot[context] table[x=n, y=gaim_mp_v2] {\curvesa};
      \addplot[context] table[x=n, y=gaim_mp_v3] {\curvesa};
      \addplot[context] table[x=n, y=gaim_mp_v4] {\curvesa};
      \addplot[context] table[x=n, y=gaim_mp_v5] {\curvesa};
      \addplot[context] table[x=n, y=gaim_omp_v1] {\curvesa};
      \addplot[context] table[x=n, y=gaim_omp_v2] {\curvesa};
      \addplot[context] table[x=n, y=gaim_omp_v3] {\curvesa};
      \addplot[context] table[x=n, y=gaim_omp_v4] {\curvesa};
      \addplot[context] table[x=n, y=gaim_omp_v5] {\curvesa};
      \addplot[context] table[x=n, y=qdaim_rff_v1] {\curvesa};
      \addplot[context] table[x=n, y=qdaim_rff_v5] {\curvesa};
      \addplot[mean value] coordinates {(-15,0.065696) (527,0.065696)};
      \addplot[v1, mark phase=2] table[x=n, y=qdaim_exact_v1] {\curvesa};
      \addplot[v5, mark phase=3] table[x=n, y=qdaim_exact_v5] {\curvesa};
    \nextgroupplot[ymin=0.06385, ymax=0.08321, title={Q-DAIM RFF}]
      \addplot[context] table[x=n, y=gaim_exact_v1] {\curvesa};
      \addplot[context] table[x=n, y=gaim_exact_v2] {\curvesa};
      \addplot[context] table[x=n, y=gaim_exact_v3] {\curvesa};
      \addplot[context] table[x=n, y=gaim_exact_v4] {\curvesa};
      \addplot[context] table[x=n, y=gaim_exact_v5] {\curvesa};
      \addplot[context] table[x=n, y=gaim_mp_v1] {\curvesa};
      \addplot[context] table[x=n, y=gaim_mp_v2] {\curvesa};
      \addplot[context] table[x=n, y=gaim_mp_v3] {\curvesa};
      \addplot[context] table[x=n, y=gaim_mp_v4] {\curvesa};
      \addplot[context] table[x=n, y=gaim_mp_v5] {\curvesa};
      \addplot[context] table[x=n, y=gaim_omp_v1] {\curvesa};
      \addplot[context] table[x=n, y=gaim_omp_v2] {\curvesa};
      \addplot[context] table[x=n, y=gaim_omp_v3] {\curvesa};
      \addplot[context] table[x=n, y=gaim_omp_v4] {\curvesa};
      \addplot[context] table[x=n, y=gaim_omp_v5] {\curvesa};
      \addplot[context] table[x=n, y=qdaim_exact_v1] {\curvesa};
      \addplot[context] table[x=n, y=qdaim_exact_v5] {\curvesa};
      \addplot[mean value] coordinates {(-15,0.065696) (527,0.065696)};
      \addplot[v1, mark phase=2] table[x=n, y=qdaim_rff_v1] {\curvesa};
      \addplot[v5, mark phase=3] table[x=n, y=qdaim_rff_v5] {\curvesa};
    \nextgroupplot[ymin=0.03871, ymax=0.04984, ylabel={relative error $\varepsilon_h^{10\%}$}]
      \addplot[context] table[x=n, y=gaim_mp_v1] {\curvesb};
      \addplot[context] table[x=n, y=gaim_mp_v2] {\curvesb};
      \addplot[context] table[x=n, y=gaim_mp_v3] {\curvesb};
      \addplot[context] table[x=n, y=gaim_mp_v4] {\curvesb};
      \addplot[context] table[x=n, y=gaim_mp_v5] {\curvesb};
      \addplot[context] table[x=n, y=gaim_omp_v1] {\curvesb};
      \addplot[context] table[x=n, y=gaim_omp_v2] {\curvesb};
      \addplot[context] table[x=n, y=gaim_omp_v3] {\curvesb};
      \addplot[context] table[x=n, y=gaim_omp_v4] {\curvesb};
      \addplot[context] table[x=n, y=gaim_omp_v5] {\curvesb};
      \addplot[context] table[x=n, y=qdaim_exact_v1] {\curvesb};
      \addplot[context] table[x=n, y=qdaim_exact_v5] {\curvesb};
      \addplot[context] table[x=n, y=qdaim_rff_v1] {\curvesb};
      \addplot[context] table[x=n, y=qdaim_rff_v5] {\curvesb};
      \addplot[mean value] coordinates {(-15,0.040823) (527,0.040823)};
      \addplot[v1, mark phase=2] table[x=n, y=gaim_exact_v1] {\curvesb};
      \addplot[v2, mark phase=4] table[x=n, y=gaim_exact_v2] {\curvesb};
      \addplot[v3, mark phase=6] table[x=n, y=gaim_exact_v3] {\curvesb};
      \addplot[v4, mark phase=8] table[x=n, y=gaim_exact_v4] {\curvesb};
      \addplot[v5, mark phase=3] table[x=n, y=gaim_exact_v5] {\curvesb};
    \nextgroupplot[ymin=0.03871, ymax=0.04984]
      \addplot[context] table[x=n, y=gaim_exact_v1] {\curvesb};
      \addplot[context] table[x=n, y=gaim_exact_v2] {\curvesb};
      \addplot[context] table[x=n, y=gaim_exact_v3] {\curvesb};
      \addplot[context] table[x=n, y=gaim_exact_v4] {\curvesb};
      \addplot[context] table[x=n, y=gaim_exact_v5] {\curvesb};
      \addplot[context] table[x=n, y=gaim_omp_v1] {\curvesb};
      \addplot[context] table[x=n, y=gaim_omp_v2] {\curvesb};
      \addplot[context] table[x=n, y=gaim_omp_v3] {\curvesb};
      \addplot[context] table[x=n, y=gaim_omp_v4] {\curvesb};
      \addplot[context] table[x=n, y=gaim_omp_v5] {\curvesb};
      \addplot[context] table[x=n, y=qdaim_exact_v1] {\curvesb};
      \addplot[context] table[x=n, y=qdaim_exact_v5] {\curvesb};
      \addplot[context] table[x=n, y=qdaim_rff_v1] {\curvesb};
      \addplot[context] table[x=n, y=qdaim_rff_v5] {\curvesb};
      \addplot[mean value] coordinates {(-15,0.040823) (527,0.040823)};
      \addplot[v1, mark phase=2] table[x=n, y=gaim_mp_v1] {\curvesb};
      \addplot[v2, mark phase=4] table[x=n, y=gaim_mp_v2] {\curvesb};
      \addplot[v3, mark phase=6] table[x=n, y=gaim_mp_v3] {\curvesb};
      \addplot[v4, mark phase=8] table[x=n, y=gaim_mp_v4] {\curvesb};
      \addplot[v5, mark phase=3] table[x=n, y=gaim_mp_v5] {\curvesb};
    \nextgroupplot[ymin=0.03871, ymax=0.04984, xlabel={number of nodes $n$}]
      \addplot[context] table[x=n, y=gaim_exact_v1] {\curvesb};
      \addplot[context] table[x=n, y=gaim_exact_v2] {\curvesb};
      \addplot[context] table[x=n, y=gaim_exact_v3] {\curvesb};
      \addplot[context] table[x=n, y=gaim_exact_v4] {\curvesb};
      \addplot[context] table[x=n, y=gaim_exact_v5] {\curvesb};
      \addplot[context] table[x=n, y=gaim_mp_v1] {\curvesb};
      \addplot[context] table[x=n, y=gaim_mp_v2] {\curvesb};
      \addplot[context] table[x=n, y=gaim_mp_v3] {\curvesb};
      \addplot[context] table[x=n, y=gaim_mp_v4] {\curvesb};
      \addplot[context] table[x=n, y=gaim_mp_v5] {\curvesb};
      \addplot[context] table[x=n, y=qdaim_exact_v1] {\curvesb};
      \addplot[context] table[x=n, y=qdaim_exact_v5] {\curvesb};
      \addplot[context] table[x=n, y=qdaim_rff_v1] {\curvesb};
      \addplot[context] table[x=n, y=qdaim_rff_v5] {\curvesb};
      \addplot[mean value] coordinates {(-15,0.040823) (527,0.040823)};
      \addplot[v1, mark phase=2] table[x=n, y=gaim_omp_v1] {\curvesb};
      \addplot[v2, mark phase=4] table[x=n, y=gaim_omp_v2] {\curvesb};
      \addplot[v3, mark phase=6] table[x=n, y=gaim_omp_v3] {\curvesb};
      \addplot[v4, mark phase=8] table[x=n, y=gaim_omp_v4] {\curvesb};
      \addplot[v5, mark phase=3] table[x=n, y=gaim_omp_v5] {\curvesb};
    \nextgroupplot[ymin=0.03871, ymax=0.04984]
      \addplot[context] table[x=n, y=gaim_exact_v1] {\curvesb};
      \addplot[context] table[x=n, y=gaim_exact_v2] {\curvesb};
      \addplot[context] table[x=n, y=gaim_exact_v3] {\curvesb};
      \addplot[context] table[x=n, y=gaim_exact_v4] {\curvesb};
      \addplot[context] table[x=n, y=gaim_exact_v5] {\curvesb};
      \addplot[context] table[x=n, y=gaim_mp_v1] {\curvesb};
      \addplot[context] table[x=n, y=gaim_mp_v2] {\curvesb};
      \addplot[context] table[x=n, y=gaim_mp_v3] {\curvesb};
      \addplot[context] table[x=n, y=gaim_mp_v4] {\curvesb};
      \addplot[context] table[x=n, y=gaim_mp_v5] {\curvesb};
      \addplot[context] table[x=n, y=gaim_omp_v1] {\curvesb};
      \addplot[context] table[x=n, y=gaim_omp_v2] {\curvesb};
      \addplot[context] table[x=n, y=gaim_omp_v3] {\curvesb};
      \addplot[context] table[x=n, y=gaim_omp_v4] {\curvesb};
      \addplot[context] table[x=n, y=gaim_omp_v5] {\curvesb};
      \addplot[context] table[x=n, y=qdaim_rff_v1] {\curvesb};
      \addplot[context] table[x=n, y=qdaim_rff_v5] {\curvesb};
      \addplot[mean value] coordinates {(-15,0.040823) (527,0.040823)};
      \addplot[v1, mark phase=2] table[x=n, y=qdaim_exact_v1] {\curvesb};
      \addplot[v5, mark phase=3] table[x=n, y=qdaim_exact_v5] {\curvesb};
    \nextgroupplot[ymin=0.03871, ymax=0.04984]
      \addplot[context] table[x=n, y=gaim_exact_v1] {\curvesb};
      \addplot[context] table[x=n, y=gaim_exact_v2] {\curvesb};
      \addplot[context] table[x=n, y=gaim_exact_v3] {\curvesb};
      \addplot[context] table[x=n, y=gaim_exact_v4] {\curvesb};
      \addplot[context] table[x=n, y=gaim_exact_v5] {\curvesb};
      \addplot[context] table[x=n, y=gaim_mp_v1] {\curvesb};
      \addplot[context] table[x=n, y=gaim_mp_v2] {\curvesb};
      \addplot[context] table[x=n, y=gaim_mp_v3] {\curvesb};
      \addplot[context] table[x=n, y=gaim_mp_v4] {\curvesb};
      \addplot[context] table[x=n, y=gaim_mp_v5] {\curvesb};
      \addplot[context] table[x=n, y=gaim_omp_v1] {\curvesb};
      \addplot[context] table[x=n, y=gaim_omp_v2] {\curvesb};
      \addplot[context] table[x=n, y=gaim_omp_v3] {\curvesb};
      \addplot[context] table[x=n, y=gaim_omp_v4] {\curvesb};
      \addplot[context] table[x=n, y=gaim_omp_v5] {\curvesb};
      \addplot[context] table[x=n, y=qdaim_exact_v1] {\curvesb};
      \addplot[context] table[x=n, y=qdaim_exact_v5] {\curvesb};
      \addplot[mean value] coordinates {(-15,0.040823) (527,0.040823)};
      \addplot[v1, mark phase=2] table[x=n, y=qdaim_rff_v1] {\curvesb};
      \addplot[v5, mark phase=3] table[x=n, y=qdaim_rff_v5] {\curvesb};
  \end{groupplot}
  \node[below=4pt, draw, inner sep=1pt, align=center] at (current bounding box.south)
    {\pgfplotslegendfromname{legend-fig-resid-exact-gh-curves}\\[-2pt]\pgfplotslegendfromname{legend-fig-resid-exact-gh-references}};
\end{tikzpicture}

%% file: figures/tikz/convergence_exact_linf_sink32.tex
\begin{tikzpicture}
  \pgfplotsset{every axis/.append style={
      scale only axis, /tikz/font=\footnotesize, label style={align=center}, title style={align=center},
      legend cell align=left,
      legend style={font=\footnotesize, /tikz/every even column/.append style={column sep=7pt}}}}
  \pgfplotsset{
    v1/.style={blue, mark=*},
    v2/.style={red, mark=square*},
    v3/.style={cyan!60!black, mark=triangle*},
    v4/.style={orange!50!black, mark=diamond*},
    v5/.style={magenta!70!black, mark=pentagon*},
  }
  \pgfplotsset{
    context/.style={gray!40, thin, mark=none},
    mean value/.style={black, thin, dashdotted, mark=none},
    zero readout/.style={black, thin, densely dotted, mark=none},
    every axis plot/.append style={very thick, mark size=1.8pt, mark options={solid}, mark repeat=8, forget plot},
    legend image code/.code={\draw[very thick, mark size=1.8pt, mark options={solid}, mark repeat=2, mark phase=2, #1]
      plot coordinates {(0pt,0pt) (10pt,0pt) (20pt,0pt)};},
  }
  \pgfplotsset{
    band/.style={draw=none, fill, fill opacity=0.12, mark=none},
    range line/.style={thin, densely dashed, mark=none},
    range legend/.style={black, thin, densely dashed, mark=none},
    band legend/.style={legend image code/.code={\fill[black, fill opacity=0.12] (0pt,-3pt) rectangle (20pt,3pt);}},
  }
  \pgfplotstableread[row sep=\\]{
    n gaim_exact_v1 gaim_exact_v2 gaim_exact_v3 gaim_exact_v4 gaim_exact_v5 gaim_mp_v1 gaim_mp_v2 gaim_mp_v3 gaim_mp_v4 gaim_mp_v5 gaim_omp_v1 gaim_omp_v2 gaim_omp_v3 gaim_omp_v4 gaim_omp_v5 qdaim_exact_v1 qdaim_exact_v5 qdaim_rff_v1 qdaim_rff_v5\\
    1 0.20723 0.20723 0.20723 0.21387 0.20723 0.2072 0.21035 0.21035 0.21035 0.2072 0.2072 0.21035 0.21035 0.21035 0.2072 0.20723 0.20723 0.20722 0.20722\\
    17 0.21317 0.19883 0.21321 0.20634 0.21317 0.20645 0.20537 0.20537 0.20537 0.20645 0.21301 0.20863 0.21183 0.20863 0.21301 0.21325 0.21325 0.21314 0.21314\\
    33 0.21233 0.19868 0.20592 0.19517 0.21233 0.20531 0.20489 0.20459 0.20441 0.20531 0.212 0.20589 0.21596 0.20572 0.212 0.21244 0.21244 0.21244 0.21244\\
    49 0.21177 0.19846 0.2049 0.19638 0.21177 0.2043 0.21007 0.20993 0.20953 0.2043 0.20975 0.21024 0.21287 0.20975 0.20975 0.21167 0.21167 0.21158 0.21158\\
    65 0.21123 0.19592 0.1972 0.19083 0.21123 0.20152 0.20918 0.20913 0.20832 0.20152 0.20364 0.20939 0.21026 0.20861 0.20364 0.21123 0.21122 0.21127 0.21127\\
    81 0.21074 0.18991 0.19514 0.18796 0.21074 0.19342 0.20747 0.2077 0.207 0.19386 0.20254 0.20794 0.20379 0.20725 0.20254 0.21048 0.21048 0.21051 0.21048\\
    97 0.21011 0.18895 0.19436 0.18633 0.21011 0.19294 0.20257 0.2023 0.20173 0.1935 0.20113 0.20759 0.2032 0.20632 0.20113 0.21456 0.21457 0.20998 0.21002\\
    113 0.20945 0.18772 0.19396 0.18138 0.20945 0.19174 0.19857 0.19772 0.19714 0.19239 0.19975 0.20703 0.20042 0.20577 0.19975 0.21304 0.21314 0.20904 0.20909\\
    129 0.20795 0.18568 0.19304 0.17988 0.20795 0.18967 0.19979 0.19904 0.19832 0.19001 0.19918 0.20233 0.19706 0.20201 0.19918 0.2081 0.20626 0.206 0.20632\\
    145 0.20724 0.18485 0.19165 0.17445 0.20724 0.1848 0.20077 0.20021 0.19919 0.1848 0.19686 0.19745 0.19469 0.19745 0.19686 0.20508 0.20538 0.20539 0.20545\\
    161 0.20619 0.19141 0.191 0.16835 0.20619 0.1815 0.1933 0.1933 0.1933 0.1815 0.1955 0.19254 0.19059 0.19254 0.1955 0.20438 0.20449 0.20478 0.20787\\
    177 0.20342 0.18511 0.19017 0.16782 0.20342 0.17933 0.18821 0.18821 0.18821 0.17933 0.19424 0.18437 0.18809 0.18447 0.19424 0.20333 0.20689 0.20411 0.20724\\
    193 0.20298 0.18384 0.18937 0.16676 0.20298 0.17631 0.186 0.18538 0.18452 0.17688 0.1932 0.18314 0.18643 0.18255 0.1932 0.20628 0.20622 0.20667 0.20647\\
    209 0.202 0.18543 0.18814 0.1637 0.202 0.17508 0.18255 0.18219 0.18046 0.17536 0.19224 0.17856 0.1858 0.17788 0.19224 0.20221 0.20545 0.2058 0.20596\\
    225 0.20164 0.18001 0.18556 0.16148 0.20164 0.17384 0.18217 0.18144 0.18127 0.17425 0.19094 0.17678 0.1818 0.17609 0.19094 0.20115 0.20087 0.20511 0.20131\\
    241 0.19562 0.17925 0.18234 0.15859 0.19562 0.17324 0.1803 0.17945 0.17938 0.17351 0.19054 0.17595 0.18097 0.17495 0.19054 0.20065 0.20019 0.20045 0.20053\\
    257 0.19468 0.17441 0.17902 0.15805 0.19468 0.17027 0.17936 0.17806 0.17767 0.17027 0.18968 0.17471 0.18027 0.17365 0.18968 0.19955 0.19913 0.20081 0.19967\\
    273 0.19341 0.17336 0.1777 0.15732 0.19341 0.1697 0.1758 0.17595 0.17545 0.1697 0.18881 0.17356 0.17978 0.17294 0.18881 0.19857 0.19894 0.20144 0.19882\\
    289 0.19286 0.17232 0.17626 0.15691 0.19286 0.16839 0.17307 0.17336 0.17267 0.16839 0.18783 0.17238 0.17641 0.17163 0.18783 0.19806 0.19842 0.19982 0.19824\\
    305 0.1918 0.16863 0.17428 0.15638 0.1918 0.16723 0.17327 0.17355 0.17292 0.16723 0.18695 0.1719 0.17577 0.16992 0.18695 0.19744 0.19778 0.19562 0.19768\\
    321 0.19098 0.16761 0.17342 0.15571 0.19098 0.16644 0.17207 0.17247 0.17178 0.16644 0.18587 0.17097 0.17518 0.16925 0.18587 0.19706 0.19562 0.19483 0.19623\\
    337 0.19026 0.16447 0.17179 0.15433 0.19026 0.166 0.16899 0.17039 0.1692 0.166 0.185 0.1703 0.17272 0.16872 0.185 0.19685 0.19141 0.18809 0.18948\\
    353 0.1897 0.16167 0.17029 0.15401 0.1897 0.16513 0.1697 0.1707 0.16945 0.16513 0.18443 0.16918 0.16911 0.16796 0.18443 0.19507 0.19151 0.18728 0.18563\\
    369 0.18822 0.15976 0.16917 0.1536 0.18822 0.16374 0.16752 0.16885 0.16683 0.16374 0.18339 0.16729 0.1683 0.16684 0.1834 0.19332 0.19104 0.18648 0.18371\\
    385 0.18747 0.1575 0.16718 0.15337 0.18747 0.16298 0.16758 0.16934 0.16739 0.16298 0.18279 0.16671 0.16702 0.16613 0.18279 0.19144 0.18984 0.18468 0.18249\\
    401 0.18665 0.15677 0.16701 0.15248 0.18665 0.16227 0.16495 0.1661 0.16468 0.16227 0.1808 0.16638 0.16599 0.16475 0.18184 0.1909 0.18934 0.18429 0.18127\\
    417 0.18573 0.15544 0.16575 0.1509 0.18573 0.16131 0.16487 0.16625 0.16484 0.16131 0.17988 0.16585 0.16525 0.16359 0.18097 0.19029 0.18846 0.18169 0.18092\\
    433 0.18474 0.15833 0.16541 0.15125 0.18474 0.1608 0.16306 0.16434 0.16298 0.1608 0.17885 0.16514 0.16444 0.16283 0.18046 0.18981 0.18965 0.18077 0.17977\\
    449 0.18374 0.15476 0.16367 0.15098 0.18374 0.15985 0.16229 0.16343 0.16203 0.15985 0.17777 0.16439 0.16349 0.16162 0.17944 0.18964 0.18654 0.17985 0.17913\\
    465 0.18332 0.15398 0.16258 0.15062 0.18332 0.15896 0.16147 0.16195 0.16044 0.15896 0.17533 0.16337 0.16256 0.16045 0.17794 0.18942 0.18467 0.17926 0.17818\\
    481 0.18272 0.15259 0.16187 0.15033 0.18272 0.15784 0.16067 0.16128 0.15963 0.15784 0.17422 0.16247 0.16179 0.16 0.1774 0.18881 0.1838 0.1783 0.17748\\
    497 0.1823 0.15326 0.16126 0.14982 0.1823 0.15684 0.15982 0.15999 0.1583 0.15684 0.17209 0.16186 0.1609 0.15908 0.17659 0.18832 0.18009 0.17778 0.1767\\
    512 0.1814 0.15363 0.16035 0.1493 0.1814 0.15648 0.15935 0.15983 0.15845 0.15648 0.16854 0.16118 0.15979 0.15846 0.1759 0.18776 0.17803 0.17683 0.17616\\
  }\curvesa
  \pgfplotstableread[row sep=\\]{
    n gaim_exact_v1 gaim_exact_v2 gaim_exact_v3 gaim_exact_v4 gaim_exact_v5 gaim_mp_v1 gaim_mp_v2 gaim_mp_v3 gaim_mp_v4 gaim_mp_v5 gaim_omp_v1 gaim_omp_v2 gaim_omp_v3 gaim_omp_v4 gaim_omp_v5 qdaim_exact_v1 qdaim_exact_v5 qdaim_rff_v1 qdaim_rff_v5\\
    1 0.16889 0.16888 0.16888 0.16892 0.16889 0.16881 0.16888 0.16888 0.16878 0.16881 0.16881 0.16888 0.16888 0.16878 0.16881 0.16892 0.16892 0.1689 0.1689\\
    17 0.16807 0.16843 0.16764 0.16792 0.16807 0.1682 0.16804 0.16824 0.16818 0.16817 0.16814 0.16782 0.16809 0.16825 0.16819 0.16811 0.16811 0.16812 0.16812\\
    33 0.16744 0.16765 0.16697 0.1675 0.16744 0.16727 0.1694 0.1693 0.16949 0.16707 0.16704 0.16721 0.16896 0.16748 0.16746 0.1674 0.1674 0.1674 0.1674\\
    49 0.16695 0.16737 0.16624 0.16873 0.16695 0.16626 0.16711 0.16718 0.16694 0.16655 0.16629 0.16461 0.16695 0.16519 0.16645 0.16701 0.16701 0.16701 0.16701\\
    65 0.16654 0.16677 0.16532 0.16808 0.16654 0.16539 0.16362 0.16385 0.16389 0.16571 0.16371 0.16317 0.16435 0.16401 0.16438 0.16632 0.16632 0.16627 0.16627\\
    81 0.16582 0.16614 0.16457 0.1665 0.16582 0.16466 0.16288 0.16341 0.16299 0.16491 0.16289 0.1623 0.15961 0.163 0.16381 0.16816 0.16816 0.16564 0.16561\\
    97 0.16508 0.16571 0.16364 0.16491 0.16508 0.16387 0.16222 0.16282 0.16253 0.16407 0.16082 0.15927 0.15569 0.15943 0.16181 0.1674 0.1675 0.16738 0.1674\\
    113 0.16468 0.16505 0.16307 0.15977 0.16468 0.16311 0.15968 0.15928 0.15968 0.16349 0.1588 0.15628 0.15144 0.15663 0.1588 0.1668 0.1668 0.16451 0.16452\\
    129 0.16415 0.16381 0.16226 0.15667 0.16415 0.16242 0.15546 0.15577 0.15546 0.16291 0.15736 0.15427 0.15074 0.15407 0.15883 0.16401 0.16408 0.16395 0.16374\\
    145 0.16354 0.16339 0.16142 0.14592 0.16354 0.16141 0.15351 0.15392 0.15382 0.16228 0.15556 0.15258 0.15032 0.15258 0.15731 0.16336 0.16538 0.16338 0.16562\\
    161 0.16277 0.16165 0.16052 0.14304 0.16277 0.16066 0.15176 0.15236 0.15221 0.16142 0.15327 0.14793 0.14825 0.14823 0.15581 0.16458 0.16478 0.1653 0.1651\\
    177 0.16244 0.16184 0.15974 0.14091 0.16244 0.16025 0.14902 0.14973 0.14877 0.1604 0.1525 0.14778 0.14705 0.14753 0.15466 0.16418 0.16414 0.16446 0.16454\\
    193 0.162 0.16036 0.15939 0.13823 0.162 0.15939 0.14817 0.14838 0.14774 0.15913 0.15156 0.1471 0.14658 0.147 0.15346 0.16352 0.16359 0.16376 0.1636\\
    209 0.15881 0.15939 0.15825 0.13801 0.15881 0.15835 0.14714 0.14743 0.14664 0.1579 0.15071 0.14682 0.14574 0.14682 0.15128 0.16073 0.16318 0.16324 0.16298\\
    225 0.15835 0.15849 0.15786 0.13779 0.15835 0.15735 0.14563 0.1454 0.14543 0.15742 0.14931 0.14535 0.14109 0.1448 0.15038 0.16025 0.16061 0.16261 0.16068\\
    241 0.15764 0.15723 0.15758 0.13508 0.15764 0.15645 0.14385 0.14364 0.14385 0.15668 0.14755 0.14174 0.14101 0.14162 0.14966 0.15988 0.16011 0.1599 0.16005\\
    257 0.15688 0.15512 0.1569 0.13217 0.15688 0.15571 0.14288 0.14273 0.14288 0.156 0.14646 0.14012 0.1404 0.14012 0.14879 0.15946 0.15921 0.16136 0.15956\\
    273 0.15599 0.15286 0.1561 0.13112 0.15599 0.15301 0.14049 0.14163 0.14135 0.15513 0.14562 0.13833 0.13938 0.13691 0.14816 0.15893 0.15878 0.16073 0.15894\\
    289 0.1557 0.15166 0.15471 0.12995 0.1557 0.15259 0.14009 0.14094 0.1406 0.15459 0.14515 0.13777 0.13822 0.13547 0.14739 0.15823 0.15802 0.16029 0.15822\\
    305 0.15498 0.15094 0.15429 0.12882 0.15498 0.15215 0.1382 0.13894 0.13785 0.15392 0.14456 0.13688 0.13624 0.13378 0.14641 0.15777 0.15749 0.15956 0.15719\\
    321 0.15416 0.14928 0.15354 0.12821 0.15416 0.1516 0.13735 0.13735 0.13735 0.15344 0.14316 0.13623 0.13585 0.13292 0.14469 0.15727 0.15694 0.1591 0.15675\\
    337 0.15384 0.1457 0.15278 0.12714 0.15384 0.1511 0.13499 0.13497 0.13499 0.15257 0.14206 0.13543 0.13529 0.13234 0.14337 0.15687 0.15631 0.15612 0.15569\\
    353 0.15335 0.14421 0.15135 0.12688 0.15335 0.15065 0.13324 0.13365 0.13296 0.15159 0.1417 0.13371 0.13489 0.13158 0.14281 0.15633 0.15561 0.15548 0.15529\\
    369 0.15215 0.14434 0.15068 0.12608 0.15215 0.14991 0.13237 0.13313 0.12938 0.1509 0.14083 0.1331 0.13339 0.13081 0.14176 0.15578 0.15484 0.15473 0.15466\\
    385 0.15159 0.14352 0.14987 0.12564 0.15159 0.14937 0.13153 0.13217 0.12876 0.15009 0.1403 0.13248 0.13304 0.1298 0.14081 0.15513 0.15422 0.15407 0.15396\\
    401 0.15117 0.14274 0.14941 0.12504 0.15117 0.1492 0.13079 0.13114 0.12849 0.1493 0.13954 0.13218 0.13138 0.12906 0.13976 0.15436 0.1536 0.15362 0.15346\\
    417 0.15061 0.1424 0.14886 0.12457 0.15061 0.14789 0.13051 0.12977 0.12789 0.14847 0.13909 0.13061 0.12823 0.12817 0.13909 0.15377 0.15223 0.15283 0.15295\\
    433 0.15014 0.14238 0.14809 0.12426 0.15014 0.14643 0.12959 0.12932 0.12714 0.14767 0.13854 0.12754 0.12802 0.12569 0.13854 0.15049 0.15133 0.15241 0.15223\\
    449 0.14873 0.14181 0.14717 0.12333 0.14873 0.14557 0.12914 0.12857 0.12708 0.14661 0.13768 0.12695 0.12743 0.1248 0.13768 0.14987 0.15084 0.15119 0.15026\\
    465 0.14706 0.14156 0.14685 0.1232 0.14706 0.14439 0.12821 0.12694 0.12632 0.14583 0.13692 0.12634 0.12674 0.12453 0.13692 0.14954 0.14935 0.15014 0.14828\\
    481 0.14639 0.14104 0.14574 0.12244 0.14639 0.14309 0.12735 0.12667 0.12547 0.143 0.13651 0.12546 0.1262 0.12369 0.13651 0.149 0.14859 0.14827 0.14764\\
    497 0.14492 0.13707 0.14351 0.12161 0.14492 0.14148 0.12691 0.12594 0.12448 0.14214 0.13547 0.1246 0.12619 0.12296 0.13552 0.14828 0.14806 0.14582 0.14496\\
    512 0.14389 0.13518 0.14237 0.12085 0.14389 0.14125 0.12623 0.12362 0.12388 0.14166 0.13508 0.12464 0.12531 0.12296 0.13518 0.14697 0.14652 0.14359 0.14388\\
  }\curvesb
  \begin{groupplot}[
      group style={group size=5 by 2, horizontal sep=6pt, vertical sep=9pt,
        xticklabels at=edge bottom, yticklabels at=edge left},
      width=77pt, height=92pt, grid=major, tick label style={font=\scriptsize},
      xmin=-15, xmax=527, xtick={0,250,500}, scaled y ticks=false, yticklabel style={/pgf/number format/fixed, /pgf/number format/fixed zerofill, /pgf/number format/precision=3}]
    \nextgroupplot[ymin=0.1466, ymax=0.2186, title={GAIM exact}, ylabel={RC off\\\scriptsize relative attention error $\varepsilon_{\Att}^{10\%}$}, legend to name=legend-convergence-exact-linf-sink32-curves, legend columns=-1, legend style={draw=none}]
      \addplot[context] table[x=n, y=gaim_mp_v1] {\curvesa};
      \addplot[context] table[x=n, y=gaim_mp_v2] {\curvesa};
      \addplot[context] table[x=n, y=gaim_mp_v3] {\curvesa};
      \addplot[context] table[x=n, y=gaim_mp_v4] {\curvesa};
      \addplot[context] table[x=n, y=gaim_mp_v5] {\curvesa};
      \addplot[context] table[x=n, y=gaim_omp_v1] {\curvesa};
      \addplot[context] table[x=n, y=gaim_omp_v2] {\curvesa};
      \addplot[context] table[x=n, y=gaim_omp_v3] {\curvesa};
      \addplot[context] table[x=n, y=gaim_omp_v4] {\curvesa};
      \addplot[context] table[x=n, y=gaim_omp_v5] {\curvesa};
      \addplot[context] table[x=n, y=qdaim_exact_v1] {\curvesa};
      \addplot[context] table[x=n, y=qdaim_exact_v5] {\curvesa};
      \addplot[context] table[x=n, y=qdaim_rff_v1] {\curvesa};
      \addplot[context] table[x=n, y=qdaim_rff_v5] {\curvesa};
      \addplot[mean value] coordinates {(-15,0.16892) (527,0.16892)};
      \addplot[v1, mark phase=2] table[x=n, y=gaim_exact_v1] {\curvesa};
      \addplot[v2, mark phase=4] table[x=n, y=gaim_exact_v2] {\curvesa};
      \addplot[v3, mark phase=6] table[x=n, y=gaim_exact_v3] {\curvesa};
      \addplot[v4, mark phase=8] table[x=n, y=gaim_exact_v4] {\curvesa};
      \addplot[v5, mark phase=3] table[x=n, y=gaim_exact_v5] {\curvesa};
      \addlegendimage{v1}\addlegendentry{v1}
      \addlegendimage{v2}\addlegendentry{v2}
      \addlegendimage{v3}\addlegendentry{v3}
      \addlegendimage{v4}\addlegendentry{v4}
      \addlegendimage{v5}\addlegendentry{v5}
      \addlegendimage{context}\addlegendentry{the other selectors}
    \nextgroupplot[ymin=0.1466, ymax=0.2186, title={GAIM-LIN}, legend to name=legend-convergence-exact-linf-sink32-references, legend columns=-1, legend style={draw=none}]
      \addplot[context] table[x=n, y=gaim_exact_v1] {\curvesa};
      \addplot[context] table[x=n, y=gaim_exact_v2] {\curvesa};
      \addplot[context] table[x=n, y=gaim_exact_v3] {\curvesa};
      \addplot[context] table[x=n, y=gaim_exact_v4] {\curvesa};
      \addplot[context] table[x=n, y=gaim_exact_v5] {\curvesa};
      \addplot[context] table[x=n, y=gaim_omp_v1] {\curvesa};
      \addplot[context] table[x=n, y=gaim_omp_v2] {\curvesa};
      \addplot[context] table[x=n, y=gaim_omp_v3] {\curvesa};
      \addplot[context] table[x=n, y=gaim_omp_v4] {\curvesa};
      \addplot[context] table[x=n, y=gaim_omp_v5] {\curvesa};
      \addplot[context] table[x=n, y=qdaim_exact_v1] {\curvesa};
      \addplot[context] table[x=n, y=qdaim_exact_v5] {\curvesa};
      \addplot[context] table[x=n, y=qdaim_rff_v1] {\curvesa};
      \addplot[context] table[x=n, y=qdaim_rff_v5] {\curvesa};
      \addplot[mean value] coordinates {(-15,0.16892) (527,0.16892)};
      \addplot[v1, mark phase=2] table[x=n, y=gaim_mp_v1] {\curvesa};
      \addplot[v2, mark phase=4] table[x=n, y=gaim_mp_v2] {\curvesa};
      \addplot[v3, mark phase=6] table[x=n, y=gaim_mp_v3] {\curvesa};
      \addplot[v4, mark phase=8] table[x=n, y=gaim_mp_v4] {\curvesa};
      \addplot[v5, mark phase=3] table[x=n, y=gaim_mp_v5] {\curvesa};
      \addlegendimage{mean value}\addlegendentry{mean vector approximation}
      \addlegendimage{empty legend}\addlegendentry{zero readout $\approx$ $0.76$}
    \nextgroupplot[ymin=0.1466, ymax=0.2186, title={GAIM-LIN-O}]
      \addplot[context] table[x=n, y=gaim_exact_v1] {\curvesa};
      \addplot[context] table[x=n, y=gaim_exact_v2] {\curvesa};
      \addplot[context] table[x=n, y=gaim_exact_v3] {\curvesa};
      \addplot[context] table[x=n, y=gaim_exact_v4] {\curvesa};
      \addplot[context] table[x=n, y=gaim_exact_v5] {\curvesa};
      \addplot[context] table[x=n, y=gaim_mp_v1] {\curvesa};
      \addplot[context] table[x=n, y=gaim_mp_v2] {\curvesa};
      \addplot[context] table[x=n, y=gaim_mp_v3] {\curvesa};
      \addplot[context] table[x=n, y=gaim_mp_v4] {\curvesa};
      \addplot[context] table[x=n, y=gaim_mp_v5] {\curvesa};
      \addplot[context] table[x=n, y=qdaim_exact_v1] {\curvesa};
      \addplot[context] table[x=n, y=qdaim_exact_v5] {\curvesa};
      \addplot[context] table[x=n, y=qdaim_rff_v1] {\curvesa};
      \addplot[context] table[x=n, y=qdaim_rff_v5] {\curvesa};
      \addplot[mean value] coordinates {(-15,0.16892) (527,0.16892)};
      \addplot[v1, mark phase=2] table[x=n, y=gaim_omp_v1] {\curvesa};
      \addplot[v2, mark phase=4] table[x=n, y=gaim_omp_v2] {\curvesa};
      \addplot[v3, mark phase=6] table[x=n, y=gaim_omp_v3] {\curvesa};
      \addplot[v4, mark phase=8] table[x=n, y=gaim_omp_v4] {\curvesa};
      \addplot[v5, mark phase=3] table[x=n, y=gaim_omp_v5] {\curvesa};
    \nextgroupplot[ymin=0.1466, ymax=0.2186, title={Q-DAIM exact}]
      \addplot[context] table[x=n, y=gaim_exact_v1] {\curvesa};
      \addplot[context] table[x=n, y=gaim_exact_v2] {\curvesa};
      \addplot[context] table[x=n, y=gaim_exact_v3] {\curvesa};
      \addplot[context] table[x=n, y=gaim_exact_v4] {\curvesa};
      \addplot[context] table[x=n, y=gaim_exact_v5] {\curvesa};
      \addplot[context] table[x=n, y=gaim_mp_v1] {\curvesa};
      \addplot[context] table[x=n, y=gaim_mp_v2] {\curvesa};
      \addplot[context] table[x=n, y=gaim_mp_v3] {\curvesa};
      \addplot[context] table[x=n, y=gaim_mp_v4] {\curvesa};
      \addplot[context] table[x=n, y=gaim_mp_v5] {\curvesa};
      \addplot[context] table[x=n, y=gaim_omp_v1] {\curvesa};
      \addplot[context] table[x=n, y=gaim_omp_v2] {\curvesa};
      \addplot[context] table[x=n, y=gaim_omp_v3] {\curvesa};
      \addplot[context] table[x=n, y=gaim_omp_v4] {\curvesa};
      \addplot[context] table[x=n, y=gaim_omp_v5] {\curvesa};
      \addplot[context] table[x=n, y=qdaim_rff_v1] {\curvesa};
      \addplot[context] table[x=n, y=qdaim_rff_v5] {\curvesa};
      \addplot[mean value] coordinates {(-15,0.16892) (527,0.16892)};
      \addplot[v1, mark phase=2] table[x=n, y=qdaim_exact_v1] {\curvesa};
      \addplot[v5, mark phase=3] table[x=n, y=qdaim_exact_v5] {\curvesa};
    \nextgroupplot[ymin=0.1466, ymax=0.2186, title={Q-DAIM RFF}]
      \addplot[context] table[x=n, y=gaim_exact_v1] {\curvesa};
      \addplot[context] table[x=n, y=gaim_exact_v2] {\curvesa};
      \addplot[context] table[x=n, y=gaim_exact_v3] {\curvesa};
      \addplot[context] table[x=n, y=gaim_exact_v4] {\curvesa};
      \addplot[context] table[x=n, y=gaim_exact_v5] {\curvesa};
      \addplot[context] table[x=n, y=gaim_mp_v1] {\curvesa};
      \addplot[context] table[x=n, y=gaim_mp_v2] {\curvesa};
      \addplot[context] table[x=n, y=gaim_mp_v3] {\curvesa};
      \addplot[context] table[x=n, y=gaim_mp_v4] {\curvesa};
      \addplot[context] table[x=n, y=gaim_mp_v5] {\curvesa};
      \addplot[context] table[x=n, y=gaim_omp_v1] {\curvesa};
      \addplot[context] table[x=n, y=gaim_omp_v2] {\curvesa};
      \addplot[context] table[x=n, y=gaim_omp_v3] {\curvesa};
      \addplot[context] table[x=n, y=gaim_omp_v4] {\curvesa};
      \addplot[context] table[x=n, y=gaim_omp_v5] {\curvesa};
      \addplot[context] table[x=n, y=qdaim_exact_v1] {\curvesa};
      \addplot[context] table[x=n, y=qdaim_exact_v5] {\curvesa};
      \addplot[mean value] coordinates {(-15,0.16892) (527,0.16892)};
      \addplot[v1, mark phase=2] table[x=n, y=qdaim_rff_v1] {\curvesa};
      \addplot[v5, mark phase=3] table[x=n, y=qdaim_rff_v5] {\curvesa};
    \nextgroupplot[ymin=0.1189, ymax=0.1714, ylabel={RC on\\\scriptsize relative attention error $\varepsilon_{\Att}^{10\%}$}]
      \addplot[context] table[x=n, y=gaim_mp_v1] {\curvesb};
      \addplot[context] table[x=n, y=gaim_mp_v2] {\curvesb};
      \addplot[context] table[x=n, y=gaim_mp_v3] {\curvesb};
      \addplot[context] table[x=n, y=gaim_mp_v4] {\curvesb};
      \addplot[context] table[x=n, y=gaim_mp_v5] {\curvesb};
      \addplot[context] table[x=n, y=gaim_omp_v1] {\curvesb};
      \addplot[context] table[x=n, y=gaim_omp_v2] {\curvesb};
      \addplot[context] table[x=n, y=gaim_omp_v3] {\curvesb};
      \addplot[context] table[x=n, y=gaim_omp_v4] {\curvesb};
      \addplot[context] table[x=n, y=gaim_omp_v5] {\curvesb};
      \addplot[context] table[x=n, y=qdaim_exact_v1] {\curvesb};
      \addplot[context] table[x=n, y=qdaim_exact_v5] {\curvesb};
      \addplot[context] table[x=n, y=qdaim_rff_v1] {\curvesb};
      \addplot[context] table[x=n, y=qdaim_rff_v5] {\curvesb};
      \addplot[mean value] coordinates {(-15,0.16892) (527,0.16892)};
      \addplot[v1, mark phase=2] table[x=n, y=gaim_exact_v1] {\curvesb};
      \addplot[v2, mark phase=4] table[x=n, y=gaim_exact_v2] {\curvesb};
      \addplot[v3, mark phase=6] table[x=n, y=gaim_exact_v3] {\curvesb};
      \addplot[v4, mark phase=8] table[x=n, y=gaim_exact_v4] {\curvesb};
      \addplot[v5, mark phase=3] table[x=n, y=gaim_exact_v5] {\curvesb};
    \nextgroupplot[ymin=0.1189, ymax=0.1714]
      \addplot[context] table[x=n, y=gaim_exact_v1] {\curvesb};
      \addplot[context] table[x=n, y=gaim_exact_v2] {\curvesb};
      \addplot[context] table[x=n, y=gaim_exact_v3] {\curvesb};
      \addplot[context] table[x=n, y=gaim_exact_v4] {\curvesb};
      \addplot[context] table[x=n, y=gaim_exact_v5] {\curvesb};
      \addplot[context] table[x=n, y=gaim_omp_v1] {\curvesb};
      \addplot[context] table[x=n, y=gaim_omp_v2] {\curvesb};
      \addplot[context] table[x=n, y=gaim_omp_v3] {\curvesb};
      \addplot[context] table[x=n, y=gaim_omp_v4] {\curvesb};
      \addplot[context] table[x=n, y=gaim_omp_v5] {\curvesb};
      \addplot[context] table[x=n, y=qdaim_exact_v1] {\curvesb};
      \addplot[context] table[x=n, y=qdaim_exact_v5] {\curvesb};
      \addplot[context] table[x=n, y=qdaim_rff_v1] {\curvesb};
      \addplot[context] table[x=n, y=qdaim_rff_v5] {\curvesb};
      \addplot[mean value] coordinates {(-15,0.16892) (527,0.16892)};
      \addplot[v1, mark phase=2] table[x=n, y=gaim_mp_v1] {\curvesb};
      \addplot[v2, mark phase=4] table[x=n, y=gaim_mp_v2] {\curvesb};
      \addplot[v3, mark phase=6] table[x=n, y=gaim_mp_v3] {\curvesb};
      \addplot[v4, mark phase=8] table[x=n, y=gaim_mp_v4] {\curvesb};
      \addplot[v5, mark phase=3] table[x=n, y=gaim_mp_v5] {\curvesb};
    \nextgroupplot[ymin=0.1189, ymax=0.1714, xlabel={number of nodes $n$}]
      \addplot[context] table[x=n, y=gaim_exact_v1] {\curvesb};
      \addplot[context] table[x=n, y=gaim_exact_v2] {\curvesb};
      \addplot[context] table[x=n, y=gaim_exact_v3] {\curvesb};
      \addplot[context] table[x=n, y=gaim_exact_v4] {\curvesb};
      \addplot[context] table[x=n, y=gaim_exact_v5] {\curvesb};
      \addplot[context] table[x=n, y=gaim_mp_v1] {\curvesb};
      \addplot[context] table[x=n, y=gaim_mp_v2] {\curvesb};
      \addplot[context] table[x=n, y=gaim_mp_v3] {\curvesb};
      \addplot[context] table[x=n, y=gaim_mp_v4] {\curvesb};
      \addplot[context] table[x=n, y=gaim_mp_v5] {\curvesb};
      \addplot[context] table[x=n, y=qdaim_exact_v1] {\curvesb};
      \addplot[context] table[x=n, y=qdaim_exact_v5] {\curvesb};
      \addplot[context] table[x=n, y=qdaim_rff_v1] {\curvesb};
      \addplot[context] table[x=n, y=qdaim_rff_v5] {\curvesb};
      \addplot[mean value] coordinates {(-15,0.16892) (527,0.16892)};
      \addplot[v1, mark phase=2] table[x=n, y=gaim_omp_v1] {\curvesb};
      \addplot[v2, mark phase=4] table[x=n, y=gaim_omp_v2] {\curvesb};
      \addplot[v3, mark phase=6] table[x=n, y=gaim_omp_v3] {\curvesb};
      \addplot[v4, mark phase=8] table[x=n, y=gaim_omp_v4] {\curvesb};
      \addplot[v5, mark phase=3] table[x=n, y=gaim_omp_v5] {\curvesb};
    \nextgroupplot[ymin=0.1189, ymax=0.1714]
      \addplot[context] table[x=n, y=gaim_exact_v1] {\curvesb};
      \addplot[context] table[x=n, y=gaim_exact_v2] {\curvesb};
      \addplot[context] table[x=n, y=gaim_exact_v3] {\curvesb};
      \addplot[context] table[x=n, y=gaim_exact_v4] {\curvesb};
      \addplot[context] table[x=n, y=gaim_exact_v5] {\curvesb};
      \addplot[context] table[x=n, y=gaim_mp_v1] {\curvesb};
      \addplot[context] table[x=n, y=gaim_mp_v2] {\curvesb};
      \addplot[context] table[x=n, y=gaim_mp_v3] {\curvesb};
      \addplot[context] table[x=n, y=gaim_mp_v4] {\curvesb};
      \addplot[context] table[x=n, y=gaim_mp_v5] {\curvesb};
      \addplot[context] table[x=n, y=gaim_omp_v1] {\curvesb};
      \addplot[context] table[x=n, y=gaim_omp_v2] {\curvesb};
      \addplot[context] table[x=n, y=gaim_omp_v3] {\curvesb};
      \addplot[context] table[x=n, y=gaim_omp_v4] {\curvesb};
      \addplot[context] table[x=n, y=gaim_omp_v5] {\curvesb};
      \addplot[context] table[x=n, y=qdaim_rff_v1] {\curvesb};
      \addplot[context] table[x=n, y=qdaim_rff_v5] {\curvesb};
      \addplot[mean value] coordinates {(-15,0.16892) (527,0.16892)};
      \addplot[v1, mark phase=2] table[x=n, y=qdaim_exact_v1] {\curvesb};
      \addplot[v5, mark phase=3] table[x=n, y=qdaim_exact_v5] {\curvesb};
    \nextgroupplot[ymin=0.1189, ymax=0.1714]
      \addplot[context] table[x=n, y=gaim_exact_v1] {\curvesb};
      \addplot[context] table[x=n, y=gaim_exact_v2] {\curvesb};
      \addplot[context] table[x=n, y=gaim_exact_v3] {\curvesb};
      \addplot[context] table[x=n, y=gaim_exact_v4] {\curvesb};
      \addplot[context] table[x=n, y=gaim_exact_v5] {\curvesb};
      \addplot[context] table[x=n, y=gaim_mp_v1] {\curvesb};
      \addplot[context] table[x=n, y=gaim_mp_v2] {\curvesb};
      \addplot[context] table[x=n, y=gaim_mp_v3] {\curvesb};
      \addplot[context] table[x=n, y=gaim_mp_v4] {\curvesb};
      \addplot[context] table[x=n, y=gaim_mp_v5] {\curvesb};
      \addplot[context] table[x=n, y=gaim_omp_v1] {\curvesb};
      \addplot[context] table[x=n, y=gaim_omp_v2] {\curvesb};
      \addplot[context] table[x=n, y=gaim_omp_v3] {\curvesb};
      \addplot[context] table[x=n, y=gaim_omp_v4] {\curvesb};
      \addplot[context] table[x=n, y=gaim_omp_v5] {\curvesb};
      \addplot[context] table[x=n, y=qdaim_exact_v1] {\curvesb};
      \addplot[context] table[x=n, y=qdaim_exact_v5] {\curvesb};
      \addplot[mean value] coordinates {(-15,0.16892) (527,0.16892)};
      \addplot[v1, mark phase=2] table[x=n, y=qdaim_rff_v1] {\curvesb};
      \addplot[v5, mark phase=3] table[x=n, y=qdaim_rff_v5] {\curvesb};
  \end{groupplot}
  \node[below=4pt, draw, inner sep=1pt, align=center] at (current bounding box.south)
    {\pgfplotslegendfromname{legend-convergence-exact-linf-sink32-curves}\\[-2pt]\pgfplotslegendfromname{legend-convergence-exact-linf-sink32-references}};
\end{tikzpicture}

%% file: tables/rc_head_linf.tex
\begin{table}[!htbp]
  \centering
  \caption{Number of query heads per layer (out of $28$) on which range correction lowers the maximum error in $L^\infty$ norm,
  $\max_{q\in Q}\|\Att(q,K,V)-\widetilde\Att_L(q,K,V)\|_\infty$.
  GAIM exact v2, nodes selected in the $L^\infty$ score norm, $\tsink=32$. In parentheses we give the heads for which the error decreases by more than $10\%$. Last column: heads improved under every variant
  and every budget.}
  \label{tab:rc_heads_improved_linf_layers}
  \begin{tabular}{llcccc}
    \toprule
    Layer & Variant & $n=32$ & $n=128$ & $n=512$ & All \\
    \midrule
    0  
       & v2 & 23 (17) & 22 (16) & 20 (7)  & \multirow{1}{*}{17}\\
    \midrule
    7  
       & v2 & 27 (10) & 26 (10) & 24 (8)  & \multirow{1}{*}{18} \\
    \midrule
    14 
       & v2 & 28 (6)  & 28 (6)  & 26 (2)  & \multirow{1}{*}{23}\\
    \midrule
    21 
       & v2 & 28 (7)  & 28 (8)  & 28 (4)  & \multirow{1}{*}{26}\\
    \bottomrule
  \end{tabular}
\end{table}

%% file: tables/temp_comp.tex
\begin{table}[!htbp]
  \centering
  \caption{Number of query heads per layer (out of $28$) on which the temperature-scaled projection lowers the maximum error in $L^\infty$ norm,
  $\max_{q\in Q}\|\Att(q,K,V)-\widetilde\Att_L(q,K,V)\|_\infty$.
  GAIM exact v4 with the nodes selected in the $L^\infty$ score norm, and Q-DAIM exact v5,
  $\tsink=32$, RC on, the projection is solved in the kernel rescaled as
in~\cite{schroderWildCatNearLinearAttention2026}, $\tau\approx2.04$. In parentheses we give the
  heads for which the error decreases by more than $10\%$. The last column is the number of heads that improved for all $n$.}
  \label{tab:temperature_heads_improved_linf}
  \begin{tabular}{l cccc cccc}
    \toprule
    & \multicolumn{4}{c}{GAIM exact v4} & \multicolumn{4}{c}{Q-DAIM exact v5} \\
    \cmidrule(lr){2-5}\cmidrule(lr){6-9}
    Layer & $n=32$ & $n=128$ & $n=512$ & All & $n=32$ & $n=128$ & $n=512$ & All \\
    \midrule
    0  & 9 (4)   & 9 (8)   & 8 (4)   & 6  & 25 (19) & 25 (19) & 16 (14) & 15 \\
    7  & 20 (6)  & 22 (11) & 24 (14) & 18 & 21 (5)  & 24 (12) & 25 (20) & 19 \\
    14 & 25 (9)  & 24 (17) & 21 (17) & 19 & 25 (0)  & 28 (7)  & 28 (16) & 25 \\
    21 & 25 (4)  & 21 (15) & 22 (17) & 17 & 28 (1)  & 28 (10) & 26 (19) & 26 \\
    \bottomrule
  \end{tabular}
\end{table}

%% file: figures/tikz/projections_gaim_exact_v4_linf_sink32.tex
\begin{tikzpicture}
  \pgfplotsset{every axis/.append style={
      scale only axis, /tikz/font=\footnotesize, label style={align=center}, title style={align=center},
      legend cell align=left,
      legend style={font=\footnotesize, /tikz/every even column/.append style={column sep=7pt}}}}
  \pgfplotsset{
    galerkin/.style={blue, mark=*},
    least squares/.style={red, mark=square*},
    revealed range/.style={cyan!60!black, mark=triangle*},
    temperature/.style={orange!50!black, mark=diamond*},
    no projection/.style={black!55, mark=x, mark size=2.4pt},
    mean value/.style={black, thin, dashdotted, mark=none},
    zero readout/.style={black, thin, densely dotted, mark=none},
    every axis plot/.append style={very thick, mark size=1.8pt, mark options={solid}, forget plot},
    legend image code/.code={\draw[very thick, mark size=1.8pt, mark options={solid}, mark repeat=2, mark phase=2, #1]
      plot coordinates {(0pt,0pt) (10pt,0pt) (20pt,0pt)};},
    band/.style={draw=none, fill, fill opacity=0.12, mark=none},
    range line/.style={thin, densely dashed, mark=none},
    range legend/.style={black, thin, densely dashed, mark=none},
    band legend/.style={legend image code/.code={\fill[black, fill opacity=0.12] (0pt,-3pt) rectangle (20pt,3pt);}},
  }
  \pgfplotstableread[row sep=\\]{
    n galerkin least_squares revealed_range temperature no_projection\\
    1 0.21387 0.21387 0.23629 0.19852 0.21387\\
    17 0.20634 0.2094 4.3534 0.16422 0.20537\\
    33 0.19517 0.18722 8.9436 0.15626 0.20169\\
    65 0.19083 0.18053 8.6938 0.13649 0.2062\\
    97 0.18633 0.16975 12.468 0.11599 0.20097\\
    129 0.17988 0.16484 34.75 0.10611 0.19878\\
    193 0.16676 0.16249 36.222 0.097639 0.18441\\
    257 0.15805 0.15589 14.623 0.09727 0.17467\\
    385 0.15337 0.15314 9.8761 0.088893 0.16382\\
    512 0.1493 0.14793 14.249 0.088729 0.15863\\
  }\curvesa
  \pgfplotstableread[row sep=\\]{
    n galerkin least_squares revealed_range temperature no_projection\\
    1 0.16892 0.16892 0.21416 0.168 0.16892\\
    17 0.16792 0.16791 5.3378 0.14102 0.16792\\
    33 0.1675 0.15794 10.807 0.14018 0.16753\\
    65 0.16808 0.16526 10.654 0.12157 0.16813\\
    97 0.16491 0.15937 15.851 0.10853 0.1671\\
    129 0.15667 0.14494 105.06 0.1029 0.16452\\
    193 0.13823 0.14245 55.443 0.096308 0.15255\\
    257 0.13217 0.12844 19.158 0.091882 0.14116\\
    385 0.12564 0.12268 21.657 0.082763 0.1365\\
    512 0.12085 0.11478 19.423 0.079966 0.12979\\
  }\curvesb
  \begin{groupplot}[
      group style={group size=2 by 1, horizontal sep=8pt, yticklabels at=edge left},
      width=198pt, height=120pt, grid=major, tick label style={font=\scriptsize},
      xmin=-15, xmax=527, xtick={0,100,...,500}, ymin=0.07461, ymax=0.2192,
      scaled y ticks=false, yticklabel style={/pgf/number format/fixed, /pgf/number format/fixed zerofill, /pgf/number format/precision=2}]
    \nextgroupplot[title={RC off}, xlabel={number of nodes $n$}, ylabel={relative attention error $\varepsilon_{\Att}^{10\%}$}, legend to name=legend-projections-gaim-exact-v4-linf-sink32, legend columns=2]
      \addplot[mean value] coordinates {(-15,0.16892) (527,0.16892)};
      \addplot[galerkin] table[x=n, y=galerkin] {\curvesa};
      \addplot[least squares] table[x=n, y=least_squares] {\curvesa};
      \addplot[temperature] table[x=n, y=temperature] {\curvesa};
      \addplot[no projection] table[x=n, y=no_projection] {\curvesa};
      \addlegendimage{galerkin}\addlegendentry{GAIM (Galerkin)}
      \addlegendimage{least squares}\addlegendentry{least squares (direct and Petrov--Galerkin)}
      \addlegendimage{empty legend}\addlegendentry{Q-DAIM revealed range $\geq 4.4$}
      \addlegendimage{temperature}\addlegendentry{WildCat temperature, $\tau\approx2.0$}
      \addlegendimage{no projection}\addlegendentry{no projection ($W = V[\mathfrak L,:]$)}
      \addlegendimage{mean value}\addlegendentry{mean vector approximation}
      \addlegendimage{empty legend}\addlegendentry{zero readout $\approx 0.76$}
    \nextgroupplot[title={RC on}, xlabel={number of nodes $n$}]
      \addplot[mean value] coordinates {(-15,0.16892) (527,0.16892)};
      \addplot[galerkin] table[x=n, y=galerkin] {\curvesb};
      \addplot[least squares] table[x=n, y=least_squares] {\curvesb};
      \addplot[temperature] table[x=n, y=temperature] {\curvesb};
      \addplot[no projection] table[x=n, y=no_projection] {\curvesb};
  \end{groupplot}
  \node[below=4pt] at (current bounding box.south) {\pgfplotslegendfromname{legend-projections-gaim-exact-v4-linf-sink32}};
\end{tikzpicture}

%% file: figures/tikz/projections_qdaim_exact_v5_linf_sink32.tex
\begin{tikzpicture}
  \pgfplotsset{every axis/.append style={
      scale only axis, /tikz/font=\footnotesize, label style={align=center}, title style={align=center},
      legend cell align=left,
      legend style={font=\footnotesize, /tikz/every even column/.append style={column sep=7pt}}}}
  \pgfplotsset{
    galerkin/.style={blue, mark=*},
    least squares/.style={red, mark=square*},
    revealed range/.style={cyan!60!black, mark=triangle*},
    temperature/.style={orange!50!black, mark=diamond*},
    no projection/.style={black!55, mark=x, mark size=2.4pt},
    mean value/.style={black, thin, dashdotted, mark=none},
    zero readout/.style={black, thin, densely dotted, mark=none},
    every axis plot/.append style={very thick, mark size=1.8pt, mark options={solid}, forget plot},
    legend image code/.code={\draw[very thick, mark size=1.8pt, mark options={solid}, mark repeat=2, mark phase=2, #1]
      plot coordinates {(0pt,0pt) (10pt,0pt) (20pt,0pt)};},
    band/.style={draw=none, fill, fill opacity=0.12, mark=none},
    range line/.style={thin, densely dashed, mark=none},
    range legend/.style={black, thin, densely dashed, mark=none},
    band legend/.style={legend image code/.code={\fill[black, fill opacity=0.12] (0pt,-3pt) rectangle (20pt,3pt);}},
  }
  \pgfplotstableread[row sep=\\]{
    n galerkin least_squares revealed_range temperature no_projection\\
    1 0.20723 0.20723 0.20723 0.19909 0.20723\\
    17 0.21325 0.21325 0.21322 0.19478 0.21325\\
    33 0.21244 0.21244 0.21222 0.18158 0.21244\\
    65 0.21122 0.21122 0.21034 0.17587 0.21123\\
    97 0.21457 0.21457 0.21289 0.16397 0.21457\\
    129 0.20626 0.20625 0.19687 0.15476 0.20629\\
    193 0.20622 0.20621 0.18757 0.14536 0.20623\\
    257 0.19913 0.19719 0.18515 0.13188 0.1992\\
    385 0.18984 0.1866 0.17849 0.11695 0.193\\
    512 0.17803 0.17607 0.17081 0.1061 0.18907\\
  }\curvesa
  \pgfplotstableread[row sep=\\]{
    n galerkin least_squares revealed_range temperature no_projection\\
    1 0.16892 0.16892 0.16892 0.1688 0.16892\\
    17 0.16811 0.16808 0.16805 0.16601 0.16811\\
    33 0.1674 0.1674 0.16727 0.16496 0.1674\\
    65 0.16632 0.16631 0.16814 0.16158 0.16632\\
    97 0.1675 0.1675 0.16627 0.15007 0.1675\\
    129 0.16408 0.16408 0.15786 0.13868 0.16516\\
    193 0.16359 0.16356 0.15645 0.12877 0.16361\\
    257 0.15921 0.1592 0.14962 0.11796 0.15924\\
    385 0.15422 0.1542 0.14019 0.10786 0.15424\\
    512 0.14652 0.14469 0.13488 0.097368 0.14772\\
  }\curvesb
  \begin{groupplot}[
      group style={group size=2 by 1, horizontal sep=8pt, yticklabels at=edge left},
      width=198pt, height=120pt, grid=major, tick label style={font=\scriptsize},
      xmin=-15, xmax=527, xtick={0,100,...,500}, ymin=0.09268, ymax=0.2193,
      scaled y ticks=false, yticklabel style={/pgf/number format/fixed, /pgf/number format/fixed zerofill, /pgf/number format/precision=2}]
    \nextgroupplot[title={RC off}, xlabel={number of nodes $n$}, ylabel={relative attention error $\varepsilon_{\Att}^{10\%}$}, legend to name=legend-projections-qdaim-exact-v5-linf-sink32, legend columns=2]
      \addplot[mean value] coordinates {(-15,0.16892) (527,0.16892)};
      \addplot[galerkin] table[x=n, y=galerkin] {\curvesa};
      \addplot[least squares] table[x=n, y=least_squares] {\curvesa};
      \addplot[revealed range] table[x=n, y=revealed_range] {\curvesa};
      \addplot[temperature] table[x=n, y=temperature] {\curvesa};
      \addplot[no projection] table[x=n, y=no_projection] {\curvesa};
      \addlegendimage{galerkin}\addlegendentry{GAIM (Galerkin)}
      \addlegendimage{least squares}\addlegendentry{least squares (direct and Petrov--Galerkin)}
      \addlegendimage{revealed range}\addlegendentry{Q-DAIM revealed range}
      \addlegendimage{temperature}\addlegendentry{WildCat temperature, $\tau\approx2.0$}
      \addlegendimage{no projection}\addlegendentry{no projection ($W = V[\mathfrak L,:]$)}
      \addlegendimage{mean value}\addlegendentry{mean vector approximation}
      \addlegendimage{empty legend}\addlegendentry{zero readout $\approx 0.76$}
    \nextgroupplot[title={RC on}, xlabel={number of nodes $n$}]
      \addplot[mean value] coordinates {(-15,0.16892) (527,0.16892)};
      \addplot[galerkin] table[x=n, y=galerkin] {\curvesb};
      \addplot[least squares] table[x=n, y=least_squares] {\curvesb};
      \addplot[revealed range] table[x=n, y=revealed_range] {\curvesb};
      \addplot[temperature] table[x=n, y=temperature] {\curvesb};
      \addplot[no projection] table[x=n, y=no_projection] {\curvesb};
  \end{groupplot}
  \node[below=4pt] at (current bounding box.south) {\pgfplotslegendfromname{legend-projections-qdaim-exact-v5-linf-sink32}};
\end{tikzpicture}

%% file: figures/tikz/global_solve_ratio_cs256_p25_sink32.tex
\begin{tikzpicture}
  \pgfplotsset{every axis/.append style={
      scale only axis, /tikz/font=\footnotesize, label style={align=center}, title style={align=center},
      legend cell align=left,
      legend style={font=\footnotesize, /tikz/every even column/.append style={column sep=7pt}}}}
  \pgfplotsset{
    spread/.style={only marks, mark=none, blue!50, error bars/.cd, x dir=both, x explicit, error mark=none, error bar style={thin}},
    quartiles/.style={only marks, mark=*, mark size=2pt, blue, error bars/.cd, x dir=both, x explicit, error mark=none,
      error bar style={line width=2pt}},
  }
  \pgfplotstableread[row sep=\\]{
    row median qlow qhigh plow phigh\\
    4 0.99246 0.01489 0.0050503 0.055502 0.0064995\\
    3 0.98889 0.021481 0.0090512 0.10811 0.010706\\
    2 0.99622 0.0068284 0.0026099 0.047314 0.003238\\
    1 0.99789 0.0041645 0.0019181 0.010656 0.0020383\\
    0 0.99716 0.009173 0.0020172 0.040711 0.0027201\\
  }\ratiosa
  \pgfplotstableread[row sep=\\]{
    row median qlow qhigh plow phigh\\
    4 0.99338 0.01431 0.0045011 0.039427 0.0056556\\
    3 0.98868 0.015217 0.009209 0.10939 0.011109\\
    2 0.99689 0.0077416 0.0022008 0.043761 0.0027086\\
    1 0.9989 0.0034211 0.00096638 0.010332 0.0010695\\
    0 0.99737 0.0070087 0.0020395 0.017199 0.0025345\\
  }\ratiosb
  \pgfplotstableread[row sep=\\]{
    row median qlow qhigh plow phigh\\
    4 1 6.698e-05 2.7653e-05 0.0010171 9.0965e-05\\
    3 0.99995 0.00013637 3.5321e-05 0.00047575 4.7471e-05\\
    2 0.99998 0.00011848 2.3534e-05 0.0010452 5.8948e-05\\
    1 1 1.4415e-05 3.4942e-05 0.00013173 0.00052346\\
    0 1 3.947e-05 3.044e-05 0.00053125 0.00052626\\
  }\ratiosc
  \begin{groupplot}[
      group style={group size=3 by 1, horizontal sep=10pt, yticklabels at=edge left},
      width=108pt, height=108pt, xmajorgrids, scaled x ticks=false, xlabel={global $\div$ segment-local},
      xticklabel style={/pgf/number format/fixed, /pgf/number format/precision=2},
      ymin=-0.6, ymax=4.6, ytick={0,...,4}, ytick style={draw=none},
      yticklabels={{Q-DAIM RFF [v5]}, {Q-DAIM exact [v5]}, {GAIM-LIN-O [v1, $L^\infty$]}, {GAIM-LIN [v1, $L^\infty$]}, {GAIM exact [v3, $L^\infty$]}}]
    \nextgroupplot[title={future queries, RC off}, xmin=0.859, xmax=1.02]
      \draw[black] (axis cs:1,-0.6) -- (axis cs:1,4.6);
      \addplot[spread] table[x=median, y=row, x error minus=plow, x error plus=phigh] {\ratiosa};
      \addplot[quartiles] table[x=median, y=row, x error minus=qlow, x error plus=qhigh] {\ratiosa};
      \node[anchor=south east, font=\tiny, inner sep=1pt] at ({rel axis cs:1,0}|-{axis cs:0,4.12}) {worse in 3\%};
      \node[anchor=south east, font=\tiny, inner sep=1pt] at ({rel axis cs:1,0}|-{axis cs:0,3.12}) {worse in 5\%};
      \node[anchor=south east, font=\tiny, inner sep=1pt] at ({rel axis cs:1,0}|-{axis cs:0,2.12}) {worse in 3\%};
      \node[anchor=south east, font=\tiny, inner sep=1pt] at ({rel axis cs:1,0}|-{axis cs:0,1.12}) {worse in 0\%};
      \node[anchor=south east, font=\tiny, inner sep=1pt] at ({rel axis cs:1,0}|-{axis cs:0,0.12}) {worse in 0\%};
    \nextgroupplot[title={future queries, RC on}, xmin=0.859, xmax=1.02]
      \draw[black] (axis cs:1,-0.6) -- (axis cs:1,4.6);
      \addplot[spread] table[x=median, y=row, x error minus=plow, x error plus=phigh] {\ratiosb};
      \addplot[quartiles] table[x=median, y=row, x error minus=qlow, x error plus=qhigh] {\ratiosb};
      \node[anchor=south east, font=\tiny, inner sep=1pt] at ({rel axis cs:1,0}|-{axis cs:0,4.12}) {worse in 4\%};
      \node[anchor=south east, font=\tiny, inner sep=1pt] at ({rel axis cs:1,0}|-{axis cs:0,3.12}) {worse in 6\%};
      \node[anchor=south east, font=\tiny, inner sep=1pt] at ({rel axis cs:1,0}|-{axis cs:0,2.12}) {worse in 1\%};
      \node[anchor=south east, font=\tiny, inner sep=1pt] at ({rel axis cs:1,0}|-{axis cs:0,1.12}) {worse in 1\%};
      \node[anchor=south east, font=\tiny, inner sep=1pt] at ({rel axis cs:1,0}|-{axis cs:0,0.12}) {worse in 1\%};
    \nextgroupplot[title={fit at the keys}, xmin=0.998694, xmax=1.00077, xticklabel style={/pgf/number format/precision=4}]
      \draw[black] (axis cs:1,-0.6) -- (axis cs:1,4.6);
      \addplot[spread] table[x=median, y=row, x error minus=plow, x error plus=phigh] {\ratiosc};
      \addplot[quartiles] table[x=median, y=row, x error minus=qlow, x error plus=qhigh] {\ratiosc};
  \end{groupplot}
\end{tikzpicture}

%% file: figures/tikz/graph3_heatmap_aug_sink32_rc.tex
\begin{tikzpicture}
  \pgfplotsset{every axis/.append style={
      scale only axis, /tikz/font=\footnotesize, label style={align=center}, title style={align=center},
      legend cell align=left,
      legend style={font=\footnotesize, /tikz/every even column/.append style={column sep=7pt}}}}
  \protected\def\sci#1#2{\shortstack{#1\\[-1.5pt]e#2}}
  \pgfplotsset{
    colormap={change}{color=(blue!65) color=(white) color=(red!65)},
    colormap={residual}{color=(blue!4) color=(blue!55)},
    cells/.style={matrix plot*, point meta=explicit, mesh/cols=5},
    numbers/.style={only marks, mark=none, point meta=explicit symbolic, nodes near coords,
      every node near coord/.append style={anchor=center, inner sep=0pt, black, font=\fontsize{5.5}{6}\selectfont}},
  }
  \pgfplotstableread[row sep=\\]{
    x y shade text\\
    0 0 1.7833 61\\
    1 0 1.6826 48\\
    2 0 1.3897 25\\
    3 0 1.3039 20\\
    4 0 1.1214 13\\
    0 1 1.8561 72\\
    1 1 1.7904 62\\
    2 1 1.563 37\\
    3 1 1.2423 17\\
    4 1 1.1146 13\\
    0 2 1.8807 76\\
    1 2 1.8308 68\\
    2 2 1.61 41\\
    3 2 1.4906 31\\
    4 2 1.2343 17\\
    0 3 1.8535 71\\
    1 3 1.7365 55\\
    2 3 1.5643 37\\
    3 3 1.3478 22\\
    4 3 1.2408 17\\
    0 4 1.8939 78\\
    1 4 1.8794 76\\
    2 4 1.7251 53\\
    3 4 1.6011 40\\
    4 4 1.3638 23\\
  }\gridablock
  \pgfplotstableread[row sep=\\]{
    x y shade text\\
    0 0 0.87315 7.5\\
    1 0 0.82099 6.6\\
    2 0 0.24592 1.8\\
    3 0 0.026281 1.1\\
    4 0 -0.23039 0.59\\
    0 1 1.0329 11\\
    1 1 0.71481 5.2\\
    2 1 0.25779 1.8\\
    3 1 0.024126 1.1\\
    4 1 -0.11128 0.77\\
    0 2 0.97294 9.4\\
    1 2 0.74777 5.6\\
    2 2 0.35937 2.3\\
    3 2 0.12709 1.3\\
    4 2 -0.1446 0.72\\
    0 3 1.0033 10\\
    1 3 0.81268 6.5\\
    2 3 0.54033 3.5\\
    3 3 0.17382 1.5\\
    4 3 -0.27347 0.53\\
    0 4 0.96553 9.2\\
    1 4 0.8514 7.1\\
    2 4 0.60453 4.0\\
    3 4 0.28937 1.9\\
    4 4 0.23968 1.7\\
  }\gridbblock
  \pgfplotstableread[row sep=\\]{
    x y shade text\\
    0 0 0.64903 4.5\\
    1 0 0.2801 1.9\\
    2 0 -0.068045 0.85\\
    3 0 -0.45922 0.35\\
    4 0 -1.0884 0.082\\
    0 1 0.73591 5.4\\
    1 1 0.32555 2.1\\
    2 1 -0.1147 0.77\\
    3 1 -0.48557 0.33\\
    4 1 -1.0574 0.088\\
    0 2 0.4674 2.9\\
    1 2 0.24547 1.8\\
    2 2 -0.046705 0.90\\
    3 2 -0.37688 0.42\\
    4 2 -0.83979 0.14\\
    0 3 0.41504 2.6\\
    1 3 0.24326 1.8\\
    2 3 0.02656 1.1\\
    3 3 -0.20744 0.62\\
    4 3 -0.45653 0.35\\
    0 4 0.5642 3.7\\
    1 4 0.36289 2.3\\
    2 4 0.22186 1.7\\
    3 4 -0.039093 0.91\\
    4 4 -0.33314 0.46\\
  }\gridcblock
  \pgfplotstableread[row sep=\\]{
    x y shade text\\
    0 0 1.3533 23\\
    1 0 1.0547 11\\
    2 0 0.99197 9.8\\
    3 0 0.87172 7.4\\
    4 0 0.85255 7.1\\
    0 1 0.79353 6.2\\
    1 1 0.43775 2.7\\
    2 1 -0.23139 0.59\\
    3 1 -0.71353 0.19\\
    4 1 -1.0672 0.086\\
    0 2 0.45596 2.9\\
    1 2 0.063975 1.2\\
    2 2 -0.2672 0.54\\
    3 2 -0.81694 0.15\\
    4 2 -1.6164 0.024\\
    0 3 0.2223 1.7\\
    1 3 -0.057243 0.88\\
    2 3 -0.49137 0.32\\
    3 3 -0.89863 0.13\\
    4 3 -1.2691 0.054\\
    0 4 0.13786 1.4\\
    1 4 -0.18384 0.65\\
    2 4 -0.53928 0.29\\
    3 4 -0.9792 0.10\\
    4 4 -1.369 0.043\\
  }\griddblock
  \pgfplotstableread[row sep=\\]{
    x y shade text\\
    0 0 1.9872 97\\
    1 0 1.9607 91\\
    2 0 1.9087 81\\
    3 0 1.6959 50\\
    4 0 1.2304 17\\
    0 1 1.4742 30\\
    1 1 1.0573 11\\
    2 1 0.36156 2.3\\
    3 1 -0.72045 0.19\\
    4 1 -1.0721 0.085\\
    0 2 0.9254 8.4\\
    1 2 0.061401 1.2\\
    2 2 -0.26453 0.54\\
    3 2 -0.80905 0.16\\
    4 2 -1.1435 0.072\\
    0 3 0.22105 1.7\\
    1 3 -0.056204 0.88\\
    2 3 -0.48199 0.33\\
    3 3 -0.84456 0.14\\
    4 3 -1.1298 0.074\\
    0 4 0.14349 1.4\\
    1 4 -0.1828 0.66\\
    2 4 -0.49016 0.32\\
    3 4 -0.78302 0.16\\
    4 4 -1.0372 0.092\\
  }\grideblock
  \pgfplotstableread[row sep=\\]{
    x y shade text\\
    0 0 3.6798 \sci{+4.8}{0}\\
    1 0 3.5722 \sci{+3.7}{0}\\
    2 0 3.5766 \sci{+3.8}{0}\\
    3 0 2.7959 \sci{+6.2}{\textminus1}\\
    4 0 -1.9275 \sci{\textminus8.4}{\textminus2}\\
    0 1 -3.0088 \sci{\textminus1.0}{0}\\
    1 1 -3.3464 \sci{\textminus2.2}{0}\\
    2 1 2.4244 \sci{+2.6}{\textminus1}\\
    3 1 3.9655 \sci{+9.2}{0}\\
    4 1 1.4937 \sci{+3.0}{\textminus2}\\
    0 2 3.1133 \sci{+1.3}{0}\\
    1 2 -3.8619 \sci{\textminus7.3}{0}\\
    2 2 3.0049 \sci{+1.0}{0}\\
    3 2 3.4413 \sci{+2.8}{0}\\
    4 2 3.0513 \sci{+1.1}{0}\\
    0 3 3.5192 \sci{+3.3}{0}\\
    1 3 3.1912 \sci{+1.6}{0}\\
    2 3 2.8982 \sci{+7.9}{\textminus1}\\
    3 3 1.9626 \sci{+9.1}{\textminus2}\\
    4 3 2.4156 \sci{+2.6}{\textminus1}\\
    0 4 -3.3529 \sci{\textminus2.3}{0}\\
    1 4 -2.923 \sci{\textminus8.4}{\textminus1}\\
    2 4 3.4144 \sci{+2.6}{0}\\
    3 4 -3.7633 \sci{\textminus5.8}{0}\\
    4 4 -3.2463 \sci{\textminus1.8}{0}\\
  }\gridachange
  \pgfplotstableread[row sep=\\]{
    x y shade text\\
    0 0 -8.676e-07 \sci{\textminus2.0}{\textminus9}\\
    1 0 -1.1648e-06 \sci{\textminus2.7}{\textminus9}\\
    2 0 -0.099505 \sci{\textminus2.6}{\textminus4}\\
    3 0 -0.90394 \sci{\textminus7.0}{\textminus3}\\
    4 0 -1.1855 \sci{\textminus1.4}{\textminus2}\\
    0 1 -0.0011878 \sci{\textminus2.7}{\textminus6}\\
    1 1 0.65627 \sci{+3.5}{\textminus3}\\
    2 1 -3.0356e-05 \sci{\textminus7.0}{\textminus8}\\
    3 1 1.0582e-06 \sci{+2.4}{\textminus9}\\
    4 1 -0.89623 \sci{\textminus6.9}{\textminus3}\\
    0 2 -0.0031053 \sci{\textminus7.2}{\textminus6}\\
    1 2 8.2013e-06 \sci{+1.9}{\textminus8}\\
    2 2 -1.4026e-05 \sci{\textminus3.2}{\textminus8}\\
    3 2 1.3728 \sci{+2.3}{\textminus2}\\
    4 2 1.0744 \sci{+1.1}{\textminus2}\\
    0 3 -0.00035735 \sci{\textminus8.2}{\textminus7}\\
    1 3 0.032292 \sci{+7.7}{\textminus5}\\
    2 3 -1.7442e-06 \sci{\textminus4.0}{\textminus9}\\
    3 3 -0.16902 \sci{\textminus4.8}{\textminus4}\\
    4 3 -0.13874 \sci{\textminus3.8}{\textminus4}\\
    0 4 2.9658e-06 \sci{+6.8}{\textminus9}\\
    1 4 -6.5269e-07 \sci{\textminus1.5}{\textminus9}\\
    2 4 -6.473e-09 \sci{\textminus1.5}{\textminus11}\\
    3 4 3.0135e-09 \sci{+6.9}{\textminus12}\\
    4 4 -7.4398e-09 \sci{\textminus1.7}{\textminus11}\\
  }\gridbchange
  \pgfplotstableread[row sep=\\]{
    x y shade text\\
    0 0 -0.0040835 \sci{\textminus9.4}{\textminus6}\\
    1 0 -2.7287e-07 \sci{\textminus6.3}{\textminus10}\\
    2 0 -7.4978e-06 \sci{\textminus1.7}{\textminus8}\\
    3 0 -0.0010408 \sci{\textminus2.4}{\textminus6}\\
    4 0 -5.2872e-05 \sci{\textminus1.2}{\textminus7}\\
    0 1 -0.0038664 \sci{\textminus8.9}{\textminus6}\\
    1 1 1.3345e-05 \sci{+3.1}{\textminus8}\\
    2 1 -1.4119e-05 \sci{\textminus3.3}{\textminus8}\\
    3 1 3.4587e-06 \sci{+8.0}{\textminus9}\\
    4 1 -0.00030214 \sci{\textminus7.0}{\textminus7}\\
    0 2 6.2925e-06 \sci{+1.4}{\textminus8}\\
    1 2 -5.7508e-07 \sci{\textminus1.3}{\textminus9}\\
    2 2 -5.9584e-07 \sci{\textminus1.4}{\textminus9}\\
    3 2 2.3508e-06 \sci{+5.4}{\textminus9}\\
    4 2 -7.1314e-05 \sci{\textminus1.6}{\textminus7}\\
    0 3 -8.4035e-07 \sci{\textminus1.9}{\textminus9}\\
    1 3 1.6056e-07 \sci{+3.7}{\textminus10}\\
    2 3 2.0781e-08 \sci{+4.8}{\textminus11}\\
    3 3 -3.9265e-07 \sci{\textminus9.0}{\textminus10}\\
    4 3 2.3395e-08 \sci{+5.4}{\textminus11}\\
    0 4 2.089e-05 \sci{+4.8}{\textminus8}\\
    1 4 -8.6602e-07 \sci{\textminus2.0}{\textminus9}\\
    2 4 -2.6384e-08 \sci{\textminus6.1}{\textminus11}\\
    3 4 -8.1739e-08 \sci{\textminus1.9}{\textminus10}\\
    4 4 7.1336e-09 \sci{+1.6}{\textminus11}\\
  }\gridcchange
  \pgfplotstableread[row sep=\\]{
    x y shade text\\
    0 0 3.3888 \sci{+2.4}{0}\\
    1 0 -2.0635 \sci{\textminus1.1}{\textminus1}\\
    2 0 -1.2369 \sci{\textminus1.6}{\textminus2}\\
    3 0 2.127 \sci{+1.3}{\textminus1}\\
    4 0 -2.2317 \sci{\textminus1.7}{\textminus1}\\
    0 1 -1.2316e-05 \sci{\textminus2.8}{\textminus8}\\
    1 1 -9.786e-10 \sci{\textminus2.3}{\textminus12}\\
    2 1 -9.235e-07 \sci{\textminus2.1}{\textminus9}\\
    3 1 3.968e-07 \sci{+9.1}{\textminus10}\\
    4 1 1.3513e-07 \sci{+3.1}{\textminus10}\\
    0 2 -7.4412e-07 \sci{\textminus1.7}{\textminus9}\\
    1 2 1.2672e-07 \sci{+2.9}{\textminus10}\\
    2 2 3.1781e-06 \sci{+7.3}{\textminus9}\\
    3 2 0.7214 \sci{+4.3}{\textminus3}\\
    4 2 1.6854 \sci{+4.7}{\textminus2}\\
    0 3 -4.842e-06 \sci{\textminus1.1}{\textminus8}\\
    1 3 1.082e-07 \sci{+2.5}{\textminus10}\\
    2 3 -0.54609 \sci{\textminus2.5}{\textminus3}\\
    3 3 0.42792 \sci{+1.7}{\textminus3}\\
    4 3 -0.29422 \sci{\textminus9.7}{\textminus4}\\
    0 4 -3.2719e-07 \sci{\textminus7.5}{\textminus10}\\
    1 4 -0.88355 \sci{\textminus6.6}{\textminus3}\\
    2 4 -0.44815 \sci{\textminus1.8}{\textminus3}\\
    3 4 0.43602 \sci{+1.7}{\textminus3}\\
    4 4 0.20917 \sci{+6.2}{\textminus4}\\
  }\griddchange
  \pgfplotstableread[row sep=\\]{
    x y shade text\\
    0 0 -0.13745 \sci{\textminus3.7}{\textminus4}\\
    1 0 -0.00082169 \sci{\textminus1.9}{\textminus6}\\
    2 0 -6.0167e-05 \sci{\textminus1.4}{\textminus7}\\
    3 0 -0.00012994 \sci{\textminus3.0}{\textminus7}\\
    4 0 0.37113 \sci{+1.4}{\textminus3}\\
    0 1 -0.00022594 \sci{\textminus5.2}{\textminus7}\\
    1 1 -2.9865 \sci{\textminus9.7}{\textminus1}\\
    2 1 4.8489e-05 \sci{+1.1}{\textminus7}\\
    3 1 3.8464e-07 \sci{+8.9}{\textminus10}\\
    4 1 3.763e-07 \sci{+8.7}{\textminus10}\\
    0 2 -7.3165e-07 \sci{\textminus1.7}{\textminus9}\\
    1 2 -1.7695e-07 \sci{\textminus4.1}{\textminus10}\\
    2 2 2.9789e-06 \sci{+6.9}{\textminus9}\\
    3 2 -0.33931 \sci{\textminus1.2}{\textminus3}\\
    4 2 -1.4747 \sci{\textminus2.9}{\textminus2}\\
    0 3 -1.3979e-06 \sci{\textminus3.2}{\textminus9}\\
    1 3 5.4446e-08 \sci{+1.3}{\textminus10}\\
    2 3 -0.94677 \sci{\textminus7.8}{\textminus3}\\
    3 3 -0.77903 \sci{\textminus5.0}{\textminus3}\\
    4 3 -0.11209 \sci{\textminus2.9}{\textminus4}\\
    0 4 -7.3197e-07 \sci{\textminus1.7}{\textminus9}\\
    1 4 -1.1361 \sci{\textminus1.3}{\textminus2}\\
    2 4 0.30298 \sci{+1.0}{\textminus3}\\
    3 4 -1.0011 \sci{\textminus9.0}{\textminus3}\\
    4 4 -0.50484 \sci{\textminus2.2}{\textminus3}\\
  }\gridechange
  \begin{groupplot}[
      group style={group size=5 by 2, horizontal sep=5.0pt, vertical sep=5.0pt,
        xticklabels at=edge bottom, yticklabels at=edge left},
      width=64.5pt, height=62.0pt, tick label style={font=\scriptsize},
      enlargelimits=false, axis on top, major tick length=0pt,
      xmin=-0.5, xmax=4.5, ymin=-0.5, ymax=4.5,
      xtick={0,...,4}, xticklabels={5,10,20,35,50},
      ytick={0,...,4}, yticklabels={128,256,512,1024,2048}]
    \nextgroupplot[colormap name=residual, point meta min=-1.6164, point meta max=1.9872, title={GAIM exact\\{[v3, $L^\infty$]}}, ylabel={block\\\scriptsize segment size $\Nc$}]
      \addplot[cells] table[x=x, y=y, meta=shade] {\gridablock};
      \addplot[numbers] table[x=x, y=y, meta=text] {\gridablock};
    \nextgroupplot[colormap name=residual, point meta min=-1.6164, point meta max=1.9872, title={GAIM-LIN\\{[v1, $L^\infty$]}}]
      \addplot[cells] table[x=x, y=y, meta=shade] {\gridbblock};
      \addplot[numbers] table[x=x, y=y, meta=text] {\gridbblock};
    \nextgroupplot[colormap name=residual, point meta min=-1.6164, point meta max=1.9872, title={GAIM-LIN-O\\{[v1, $L^\infty$]}}]
      \addplot[cells] table[x=x, y=y, meta=shade] {\gridcblock};
      \addplot[numbers] table[x=x, y=y, meta=text] {\gridcblock};
    \nextgroupplot[colormap name=residual, point meta min=-1.6164, point meta max=1.9872, title={Q-DAIM exact\\{[v5]}}]
      \addplot[cells] table[x=x, y=y, meta=shade] {\griddblock};
      \addplot[numbers] table[x=x, y=y, meta=text] {\griddblock};
    \nextgroupplot[colormap name=residual, point meta min=-1.6164, point meta max=1.9872, title={Q-DAIM RFF\\{[v5]}}]
      \addplot[cells] table[x=x, y=y, meta=shade] {\grideblock};
      \addplot[numbers] table[x=x, y=y, meta=text] {\grideblock};
    \nextgroupplot[colormap name=change, point meta min=-3.965, point meta max=3.965, ylabel={streaming $-$ block\\\scriptsize segment size $\Nc$}]
      \addplot[cells] table[x=x, y=y, meta=shade] {\gridachange};
      \addplot[numbers] table[x=x, y=y, meta=text] {\gridachange};
    \nextgroupplot[colormap name=change, point meta min=-3.965, point meta max=3.965]
      \addplot[cells] table[x=x, y=y, meta=shade] {\gridbchange};
      \addplot[numbers] table[x=x, y=y, meta=text] {\gridbchange};
    \nextgroupplot[colormap name=change, point meta min=-3.965, point meta max=3.965, xlabel={nodes per segment $\nc/\Nc$ (\%)}]
      \addplot[cells] table[x=x, y=y, meta=shade] {\gridcchange};
      \addplot[numbers] table[x=x, y=y, meta=text] {\gridcchange};
    \nextgroupplot[colormap name=change, point meta min=-3.965, point meta max=3.965]
      \addplot[cells] table[x=x, y=y, meta=shade] {\griddchange};
      \addplot[numbers] table[x=x, y=y, meta=text] {\griddchange};
    \nextgroupplot[colormap name=change, point meta min=-3.965, point meta max=3.965]
      \addplot[cells] table[x=x, y=y, meta=shade] {\gridechange};
      \addplot[numbers] table[x=x, y=y, meta=text] {\gridechange};
  \end{groupplot}
  \begin{axis}[hide axis, at={(group c5r1.north east)}, anchor=north west,
      width=1pt, height=62.0pt, xmin=0, xmax=1, ymin=0, ymax=1,
      colormap name=residual, point meta min=-1.6164, point meta max=1.9872,
      colorbar, colorbar style={at={(parent axis.north east)}, anchor=north west, xshift=4pt, width=7pt,
        tick label style={font=\scriptsize}, ytick={-1,0,1}, yticklabels={0.1,1,10}, ylabel={operator norm (\%)}, ylabel style={font=\scriptsize}}]
  \end{axis}
  \begin{axis}[hide axis, at={(group c5r2.north east)}, anchor=north west,
      width=1pt, height=62.0pt, xmin=0, xmax=1, ymin=0, ymax=1,
      colormap name=change, point meta min=-3.9655, point meta max=3.9655,
      colorbar, colorbar style={at={(parent axis.north east)}, anchor=north west, xshift=4pt, width=7pt,
        tick label style={font=\scriptsize}, ytick={-3,-1.041,0,1.041,3}, yticklabels={$-1$,$-10^{-2}$,$0$,$+10^{-2}$,$+1$}, ylabel={difference}, ylabel style={font=\scriptsize}}]
  \end{axis}
\end{tikzpicture}

%% file: figures/tikz/graph3_realq_heatmap_aug_sink32_rc_recent32.tex
\begin{tikzpicture}
  \pgfplotsset{every axis/.append style={
      scale only axis, /tikz/font=\footnotesize, label style={align=center}, title style={align=center},
      legend cell align=left,
      legend style={font=\footnotesize, /tikz/every even column/.append style={column sep=7pt}}}}
  \protected\def\sci#1#2{\shortstack{#1\\[-1.5pt]e#2}}
  \pgfplotsset{
    colormap={change}{color=(blue!65) color=(white) color=(red!65)},
    colormap={residual}{color=(blue!4) color=(blue!55)},
    cells/.style={matrix plot*, point meta=explicit, mesh/cols=5},
    numbers/.style={only marks, mark=none, point meta=explicit symbolic, nodes near coords,
      every node near coord/.append style={anchor=center, inner sep=0pt, black, font=\fontsize{5.5}{6}\selectfont}},
  }
  \pgfplotstableread[row sep=\\]{
    x y shade text\\
    0 0 1.7241 1.7\\
    1 0 1.5921 1.6\\
    2 0 1.3471 1.3\\
    3 0 1.0957 1.1\\
    4 0 0.68031 0.7\\
    0 1 1.723 1.7\\
    1 1 1.6073 1.6\\
    2 1 1.3939 1.4\\
    3 1 1.1512 1.2\\
    4 1 0.70516 0.7\\
    0 2 1.7322 1.7\\
    1 2 1.6298 1.6\\
    2 2 1.4342 1.4\\
    3 2 1.1545 1.2\\
    4 2 0.747 0.7\\
    0 3 1.732 1.7\\
    1 3 1.6186 1.6\\
    2 3 1.4415 1.4\\
    3 3 1.1553 1.2\\
    4 3 0.77033 0.8\\
    0 4 1.726 1.7\\
    1 4 1.6213 1.6\\
    2 4 1.4565 1.5\\
    3 4 1.1632 1.2\\
    4 4 0.82803 0.8\\
  }\gridablock
  \pgfplotstableread[row sep=\\]{
    x y shade text\\
    0 0 1.7258 1.7\\
    1 0 1.6023 1.6\\
    2 0 1.4004 1.4\\
    3 0 1.0868 1.1\\
    4 0 0.81497 0.8\\
    0 1 1.728 1.7\\
    1 1 1.6229 1.6\\
    2 1 1.414 1.4\\
    3 1 1.1037 1.1\\
    4 1 0.82098 0.8\\
    0 2 1.7431 1.7\\
    1 2 1.6262 1.6\\
    2 2 1.4252 1.4\\
    3 2 1.1008 1.1\\
    4 2 0.81203 0.8\\
    0 3 1.7392 1.7\\
    1 3 1.6355 1.6\\
    2 3 1.4384 1.4\\
    3 3 1.1254 1.1\\
    4 3 0.82861 0.8\\
    0 4 1.7354 1.7\\
    1 4 1.619 1.6\\
    2 4 1.4271 1.4\\
    3 4 1.1215 1.1\\
    4 4 0.83711 0.8\\
  }\gridbblock
  \pgfplotstableread[row sep=\\]{
    x y shade text\\
    0 0 1.728 1.7\\
    1 0 1.6293 1.6\\
    2 0 1.3505 1.4\\
    3 0 1.0778 1.1\\
    4 0 0.81242 0.8\\
    0 1 1.7244 1.7\\
    1 1 1.6301 1.6\\
    2 1 1.3882 1.4\\
    3 1 1.0953 1.1\\
    4 1 0.84648 0.8\\
    0 2 1.7175 1.7\\
    1 2 1.6227 1.6\\
    2 2 1.4114 1.4\\
    3 2 1.1696 1.2\\
    4 2 0.87514 0.9\\
    0 3 1.7285 1.7\\
    1 3 1.6199 1.6\\
    2 3 1.403 1.4\\
    3 3 1.1837 1.2\\
    4 3 0.87525 0.9\\
    0 4 1.7246 1.7\\
    1 4 1.6138 1.6\\
    2 4 1.4113 1.4\\
    3 4 1.1896 1.2\\
    4 4 0.91134 0.9\\
  }\gridcblock
  \pgfplotstableread[row sep=\\]{
    x y shade text\\
    0 0 1.7284 1.7\\
    1 0 1.6226 1.6\\
    2 0 1.3558 1.4\\
    3 0 1.1118 1.1\\
    4 0 0.85365 0.9\\
    0 1 1.727 1.7\\
    1 1 1.6415 1.6\\
    2 1 1.4141 1.4\\
    3 1 1.1181 1.1\\
    4 1 0.81235 0.8\\
    0 2 1.7462 1.7\\
    1 2 1.665 1.7\\
    2 2 1.4804 1.5\\
    3 2 1.1776 1.2\\
    4 2 0.84161 0.8\\
    0 3 1.7544 1.8\\
    1 3 1.6637 1.7\\
    2 3 1.5129 1.5\\
    3 3 1.1842 1.2\\
    4 3 0.86428 0.9\\
    0 4 1.7516 1.8\\
    1 4 1.6644 1.7\\
    2 4 1.5217 1.5\\
    3 4 1.2352 1.2\\
    4 4 0.96261 1.0\\
  }\griddblock
  \pgfplotstableread[row sep=\\]{
    x y shade text\\
    0 0 1.743 1.7\\
    1 0 1.6456 1.6\\
    2 0 1.4326 1.4\\
    3 0 1.1573 1.2\\
    4 0 0.87993 0.9\\
    0 1 1.7393 1.7\\
    1 1 1.6331 1.6\\
    2 1 1.4075 1.4\\
    3 1 1.1153 1.1\\
    4 1 0.81277 0.8\\
    0 2 1.7447 1.7\\
    1 2 1.6563 1.7\\
    2 2 1.4874 1.5\\
    3 2 1.1108 1.1\\
    4 2 0.82751 0.8\\
    0 3 1.757 1.8\\
    1 3 1.6533 1.7\\
    2 3 1.4989 1.5\\
    3 3 1.1345 1.1\\
    4 3 0.82233 0.8\\
    0 4 1.7523 1.8\\
    1 4 1.6419 1.6\\
    2 4 1.4177 1.4\\
    3 4 1.1393 1.1\\
    4 4 0.85458 0.9\\
  }\grideblock
  \pgfplotstableread[row sep=\\]{
    x y shade text\\
    0 0 -3.0107 \sci{\textminus1.0}{\textminus2}\\
    1 0 -2.6196 \sci{\textminus4.2}{\textminus3}\\
    2 0 -3.2544 \sci{\textminus1.8}{\textminus2}\\
    3 0 -3.376 \sci{\textminus2.4}{\textminus2}\\
    4 0 -2.8157 \sci{\textminus6.5}{\textminus3}\\
    0 1 -2.6971 \sci{\textminus5.0}{\textminus3}\\
    1 1 -2.4404 \sci{\textminus2.7}{\textminus3}\\
    2 1 -2.6808 \sci{\textminus4.8}{\textminus3}\\
    3 1 -2.5495 \sci{\textminus3.5}{\textminus3}\\
    4 1 3.1564 \sci{+1.4}{\textminus2}\\
    0 2 -2.0594 \sci{\textminus1.1}{\textminus3}\\
    1 2 -2.8009 \sci{\textminus6.3}{\textminus3}\\
    2 2 -2.65 \sci{\textminus4.5}{\textminus3}\\
    3 2 2.7327 \sci{+5.4}{\textminus3}\\
    4 2 3.3695 \sci{+2.3}{\textminus2}\\
    0 3 -2.9981 \sci{\textminus9.9}{\textminus3}\\
    1 3 -2.807 \sci{\textminus6.4}{\textminus3}\\
    2 3 2.5574 \sci{+3.6}{\textminus3}\\
    3 3 2.5977 \sci{+3.9}{\textminus3}\\
    4 3 -2.9198 \sci{\textminus8.3}{\textminus3}\\
    0 4 -1.2061 \sci{\textminus1.5}{\textminus4}\\
    1 4 -2.3031 \sci{\textminus2.0}{\textminus3}\\
    2 4 2.5423 \sci{+3.5}{\textminus3}\\
    3 4 2.2991 \sci{+2.0}{\textminus3}\\
    4 4 -2.8582 \sci{\textminus7.2}{\textminus3}\\
  }\gridachange
  \pgfplotstableread[row sep=\\]{
    x y shade text\\
    0 0 1.74 \sci{+5.4}{\textminus4}\\
    1 0 -0.0052445 \sci{\textminus1.2}{\textminus7}\\
    2 0 -0.019186 \sci{\textminus4.5}{\textminus7}\\
    3 0 -0.017381 \sci{\textminus4.1}{\textminus7}\\
    4 0 -0.66287 \sci{\textminus3.6}{\textminus5}\\
    0 1 -0.42987 \sci{\textminus1.7}{\textminus5}\\
    1 1 -0.0029002 \sci{\textminus6.7}{\textminus8}\\
    2 1 -0.0026343 \sci{\textminus6.1}{\textminus8}\\
    3 1 -0.0019469 \sci{\textminus4.5}{\textminus8}\\
    4 1 -0.22178 \sci{\textminus6.7}{\textminus6}\\
    0 2 -2.337 \sci{\textminus2.2}{\textminus3}\\
    1 2 -1.2131 \sci{\textminus1.5}{\textminus4}\\
    2 2 -2.6629 \sci{\textminus4.6}{\textminus3}\\
    3 2 -0.029545 \sci{\textminus7.0}{\textminus7}\\
    4 2 -0.070083 \sci{\textminus1.8}{\textminus6}\\
    0 3 2.3543 \sci{+2.3}{\textminus3}\\
    1 3 1.6993 \sci{+4.9}{\textminus4}\\
    2 3 -2.6898 \sci{\textminus4.9}{\textminus3}\\
    3 3 2.7735 \sci{+5.9}{\textminus3}\\
    4 3 2.5314 \sci{+3.4}{\textminus3}\\
    0 4 -1.2689 \sci{\textminus1.8}{\textminus4}\\
    1 4 -1.4326 \sci{\textminus2.6}{\textminus4}\\
    2 4 -2.951 \sci{\textminus8.9}{\textminus3}\\
    3 4 2.3228 \sci{+2.1}{\textminus3}\\
    4 4 -2.2726 \sci{\textminus1.9}{\textminus3}\\
  }\gridbchange
  \pgfplotstableread[row sep=\\]{
    x y shade text\\
    0 0 -0.20625 \sci{\textminus6.1}{\textminus6}\\
    1 0 -0.65734 \sci{\textminus3.5}{\textminus5}\\
    2 0 -0.532 \sci{\textminus2.4}{\textminus5}\\
    3 0 -0.29284 \sci{\textminus9.6}{\textminus6}\\
    4 0 -0.070308 \sci{\textminus1.8}{\textminus6}\\
    0 1 -0.0041899 \sci{\textminus9.7}{\textminus8}\\
    1 1 -0.82557 \sci{\textminus5.7}{\textminus5}\\
    2 1 -0.26865 \sci{\textminus8.6}{\textminus6}\\
    3 1 -0.16883 \sci{\textminus4.8}{\textminus6}\\
    4 1 -1.4558 \sci{\textminus2.8}{\textminus4}\\
    0 2 -0.088298 \sci{\textminus2.3}{\textminus6}\\
    1 2 -0.0082827 \sci{\textminus1.9}{\textminus7}\\
    2 2 1.1697 \sci{+1.4}{\textminus4}\\
    3 2 2.0397 \sci{+1.1}{\textminus3}\\
    4 2 -2.2341 \sci{\textminus1.7}{\textminus3}\\
    0 3 -0.010629 \sci{\textminus2.5}{\textminus7}\\
    1 3 -1.8292 \sci{\textminus6.6}{\textminus4}\\
    2 3 2.1023 \sci{+1.3}{\textminus3}\\
    3 3 2.1803 \sci{+1.5}{\textminus3}\\
    4 3 2.522 \sci{+3.3}{\textminus3}\\
    0 4 -4.0821e-05 \sci{\textminus9.4}{\textminus10}\\
    1 4 -1.5612 \sci{\textminus3.5}{\textminus4}\\
    2 4 0.76128 \sci{+4.8}{\textminus5}\\
    3 4 -0.013653 \sci{\textminus3.2}{\textminus7}\\
    4 4 -7.6471e-09 \sci{\textminus1.8}{\textminus13}\\
  }\gridcchange
  \pgfplotstableread[row sep=\\]{
    x y shade text\\
    0 0 2.6599 \sci{+4.6}{\textminus3}\\
    1 0 -2.8786 \sci{\textminus7.6}{\textminus3}\\
    2 0 3.188 \sci{+1.5}{\textminus2}\\
    3 0 -2.6475 \sci{\textminus4.4}{\textminus3}\\
    4 0 -1.6254 \sci{\textminus4.1}{\textminus4}\\
    0 1 -0.00032964 \sci{\textminus7.6}{\textminus9}\\
    1 1 1.5928 \sci{+3.8}{\textminus4}\\
    2 1 -1.8185 \sci{\textminus6.5}{\textminus4}\\
    3 1 -1.6951 \sci{\textminus4.9}{\textminus4}\\
    4 1 -0.52377 \sci{\textminus2.3}{\textminus5}\\
    0 2 -0.27617 \sci{\textminus8.9}{\textminus6}\\
    1 2 -0.068049 \sci{\textminus1.7}{\textminus6}\\
    2 2 -0.16609 \sci{\textminus4.7}{\textminus6}\\
    3 2 2.8941 \sci{+7.8}{\textminus3}\\
    4 2 -3.127 \sci{\textminus1.3}{\textminus2}\\
    0 3 0.05498 \sci{+1.3}{\textminus6}\\
    1 3 0.75297 \sci{+4.7}{\textminus5}\\
    2 3 -2.8599 \sci{\textminus7.2}{\textminus3}\\
    3 3 3.6237 \sci{+4.2}{\textminus2}\\
    4 3 -2.2152 \sci{\textminus1.6}{\textminus3}\\
    0 4 -0.098296 \sci{\textminus2.5}{\textminus6}\\
    1 4 2.1366 \sci{+1.4}{\textminus3}\\
    2 4 -2.1076 \sci{\textminus1.3}{\textminus3}\\
    3 4 -2.7969 \sci{\textminus6.3}{\textminus3}\\
    4 4 -2.3472 \sci{\textminus2.2}{\textminus3}\\
  }\griddchange
  \pgfplotstableread[row sep=\\]{
    x y shade text\\
    0 0 -2.0237 \sci{\textminus1.0}{\textminus3}\\
    1 0 -2.1343 \sci{\textminus1.4}{\textminus3}\\
    2 0 -1.5893 \sci{\textminus3.8}{\textminus4}\\
    3 0 2.0488 \sci{+1.1}{\textminus3}\\
    4 0 1.646 \sci{+4.3}{\textminus4}\\
    0 1 0.037493 \sci{+9.0}{\textminus7}\\
    1 1 1.581 \sci{+3.7}{\textminus4}\\
    2 1 -0.47365 \sci{\textminus2.0}{\textminus5}\\
    3 1 1.9927 \sci{+9.7}{\textminus4}\\
    4 1 -0.13073 \sci{\textminus3.5}{\textminus6}\\
    0 2 -1.5726 \sci{\textminus3.6}{\textminus4}\\
    1 2 -0.84249 \sci{\textminus6.0}{\textminus5}\\
    2 2 0.04024 \sci{+9.7}{\textminus7}\\
    3 2 2.628 \sci{+4.2}{\textminus3}\\
    4 2 1.1343 \sci{+1.3}{\textminus4}\\
    0 3 -0.010892 \sci{\textminus2.5}{\textminus7}\\
    1 3 0.78597 \sci{+5.1}{\textminus5}\\
    2 3 -2.1634 \sci{\textminus1.4}{\textminus3}\\
    3 3 3.4273 \sci{+2.7}{\textminus2}\\
    4 3 -3.0711 \sci{\textminus1.2}{\textminus2}\\
    0 4 -0.38215 \sci{\textminus1.4}{\textminus5}\\
    1 4 0.97233 \sci{+8.4}{\textminus5}\\
    2 4 3.6051 \sci{+4.0}{\textminus2}\\
    3 4 3.0745 \sci{+1.2}{\textminus2}\\
    4 4 3.1563 \sci{+1.4}{\textminus2}\\
  }\gridechange
  \begin{groupplot}[
      group style={group size=5 by 2, horizontal sep=5.0pt, vertical sep=5.0pt,
        xticklabels at=edge bottom, yticklabels at=edge left},
      width=64.5pt, height=62.0pt, tick label style={font=\scriptsize},
      enlargelimits=false, axis on top, major tick length=0pt,
      xmin=-0.5, xmax=4.5, ymin=-0.5, ymax=4.5,
      xtick={0,...,4}, xticklabels={5,10,20,35,50},
      ytick={0,...,4}, yticklabels={128,256,512,1024,2048}]
    \nextgroupplot[colormap name=residual, point meta min=0.68031, point meta max=1.757, title={GAIM exact\\{[v3, $L^\infty$]}}, ylabel={block\\\scriptsize segment size $\Nc$}]
      \addplot[cells] table[x=x, y=y, meta=shade] {\gridablock};
      \addplot[numbers] table[x=x, y=y, meta=text] {\gridablock};
    \nextgroupplot[colormap name=residual, point meta min=0.68031, point meta max=1.757, title={GAIM-LIN\\{[v1, $L^\infty$]}}]
      \addplot[cells] table[x=x, y=y, meta=shade] {\gridbblock};
      \addplot[numbers] table[x=x, y=y, meta=text] {\gridbblock};
    \nextgroupplot[colormap name=residual, point meta min=0.68031, point meta max=1.757, title={GAIM-LIN-O\\{[v1, $L^\infty$]}}]
      \addplot[cells] table[x=x, y=y, meta=shade] {\gridcblock};
      \addplot[numbers] table[x=x, y=y, meta=text] {\gridcblock};
    \nextgroupplot[colormap name=residual, point meta min=0.68031, point meta max=1.757, title={Q-DAIM exact\\{[v5]}}]
      \addplot[cells] table[x=x, y=y, meta=shade] {\griddblock};
      \addplot[numbers] table[x=x, y=y, meta=text] {\griddblock};
    \nextgroupplot[colormap name=residual, point meta min=0.68031, point meta max=1.757, title={Q-DAIM RFF\\{[v5]}}]
      \addplot[cells] table[x=x, y=y, meta=shade] {\grideblock};
      \addplot[numbers] table[x=x, y=y, meta=text] {\grideblock};
    \nextgroupplot[colormap name=change, point meta min=-3.624, point meta max=3.624, ylabel={streaming $-$ block\\\scriptsize segment size $\Nc$}]
      \addplot[cells] table[x=x, y=y, meta=shade] {\gridachange};
      \addplot[numbers] table[x=x, y=y, meta=text] {\gridachange};
    \nextgroupplot[colormap name=change, point meta min=-3.624, point meta max=3.624]
      \addplot[cells] table[x=x, y=y, meta=shade] {\gridbchange};
      \addplot[numbers] table[x=x, y=y, meta=text] {\gridbchange};
    \nextgroupplot[colormap name=change, point meta min=-3.624, point meta max=3.624, xlabel={nodes per segment $\nc/\Nc$ (\%)}]
      \addplot[cells] table[x=x, y=y, meta=shade] {\gridcchange};
      \addplot[numbers] table[x=x, y=y, meta=text] {\gridcchange};
    \nextgroupplot[colormap name=change, point meta min=-3.624, point meta max=3.624]
      \addplot[cells] table[x=x, y=y, meta=shade] {\griddchange};
      \addplot[numbers] table[x=x, y=y, meta=text] {\griddchange};
    \nextgroupplot[colormap name=change, point meta min=-3.624, point meta max=3.624]
      \addplot[cells] table[x=x, y=y, meta=shade] {\gridechange};
      \addplot[numbers] table[x=x, y=y, meta=text] {\gridechange};
  \end{groupplot}
  \begin{axis}[hide axis, at={(group c5r1.north east)}, anchor=north west,
      width=1pt, height=62.0pt, xmin=0, xmax=1, ymin=0, ymax=1,
      colormap name=residual, point meta min=0.68031, point meta max=1.757,
      colorbar, colorbar style={at={(parent axis.north east)}, anchor=north west, xshift=4pt, width=7pt,
        tick label style={font=\scriptsize}, ylabel={error $\varepsilon_{\Att}^{10\%}$ (\%)}, ylabel style={font=\scriptsize}}]
  \end{axis}
  \begin{axis}[hide axis, at={(group c5r2.north east)}, anchor=north west,
      width=1pt, height=62.0pt, xmin=0, xmax=1, ymin=0, ymax=1,
      colormap name=change, point meta min=-3.6237, point meta max=3.6237,
      colorbar, colorbar style={at={(parent axis.north east)}, anchor=north west, xshift=4pt, width=7pt,
        tick label style={font=\scriptsize}, ytick={-3,-1.041,0,1.041,3}, yticklabels={$-10^{-2}$,$-10^{-4}$,$0$,$+10^{-4}$,$+10^{-2}$}, ylabel={difference}, ylabel style={font=\scriptsize}}]
  \end{axis}
\end{tikzpicture}

%% file: tables/longbench_e_ledger.tex
\begin{table*}[t]
\centering
\caption{LongBench-E - 25 prompts per task at 25\% context compression.  Key centers are given as the tuple (selection, solve), c = segment mean, g = global mean. The Exact and \textsc{CompressKV} rows are the values reported in Table~4 of~\cite{schroderWildCatNearLinearAttention2026} (25\% compression) on the complete benchmark (compared to our 25 samples/task). For comparison, the percentages rows give, per task, our best method on average (Q-DAIM exact, $\Nc{=}64$) and the best score per task relative to Dense, and \textsc{CompressKV} relative to Exact. Proj.: Galerkin/basis projection (Gal.), none (each node keeps its own value with a unit weight, $[V_L \,|\, 1]$), or Q-DAIM's revealed-range projection restricted to the revealed rank (range$_r$). Methods optionally use the $\tau$ temperature scaled kernel for comparison with WildCat~\cite{schroderWildCatNearLinearAttention2026}. RP-Cholesky corresponds to an implementation of the WildCat selection and projection on our harness. $^\dagger$: run on PyTorch 2.14, whose GPU (MPS) attention kernels round differently. }
\label{tab:longbench_e_ledger}
\resizebox{\textwidth}{!}{%
\begin{tabular}{llrllccccrrrrrrrrrrrrrr}
\toprule
Selection method  & $\Nc$ & Deflation & \begin{tabular}[b]{@{}l@{}}RoPE\\Strip.\end{tabular} & Centering & $\tau$ & RC & Proj. & \rotatebox{60}{Qasper} & \rotatebox{60}{MFQA} & \rotatebox{60}{HotpotQA} & \rotatebox{60}{2Wiki} & \rotatebox{60}{GovRep} & \rotatebox{60}{MultiNews} & \rotatebox{60}{TREC} & \rotatebox{60}{TriviaQA} & \rotatebox{60}{SAMSum} & \rotatebox{60}{PCount} & \rotatebox{60}{PRetr} & \rotatebox{60}{LCC} & \rotatebox{60}{RepoB-P} & \rotatebox{60}{Mean} & \rotatebox{60}{\% dense} \\
\midrule
Dense  & -- & -- & -- & -- & -- & -- & -- & 54.9 & 61.8 & 57.1 & 60.8 & 34.7 & 26.9 & 72.0 & 86.7 & 46.7 & 16.0 & 100.0 & 68.6 & 65.6 & 57.84 & 100.0 \\
\midrule
GAIM exact [v3, $L^\infty$]  & 128 & range & off & (c,c) & 1 & on & Gal. & 45.8 & 58.4 & 48.2 & 36.0 & 31.1 & 23.2 & 64.0 & 85.5 & 41.3 & 4.0 & 80.0 & 52.7 & 56.1 & 48.18 & 83.3 \\
\midrule
\multirow{8}{*}{GAIM-LIN [v1, $L^\infty$]} & 128$^\dagger$ & none & off & (c,c) & 1 & on & Gal. & \textbf{48.8} & \textbf{59.3} & 52.2 & 36.4 & 30.0 & 22.7 & 64.0 & 88.8 & 41.2 & 8.0 & 84.0 & 47.2 & 53.0 & 48.88 & 84.5 \\
 & 128$^\dagger$ & none & off & (c,c) & 1 & on & none & 40.5 & 57.7 & 49.2 & 47.3 & 30.9 & 23.6 & 68.0 & 85.0 & 37.5 & 8.0 & 92.0 & 44.2 & 56.6 & 49.27 & 85.2 \\
 & 128$^\dagger$ & none & off & (c,c) & 1 & off & Gal. & 43.9 & 52.3 & 48.0 & 43.7 & 30.3 & 23.1 & 60.0 & 86.8 & 39.0 & 8.0 & 92.0 & 52.5 & 55.3 & 48.84 & 84.4 \\
 & 128$^\dagger$ & none & off & (c,c) & 1 & off & none & 46.5 & 54.3 & 45.8 & 38.5 & 31.1 & 23.2 & 68.0 & 84.9 & 37.8 & 4.0 & 80.0 & 51.4 & \textbf{61.4} & 48.23 & 83.4 \\
 & 128 & none & off & (c,c) & $\tau$ & on & Gal. & 35.5 & 49.8 & 54.8 & 47.6 & 29.9 & 23.2 & 64.0 & 89.3 & 38.6 & 12.0 & 80.0 & 51.1 & 56.0 & 48.60 & 84.0 \\
 & 512 & none & on & (c,c) & 1 & on & Gal. & 37.1 & 49.5 & 50.6 & 36.5 & 29.1 & 22.9 & 64.0 & 90.2 & 39.2 & \textbf{16.0} & 84.0 & 53.1 & 58.1 & 48.48 & 83.8 \\
   & 512 & range & on & (c,c) & 1 & on & Gal. & 35.2 & 48.7 & 49.8 & 36.3 & 28.7 & 22.8 & 64.0 & 88.5 & 41.3 & 12.0 & 64.0 & \textbf{59.4} & 58.0 & 46.83 & 81.0 \\
   & 512 & basis + refit & off & (g,g) & 1 & on & Gal. & 37.8 & 47.1 & 50.2 & 35.3 & 29.0 & 22.1 & 64.0 & 86.9 & 39.2 & 12.0 & 48.0 & 49.8 & 60.9 & 44.80 & 77.5 \\
\midrule
\multirow{6}{*}{Q-DAIM exact [v5]} & 64 & range & off & (c,c) & 1 & on & Gal. & 41.5 & 55.4 & 53.0 & 35.0 & 31.7 & \textbf{24.0} & 64.0 & 90.3 & 42.5 & 12.0 & 88.0 & 53.8 & 52.4 & 49.50 & 85.6 \\
 & 64 & range & off & (c,c) & $\tau$ & on & Gal. & 38.4 & 58.1 & \textbf{56.1} & 39.8 & 30.8 & 22.9 & 60.0 & 81.7 & 40.3 & 4.0 & 92.0 & 55.8 & 38.3 & 47.56 & 82.2 \\
 & 64$^\dagger$ & range & off & (c,c) & 1 & on & none & 34.8 & 48.6 & 52.6 & 40.0 & 31.4 & 23.5 & 56.0 & 91.3 & 42.4 & 8.0 & 88.0 & 52.1 & 54.2 & 47.93 & 82.9 \\
 & 64$^\dagger$ & range & off & (c,c) & 1 & off & Gal. & 38.3 & 52.0 & 49.5 & 45.9 & \textbf{32.5} & 22.1 & 56.0 & 88.7 & 39.9 & 4.0 & 92.0 & 50.1 & 52.6 & 47.97 & 82.9 \\
 & 64$^\dagger$ & range & off & (c,c) & 1 & off & none & 38.2 & 53.3 & 50.7 & \textbf{60.5} & 31.8 & 22.5 & 52.0 & 92.0 & 41.3 & 8.0 & \textbf{96.0} & 53.5 & 55.5 & \textbf{50.40} & 87.1 \\
 & 64$^\dagger$ & range & off & (c,c) & 1 & on & range$_r$ & 39.4 & 52.1 & 54.7 & 43.6 & 31.6 & 22.1 & 60.0 & \textbf{95.3} & \textbf{42.9} & 8.0 & \textbf{96.0} & 54.8 & 49.8 & 50.02 & 86.5 \\
\midrule
\multirow{2}{*}{Q-DAIM RFF [v5]}  & 512 & none & on & (c,c) & 1 & on & Gal. & 39.2 & 47.0 & 48.3 & 45.6 & 29.0 & 21.7 & 68.0 & 81.6 & 41.4 & 12.0 & 80.0 & 53.9 & 52.0 & 47.67 & 82.4 \\
  & 2048 & none & on & (c,c) & 1 & on & Gal. & 42.8 & 39.4 & 39.3 & 42.5 & 28.2 & 22.4 & \textbf{72.0} & 88.9 & 40.8 & 12.0 & 40.0 & 58.4 & 60.5 & 45.17 & 78.1 \\
\midrule
\multirow{3}{*}{RP-Cholesky[\cite{schroderWildCatNearLinearAttention2026}-based]} & 128 & none & off & (c,c) & $\tau$ & on & Gal. & 33.8 & 58.3 & 48.8 & 53.6 & 29.4 & 22.6 & 60.0 & 88.9 & 41.6 & 8.0 & 76.0 & 53.9 & 55.4 & 48.49 & 83.8 \\
 & 128 & none & off & (c,c) & $\tau$ & off & Gal. & 36.8 & 58.6 & 43.2 & 53.6 & 29.8 & 23.4 & 64.0 & 88.9 & 37.4 & 12.0 & 76.0 & 51.6 & 53.6 & 48.38 & 83.7 \\
 & 128$^\dagger$ & none & off & (c,c) & $\tau$ & off & none & 41.1 & 57.9 & 54.5 & 42.9 & 31.7 & 22.7 & 60.0 & 91.3 & 38.1 & 12.0 & 84.0 & 55.6 & 59.0 & 50.06 & 86.6 \\
\midrule
\begin{tabular}[c]{@{}l@{}}Q-DAIM exact [v5],\\ \% of Dense\end{tabular} & 64$^\dagger$ & range & off & (c,c) & 1 & off & none & 69.5 & 86.2 & 88.8 & 99.6 & 91.7 & 83.4 & 72.2 & 106.2 & 88.4 & 50.0 & 96.0 & 77.9 & 84.6 & 84.2 & 87.1 \\
\begin{tabular}[c]{@{}l@{}}Best score,\\ \% of Dense\end{tabular}  & -- & -- & -- & -- & -- & -- & -- & 88.9 & 95.9 & 98.2 & 99.6 & 93.7 & 89.0 & 100.0 & 110.0 & 91.8 & 100.0 & 96.0 & 86.5 & 93.6 & 95.6 & 96.3 \\
\midrule\midrule
Exact~\cite{schroderWildCatNearLinearAttention2026} & -- & -- & -- & -- & -- & -- & -- & 43.8 & 50.1 & 56.4 & 43.7 & 34.2 & 24.3 & 64.0 & 87.5 & 38.6 & 14.7 & 99.7 & 70.0 & 62.2 & 53.00 & 100.0 \\
\textsc{CompressKV}~\cite{schroderWildCatNearLinearAttention2026} & -- & -- & -- & -- & -- & -- & -- & 33.2 & 38.1 & 43.4 & 33.4 & 30.3 & 22.3 & 54.3 & 86.2 & 35.4 & 14.3 & 98.0 & 64.8 & 60.4 & 47.25 & 89.1 \\
\begin{tabular}[c]{@{}l@{}}\textsc{CompressKV},\\ \% of Exact\end{tabular} & -- & -- & -- & -- & -- & -- & -- & 75.9 & 76.1 & 77.0 & 76.3 & 88.7 & 91.5 & 84.9 & 98.4 & 91.8 & 97.7 & 98.3 & 92.7 & 97.2 & 88.2 & 89.1 \\
\bottomrule
\end{tabular}}
\end{table*}

%% file: figures/tikz/projections_gaim_exact_v4_linf_sink32_iqr.tex
\begin{tikzpicture}
  \pgfplotsset{every axis/.append style={
      scale only axis, /tikz/font=\footnotesize, label style={align=center}, title style={align=center},
      legend cell align=left,
      legend style={font=\footnotesize, /tikz/every even column/.append style={column sep=7pt}}}}
  \pgfplotsset{
    galerkin/.style={blue, mark=*},
    least squares/.style={red, mark=square*},
    revealed range/.style={cyan!60!black, mark=triangle*},
    temperature/.style={orange!50!black, mark=diamond*},
    no projection/.style={black!55, mark=x, mark size=2.4pt},
    mean value/.style={black, thin, dashdotted, mark=none},
    zero readout/.style={black, thin, densely dotted, mark=none},
    every axis plot/.append style={very thick, mark size=1.8pt, mark options={solid}, forget plot},
    legend image code/.code={\draw[very thick, mark size=1.8pt, mark options={solid}, mark repeat=2, mark phase=2, #1]
      plot coordinates {(0pt,0pt) (10pt,0pt) (20pt,0pt)};},
    band/.style={draw=none, fill, fill opacity=0.12, mark=none},
    range line/.style={thin, densely dashed, mark=none},
    range legend/.style={black, thin, densely dashed, mark=none},
    band legend/.style={legend image code/.code={\fill[black, fill opacity=0.12] (0pt,-3pt) rectangle (20pt,3pt);}},
  }
  \pgfplotstableread[row sep=\\]{
    n galerkin least_squares revealed_range temperature no_projection\\
    1 0.21387 0.21387 0.23629 0.19852 0.21387\\
    17 0.20634 0.2094 4.3534 0.16422 0.20537\\
    33 0.19517 0.18722 8.9436 0.15626 0.20169\\
    65 0.19083 0.18053 8.6938 0.13649 0.2062\\
    97 0.18633 0.16975 12.468 0.11599 0.20097\\
    129 0.17988 0.16484 34.75 0.10611 0.19878\\
    193 0.16676 0.16249 36.222 0.097639 0.18441\\
    257 0.15805 0.15589 14.623 0.09727 0.17467\\
    385 0.15337 0.15314 9.8761 0.088893 0.16382\\
    512 0.1493 0.14793 14.249 0.088729 0.15863\\
  }\curvesa
  \pgfplotstableread[row sep=\\]{
    n galerkin least_squares revealed_range temperature no_projection\\
    1 0.16892 0.16892 0.21416 0.168 0.16892\\
    17 0.16792 0.16791 5.3378 0.14102 0.16792\\
    33 0.1675 0.15794 10.807 0.14018 0.16753\\
    65 0.16808 0.16526 10.654 0.12157 0.16813\\
    97 0.16491 0.15937 15.851 0.10853 0.1671\\
    129 0.15667 0.14494 105.06 0.1029 0.16452\\
    193 0.13823 0.14245 55.443 0.096308 0.15255\\
    257 0.13217 0.12844 19.158 0.091882 0.14116\\
    385 0.12564 0.12268 21.657 0.082763 0.1365\\
    512 0.12085 0.11478 19.423 0.079966 0.12979\\
  }\curvesb
  \begin{groupplot}[
      group style={group size=2 by 1, horizontal sep=8pt, yticklabels at=edge left},
      width=198pt, height=120pt, grid=major, tick label style={font=\scriptsize},
      xmin=-15, xmax=527, xtick={0,100,...,500}, ymin=0.02241, ymax=0.4625,
      scaled y ticks=false, yticklabel style={/pgf/number format/fixed, /pgf/number format/fixed zerofill, /pgf/number format/precision=2}]
    \nextgroupplot[title={RC off}, xlabel={number of nodes $n$}, ylabel={relative attention error $\varepsilon_{\Att}^{10\%}$}, legend to name=legend-projections-gaim-exact-v4-linf-sink32-iqr, legend columns=2]
      \addplot[galerkin, range line] coordinates {(1,0.089044) (17,0.087991) (33,0.088935) (65,0.085049) (97,0.080632) (129,0.07852) (193,0.075038) (257,0.073266) (385,0.068088) (512,0.065032)};
      \addplot[galerkin, range line] coordinates {(1,0.43378) (17,0.39796) (33,0.3887) (65,0.37829) (97,0.36019) (129,0.34544) (193,0.33557) (257,0.32468) (385,0.30745) (512,0.29275)};
      \addplot[least squares, range line] coordinates {(1,0.089043) (17,0.08776) (33,0.088917) (65,0.084781) (97,0.078283) (129,0.077121) (193,0.072856) (257,0.071356) (385,0.066281) (512,0.064448)};
      \addplot[least squares, range line] coordinates {(1,0.42778) (17,0.37197) (33,0.37659) (65,0.3688) (97,0.3415) (129,0.33837) (193,0.33456) (257,0.32625) (385,0.30632) (512,0.30323)};
      \addplot[temperature, range line] coordinates {(1,0.08739) (17,0.081121) (33,0.077419) (65,0.063953) (97,0.053869) (129,0.051767) (193,0.050424) (257,0.046538) (385,0.04499) (512,0.040446)};
      \addplot[temperature, range line] coordinates {(1,0.39041) (17,0.29212) (33,0.2746) (65,0.25145) (97,0.2424) (129,0.23433) (193,0.22142) (257,0.21291) (385,0.20844) (512,0.19353)};
      \addplot[no projection, range line] coordinates {(1,0.089045) (17,0.0881) (33,0.088936) (65,0.086451) (97,0.082471) (129,0.080706) (193,0.076961) (257,0.074722) (385,0.07077) (512,0.067433)};
      \addplot[no projection, range line] coordinates {(1,0.44615) (17,0.41331) (33,0.40287) (65,0.38653) (97,0.37734) (129,0.36211) (193,0.34366) (257,0.33235) (385,0.31836) (512,0.30808)};
      \addplot[mean value] coordinates {(-15,0.16892) (527,0.16892)};
      \addplot[galerkin] table[x=n, y=galerkin] {\curvesa};
      \addplot[least squares] table[x=n, y=least_squares] {\curvesa};
      \addplot[temperature] table[x=n, y=temperature] {\curvesa};
      \addplot[no projection] table[x=n, y=no_projection] {\curvesa};
      \addlegendimage{galerkin}\addlegendentry{GAIM (Galerkin)}
      \addlegendimage{least squares}\addlegendentry{least squares (direct and Petrov--Galerkin)}
      \addlegendimage{empty legend}\addlegendentry{Q-DAIM revealed range $\geq 4.4$}
      \addlegendimage{temperature}\addlegendentry{WildCat temperature, $\tau\approx2.0$}
      \addlegendimage{no projection}\addlegendentry{no projection ($W = V[\mathfrak L,:]$)}
      \addlegendimage{mean value}\addlegendentry{mean vector approximation}
      \addlegendimage{empty legend}\addlegendentry{zero readout $\approx 0.76$}
      \addlegendimage{range legend}\addlegendentry{interquartile range}
    \nextgroupplot[title={RC on}, xlabel={number of nodes $n$}]
      \addplot[galerkin, range line] coordinates {(1,0.080459) (17,0.079236) (33,0.082341) (65,0.073138) (97,0.071519) (129,0.068892) (193,0.06522) (257,0.063558) (385,0.059497) (512,0.058152)};
      \addplot[galerkin, range line] coordinates {(1,0.292) (17,0.28544) (33,0.28354) (65,0.2812) (97,0.28181) (129,0.27064) (193,0.26733) (257,0.26608) (385,0.26084) (512,0.24841)};
      \addplot[least squares, range line] coordinates {(1,0.080459) (17,0.078899) (33,0.081146) (65,0.071658) (97,0.069003) (129,0.065671) (193,0.064145) (257,0.06261) (385,0.059211) (512,0.057807)};
      \addplot[least squares, range line] coordinates {(1,0.292) (17,0.28533) (33,0.28299) (65,0.2798) (97,0.28036) (129,0.27018) (193,0.26668) (257,0.26523) (385,0.26618) (512,0.26469)};
      \addplot[temperature, range line] coordinates {(1,0.079024) (17,0.072371) (33,0.069924) (65,0.06086) (97,0.054105) (129,0.050638) (193,0.050808) (257,0.046987) (385,0.040566) (512,0.038708)};
      \addplot[temperature, range line] coordinates {(1,0.2834) (17,0.25587) (33,0.25341) (65,0.24102) (97,0.23402) (129,0.23115) (193,0.21802) (257,0.21008) (385,0.20145) (512,0.19232)};
      \addplot[no projection, range line] coordinates {(1,0.080459) (17,0.079927) (33,0.083189) (65,0.077064) (97,0.074652) (129,0.07224) (193,0.06907) (257,0.066528) (385,0.062715) (512,0.059322)};
      \addplot[no projection, range line] coordinates {(1,0.292) (17,0.28558) (33,0.28416) (65,0.29014) (97,0.2837) (129,0.27873) (193,0.26875) (257,0.26711) (385,0.26654) (512,0.26515)};
      \addplot[mean value] coordinates {(-15,0.16892) (527,0.16892)};
      \addplot[galerkin] table[x=n, y=galerkin] {\curvesb};
      \addplot[least squares] table[x=n, y=least_squares] {\curvesb};
      \addplot[temperature] table[x=n, y=temperature] {\curvesb};
      \addplot[no projection] table[x=n, y=no_projection] {\curvesb};
  \end{groupplot}
  \node[below=4pt] at (current bounding box.south) {\pgfplotslegendfromname{legend-projections-gaim-exact-v4-linf-sink32-iqr}};
\end{tikzpicture}

%% file: figures/tikz/projections_qdaim_exact_v5_linf_sink32_iqr.tex
\begin{tikzpicture}
  \pgfplotsset{every axis/.append style={
      scale only axis, /tikz/font=\footnotesize, label style={align=center}, title style={align=center},
      legend cell align=left,
      legend style={font=\footnotesize, /tikz/every even column/.append style={column sep=7pt}}}}
  \pgfplotsset{
    galerkin/.style={blue, mark=*},
    least squares/.style={red, mark=square*},
    revealed range/.style={cyan!60!black, mark=triangle*},
    temperature/.style={orange!50!black, mark=diamond*},
    no projection/.style={black!55, mark=x, mark size=2.4pt},
    mean value/.style={black, thin, dashdotted, mark=none},
    zero readout/.style={black, thin, densely dotted, mark=none},
    every axis plot/.append style={very thick, mark size=1.8pt, mark options={solid}, forget plot},
    legend image code/.code={\draw[very thick, mark size=1.8pt, mark options={solid}, mark repeat=2, mark phase=2, #1]
      plot coordinates {(0pt,0pt) (10pt,0pt) (20pt,0pt)};},
    band/.style={draw=none, fill, fill opacity=0.12, mark=none},
    range line/.style={thin, densely dashed, mark=none},
    range legend/.style={black, thin, densely dashed, mark=none},
    band legend/.style={legend image code/.code={\fill[black, fill opacity=0.12] (0pt,-3pt) rectangle (20pt,3pt);}},
  }
  \pgfplotstableread[row sep=\\]{
    n galerkin least_squares revealed_range temperature no_projection\\
    1 0.20723 0.20723 0.20723 0.19909 0.20723\\
    17 0.21325 0.21325 0.21322 0.19478 0.21325\\
    33 0.21244 0.21244 0.21222 0.18158 0.21244\\
    65 0.21122 0.21122 0.21034 0.17587 0.21123\\
    97 0.21457 0.21457 0.21289 0.16397 0.21457\\
    129 0.20626 0.20625 0.19687 0.15476 0.20629\\
    193 0.20622 0.20621 0.18757 0.14536 0.20623\\
    257 0.19913 0.19719 0.18515 0.13188 0.1992\\
    385 0.18984 0.1866 0.17849 0.11695 0.193\\
    512 0.17803 0.17607 0.17081 0.1061 0.18907\\
  }\curvesa
  \pgfplotstableread[row sep=\\]{
    n galerkin least_squares revealed_range temperature no_projection\\
    1 0.16892 0.16892 0.16892 0.1688 0.16892\\
    17 0.16811 0.16808 0.16805 0.16601 0.16811\\
    33 0.1674 0.1674 0.16727 0.16496 0.1674\\
    65 0.16632 0.16631 0.16814 0.16158 0.16632\\
    97 0.1675 0.1675 0.16627 0.15007 0.1675\\
    129 0.16408 0.16408 0.15786 0.13868 0.16516\\
    193 0.16359 0.16356 0.15645 0.12877 0.16361\\
    257 0.15921 0.1592 0.14962 0.11796 0.15924\\
    385 0.15422 0.1542 0.14019 0.10786 0.15424\\
    512 0.14652 0.14469 0.13488 0.097368 0.14772\\
  }\curvesb
  \begin{groupplot}[
      group style={group size=2 by 1, horizontal sep=8pt, yticklabels at=edge left},
      width=198pt, height=120pt, grid=major, tick label style={font=\scriptsize},
      xmin=-15, xmax=527, xtick={0,100,...,500}, ymin=0.03367, ymax=0.5041,
      scaled y ticks=false, yticklabel style={/pgf/number format/fixed, /pgf/number format/fixed zerofill, /pgf/number format/precision=2}]
    \nextgroupplot[title={RC off}, xlabel={number of nodes $n$}, ylabel={relative attention error $\varepsilon_{\Att}^{10\%}$}, legend to name=legend-projections-qdaim-exact-v5-linf-sink32-iqr, legend columns=2]
      \addplot[galerkin, range line] coordinates {(1,0.089273) (17,0.089142) (33,0.089061) (65,0.089674) (97,0.089404) (129,0.088653) (193,0.087862) (257,0.085982) (385,0.083726) (512,0.080324)};
      \addplot[galerkin, range line] coordinates {(1,0.43499) (17,0.42112) (33,0.44596) (65,0.43518) (97,0.42568) (129,0.41967) (193,0.40851) (257,0.39379) (385,0.37549) (512,0.35889)};
      \addplot[least squares, range line] coordinates {(1,0.089273) (17,0.089142) (33,0.089061) (65,0.08967) (97,0.089401) (129,0.08865) (193,0.087816) (257,0.085976) (385,0.08367) (512,0.080284)};
      \addplot[least squares, range line] coordinates {(1,0.43499) (17,0.41953) (33,0.44323) (65,0.43489) (97,0.42256) (129,0.41733) (193,0.40573) (257,0.39035) (385,0.37542) (512,0.35818)};
      \addplot[revealed range, range line] coordinates {(1,0.089273) (17,0.08914) (33,0.089056) (65,0.089317) (97,0.088885) (129,0.089255) (193,0.086243) (257,0.084216) (385,0.081575) (512,0.078523)};
      \addplot[revealed range, range line] coordinates {(1,0.43497) (17,0.46752) (33,0.4859) (65,0.48096) (97,0.48672) (129,0.41597) (193,0.39097) (257,0.39973) (385,0.38333) (512,0.40782)};
      \addplot[temperature, range line] coordinates {(1,0.089247) (17,0.088924) (33,0.087231) (65,0.08592) (97,0.082133) (129,0.079625) (193,0.076124) (257,0.070127) (385,0.062703) (512,0.054765)};
      \addplot[temperature, range line] coordinates {(1,0.40327) (17,0.40243) (33,0.40083) (65,0.38016) (97,0.35155) (129,0.33894) (193,0.30503) (257,0.27244) (385,0.23978) (512,0.21526)};
      \addplot[no projection, range line] coordinates {(1,0.089273) (17,0.089142) (33,0.089062) (65,0.089677) (97,0.089406) (129,0.088657) (193,0.087897) (257,0.086498) (385,0.083887) (512,0.081431)};
      \addplot[no projection, range line] coordinates {(1,0.43499) (17,0.42229) (33,0.44736) (65,0.43558) (97,0.42976) (129,0.42167) (193,0.41079) (257,0.39687) (385,0.37563) (512,0.3594)};
      \addplot[mean value] coordinates {(-15,0.16892) (527,0.16892)};
      \addplot[galerkin] table[x=n, y=galerkin] {\curvesa};
      \addplot[least squares] table[x=n, y=least_squares] {\curvesa};
      \addplot[revealed range] table[x=n, y=revealed_range] {\curvesa};
      \addplot[temperature] table[x=n, y=temperature] {\curvesa};
      \addplot[no projection] table[x=n, y=no_projection] {\curvesa};
      \addlegendimage{galerkin}\addlegendentry{GAIM (Galerkin)}
      \addlegendimage{least squares}\addlegendentry{least squares (direct and Petrov--Galerkin)}
      \addlegendimage{revealed range}\addlegendentry{Q-DAIM revealed range}
      \addlegendimage{temperature}\addlegendentry{WildCat temperature, $\tau\approx2.0$}
      \addlegendimage{no projection}\addlegendentry{no projection ($W = V[\mathfrak L,:]$)}
      \addlegendimage{mean value}\addlegendentry{mean vector approximation}
      \addlegendimage{empty legend}\addlegendentry{zero readout $\approx 0.76$}
      \addlegendimage{range legend}\addlegendentry{interquartile range}
    \nextgroupplot[title={RC on}, xlabel={number of nodes $n$}]
      \addplot[galerkin, range line] coordinates {(1,0.080456) (17,0.080112) (33,0.07973) (65,0.083247) (97,0.082731) (129,0.081894) (193,0.080511) (257,0.075099) (385,0.071742) (512,0.072824)};
      \addplot[galerkin, range line] coordinates {(1,0.29004) (17,0.29888) (33,0.29582) (65,0.29065) (97,0.28682) (129,0.28674) (193,0.28398) (257,0.28838) (385,0.2891) (512,0.28291)};
      \addplot[least squares, range line] coordinates {(1,0.080456) (17,0.080113) (33,0.079727) (65,0.083247) (97,0.08273) (129,0.081892) (193,0.080498) (257,0.075035) (385,0.071728) (512,0.071962)};
      \addplot[least squares, range line] coordinates {(1,0.29004) (17,0.29888) (33,0.29579) (65,0.29047) (97,0.28682) (129,0.28673) (193,0.28395) (257,0.28835) (385,0.28902) (512,0.28276)};
      \addplot[revealed range, range line] coordinates {(1,0.080456) (17,0.080113) (33,0.079711) (65,0.083182) (97,0.082459) (129,0.08152) (193,0.079136) (257,0.074277) (385,0.071434) (512,0.070043)};
      \addplot[revealed range, range line] coordinates {(1,0.29004) (17,0.29888) (33,0.29369) (65,0.29797) (97,0.31071) (129,0.33866) (193,0.30325) (257,0.2986) (385,0.29972) (512,0.28886)};
      \addplot[temperature, range line] coordinates {(1,0.080444) (17,0.079621) (33,0.082565) (65,0.079665) (97,0.075763) (129,0.072747) (193,0.067936) (257,0.06331) (385,0.056903) (512,0.051093)};
      \addplot[temperature, range line] coordinates {(1,0.29005) (17,0.29532) (33,0.27871) (65,0.27936) (97,0.27033) (129,0.26592) (193,0.24581) (257,0.22679) (385,0.20766) (512,0.18756)};
      \addplot[no projection, range line] coordinates {(1,0.080456) (17,0.080112) (33,0.079732) (65,0.083247) (97,0.082732) (129,0.081897) (193,0.080522) (257,0.079201) (385,0.07614) (512,0.072862)};
      \addplot[no projection, range line] coordinates {(1,0.29004) (17,0.29888) (33,0.2959) (65,0.29072) (97,0.28682) (129,0.28675) (193,0.28399) (257,0.28841) (385,0.28919) (512,0.28316)};
      \addplot[mean value] coordinates {(-15,0.16892) (527,0.16892)};
      \addplot[galerkin] table[x=n, y=galerkin] {\curvesb};
      \addplot[least squares] table[x=n, y=least_squares] {\curvesb};
      \addplot[revealed range] table[x=n, y=revealed_range] {\curvesb};
      \addplot[temperature] table[x=n, y=temperature] {\curvesb};
      \addplot[no projection] table[x=n, y=no_projection] {\curvesb};
  \end{groupplot}
  \node[below=4pt] at (current bounding box.south) {\pgfplotslegendfromname{legend-projections-qdaim-exact-v5-linf-sink32-iqr}};
\end{tikzpicture}

%% file: figures/tikz/projections_operator_gaim_exact_v1_linf_sink32.tex
\begin{tikzpicture}
  \pgfplotsset{every axis/.append style={
      scale only axis, /tikz/font=\footnotesize, label style={align=center}, title style={align=center},
      legend cell align=left,
      legend style={font=\footnotesize, /tikz/every even column/.append style={column sep=7pt}}}}
  \pgfplotsset{
    galerkin/.style={blue, mark=*},
    least squares/.style={red, mark=square*},
    revealed range/.style={cyan!60!black, mark=triangle*},
    temperature/.style={orange!50!black, mark=diamond*},
    no projection/.style={black!55, mark=x, mark size=2.4pt},
    mean value/.style={black, thin, dashdotted, mark=none},
    zero readout/.style={black, thin, densely dotted, mark=none},
    every axis plot/.append style={very thick, mark size=1.8pt, mark options={solid}, forget plot},
    legend image code/.code={\draw[very thick, mark size=1.8pt, mark options={solid}, mark repeat=2, mark phase=2, #1]
      plot coordinates {(0pt,0pt) (10pt,0pt) (20pt,0pt)};},
    band/.style={draw=none, fill, fill opacity=0.12, mark=none},
    range line/.style={thin, densely dashed, mark=none},
    range legend/.style={black, thin, densely dashed, mark=none},
    band legend/.style={legend image code/.code={\fill[black, fill opacity=0.12] (0pt,-3pt) rectangle (20pt,3pt);}},
  }
  \pgfplotstableread[row sep=\\]{
    n galerkin least_squares revealed_range temperature no_projection\\
    1 0.57162 0.57165 0.61367 1.0351 0.57165\\
    17 0.063941 0.064005 0.064894 0.46297 0.064007\\
    33 0.03785 0.03792 0.04074 0.30574 0.037924\\
    65 0.01919 0.019208 0.033379 0.19849 0.019211\\
    97 0.011167 0.011172 0.062042 0.17751 0.011172\\
    129 0.0081923 0.0081969 2.5313 0.14355 0.0081972\\
    193 0.0048531 0.0048566 0.033497 0.10209 0.004857\\
    257 0.0034443 0.0034471 0.017145 0.073033 0.0034474\\
    385 0.0020034 0.0020042 0.0088883 0.045846 0.0020042\\
    512 0.0013239 0.0013245 0.0029144 0.028637 0.0013245\\
  }\curvesa
  \pgfplotstableread[row sep=\\]{
    n galerkin least_squares revealed_range temperature no_projection\\
    1 0.61012 0.61012 0.62195 3.7768 0.61012\\
    17 0.084643 0.08482 0.085976 5.7514 0.084821\\
    33 0.053719 0.053884 0.060351 5.0777 0.053896\\
    65 0.021984 0.02202 0.04908 4.371 0.022026\\
    97 0.013368 0.013377 0.11933 4.3704 0.013377\\
    129 0.0099216 0.0099276 5.1916 3.5853 0.0099279\\
    193 0.0060738 0.0060795 0.086367 2.5421 0.0060799\\
    257 0.0043538 0.0043589 0.043727 1.7939 0.004359\\
    385 0.0025679 0.0025697 0.014961 1.0697 0.0025696\\
    512 0.0017468 0.0017483 0.0056793 0.6166 0.0017482\\
  }\curvesb
  \begin{groupplot}[
      group style={group size=2 by 1, horizontal sep=46pt},
      width=176pt, height=120pt, grid=major, tick label style={font=\scriptsize},
      xmin=-15, xmax=527, xtick={0,100,...,500}, ymode=log, ymin=0.001, ymax=10,]
    \nextgroupplot[ylabel={$\|M_{\Delta,\eta}\|_2^2\,/\,\|M_{\eta}\|_2^2$}, xlabel={number of nodes $n$}, legend to name=legend-projections-operator-gaim-exact-v1-linf-sink32, legend columns=2]
      \addplot[galerkin] table[x=n, y=galerkin] {\curvesa};
      \addplot[least squares] table[x=n, y=least_squares] {\curvesa};
      \addplot[revealed range] table[x=n, y=revealed_range] {\curvesa};
      \addplot[temperature] table[x=n, y=temperature] {\curvesa};
      \addplot[no projection] table[x=n, y=no_projection] {\curvesa};
      \addlegendimage{galerkin}\addlegendentry{GAIM (Galerkin)}
      \addlegendimage{least squares}\addlegendentry{least squares (direct and Petrov--Galerkin)}
      \addlegendimage{revealed range}\addlegendentry{Q-DAIM revealed range}
      \addlegendimage{temperature}\addlegendentry{WildCat temperature, $\tau\approx2.0$}
      \addlegendimage{no projection}\addlegendentry{no projection ($W = V[\mathfrak L,:]$)}
    \nextgroupplot[ylabel={$\|M^{h}_{\Delta,\eta}\|_2^2\,/\,\|M^{h}_{\eta}\|_2^2$}, xlabel={number of nodes $n$}]
      \addplot[galerkin] table[x=n, y=galerkin] {\curvesb};
      \addplot[least squares] table[x=n, y=least_squares] {\curvesb};
      \addplot[revealed range] table[x=n, y=revealed_range] {\curvesb};
      \addplot[temperature] table[x=n, y=temperature] {\curvesb};
      \addplot[no projection] table[x=n, y=no_projection] {\curvesb};
  \end{groupplot}
  \node[below=4pt] at (current bounding box.south) {\pgfplotslegendfromname{legend-projections-operator-gaim-exact-v1-linf-sink32}};
\end{tikzpicture}

%% file: figures/tikz/projections_operator_qdaim_exact_v5_linf_sink32.tex
\begin{tikzpicture}
  \pgfplotsset{every axis/.append style={
      scale only axis, /tikz/font=\footnotesize, label style={align=center}, title style={align=center},
      legend cell align=left,
      legend style={font=\footnotesize, /tikz/every even column/.append style={column sep=7pt}}}}
  \pgfplotsset{
    galerkin/.style={blue, mark=*},
    least squares/.style={red, mark=square*},
    revealed range/.style={cyan!60!black, mark=triangle*},
    temperature/.style={orange!50!black, mark=diamond*},
    no projection/.style={black!55, mark=x, mark size=2.4pt},
    mean value/.style={black, thin, dashdotted, mark=none},
    zero readout/.style={black, thin, densely dotted, mark=none},
    every axis plot/.append style={very thick, mark size=1.8pt, mark options={solid}, forget plot},
    legend image code/.code={\draw[very thick, mark size=1.8pt, mark options={solid}, mark repeat=2, mark phase=2, #1]
      plot coordinates {(0pt,0pt) (10pt,0pt) (20pt,0pt)};},
    band/.style={draw=none, fill, fill opacity=0.12, mark=none},
    range line/.style={thin, densely dashed, mark=none},
    range legend/.style={black, thin, densely dashed, mark=none},
    band legend/.style={legend image code/.code={\fill[black, fill opacity=0.12] (0pt,-3pt) rectangle (20pt,3pt);}},
  }
  \pgfplotstableread[row sep=\\]{
    n galerkin least_squares revealed_range temperature no_projection\\
    1 0.61998 0.61998 0.61999 1.0332 0.61998\\
    17 0.19285 0.19299 0.19449 0.82407 0.19299\\
    33 0.12836 0.12838 0.13403 0.63449 0.12838\\
    65 0.07945 0.07946 0.10863 0.30924 0.07946\\
    97 0.059886 0.059929 0.095657 0.17587 0.059929\\
    129 0.050999 0.05105 0.27666 0.1553 0.051055\\
    193 0.032909 0.032948 0.063326 0.10632 0.032955\\
    257 0.021668 0.021671 0.033744 0.091484 0.021671\\
    385 0.014211 0.014213 0.018485 0.076254 0.014213\\
    512 0.011971 0.011975 0.014617 0.062259 0.011975\\
  }\curvesa
  \pgfplotstableread[row sep=\\]{
    n galerkin least_squares revealed_range temperature no_projection\\
    1 0.6396 0.63961 0.63961 1.4354 0.63961\\
    17 0.21488 0.21504 0.21696 3.545 0.21504\\
    33 0.1584 0.15842 0.16475 3.802 0.15842\\
    65 0.10346 0.10348 0.14207 3.3268 0.10347\\
    97 0.076331 0.076346 0.11686 2.6815 0.076347\\
    129 0.070813 0.07084 0.34684 2.9146 0.070841\\
    193 0.041728 0.041733 0.099232 2.2125 0.041733\\
    257 0.028255 0.028261 0.062102 1.6776 0.028261\\
    385 0.01788 0.017885 0.029948 1.0129 0.017885\\
    512 0.01505 0.015056 0.021836 0.56942 0.015056\\
  }\curvesb
  \begin{groupplot}[
      group style={group size=2 by 1, horizontal sep=46pt},
      width=176pt, height=120pt, grid=major, tick label style={font=\scriptsize},
      xmin=-15, xmax=527, xtick={0,100,...,500}, ymode=log, ymin=0.009577, ymax=4.753,]
    \nextgroupplot[ylabel={$\|M_{\Delta,\eta}\|_2^2\,/\,\|M_{\eta}\|_2^2$}, xlabel={number of nodes $n$}, legend to name=legend-projections-operator-qdaim-exact-v5-linf-sink32, legend columns=2]
      \addplot[galerkin] table[x=n, y=galerkin] {\curvesa};
      \addplot[least squares] table[x=n, y=least_squares] {\curvesa};
      \addplot[revealed range] table[x=n, y=revealed_range] {\curvesa};
      \addplot[temperature] table[x=n, y=temperature] {\curvesa};
      \addplot[no projection] table[x=n, y=no_projection] {\curvesa};
      \addlegendimage{galerkin}\addlegendentry{GAIM (Galerkin)}
      \addlegendimage{least squares}\addlegendentry{least squares (direct and Petrov--Galerkin)}
      \addlegendimage{revealed range}\addlegendentry{Q-DAIM revealed range}
      \addlegendimage{temperature}\addlegendentry{WildCat temperature, $\tau\approx2.0$}
      \addlegendimage{no projection}\addlegendentry{no projection ($W = V[\mathfrak L,:]$)}
    \nextgroupplot[ylabel={$\|M^{h}_{\Delta,\eta}\|_2^2\,/\,\|M^{h}_{\eta}\|_2^2$}, xlabel={number of nodes $n$}]
      \addplot[galerkin] table[x=n, y=galerkin] {\curvesb};
      \addplot[least squares] table[x=n, y=least_squares] {\curvesb};
      \addplot[revealed range] table[x=n, y=revealed_range] {\curvesb};
      \addplot[temperature] table[x=n, y=temperature] {\curvesb};
      \addplot[no projection] table[x=n, y=no_projection] {\curvesb};
  \end{groupplot}
  \node[below=4pt] at (current bounding box.south) {\pgfplotslegendfromname{legend-projections-operator-qdaim-exact-v5-linf-sink32}};
\end{tikzpicture}

%% file: figures/tikz/gram_conditioning_per_layer.tex
\begin{tikzpicture}
  \pgfplotsset{every axis/.append style={
      scale only axis, /tikz/font=\footnotesize, label style={align=center}, title style={align=center},
      legend cell align=left,
      legend style={font=\footnotesize, /tikz/every even column/.append style={column sep=7pt}}}}
  \definecolor{kgauss}{HTML}{1F77B4}\definecolor{korigin}{HTML}{D62728}\definecolor{kmean}{HTML}{FF7F0E}
  \pgfplotsset{
    skgauss/.style={kgauss, mark=*}, skorigin/.style={korigin, mark=square*}, skmean/.style={kmean, mark=diamond*},
    median/.style={mark=none, thick, forget plot},
    head/.style={only marks, mark size=1.1pt, fill opacity=0.55, draw opacity=0, forget plot},
    bracket/.style={only marks, mark size=1.3pt, fill=white, error bars/.cd, y dir=plus, y explicit, error mark=none,
      error bar style={thin}, /pgfplots/forget plot},
    offscale/.style={only marks, mark=triangle*, mark size=1.8pt, forget plot},
    resolution/.style={densely dotted, gray},
  }
  \pgfplotstableread[row sep=\\]{
    layer median\\
    0 1.8935\\
    1 0.2927\\
    2 0.86857\\
    3 0.61517\\
    4 0.3792\\
    5 0.41568\\
    6 0.42811\\
    7 0.32794\\
    8 0.32885\\
    9 0.34248\\
    10 0.20824\\
    11 0.30037\\
    12 0.46024\\
    13 0.41578\\
    14 0.42083\\
    15 0.42446\\
    16 0.38321\\
    17 0.34937\\
    18 0.49922\\
    19 0.56411\\
    20 0.5411\\
    21 0.46157\\
    22 0.49486\\
    23 0.45286\\
    24 0.45589\\
    25 0.3456\\
    26 0.5731\\
    27 0.78602\\
  }\condakgaussmedian
  \pgfplotstableread[row sep=\\]{
    layer value up\\
    0 0.35299 0\\
    0 1.52 0\\
    0 2.267 0\\
    0 4.1144 0\\
    1 1.323 0\\
    1 0.46001 0\\
    1 5.1433e-05 0\\
    1 0.12538 0\\
    2 2.1132 0\\
    2 0.77781 0\\
    2 0.95932 0\\
    2 0.32946 0\\
    3 0.53955 0\\
    3 0.70197 0\\
    3 0.69078 0\\
    3 0.16455 0\\
    4 0.64989 0\\
    4 0.35117 0\\
    4 0.40723 0\\
    4 0.1025 0\\
    5 0.46923 0\\
    5 0.86875 0\\
    5 0.36214 0\\
    5 0.16779 0\\
    6 0.070596 0\\
    6 0.78929 0\\
    6 0.706 0\\
    6 0.15022 0\\
    7 0.21908 0\\
    7 0.2393 0\\
    7 0.41657 0\\
    7 0.55992 0\\
    8 0.37927 0\\
    8 0.35031 0\\
    8 0.12823 0\\
    8 0.30738 0\\
    9 0.28757 0\\
    9 0.28633 0\\
    9 0.41754 0\\
    9 0.39739 0\\
    10 0.32287 0\\
    10 0.16369 0\\
    10 0.25279 0\\
    10 0.1498 0\\
    11 0.24715 0\\
    11 0.40931 0\\
    11 0.15455 0\\
    11 0.35359 0\\
    12 0.55786 0\\
    12 0.63738 0\\
    12 0.36262 0\\
    12 0.20964 0\\
    13 0.45215 0\\
    13 0.14545 0\\
    13 0.37942 0\\
    13 0.5603 0\\
    14 0.40236 0\\
    14 0.50004 0\\
    14 0.33964 0\\
    14 0.4393 0\\
    15 0.39669 0\\
    15 0.1815 0\\
    15 0.50152 0\\
    15 0.45223 0\\
    16 0.44827 0\\
    16 0.5819 0\\
    16 0.31815 0\\
    16 0.31308 0\\
    17 0.23769 0\\
    17 0.30558 0\\
    17 0.5008 0\\
    17 0.39316 0\\
    18 0.60895 0\\
    18 0.42629 0\\
    18 0.31328 0\\
    18 0.57214 0\\
    19 0.71523 0\\
    19 0.23242 0\\
    19 0.6287 0\\
    19 0.49951 0\\
    20 0.51407 0\\
    20 0.57362 0\\
    20 0.56813 0\\
    20 0.51194 0\\
    21 0.54266 0\\
    21 0.59557 0\\
    21 0.14447 0\\
    21 0.38048 0\\
    22 0.20317 0\\
    22 0.29375 0\\
    22 0.69598 0\\
    22 0.87078 0\\
    23 0.58259 0\\
    23 0.32313 0\\
    23 0.24796 0\\
    23 0.71831 0\\
    24 0.14088 0\\
    24 0.91893 0\\
    24 0.3088 0\\
    24 0.60298 0\\
    25 0.12292 0\\
    25 0.78631 0\\
    25 0.077585 0\\
    25 0.56828 0\\
    26 0.85412 0\\
    26 0.31021 0\\
    26 0.83598 0\\
    26 0.10563 0\\
    27 0.90654 0\\
    27 0.16767 0\\
    27 1.0157 0\\
    27 0.6655 0\\
  }\condakgausshead
  \pgfplotstableread[row sep=\\]{
    layer median\\
    0 60.345\\
    1 14.052\\
    2 19.483\\
    3 23.472\\
    4 15.342\\
    5 13.815\\
    6 11.711\\
    7 15.448\\
    8 16.351\\
    9 19.811\\
    10 14.414\\
    11 15.184\\
    12 15.7\\
    13 27.258\\
    14 20.31\\
    15 23.222\\
    16 16.431\\
    17 14.374\\
    18 15.84\\
    19 16.266\\
    20 16.194\\
    21 17.107\\
    22 18.642\\
    23 17.141\\
    24 17.564\\
    25 16.648\\
    26 15.767\\
    27 18.517\\
  }\condakoriginmedian
  \pgfplotstableread[row sep=\\]{
    layer value up\\
    3 10.805 0\\
    6 8.7358 0\\
    6 10.089 0\\
    10 9.6333 0\\
    13 10.229 0\\
    21 11.242 0\\
    25 9.7255 0\\
  }\condakoriginhead
  \pgfplotstableread[row sep=\\]{
    layer value up\\
    1 25.746 1.323\\
    1 14.813 0.46001\\
    1 12.688 5.1433e-05\\
    1 13.292 0.12538\\
    2 15.53 2.1132\\
    2 34.944 0.77781\\
    2 17.167 0.95932\\
    2 21.799 0.32946\\
    3 39.402 0.53955\\
    3 20.933 0.69078\\
    3 26.011 0.16455\\
    4 16.901 0.64989\\
    4 17.935 0.35117\\
    4 12.586 0.40723\\
    4 13.783 0.1025\\
    5 20.356 0.46923\\
    5 12.968 0.86875\\
    5 14.662 0.36214\\
    5 12.905 0.16779\\
    6 18.12 0.78929\\
    6 13.333 0.15022\\
    7 12.522 0.21908\\
    7 13.403 0.2393\\
    7 17.493 0.41657\\
    7 19.344 0.55992\\
    8 12.72 0.37927\\
    8 16.883 0.35031\\
    8 15.818 0.12823\\
    8 18.983 0.30738\\
    9 20.851 0.28757\\
    9 14.182 0.28633\\
    9 18.771 0.41754\\
    9 26.019 0.39739\\
    10 16.387 0.16369\\
    10 16.609 0.25279\\
    10 12.441 0.1498\\
    11 14.212 0.24715\\
    11 16.156 0.40931\\
    11 13.804 0.15455\\
    11 20.295 0.35359\\
    12 15.145 0.55786\\
    12 13.672 0.63738\\
    12 16.256 0.36262\\
    12 17.354 0.20964\\
    13 30.127 0.37942\\
    13 24.39 0.5603\\
    14 31.574 0.40236\\
    14 14.047 0.50004\\
    14 19.315 0.33964\\
    14 21.306 0.4393\\
    15 24.947 0.39669\\
    15 12.5 0.1815\\
    15 27.052 0.50152\\
    15 21.497 0.45223\\
    16 17.543 0.44827\\
    16 15.318 0.5819\\
    16 19.256 0.31815\\
    16 12.803 0.31308\\
    17 12.217 0.23769\\
    17 14.521 0.30558\\
    17 14.228 0.5008\\
    17 17.829 0.39316\\
    18 13.258 0.60895\\
    18 16.297 0.42629\\
    18 18.795 0.31328\\
    18 15.384 0.57214\\
    19 13.552 0.23242\\
    19 14.982 0.6287\\
    19 17.551 0.49951\\
    20 18.929 0.51407\\
    20 13.999 0.57362\\
    20 16.252 0.56813\\
    20 16.136 0.51194\\
    21 21.242 0.54266\\
    21 16.481 0.59557\\
    21 17.732 0.38048\\
    22 18.421 0.20317\\
    22 18.863 0.29375\\
    22 19.299 0.69598\\
    22 17.633 0.87078\\
    23 21.356 0.58259\\
    23 15.887 0.32313\\
    23 15.138 0.24796\\
    23 18.396 0.71831\\
    24 16.654 0.14088\\
    24 17.495 0.91893\\
    24 18.085 0.3088\\
    24 17.632 0.60298\\
    25 16.611 0.12292\\
    25 16.908 0.78631\\
    25 16.685 0.56828\\
    26 15.337 0.85412\\
    26 27.702 0.31021\\
    26 15.026 0.83598\\
    26 16.197 0.10563\\
    27 18.36 0.90654\\
    27 16.214 1.0157\\
    27 18.673 0.6655\\
  }\condakoriginbracket
  \pgfplotstableread[row sep=\\]{
    layer median\\
    0 9.8958\\
    1 12.538\\
    2 12.757\\
    3 11.844\\
    4 9.9959\\
    5 10.713\\
    6 7.4843\\
    7 7.7622\\
    8 11.366\\
    9 9.8263\\
    10 8.5865\\
    11 12.128\\
    12 12.111\\
    13 15.065\\
    14 14.114\\
    15 15.902\\
    16 13.084\\
    17 14.651\\
    18 11.839\\
    19 13.573\\
    20 12.064\\
    21 12.101\\
    22 13.009\\
    23 11.176\\
    24 11.259\\
    25 11.161\\
    26 11.321\\
    27 11.388\\
  }\condakmeanmedian
  \pgfplotstableread[row sep=\\]{
    layer value up\\
    0 8.4911 0\\
    0 8.2394 0\\
    0 11.301 0\\
    1 11.601 0\\
    1 11.881 0\\
    2 9.4869 0\\
    2 11.927 0\\
    3 11.932 0\\
    3 11.756 0\\
    3 10.19 0\\
    4 9.5705 0\\
    4 9.4126 0\\
    4 10.421 0\\
    5 10.984 0\\
    5 6.885 0\\
    5 10.443 0\\
    5 11.215 0\\
    6 6.3773 0\\
    6 8.5912 0\\
    6 5.578 0\\
    6 8.6695 0\\
    7 7.7307 0\\
    7 7.7938 0\\
    7 8.3304 0\\
    7 6.9755 0\\
    8 7.8288 0\\
    8 11.027 0\\
    8 11.704 0\\
    9 10.608 0\\
    9 9.045 0\\
    9 6.9797 0\\
    9 11.424 0\\
    10 6.4329 0\\
    10 10.307 0\\
    10 8.4417 0\\
    10 8.7314 0\\
    11 11.459 0\\
    11 9.745 0\\
    12 11.165 0\\
    12 10.793 0\\
    13 8.0025 0\\
    15 9.4253 0\\
    15 11.062 0\\
    17 10.125 0\\
    18 10.995 0\\
    18 8.1977 0\\
    19 8.0729 0\\
    20 11.524 0\\
    20 12.036 0\\
    21 10.386 0\\
    21 7.8139 0\\
    22 11.242 0\\
    23 10.909 0\\
    23 11.443 0\\
    23 9.3595 0\\
    24 9.9568 0\\
    24 10.483 0\\
    24 12.035 0\\
    25 9.9988 0\\
    25 9.1497 0\\
    26 9.4024 0\\
    26 10.493 0\\
    27 11.579 0\\
    27 10.898 0\\
    27 11.197 0\\
  }\condakmeanhead
  \pgfplotstableread[row sep=\\]{
    layer value up\\
    0 20.981 1.52\\
    1 34.029 1.323\\
    1 13.195 0.46001\\
    2 13.588 2.1132\\
    2 19.308 0.77781\\
    3 13.398 0.70197\\
    4 13.398 0.35117\\
    8 15.893 0.30738\\
    11 12.796 0.40931\\
    11 16.395 0.35359\\
    12 13.057 0.63738\\
    12 14.631 0.20964\\
    13 13.508 0.45215\\
    13 18.994 0.37942\\
    13 16.623 0.5603\\
    14 27.055 0.40236\\
    14 14.757 0.50004\\
    14 13.471 0.33964\\
    14 12.365 0.4393\\
    15 20.741 0.39669\\
    15 22.978 0.50152\\
    16 14.569 0.44827\\
    16 13.292 0.5819\\
    16 12.545 0.31815\\
    16 12.877 0.31308\\
    17 14.7 0.23769\\
    17 14.602 0.5008\\
    17 15.573 0.39316\\
    18 12.684 0.42629\\
    18 13.995 0.31328\\
    19 13.847 0.71523\\
    19 13.299 0.6287\\
    19 15.046 0.49951\\
    20 13.759 0.51407\\
    20 12.091 0.56813\\
    21 15.789 0.54266\\
    21 13.816 0.38048\\
    22 17.04 0.20317\\
    22 12.715 0.29375\\
    22 13.303 0.87078\\
    23 15.02 0.58259\\
    24 14.51 0.60298\\
    25 14.467 0.12292\\
    25 12.323 0.56828\\
    26 20.791 0.31021\\
    26 12.149 0.10563\\
    27 15.866 0.16767\\
  }\condakmeanbracket
  \pgfplotstableread[row sep=\\]{
    layer median\\
    0 1.4913\\
    1 0.17506\\
    2 0.84494\\
    3 0.42632\\
    4 0.18796\\
    5 0.3764\\
    6 0.328\\
    7 0.24051\\
    8 0.32884\\
    9 0.31462\\
    10 0.17766\\
    11 0.26264\\
    12 0.33875\\
    13 0.31243\\
    14 0.29961\\
    15 0.37469\\
    16 0.31223\\
    17 0.28766\\
    18 0.32689\\
    19 0.40772\\
    20 0.52233\\
    21 0.33849\\
    22 0.31526\\
    23 0.34325\\
    24 0.29083\\
    25 0.23531\\
    26 0.44993\\
    27 0.58131\\
  }\condbkgaussmedian
  \pgfplotstableread[row sep=\\]{
    layer value up\\
    0 0.20508 0\\
    0 1.52 0\\
    0 1.4626 0\\
    0 2.3889 0\\
    1 1.1954 0\\
    1 0.25228 0\\
    1 2.5781e-05 0\\
    1 0.097836 0\\
    2 1.7502 0\\
    2 0.7692 0\\
    2 0.92068 0\\
    2 0.31225 0\\
    3 0.5379 0\\
    3 0.70093 0\\
    3 0.31474 0\\
    3 0.16455 0\\
    4 0.64983 0\\
    4 0.21917 0\\
    4 0.15676 0\\
    4 0.066339 0\\
    5 0.39081 0\\
    5 0.55322 0\\
    5 0.36198 0\\
    5 0.10162 0\\
    6 0.070596 0\\
    6 0.78482 0\\
    6 0.50578 0\\
    6 0.15022 0\\
    7 0.16507 0\\
    7 0.2393 0\\
    7 0.24172 0\\
    7 0.47109 0\\
    8 0.36436 0\\
    8 0.35031 0\\
    8 0.084601 0\\
    8 0.30737 0\\
    9 0.28737 0\\
    9 0.24195 0\\
    9 0.34188 0\\
    9 0.38581 0\\
    10 0.31811 0\\
    10 0.11928 0\\
    10 0.21563 0\\
    10 0.1397 0\\
    11 0.20951 0\\
    11 0.31576 0\\
    11 0.1383 0\\
    11 0.34041 0\\
    12 0.54258 0\\
    12 0.37216 0\\
    12 0.30535 0\\
    12 0.18851 0\\
    13 0.37214 0\\
    13 0.09986 0\\
    13 0.25272 0\\
    13 0.54807 0\\
    14 0.24395 0\\
    14 0.33837 0\\
    14 0.26084 0\\
    14 0.43928 0\\
    15 0.3946 0\\
    15 0.1815 0\\
    15 0.50009 0\\
    15 0.35478 0\\
    16 0.41049 0\\
    16 0.30433 0\\
    16 0.31808 0\\
    16 0.30637 0\\
    17 0.23739 0\\
    17 0.16456 0\\
    17 0.42155 0\\
    17 0.33793 0\\
    18 0.48505 0\\
    18 0.34097 0\\
    18 0.3128 0\\
    18 0.25249 0\\
    19 0.68542 0\\
    19 0.23227 0\\
    19 0.58316 0\\
    19 0.22747 0\\
    20 0.45377 0\\
    20 0.54212 0\\
    20 0.56769 0\\
    20 0.50254 0\\
    21 0.33719 0\\
    21 0.57669 0\\
    21 0.12161 0\\
    21 0.33978 0\\
    22 0.12103 0\\
    22 0.28203 0\\
    22 0.66704 0\\
    22 0.34849 0\\
    23 0.53647 0\\
    23 0.10002 0\\
    23 0.15004 0\\
    23 0.7183 0\\
    24 0.14088 0\\
    24 0.27287 0\\
    24 0.3088 0\\
    24 0.43237 0\\
    25 0.054511 0\\
    25 0.40994 0\\
    25 0.060693 0\\
    25 0.4678 0\\
    26 0.58966 0\\
    26 0.3102 0\\
    26 0.64674 0\\
    26 0.062104 0\\
    27 0.62529 0\\
    27 0.084292 0\\
    27 0.65396 0\\
    27 0.53733 0\\
  }\condbkgausshead
  \pgfplotstableread[row sep=\\]{
    layer median\\
    0 50.477\\
    1 11.614\\
    2 16.015\\
    3 17.574\\
    4 13.099\\
    5 12.958\\
    6 10.334\\
    7 13.766\\
    8 15.209\\
    9 17.517\\
    10 11.589\\
    11 14.975\\
    12 13.153\\
    13 24.548\\
    14 17.363\\
    15 17.785\\
    16 15.265\\
    17 12.407\\
    18 12.724\\
    19 14.487\\
    20 14.32\\
    21 16.131\\
    22 14.9\\
    23 15.828\\
    24 13.625\\
    25 14.602\\
    26 13.064\\
    27 14.608\\
  }\condbkoriginmedian
  \pgfplotstableread[row sep=\\]{
    layer value up\\
    1 12.456 0\\
    1 10.772 0\\
    1 10.487 0\\
    3 9.4689 0\\
    4 10.626 0\\
    4 12.544 0\\
    5 11.693 0\\
    6 8.5 0\\
    6 7.1155 0\\
    6 12.168 0\\
    7 12.526 0\\
    7 12.839 0\\
    8 11.471 0\\
    10 8.2376 0\\
    10 12.048 0\\
    10 11.13 0\\
    11 12.427 0\\
    12 11.044 0\\
    12 10.979 0\\
    13 9.0924 0\\
    14 9.501 0\\
    15 12.303 0\\
    16 9.3851 0\\
    17 10.789 0\\
    17 11.476 0\\
    18 12.594 0\\
    18 12.854 0\\
    18 10.791 0\\
    19 12.515 0\\
    20 12.911 0\\
    21 10.858 0\\
    22 12.884 0\\
    23 10.917 0\\
    24 12.91 0\\
    24 12.88 0\\
    25 9.0191 0\\
    26 10.621 0\\
    26 11.107 0\\
  }\condbkoriginhead
  \pgfplotstableread[row sep=\\]{
    layer value up\\
    0 38.698 1.302\\
    1 21.646 1.1954\\
    2 12.944 1.5785\\
    2 16.934 0.7692\\
    2 15.13 0.92068\\
    2 16.9 0.31225\\
    3 37.836 0.5379\\
    3 14.767 0.31474\\
    3 20.381 0.16455\\
    4 13.654 0.64983\\
    4 15.234 0.21917\\
    5 17.944 0.39081\\
    5 12.944 0.55069\\
    5 12.971 0.36198\\
    6 17.729 0.78482\\
    7 14.693 0.24172\\
    7 14.931 0.47109\\
    8 14.6 0.35031\\
    8 15.818 0.084601\\
    8 16.528 0.30737\\
    9 20.851 0.28737\\
    9 14.182 0.24195\\
    9 13.895 0.34188\\
    9 21.747 0.38581\\
    10 14.278 0.11928\\
    11 16.146 0.31576\\
    11 13.804 0.1383\\
    11 20.295 0.34041\\
    12 15.263 0.30535\\
    12 17.354 0.18851\\
    13 28.6 0.25272\\
    13 20.497 0.54807\\
    14 30.488 0.24395\\
    14 16.568 0.26084\\
    14 18.159 0.43928\\
    15 18.803 0.3946\\
    15 25.138 0.50009\\
    15 16.767 0.35478\\
    16 15.676 0.41049\\
    16 14.854 0.30433\\
    16 19.256 0.31808\\
    17 13.338 0.42155\\
    17 13.664 0.33793\\
    18 18.795 0.3128\\
    19 32.228 0.68542\\
    19 14.135 0.58316\\
    19 14.838 0.22747\\
    20 15.595 0.45377\\
    20 14.991 0.56769\\
    20 13.649 0.50254\\
    21 21.242 0.33719\\
    21 14.629 0.57669\\
    21 17.633 0.33978\\
    22 16.384 0.12103\\
    22 17.009 0.66704\\
    22 13.417 0.34849\\
    23 21.356 0.53647\\
    23 13.957 0.10002\\
    23 17.698 0.7183\\
    24 17.187 0.3088\\
    24 14.34 0.43237\\
    25 16.448 0.054511\\
    25 14.969 0.40994\\
    25 14.236 0.4678\\
    26 24.982 0.3102\\
    26 15.021 0.062104\\
    27 13.935 0.62529\\
    27 14.005 0.65396\\
    27 15.211 0.53733\\
  }\condbkoriginbracket
  \pgfplotstableread[row sep=\\]{
    layer median\\
    0 7.8642\\
    1 11.723\\
    2 11.652\\
    3 11.41\\
    4 9.8817\\
    5 9.5652\\
    6 6.9233\\
    7 7.4674\\
    8 10.6\\
    9 9.5656\\
    10 8.4461\\
    11 10.892\\
    12 11.219\\
    13 12.785\\
    14 12.785\\
    15 13.948\\
    16 11.749\\
    17 12.307\\
    18 11.222\\
    19 12.384\\
    20 11.397\\
    21 11.675\\
    22 12.097\\
    23 10.758\\
    24 10.759\\
    25 9.9843\\
    26 10.432\\
    27 10.122\\
  }\condbkmeanmedian
  \pgfplotstableread[row sep=\\]{
    layer value up\\
    0 7.3968 0\\
    0 7.3699 0\\
    0 8.3317 0\\
    1 11.62 0\\
    1 10.484 0\\
    1 11.827 0\\
    2 12.337 0\\
    2 8.5981 0\\
    2 10.967 0\\
    3 11.77 0\\
    3 11.05 0\\
    3 9.2997 0\\
    4 9.57 0\\
    4 12.917 0\\
    4 7.8205 0\\
    4 10.193 0\\
    5 9.4406 0\\
    5 6.2282 0\\
    5 9.6898 0\\
    5 10.102 0\\
    6 6.2207 0\\
    6 7.6259 0\\
    6 4.6207 0\\
    6 8.4734 0\\
    7 7.2012 0\\
    7 7.7336 0\\
    7 8.3151 0\\
    7 6.5058 0\\
    8 7.0973 0\\
    8 9.6412 0\\
    8 11.558 0\\
    9 10.087 0\\
    9 9.0438 0\\
    9 6.6713 0\\
    9 10.301 0\\
    10 5.8717 0\\
    10 10.182 0\\
    10 8.1608 0\\
    10 8.7314 0\\
    11 9.0233 0\\
    11 12.574 0\\
    11 9.2098 0\\
    12 11.154 0\\
    12 11.285 0\\
    12 9.5175 0\\
    13 6.9896 0\\
    13 12.469 0\\
    14 12.298 0\\
    14 11.556 0\\
    15 9.1548 0\\
    15 11.061 0\\
    16 11.695 0\\
    16 11.861 0\\
    16 9.871 0\\
    16 11.803 0\\
    17 12.134 0\\
    17 9.1391 0\\
    17 12.481 0\\
    18 10.558 0\\
    18 11.886 0\\
    18 12.834 0\\
    18 7.1376 0\\
    19 12.594 0\\
    19 8.0728 0\\
    19 12.173 0\\
    20 11.227 0\\
    20 10.982 0\\
    20 11.567 0\\
    21 9.5334 0\\
    21 6.7069 0\\
    22 11.366 0\\
    22 11.082 0\\
    22 12.827 0\\
    23 10.232 0\\
    23 11.284 0\\
    23 9.0198 0\\
    24 9.4827 0\\
    24 6.6603 0\\
    24 12.035 0\\
    24 12.793 0\\
    25 8.569 0\\
    25 7.7142 0\\
    25 11.4 0\\
    26 7.9816 0\\
    26 8.8556 0\\
    26 12.008 0\\
    27 8.9256 0\\
    27 10.126 0\\
    27 10.118 0\\
  }\condbkmeanhead
  \pgfplotstableread[row sep=\\]{
    layer value up\\
    0 20.455 1.52\\
    1 15.21 1.1954\\
    2 17.204 0.7692\\
    3 12.944 0.6147\\
    8 14.307 0.30737\\
    11 15.302 0.34041\\
    12 14.063 0.18851\\
    13 13.1 0.37214\\
    13 15.372 0.25272\\
    14 25.259 0.24395\\
    14 13.272 0.26084\\
    15 16.834 0.3946\\
    15 19.139 0.50009\\
    17 13.523 0.33793\\
    19 14.378 0.22747\\
    20 12.963 0.45377\\
    21 15.789 0.33719\\
    21 13.816 0.33978\\
    22 15.664 0.12103\\
    23 15.02 0.53647\\
    25 12.979 0.054511\\
    26 18.685 0.3102\\
    27 13.8 0.084292\\
  }\condbkmeanbracket
  \pgfplotstableread[row sep=\\]{
    layer median\\
    0 0.15587\\
    1 0.021965\\
    2 0.050378\\
    3 0.020689\\
    4 0.10902\\
    5 0.20329\\
    6 0.1725\\
    7 0.27541\\
    8 0.16796\\
    9 0.22487\\
    10 0.11385\\
    11 0.10827\\
    12 0.18463\\
    13 0.15222\\
    14 0.23798\\
    15 0.15881\\
    16 0.28335\\
    17 0.1458\\
    18 0.24836\\
    19 0.21295\\
    20 0.12225\\
    21 0.10022\\
    22 0.11465\\
    23 0.093668\\
    24 0.15728\\
    25 0.074288\\
    26 0.17663\\
    27 0.20359\\
  }\condckgaussmedian
  \pgfplotstableread[row sep=\\]{
    layer value up\\
    0 0.20491 0\\
    0 0.025744 0\\
    0 0.34977 0\\
    0 0.10683 0\\
    1 1.1968 0\\
    1 0.002027 0\\
    1 1.1355e-05 0\\
    1 0.041903 0\\
    2 0.0045464 0\\
    2 0.0064605 0\\
    2 0.23009 0\\
    2 0.094296 0\\
    3 0.015799 0\\
    3 0.011465 0\\
    3 0.070213 0\\
    3 0.02558 0\\
    4 0.28951 0\\
    4 0.043467 0\\
    4 0.15796 0\\
    4 0.060079 0\\
    5 0.29323 0\\
    5 0.5596 0\\
    5 0.11336 0\\
    5 0.08539 0\\
    6 0.015218 0\\
    6 0.22684 0\\
    6 0.31894 0\\
    6 0.11816 0\\
    7 0.20482 0\\
    7 0.071601 0\\
    7 0.34601 0\\
    7 0.37679 0\\
    8 0.18134 0\\
    8 0.15457 0\\
    8 0.040734 0\\
    8 0.18571 0\\
    9 0.25711 0\\
    9 0.19262 0\\
    9 0.31226 0\\
    9 0.16126 0\\
    10 0.31797 0\\
    10 0.059817 0\\
    10 0.13495 0\\
    10 0.092754 0\\
    11 0.1093 0\\
    11 0.11441 0\\
    11 0.10725 0\\
    11 0.093226 0\\
    12 0.19975 0\\
    12 0.16951 0\\
    12 0.26404 0\\
    12 0.11691 0\\
    13 0.15922 0\\
    13 0.14521 0\\
    13 0.10372 0\\
    13 0.16817 0\\
    14 0.2307 0\\
    14 0.30946 0\\
    14 0.17907 0\\
    14 0.24526 0\\
    15 0.21097 0\\
    15 0.066045 0\\
    15 0.10665 0\\
    15 0.22217 0\\
    16 0.32788 0\\
    16 0.39345 0\\
    16 0.18221 0\\
    16 0.23881 0\\
    17 0.11221 0\\
    17 0.085543 0\\
    17 0.28467 0\\
    17 0.17939 0\\
    18 0.19001 0\\
    18 0.30671 0\\
    18 0.13615 0\\
    18 0.35426 0\\
    19 0.58587 0\\
    19 0.099021 0\\
    19 0.24692 0\\
    19 0.17899 0\\
    20 0.061706 0\\
    20 0.16362 0\\
    20 0.12235 0\\
    20 0.12215 0\\
    21 0.11336 0\\
    21 0.10672 0\\
    21 0.09372 0\\
    21 0.059932 0\\
    22 0.1051 0\\
    22 0.090562 0\\
    22 0.12421 0\\
    22 0.13265 0\\
    23 0.17084 0\\
    23 0.061766 0\\
    23 0.12557 0\\
    23 0.026896 0\\
    24 0.021676 0\\
    24 0.46498 0\\
    24 0.040463 0\\
    24 0.2741 0\\
    25 0.021333 0\\
    25 0.40288 0\\
    25 0.032642 0\\
    25 0.11593 0\\
    26 0.20353 0\\
    26 0.17361 0\\
    26 0.17964 0\\
    26 0.064398 0\\
    27 0.19187 0\\
    27 0.072026 0\\
    27 0.23115 0\\
    27 0.21532 0\\
  }\condckgausshead
  \pgfplotstableread[row sep=\\]{
    layer median\\
    0 50.582\\
    1 10.803\\
    2 15.797\\
    3 16.93\\
    4 7.9479\\
    5 10.382\\
    6 8.1724\\
    7 11.593\\
    8 9.4411\\
    9 16.395\\
    10 8.2711\\
    11 8.2751\\
    12 11.303\\
    13 24.417\\
    14 17.09\\
    15 18.084\\
    16 13.452\\
    17 10.363\\
    18 11.207\\
    19 12.588\\
    20 10.959\\
    21 10.138\\
    22 13.399\\
    23 11.281\\
    24 11.612\\
    25 10.365\\
    26 10.929\\
    27 14.031\\
  }\condckoriginmedian
  \pgfplotstableread[row sep=\\]{
    layer value up\\
    1 10.632 0\\
    1 10.479 0\\
    1 10.975 0\\
    2 8.5402 0\\
    3 6.962 0\\
    3 12.724 0\\
    4 8.603 0\\
    4 7.2929 0\\
    4 6.6222 0\\
    5 11.33 0\\
    5 10.352 0\\
    5 10.411 0\\
    5 6.209 0\\
    6 4.2979 0\\
    6 12.404 0\\
    6 8.3885 0\\
    6 7.9562 0\\
    7 8.1498 0\\
    7 8.6345 0\\
    8 9.6698 0\\
    8 9.8136 0\\
    8 8.1579 0\\
    8 9.2124 0\\
    9 10.481 0\\
    10 6.6972 0\\
    10 10.497 0\\
    10 8.8786 0\\
    10 7.6637 0\\
    11 5.9686 0\\
    11 9.3486 0\\
    11 7.2015 0\\
    12 8.0683 0\\
    12 8.3674 0\\
    13 6.1511 0\\
    14 11.115 0\\
    14 12.875 0\\
    15 4.3662 0\\
    16 11.281 0\\
    16 12.242 0\\
    17 10.104 0\\
    17 9.311 0\\
    17 10.622 0\\
    17 12.117 0\\
    18 9.1592 0\\
    18 12.681 0\\
    18 9.7342 0\\
    19 7.599 0\\
    19 9.7343 0\\
    20 11.343 0\\
    20 11.205 0\\
    20 10.713 0\\
    20 9.7474 0\\
    21 9.9181 0\\
    21 7.6649 0\\
    21 10.358 0\\
    22 11.879 0\\
    23 11.096 0\\
    23 11.465 0\\
    23 10.569 0\\
    24 11.476 0\\
    24 10.774 0\\
    24 11.748 0\\
    25 10.379 0\\
    25 7.878 0\\
    25 10.352 0\\
    26 10.853 0\\
    26 10.738 0\\
    26 11.005 0\\
  }\condckoriginhead
  \pgfplotstableread[row sep=\\]{
    layer value up\\
    1 24.107 1.1968\\
    2 29.624 0.0064605\\
    2 13.521 0.23009\\
    2 18.073 0.094296\\
    3 38.783 0.015799\\
    3 21.137 0.02558\\
    4 13.297 0.28951\\
    7 14.551 0.34601\\
    7 19.344 0.37679\\
    9 17.24 0.25711\\
    9 15.551 0.31226\\
    9 20.886 0.16126\\
    11 13.978 0.093226\\
    12 14.238 0.19975\\
    12 15.443 0.26404\\
    13 35.819 0.15922\\
    13 25.957 0.10372\\
    13 22.876 0.16817\\
    14 22.305 0.2307\\
    14 21.306 0.24526\\
    15 15.484 0.21097\\
    15 20.684 0.10665\\
    15 21.497 0.22217\\
    16 14.663 0.32788\\
    16 15.37 0.18221\\
    18 13.374 0.30671\\
    19 15.442 0.17899\\
    21 17.073 0.11336\\
    22 13.768 0.1051\\
    22 13.029 0.090562\\
    22 16.58 0.12421\\
    23 12.974 0.17084\\
    24 15.114 0.46498\\
    25 13.199 0.40288\\
    26 22.743 0.17361\\
    27 13.633 0.19187\\
    27 13.818 0.23115\\
    27 14.244 0.21532\\
  }\condckoriginbracket
  \pgfplotstableread[row sep=\\]{
    layer median\\
    0 6.1709\\
    1 6.4217\\
    2 6.0973\\
    3 4.533\\
    4 5.1593\\
    5 4.8653\\
    6 3.4959\\
    7 4.1907\\
    8 4.8313\\
    9 5.4315\\
    10 3.9716\\
    11 5.4736\\
    12 6.4114\\
    13 8.3869\\
    14 6.54\\
    15 9.1569\\
    16 6.9061\\
    17 7.6675\\
    18 6.7585\\
    19 6.6197\\
    20 5.382\\
    21 5.6054\\
    22 5.6235\\
    23 5.9168\\
    24 5.9831\\
    25 5.5747\\
    26 6.1479\\
    27 6.796\\
  }\condckmeanmedian
  \pgfplotstableread[row sep=\\]{
    layer value up\\
    0 5.2246 0\\
    0 5.8545 0\\
    0 6.4873 0\\
    1 10.853 0\\
    1 4.3888 0\\
    1 4.4501 0\\
    1 8.3932 0\\
    2 4.2179 0\\
    2 9.1023 0\\
    2 5.9402 0\\
    2 6.2544 0\\
    3 4.0441 0\\
    3 4.4474 0\\
    3 5.3942 0\\
    3 4.6186 0\\
    4 5.0491 0\\
    4 5.3584 0\\
    4 5.2695 0\\
    4 4.7098 0\\
    5 4.9565 0\\
    5 4.2296 0\\
    5 4.8211 0\\
    5 4.9094 0\\
    6 2.9979 0\\
    6 4.8104 0\\
    6 2.6771 0\\
    6 3.9938 0\\
    7 4.6445 0\\
    7 3.7369 0\\
    7 5.7535 0\\
    7 3.549 0\\
    8 3.8746 0\\
    8 4.6705 0\\
    8 4.9921 0\\
    8 5.9512 0\\
    9 6.0353 0\\
    9 4.8276 0\\
    9 3.9884 0\\
    9 6.5898 0\\
    10 3.0697 0\\
    10 3.9148 0\\
    10 4.0285 0\\
    10 5.0696 0\\
    11 5.5154 0\\
    11 5.4317 0\\
    11 4.289 0\\
    11 8.8666 0\\
    12 7.2837 0\\
    12 6.9847 0\\
    12 5.0774 0\\
    12 5.838 0\\
    13 7.6319 0\\
    13 3.6836 0\\
    13 11.912 0\\
    13 9.142 0\\
    14 6.2399 0\\
    14 6.4668 0\\
    14 6.6131 0\\
    15 12.062 0\\
    15 2.7705 0\\
    15 6.2515 0\\
    16 6.9431 0\\
    16 7.4341 0\\
    16 5.642 0\\
    16 6.8691 0\\
    17 7.9648 0\\
    17 4.2658 0\\
    17 7.3701 0\\
    17 8.0266 0\\
    18 6.0416 0\\
    18 7.4753 0\\
    18 7.5001 0\\
    18 3.1385 0\\
    19 8.1229 0\\
    19 3.3911 0\\
    19 5.2616 0\\
    19 7.9778 0\\
    20 7.3989 0\\
    20 5.1444 0\\
    20 5.202 0\\
    20 5.562 0\\
    21 9.7918 0\\
    21 4.2598 0\\
    21 3.5294 0\\
    21 6.951 0\\
    22 10.882 0\\
    22 5.9195 0\\
    22 4.8176 0\\
    22 5.3275 0\\
    23 6.9559 0\\
    23 5.5719 0\\
    23 6.2617 0\\
    23 4.4427 0\\
    24 5.2005 0\\
    24 6.4073 0\\
    24 5.5588 0\\
    24 6.9174 0\\
    25 8.5647 0\\
    25 5.0498 0\\
    25 5.2136 0\\
    25 5.9357 0\\
    26 5.5015 0\\
    26 5.8474 0\\
    26 6.4484 0\\
    27 6.8588 0\\
    27 8.478 0\\
    27 6.7332 0\\
    27 6.5028 0\\
  }\condckmeanhead
  \pgfplotstableread[row sep=\\]{
    layer value up\\
    0 13.235 0.025744\\
    14 18.203 0.2307\\
    15 15.413 0.10665\\
    26 15.572 0.17361\\
  }\condckmeanbracket
  \begin{groupplot}[
      group style={group size=3 by 1, horizontal sep=8pt, yticklabels at=edge left},
      width=128pt, height=118pt, xmin=-1, xmax=28, ymin=-1, ymax=40, clip mode=individual,
      xtick={0,5,...,25}, ytick={0,10,...,40}, xmajorgrids, ymajorgrids, grid style={gray!25},
      xlabel={layer},
      title style={font=\footnotesize}]
    \nextgroupplot[title={Context keys $E_{KK}$, $N = 4064$}, legend to name=legend-gram-conditioning-per-layer, legend columns=3, ylabel={$\log_{10}\kappa_2$}]
      \addplot[resolution, forget plot] coordinates {(-1,12.045) (28,12.045)};
      \addplot[skgauss, head] table[x=layer, y=value] {\condakgausshead};
      \addplot[skgauss, median] table[x=layer, y=median] {\condakgaussmedian};
      \addplot[skorigin, head] table[x=layer, y=value] {\condakoriginhead};
      \addplot[skorigin, bracket] table[x=layer, y=value, y error=up] {\condakoriginbracket};
      \addplot[skorigin, median] table[x=layer, y=median] {\condakoriginmedian};
      \addplot[skorigin, offscale] coordinates {(0,40)};
      \addplot[skorigin, offscale] coordinates {(13,40)};
      \addplot[skorigin, offscale] coordinates {(19,40)};
      \addplot[skorigin, offscale] coordinates {(27,40)};
      \addplot[skmean, head] table[x=layer, y=value] {\condakmeanhead};
      \addplot[skmean, bracket] table[x=layer, y=value, y error=up] {\condakmeanbracket};
      \addplot[skmean, median] table[x=layer, y=median] {\condakmeanmedian};
      \node[font=\tiny, anchor=north west, inner sep=1pt, yshift=-3pt, text=korigin] at (axis cs:0,40) {$10^{116}$};
      \node[font=\tiny, anchor=north, inner sep=1pt, yshift=-3pt, text=korigin] at (axis cs:13,40) {$10^{51}$};
      \node[font=\tiny, anchor=north, inner sep=1pt, yshift=-3pt, text=korigin] at (axis cs:19,40) {$10^{42}$};
      \node[font=\tiny, anchor=north east, inner sep=1pt, yshift=-3pt, text=korigin] at (axis cs:27,40) {$10^{354}$};
      \addlegendimage{skgauss, thick, mark size=1.1pt}\addlegendentry{Gaussian $\breve E$}
      \addlegendimage{skorigin, thick, mark size=1.1pt}\addlegendentry{exponential $E$, origin}
      \addlegendimage{skmean, thick, mark size=1.1pt}\addlegendentry{exponential $E$, centred on $\mu_g$}
      \addlegendimage{only marks, mark=*, mark size=1.1pt, gray, fill opacity=0.55, draw opacity=0}\addlegendentry{head, fp64 $\kappa_2$}
      \addlegendimage{only marks, mark=*, mark size=1.3pt, gray, fill=white}\addlegendentry{head past fp64: $\kappa(D)$ (bar to $\kappa(D)\,\kappa(\breve E)$)}
      \addlegendimage{resolution}\addlegendentry{fp64 resolution $1/(n\epsilon)$}
    \nextgroupplot[title={GAIM exact nodes $E_{LL}$, $n = 512$}]
      \addplot[resolution, forget plot] coordinates {(-1,12.944) (28,12.944)};
      \addplot[skgauss, head] table[x=layer, y=value] {\condbkgausshead};
      \addplot[skgauss, median] table[x=layer, y=median] {\condbkgaussmedian};
      \addplot[skorigin, head] table[x=layer, y=value] {\condbkoriginhead};
      \addplot[skorigin, bracket] table[x=layer, y=value, y error=up] {\condbkoriginbracket};
      \addplot[skorigin, median] table[x=layer, y=median] {\condbkoriginmedian};
      \addplot[skorigin, offscale] coordinates {(0,40)};
      \addplot[skorigin, offscale] coordinates {(13,40)};
      \addplot[skorigin, offscale] coordinates {(27,40)};
      \addplot[skmean, head] table[x=layer, y=value] {\condbkmeanhead};
      \addplot[skmean, bracket] table[x=layer, y=value, y error=up] {\condbkmeanbracket};
      \addplot[skmean, median] table[x=layer, y=median] {\condbkmeanmedian};
      \node[font=\tiny, anchor=north west, inner sep=1pt, yshift=-3pt, text=korigin] at (axis cs:0,40) {$10^{86}$};
      \node[font=\tiny, anchor=north, inner sep=1pt, yshift=-3pt, text=korigin] at (axis cs:13,40) {$10^{50}$};
      \node[font=\tiny, anchor=north east, inner sep=1pt, yshift=-3pt, text=korigin] at (axis cs:27,40) {$10^{340}$};
    \nextgroupplot[title={Q-DAIM nodes $E_{LL}$, $n = 512$}]
      \addplot[resolution, forget plot] coordinates {(-1,12.944) (28,12.944)};
      \addplot[skgauss, head] table[x=layer, y=value] {\condckgausshead};
      \addplot[skgauss, median] table[x=layer, y=median] {\condckgaussmedian};
      \addplot[skorigin, head] table[x=layer, y=value] {\condckoriginhead};
      \addplot[skorigin, bracket] table[x=layer, y=value, y error=up] {\condckoriginbracket};
      \addplot[skorigin, median] table[x=layer, y=median] {\condckoriginmedian};
      \addplot[skorigin, offscale] coordinates {(0,40)};
      \addplot[skorigin, offscale] coordinates {(19,40)};
      \addplot[skorigin, offscale] coordinates {(27,40)};
      \addplot[skmean, head] table[x=layer, y=value] {\condckmeanhead};
      \addplot[skmean, bracket] table[x=layer, y=value, y error=up] {\condckmeanbracket};
      \addplot[skmean, median] table[x=layer, y=median] {\condckmeanmedian};
      \node[font=\tiny, anchor=north west, inner sep=1pt, yshift=-3pt, text=korigin] at (axis cs:0,40) {$10^{90}$};
      \node[font=\tiny, anchor=north, inner sep=1pt, yshift=-3pt, text=korigin] at (axis cs:19,40) {$10^{40}$};
      \node[font=\tiny, anchor=north east, inner sep=1pt, yshift=-3pt, text=korigin] at (axis cs:27,40) {$10^{326}$};
  \end{groupplot}
  \node[below=4pt] at (current bounding box.south) {\pgfplotslegendfromname{legend-gram-conditioning-per-layer}};
\end{tikzpicture}

%% file: figures/tikz/qdaim_restart_comparison.tex
\begin{tikzpicture}
  \pgfplotsset{every axis/.append style={
      scale only axis, /tikz/font=\footnotesize, label style={align=center}, title style={align=center},
      legend cell align=left,
      legend style={font=\footnotesize, /tikz/every even column/.append style={column sep=7pt}}}}
  \pgfplotsset{
    v1/.style={blue, mark=*},
    v2/.style={red, mark=square*},
    v3/.style={cyan!60!black, mark=triangle*},
    v4/.style={orange!50!black, mark=diamond*},
    v5/.style={magenta!70!black, mark=pentagon*},
  }
  \pgfplotsset{
    single pass/.style={thin, densely dashed, mark=none},
    restart mark/.style={gray!80, thin, densely dotted, mark=none},
    mean value/.style={black, thin, dashdotted, mark=none},
    every axis plot/.append style={very thick, mark size=1.8pt, mark options={solid}, mark repeat=8, forget plot},
    legend image code/.code={\draw[very thick, mark size=1.8pt, mark options={solid}, mark repeat=2, mark phase=2, #1]
      plot coordinates {(0pt,0pt) (10pt,0pt) (20pt,0pt)};},
  }
  \pgfplotstableread[row sep=\\]{
    n qdaim_exact_single_v1 qdaim_exact_single_v5 qdaim_rff_single_v1 qdaim_rff_single_v5 qdaim_exact_restart_v1 qdaim_exact_restart_v5 qdaim_rff_restart_v1 qdaim_rff_restart_v5\\
    1 0.61627 0.61627 0.97563 0.97563 0.61627 0.61627 0.97563 0.97563\\
    17 0.083694 0.083694 0.084589 0.084589 0.083694 0.083694 0.084589 0.084589\\
    33 0.041902 0.041902 0.041124 0.041124 0.041902 0.041902 0.041124 0.041124\\
    49 0.03561 0.030647 0.035452 0.035566 0.03561 0.030647 0.035452 0.035566\\
    65 0.025152 0.025289 0.033553 0.033591 0.025152 0.025289 0.033553 0.033591\\
    81 0.02112 0.02112 0.0211 0.03118 0.020966 0.020893 0.03118 0.03118\\
    97 0.01999 0.01999 0.02003 0.025492 0.018571 0.018565 0.030109 0.030105\\
    113 0.018887 0.018934 0.019021 0.024489 0.016478 0.016532 0.0291 0.029102\\
    129 0.018019 0.017734 0.01807 0.023674 0.014981 0.014838 0.026974 0.027117\\
    145 0.017226 0.015881 0.017197 0.023175 0.01235 0.012075 0.025659 0.015655\\
    161 0.016426 0.015364 0.016519 0.021149 0.011372 0.011126 0.014982 0.015049\\
    177 0.014725 0.014758 0.01593 0.019889 0.010544 0.010226 0.014505 0.014557\\
    193 0.014228 0.013889 0.015471 0.019214 0.0096198 0.0096113 0.014055 0.014118\\
    209 0.013912 0.013466 0.013964 0.018687 0.0090249 0.0090506 0.013632 0.013743\\
    225 0.013527 0.013068 0.013621 0.018353 0.0085347 0.008517 0.00869 0.012988\\
    241 0.013218 0.012745 0.013388 0.018036 0.0081344 0.0080767 0.0083795 0.012581\\
    257 0.012871 0.012476 0.013121 0.0178 0.0076894 0.0077386 0.0081387 0.007719\\
    273 0.012606 0.012196 0.012857 0.017639 0.0072457 0.0072779 0.0077655 0.007333\\
    289 0.012349 0.011993 0.012569 0.017408 0.0068613 0.0068922 0.0074685 0.0070235\\
    305 0.012159 0.0070398 0.012387 0.017187 0.0065258 0.0065525 0.0072185 0.0066408\\
    321 0.011898 0.0068253 0.012186 0.017075 0.0062417 0.0062573 0.0063721 0.0064161\\
    337 0.011689 0.0066441 0.011998 0.016911 0.0059617 0.0059994 0.0061581 0.0061976\\
    353 0.011535 0.0064708 0.011826 0.016745 0.0057288 0.0057346 0.0059306 0.0060088\\
    369 0.011348 0.0062755 0.011678 0.016587 0.0054849 0.0054948 0.0054099 0.0058351\\
    385 0.010513 0.0060779 0.011524 0.01643 0.0052655 0.0052857 0.005234 0.005687\\
    401 0.010295 0.0057673 0.011354 0.016255 0.0050095 0.0050447 0.0049779 0.0049749\\
    417 0.010153 0.0055893 0.011185 0.016153 0.0047715 0.0047818 0.0047428 0.0047569\\
    433 0.0096187 0.0054392 0.011024 0.016015 0.0046026 0.0046054 0.0045524 0.0045592\\
    449 0.009445 0.0052633 0.010858 0.015709 0.0044078 0.0044324 0.0043853 0.0043857\\
    465 0.0093117 0.0047193 0.010774 0.015568 0.0042433 0.0042556 0.0042147 0.0042558\\
    481 0.0091814 0.0045852 0.01068 0.015417 0.0040778 0.0040998 0.0040817 0.0040651\\
    497 0.0090361 0.0044365 0.010581 0.015036 0.0039579 0.0039633 0.003959 0.003947\\
    512 0.00891 0.0043187 0.010518 0.014733 0.0038476 0.0038362 0.0038608 0.0038478\\
  }\curvesa
  \pgfplotstableread[row sep=\\]{
    n qdaim_exact_single_v1 qdaim_exact_single_v5 qdaim_rff_single_v1 qdaim_rff_single_v5 qdaim_exact_restart_v1 qdaim_exact_restart_v5 qdaim_rff_restart_v1 qdaim_rff_restart_v5\\
    1 0.20723 0.20723 0.20722 0.20722 0.20723 0.20723 0.20722 0.20722\\
    17 0.21325 0.21325 0.21314 0.21314 0.21325 0.21325 0.21314 0.21314\\
    33 0.21244 0.21244 0.21244 0.21244 0.21244 0.21244 0.21244 0.21244\\
    49 0.21167 0.21167 0.21158 0.21158 0.21167 0.21167 0.21158 0.21158\\
    65 0.21123 0.21122 0.21127 0.21127 0.21123 0.21122 0.21127 0.21127\\
    81 0.21048 0.21048 0.21051 0.21048 0.21048 0.21048 0.21051 0.21048\\
    97 0.21456 0.21457 0.20998 0.21002 0.21456 0.21457 0.20998 0.21002\\
    113 0.21304 0.21314 0.20904 0.20909 0.21304 0.21314 0.20904 0.20909\\
    129 0.2081 0.20626 0.206 0.20632 0.20804 0.20626 0.20599 0.20628\\
    145 0.20508 0.20538 0.20539 0.20545 0.20538 0.20519 0.20553 0.20581\\
    161 0.20438 0.20449 0.20478 0.20787 0.20464 0.20465 0.20457 0.20458\\
    177 0.20333 0.20689 0.20411 0.20724 0.20535 0.20716 0.20693 0.20702\\
    193 0.20628 0.20622 0.20667 0.20647 0.20636 0.20639 0.20627 0.2062\\
    209 0.20221 0.20545 0.2058 0.20596 0.20515 0.20573 0.20564 0.20562\\
    225 0.20115 0.20087 0.20511 0.20131 0.20136 0.20498 0.20471 0.20487\\
    241 0.20065 0.20019 0.20045 0.20053 0.19999 0.19994 0.20431 0.20415\\
    257 0.19955 0.19913 0.20081 0.19967 0.19955 0.19907 0.19964 0.20023\\
    273 0.19857 0.19894 0.20144 0.19882 0.19773 0.19737 0.19758 0.1995\\
    289 0.19806 0.19842 0.19982 0.19824 0.1968 0.19627 0.20129 0.19708\\
    305 0.19744 0.19778 0.19562 0.19768 0.19559 0.1956 0.2009 0.19598\\
    321 0.19706 0.19562 0.19483 0.19623 0.19526 0.19531 0.19989 0.19528\\
    337 0.19685 0.19141 0.18809 0.18948 0.19501 0.19446 0.19918 0.19425\\
    353 0.19507 0.19151 0.18728 0.18563 0.19419 0.19318 0.19843 0.19355\\
    369 0.19332 0.19104 0.18648 0.18371 0.19304 0.1921 0.19227 0.19276\\
    385 0.19144 0.18984 0.18468 0.18249 0.19027 0.19182 0.19187 0.19173\\
    401 0.1909 0.18934 0.18429 0.18127 0.19097 0.19138 0.19134 0.19118\\
    417 0.19029 0.18846 0.18169 0.18092 0.19121 0.19098 0.19095 0.19073\\
    433 0.18981 0.18965 0.18077 0.17977 0.19072 0.19065 0.19003 0.19024\\
    449 0.18964 0.18654 0.17985 0.17913 0.18959 0.1898 0.18917 0.18952\\
    465 0.18942 0.18467 0.17926 0.17818 0.18948 0.18941 0.18901 0.1889\\
    481 0.18881 0.1838 0.1783 0.17748 0.18762 0.1868 0.18856 0.18851\\
    497 0.18832 0.18009 0.17778 0.1767 0.1861 0.18699 0.18791 0.18774\\
    512 0.18776 0.17803 0.17683 0.17616 0.18507 0.18437 0.18731 0.18667\\
  }\curvesb
  \pgfplotstableread[row sep=\\]{
    n qdaim_exact_single_v1 qdaim_exact_single_v5 qdaim_rff_single_v1 qdaim_rff_single_v5 qdaim_exact_restart_v1 qdaim_exact_restart_v5 qdaim_rff_restart_v1 qdaim_rff_restart_v5\\
    1 0.16892 0.16892 0.1689 0.1689 0.16892 0.16892 0.1689 0.1689\\
    17 0.16811 0.16811 0.16812 0.16812 0.16811 0.16811 0.16812 0.16812\\
    33 0.1674 0.1674 0.1674 0.1674 0.1674 0.1674 0.1674 0.1674\\
    49 0.16701 0.16701 0.16701 0.16701 0.16701 0.16701 0.16701 0.16701\\
    65 0.16632 0.16632 0.16627 0.16627 0.16632 0.16632 0.16627 0.16627\\
    81 0.16816 0.16816 0.16564 0.16561 0.16816 0.16816 0.16564 0.16561\\
    97 0.1674 0.1675 0.16738 0.1674 0.1674 0.1675 0.16738 0.1674\\
    113 0.1668 0.1668 0.16451 0.16452 0.1668 0.1668 0.16451 0.16452\\
    129 0.16401 0.16408 0.16395 0.16374 0.16402 0.16408 0.16393 0.16374\\
    145 0.16336 0.16538 0.16338 0.16562 0.16569 0.16513 0.16348 0.16585\\
    161 0.16458 0.16478 0.1653 0.1651 0.16484 0.16492 0.1652 0.16506\\
    177 0.16418 0.16414 0.16446 0.16454 0.16419 0.16426 0.16473 0.1645\\
    193 0.16352 0.16359 0.16376 0.1636 0.16366 0.16354 0.16384 0.1638\\
    209 0.16073 0.16318 0.16324 0.16298 0.16309 0.16306 0.16341 0.16297\\
    225 0.16025 0.16061 0.16261 0.16068 0.1606 0.16244 0.16241 0.16249\\
    241 0.15988 0.16011 0.1599 0.16005 0.16023 0.16022 0.16189 0.16196\\
    257 0.15946 0.15921 0.16136 0.15956 0.1597 0.15944 0.16125 0.15992\\
    273 0.15893 0.15878 0.16073 0.15894 0.15934 0.15852 0.16067 0.15891\\
    289 0.15823 0.15802 0.16029 0.15822 0.15845 0.15815 0.16025 0.1584\\
    305 0.15777 0.15749 0.15956 0.15719 0.15791 0.15737 0.1597 0.15796\\
    321 0.15727 0.15694 0.1591 0.15675 0.15735 0.15664 0.15923 0.1594\\
    337 0.15687 0.15631 0.15612 0.15569 0.15653 0.15626 0.15876 0.15836\\
    353 0.15633 0.15561 0.15548 0.15529 0.15566 0.1555 0.15767 0.15764\\
    369 0.15578 0.15484 0.15473 0.15466 0.15507 0.15481 0.15474 0.15706\\
    385 0.15513 0.15422 0.15407 0.15396 0.15408 0.15445 0.15367 0.15475\\
    401 0.15436 0.1536 0.15362 0.15346 0.15365 0.15379 0.1531 0.154\\
    417 0.15377 0.15223 0.15283 0.15295 0.15321 0.15322 0.15265 0.15328\\
    433 0.15049 0.15133 0.15241 0.15223 0.15259 0.15251 0.15006 0.15265\\
    449 0.14987 0.15084 0.15119 0.15026 0.15179 0.15203 0.1495 0.15205\\
    465 0.14954 0.14935 0.15014 0.14828 0.1506 0.14951 0.14871 0.14941\\
    481 0.149 0.14859 0.14827 0.14764 0.14979 0.14889 0.14826 0.14876\\
    497 0.14828 0.14806 0.14582 0.14496 0.14825 0.14846 0.14791 0.14818\\
    512 0.14697 0.14652 0.14359 0.14388 0.1474 0.14754 0.14719 0.14753\\
  }\curvesc
  \begin{groupplot}[
      group style={group size=2 by 3, horizontal sep=8pt, vertical sep=9pt,
        xticklabels at=edge bottom, yticklabels at=edge left},
      width=170pt, height=92pt, grid=major, tick label style={font=\scriptsize},
      xmin=-15, xmax=527, xtick={0,100,...,500}]
    \nextgroupplot[ymode=log, ymin=0.001, ymax=1.4, title={Q-DAIM exact}, ylabel={$\|M_{\Delta,\eta}\|_2^2\,/\,\|M_{\eta}\|_2^2$}, legend to name=legend-qdaim-restart-comparison, legend columns=-1]
      \addplot[restart mark] coordinates {(128,0.001) (128,1.4)};
      \addplot[restart mark] coordinates {(256,0.001) (256,1.4)};
      \addplot[restart mark] coordinates {(384,0.001) (384,1.4)};
      \addplot[v1, single pass] table[x=n, y=qdaim_exact_single_v1] {\curvesa};
      \addplot[v1, mark phase=2] table[x=n, y=qdaim_exact_restart_v1] {\curvesa};
      \addplot[v5, single pass] table[x=n, y=qdaim_exact_single_v5] {\curvesa};
      \addplot[v5, mark phase=3] table[x=n, y=qdaim_exact_restart_v5] {\curvesa};
      \addlegendimage{v1}\addlegendentry{v1}
      \addlegendimage{v5}\addlegendentry{v5}
      \addlegendimage{black, very thick, mark=none}\addlegendentry{restarted}
      \addlegendimage{black, thin, densely dashed, mark=none}\addlegendentry{single pass}
      \addlegendimage{mean value}\addlegendentry{mean vector approximation}
      \addlegendimage{restart mark}\addlegendentry{restart}
    \nextgroupplot[ymode=log, ymin=0.001, ymax=1.4, title={Q-DAIM RFF}]
      \addplot[restart mark] coordinates {(128,0.001) (128,1.4)};
      \addplot[restart mark] coordinates {(256,0.001) (256,1.4)};
      \addplot[restart mark] coordinates {(384,0.001) (384,1.4)};
      \addplot[v1, single pass] table[x=n, y=qdaim_rff_single_v1] {\curvesa};
      \addplot[v1, mark phase=2] table[x=n, y=qdaim_rff_restart_v1] {\curvesa};
      \addplot[v5, single pass] table[x=n, y=qdaim_rff_single_v5] {\curvesa};
      \addplot[v5, mark phase=3] table[x=n, y=qdaim_rff_restart_v5] {\curvesa};
    \nextgroupplot[ymin=0.1671, ymax=0.2164, scaled y ticks=false, yticklabel style={/pgf/number format/fixed, /pgf/number format/fixed zerofill, /pgf/number format/precision=3}, ylabel={RC off\\\scriptsize relative attention error $\varepsilon_{\Att}^{10\%}$}]
      \addplot[restart mark] coordinates {(128,0.1671) (128,0.2164)};
      \addplot[restart mark] coordinates {(256,0.1671) (256,0.2164)};
      \addplot[restart mark] coordinates {(384,0.1671) (384,0.2164)};
      \addplot[mean value] coordinates {(-15,0.16892) (527,0.16892)};
      \addplot[v1, single pass] table[x=n, y=qdaim_exact_single_v1] {\curvesb};
      \addplot[v1, mark phase=2] table[x=n, y=qdaim_exact_restart_v1] {\curvesb};
      \addplot[v5, single pass] table[x=n, y=qdaim_exact_single_v5] {\curvesb};
      \addplot[v5, mark phase=3] table[x=n, y=qdaim_exact_restart_v5] {\curvesb};
    \nextgroupplot[ymin=0.1671, ymax=0.2164, scaled y ticks=false, yticklabel style={/pgf/number format/fixed, /pgf/number format/fixed zerofill, /pgf/number format/precision=3}]
      \addplot[restart mark] coordinates {(128,0.1671) (128,0.2164)};
      \addplot[restart mark] coordinates {(256,0.1671) (256,0.2164)};
      \addplot[restart mark] coordinates {(384,0.1671) (384,0.2164)};
      \addplot[mean value] coordinates {(-15,0.16892) (527,0.16892)};
      \addplot[v1, single pass] table[x=n, y=qdaim_rff_single_v1] {\curvesb};
      \addplot[v1, mark phase=2] table[x=n, y=qdaim_rff_restart_v1] {\curvesb};
      \addplot[v5, single pass] table[x=n, y=qdaim_rff_single_v5] {\curvesb};
      \addplot[v5, mark phase=3] table[x=n, y=qdaim_rff_restart_v5] {\curvesb};
    \nextgroupplot[ymin=0.1426, ymax=0.1699, scaled y ticks=false, yticklabel style={/pgf/number format/fixed, /pgf/number format/fixed zerofill, /pgf/number format/precision=3}, ylabel={RC on\\\scriptsize relative attention error $\varepsilon_{\Att}^{10\%}$}, xlabel={number of nodes $n$}]
      \addplot[restart mark] coordinates {(128,0.1426) (128,0.1699)};
      \addplot[restart mark] coordinates {(256,0.1426) (256,0.1699)};
      \addplot[restart mark] coordinates {(384,0.1426) (384,0.1699)};
      \addplot[mean value] coordinates {(-15,0.16892) (527,0.16892)};
      \addplot[v1, single pass] table[x=n, y=qdaim_exact_single_v1] {\curvesc};
      \addplot[v1, mark phase=2] table[x=n, y=qdaim_exact_restart_v1] {\curvesc};
      \addplot[v5, single pass] table[x=n, y=qdaim_exact_single_v5] {\curvesc};
      \addplot[v5, mark phase=3] table[x=n, y=qdaim_exact_restart_v5] {\curvesc};
    \nextgroupplot[ymin=0.1426, ymax=0.1699, scaled y ticks=false, yticklabel style={/pgf/number format/fixed, /pgf/number format/fixed zerofill, /pgf/number format/precision=3}, xlabel={number of nodes $n$}]
      \addplot[restart mark] coordinates {(128,0.1426) (128,0.1699)};
      \addplot[restart mark] coordinates {(256,0.1426) (256,0.1699)};
      \addplot[restart mark] coordinates {(384,0.1426) (384,0.1699)};
      \addplot[mean value] coordinates {(-15,0.16892) (527,0.16892)};
      \addplot[v1, single pass] table[x=n, y=qdaim_rff_single_v1] {\curvesc};
      \addplot[v1, mark phase=2] table[x=n, y=qdaim_rff_restart_v1] {\curvesc};
      \addplot[v5, single pass] table[x=n, y=qdaim_rff_single_v5] {\curvesc};
      \addplot[v5, mark phase=3] table[x=n, y=qdaim_rff_restart_v5] {\curvesc};
  \end{groupplot}
  \node[below=4pt] at (current bounding box.south) {\pgfplotslegendfromname{legend-qdaim-restart-comparison}};
\end{tikzpicture}

%% file: references.bib
@misc{ainslieGQATrainingGeneralized2023,
  title = {{{GQA}}: {{Training Generalized Multi-Query Transformer Models}} from {{Multi-Head Checkpoints}}},
  shorttitle = {{{GQA}}},
  author = {Ainslie, Joshua and {Lee-Thorp}, James and de Jong, Michiel and Zemlyanskiy, Yury and Lebr{\'o}n, Federico and Sanghai, Sumit},
  year = 2023,
  month = dec,
  number = {arXiv:2305.13245},
  eprint = {2305.13245},
  primaryclass = {cs},
  publisher = {arXiv},
  doi = {10.48550/arXiv.2305.13245},
  archiveprefix = {arXiv}
}

@misc{chenSkyformerRemodelSelfAttention2021,
  title = {Skyformer: {{Remodel Self-Attention}} with {{Gaussian Kernel}} and {{Nystr\"om Method}}},
  shorttitle = {Skyformer},
  author = {Chen, Yifan and Zeng, Qi and Ji, Heng and Yang, Yun},
  year = 2021,
  month = oct,
  number = {arXiv:2111.00035},
  eprint = {2111.00035},
  primaryclass = {cs},
  publisher = {arXiv},
  doi = {10.48550/arXiv.2111.00035},
  archiveprefix = {arXiv}
}

@misc{choromanskiRethinkingAttentionPerformers2022,
  title = {Rethinking {{Attention}} with {{Performers}}},
  author = {Choromanski, Krzysztof and Likhosherstov, Valerii and Dohan, David and Song, Xingyou and Gane, Andreea and Sarlos, Tamas and Hawkins, Peter and Davis, Jared and Mohiuddin, Afroz and Kaiser, Lukasz and Belanger, David and Colwell, Lucy and Weller, Adrian},
  year = 2022,
  month = nov,
  number = {arXiv:2009.14794},
  eprint = {2009.14794},
  primaryclass = {cs},
  publisher = {arXiv},
  doi = {10.48550/arXiv.2009.14794},
  archiveprefix = {arXiv}
}

@misc{comonSymmetricTensorsSymmetric2008,
  title = {Symmetric Tensors and Symmetric Tensor Rank},
  author = {Comon, Pierre and Golub, Gene and Lim, Lek-Heng and Mourrain, Bernard},
  year = 2008,
  month = sep,
  number = {arXiv:0802.1681},
  eprint = {0802.1681},
  primaryclass = {math},
  publisher = {arXiv},
  doi = {10.48550/arXiv.0802.1681},
  archiveprefix = {arXiv},
  langid = {english}
}

@misc{daoFlashAttention2FasterAttention2023,
  title = {{{FlashAttention-2}}: {{Faster Attention}} with {{Better Parallelism}} and {{Work Partitioning}}},
  shorttitle = {{{FlashAttention-2}}},
  author = {Dao, Tri},
  year = 2023,
  month = jul,
  number = {arXiv:2307.08691},
  eprint = {2307.08691},
  primaryclass = {cs},
  publisher = {arXiv},
  doi = {10.48550/arXiv.2307.08691},
  archiveprefix = {arXiv}
}

@misc{daoFlashAttentionFastMemoryEfficient2022,
  title = {{{FlashAttention}}: {{Fast}} and {{Memory-Efficient Exact Attention}} with {{IO-Awareness}}},
  shorttitle = {{{FlashAttention}}},
  author = {Dao, Tri and Fu, Daniel Y. and Ermon, Stefano and Rudra, Atri and R{\'e}, Christopher},
  year = 2022,
  month = jun,
  number = {arXiv:2205.14135},
  eprint = {2205.14135},
  primaryclass = {cs},
  publisher = {arXiv},
  doi = {10.48550/arXiv.2205.14135},
  archiveprefix = {arXiv}
}

@misc{deepseek-aiDeepSeekV2StrongEconomical2024,
  title = {{{DeepSeek-V2}}: {{A Strong}}, {{Economical}}, and {{Efficient Mixture-of-Experts Language Model}}},
  shorttitle = {{{DeepSeek-V2}}},
  author = {{DeepSeek-AI} and Liu, Aixin and Feng, Bei and Wang, Bin and Wang, Bingxuan and Liu, Bo and Zhao, Chenggang and Dengr, Chengqi and Ruan, Chong and Dai, Damai and Guo, Daya and Yang, Dejian and Chen, Deli and Ji, Dongjie and Li, Erhang and Lin, Fangyun and Luo, Fuli and Hao, Guangbo and Chen, Guanting and Li, Guowei and Zhang, H. and Xu, Hanwei and Yang, Hao and Zhang, Haowei and Ding, Honghui and Xin, Huajian and Gao, Huazuo and Li, Hui and Qu, Hui and Cai, J. L. and Liang, Jian and Guo, Jianzhong and Ni, Jiaqi and Li, Jiashi and Chen, Jin and Yuan, Jingyang and Qiu, Junjie and Song, Junxiao and Dong, Kai and Gao, Kaige and Guan, Kang and Wang, Lean and Zhang, Lecong and Xu, Lei and Xia, Leyi and Zhao, Liang and Zhang, Liyue and Li, Meng and Wang, Miaojun and Zhang, Mingchuan and Zhang, Minghua and Tang, Minghui and Li, Mingming and Tian, Ning and Huang, Panpan and Wang, Peiyi and Zhang, Peng and Zhu, Qihao and Chen, Qinyu and Du, Qiushi and Chen, R. J. and Jin, R. L. and Ge, Ruiqi and Pan, Ruizhe and Xu, Runxin and Chen, Ruyi and Li, S. S. and Lu, Shanghao and Zhou, Shangyan and Chen, Shanhuang and Wu, Shaoqing and Ye, Shengfeng and Ma, Shirong and Wang, Shiyu and Zhou, Shuang and Yu, Shuiping and Zhou, Shunfeng and Zheng, Size and Wang, T. and Pei, Tian and Yuan, Tian and Sun, Tianyu and Xiao, W. L. and Zeng, Wangding and An, Wei and Liu, Wen and Liang, Wenfeng and Gao, Wenjun and Zhang, Wentao and Li, X. Q. and Jin, Xiangyue and Wang, Xianzu and Bi, Xiao and Liu, Xiaodong and Wang, Xiaohan and Shen, Xiaojin and Chen, Xiaokang and Chen, Xiaosha and Nie, Xiaotao and Sun, Xiaowen and Wang, Xiaoxiang and Liu, Xin and Xie, Xin and Yu, Xingkai and Song, Xinnan and Zhou, Xinyi and Yang, Xinyu and Lu, Xuan and Su, Xuecheng and Wu, Y. and Li, Y. K. and Wei, Y. X. and Zhu, Y. X. and Xu, Yanhong and Huang, Yanping and Li, Yao and Zhao, Yao and Sun, Yaofeng and Li, Yaohui and Wang, Yaohui and Zheng, Yi and Zhang, Yichao and Xiong, Yiliang and Zhao, Yilong and He, Ying and Tang, Ying and Piao, Yishi and Dong, Yixin and Tan, Yixuan and Liu, Yiyuan and Wang, Yongji and Guo, Yongqiang and Zhu, Yuchen and Wang, Yuduan and Zou, Yuheng and Zha, Yukun and Ma, Yunxian and Yan, Yuting and You, Yuxiang and Liu, Yuxuan and Ren, Z. Z. and Ren, Zehui and Sha, Zhangli and Fu, Zhe and Huang, Zhen and Zhang, Zhen and Xie, Zhenda and Hao, Zhewen and Shao, Zhihong and Wen, Zhiniu and Xu, Zhipeng and Zhang, Zhongyu and Li, Zhuoshu and Wang, Zihan and Gu, Zihui and Li, Zilin and Xie, Ziwei},
  year = 2024,
  month = jun,
  number = {arXiv:2405.04434},
  eprint = {2405.04434},
  primaryclass = {cs},
  publisher = {arXiv},
  doi = {10.48550/arXiv.2405.04434},
  archiveprefix = {arXiv}
}

@misc{drmacDiscreteEmpiricalInterpolation2018,
  title = {The {{Discrete Empirical Interpolation Method}}: {{Canonical Structure}} and {{Formulation}} in {{Weighted Inner Product Spaces}}},
  shorttitle = {The {{Discrete Empirical Interpolation Method}}},
  author = {Drma{\v c}, Zlatko and Saibaba, Arvind K.},
  year = 2018,
  month = feb,
  number = {arXiv:1704.06606},
  eprint = {1704.06606},
  primaryclass = {math},
  publisher = {arXiv},
  doi = {10.48550/arXiv.1704.06606},
  archiveprefix = {arXiv}
}

@misc{epperlyEmbraceRejectionKernel2025,
  title = {Embrace Rejection: {{Kernel}} Matrix Approximation by Accelerated Randomly Pivoted {{Cholesky}}},
  shorttitle = {Embrace Rejection},
  author = {Epperly, Ethan N. and Tropp, Joel A. and Webber, Robert J.},
  year = 2025,
  month = apr,
  number = {arXiv:2410.03969},
  eprint = {2410.03969},
  primaryclass = {math},
  publisher = {arXiv},
  doi = {10.48550/arXiv.2410.03969},
  archiveprefix = {arXiv}
}

@misc{hanHyperAttentionLongcontextAttention2023,
  title = {{{HyperAttention}}: {{Long-context Attention}} in {{Near-Linear Time}}},
  shorttitle = {{{HyperAttention}}},
  author = {Han, Insu and Jayaram, Rajesh and Karbasi, Amin and Mirrokni, Vahab and Woodruff, David P. and Zandieh, Amir},
  year = 2023,
  month = dec,
  number = {arXiv:2310.05869},
  eprint = {2310.05869},
  primaryclass = {cs},
  publisher = {arXiv},
  doi = {10.48550/arXiv.2310.05869},
  archiveprefix = {arXiv}
}

@misc{hanStreamingAttentionApproximation2025,
  title = {Streaming {{Attention Approximation}} via {{Discrepancy Theory}}},
  author = {Han, Insu and Kapralov, Michael and Kochetkova, Ekaterina and Sheth, Kshiteej and Zandieh, Amir},
  year = 2025,
  month = may,
  number = {arXiv:2502.07861},
  eprint = {2502.07861},
  primaryclass = {cs},
  publisher = {arXiv},
  doi = {10.48550/arXiv.2502.07861},
  archiveprefix = {arXiv},
  langid = {english}
}

@misc{hooperMultipoleAttentionEfficient2025,
  title = {Multipole {{Attention}} for {{Efficient Long Context Reasoning}}},
  author = {Hooper, Coleman and Zhao, Sebastian and Manolache, Luca and Kim, Sehoon and Mahoney, Michael W. and Shao, Yakun Sophia and Keutzer, Kurt and Gholami, Amir},
  year = 2025,
  month = jun,
  number = {arXiv:2506.13059},
  eprint = {2506.13059},
  primaryclass = {cs},
  publisher = {arXiv},
  doi = {10.48550/arXiv.2506.13059},
  archiveprefix = {arXiv}
}

@inproceedings{kachamPolySketchFormerFastTransformers2024,  
  title = 	 {{P}oly{S}ketch{F}ormer: Fast Transformers via Sketching Polynomial Kernels},
  author =       {Kacham, Praneeth and Mirrokni, Vahab and Zhong, Peilin},
  booktitle = 	 {Proceedings of the 41st International Conference on Machine Learning},
  pages = 	 {22748--22770},
  year = 	 {2024},
  editor = 	 {Salakhutdinov, Ruslan and Kolter, Zico and Heller, Katherine and Weller, Adrian and Oliver, Nuria and Scarlett, Jonathan and Berkenkamp, Felix},
  volume = 	 {235},
  series = 	 {Proceedings of Machine Learning Research},
  month = 	 jul,
  publisher =    {PMLR},
  url = 	 {https://proceedings.mlr.press/v235/kacham24a.html},
}

@misc{kitaevReformerEfficientTransformer2020,
  title = {Reformer: {{The Efficient Transformer}}},
  shorttitle = {Reformer},
  author = {Kitaev, Nikita and Kaiser, {\L}ukasz and Levskaya, Anselm},
  year = 2020,
  month = feb,
  number = {arXiv:2001.04451},
  eprint = {2001.04451},
  primaryclass = {cs},
  publisher = {arXiv},
  doi = {10.48550/arXiv.2001.04451},
  archiveprefix = {arXiv}
}

@misc{kressnerRandomizedSVDInfinite2025,
  title = {On the Randomized {{SVD}} in Infinite Dimensions},
  author = {Kressner, Daniel and Persson, David and Uschmajew, Andr{\'e}},
  year = 2025,
  month = jun,
  number = {arXiv:2506.06882},
  eprint = {2506.06882},
  primaryclass = {math},
  publisher = {arXiv},
  doi = {10.48550/arXiv.2506.06882},
  archiveprefix = {arXiv}
}

@article{leeProxyformerNystromBasedLinear2024,
  title = {Proxyformer: {{Nystr\"om-Based Linear Transformer}} with {{Trainable Proxy Tokens}}},
  shorttitle = {Proxyformer},
  author = {Lee, Sangho and Lee, Hayun and Shin, Dongkun},
  year = 2024,
  month = mar,
  journal = {AAAI},
  volume = {38},
  number = {12},
  pages = {13418--13426},
  issn = {2374-3468, 2159-5399},
  doi = {10.1609/aaai.v38i12.29244},
  langid = {english}
}

@incollection{madayGeneralizedEmpiricalInterpolation2013,
  title = {A {{Generalized Empirical Interpolation Method}}: Application of Reduced Basis Techniques to Data Assimilation},
  shorttitle = {A {{Generalized Empirical Interpolation Method}}},
  author = {Maday, Y. and Mula, O.},
  year = 2013,
  volume = {4},
  eprint = {1512.00683},
  primaryclass = {math},
  pages = {221--235},
  doi = {10.1007/978-88-470-2592-9_13},
  archiveprefix = {arXiv},
  langid = {english}
}

@unpublished{madayGeneralMultipurposeInterpolation2007,
  TITLE = {{A general, multipurpose interpolation procedure: the magic points}},
  AUTHOR = {Maday, Yvon and Nguyen, Ngoc Cuong and Patera, Anthony T. and Pau, George S.H.},
  URL = {https://hal.science/hal-00174797},
  NOTE = {working paper or preprint},
  YEAR = {2007},
  MONTH = Sep,
  HAL_ID = {hal-00174797},
  HAL_VERSION = {v1},
}

@misc{mengTransMLAMultiHeadLatent2025,
  title = {{{TransMLA}}: {{Multi-Head Latent Attention Is All You Need}}},
  shorttitle = {{{TransMLA}}},
  author = {Meng, Fanxu and Yao, Zengwei and Zhang, Muhan},
  year = 2025,
  month = feb,
  number = {arXiv:2502.07864},
  eprint = {2502.07864},
  primaryclass = {cs},
  publisher = {arXiv},
  doi = {10.48550/arXiv.2502.07864},
  archiveprefix = {arXiv}
}

@misc{phamTensorSketchFast2025,
  title = {Tensor {{Sketch}}: {{Fast}} and {{Scalable Polynomial Kernel Approximation}}},
  shorttitle = {Tensor {{Sketch}}},
  author = {Pham, Ninh and Pagh, Rasmus},
  year = 2025,
  month = may,
  number = {arXiv:2505.08146},
  eprint = {2505.08146},
  primaryclass = {cs},
  publisher = {arXiv},
  doi = {10.48550/arXiv.2505.08146},
  archiveprefix = {arXiv}
}

@misc{qwenQwen25TechnicalReport2025,
  title = {Qwen2.5 {{Technical Report}}},
  author = {Qwen and Yang, An and Yang, Baosong and Zhang, Beichen and Hui, Binyuan and Zheng, Bo and Yu, Bowen and Li, Chengyuan and Liu, Dayiheng and Huang, Fei and Wei, Haoran and Lin, Huan and Yang, Jian and Tu, Jianhong and Zhang, Jianwei and Yang, Jianxin and Yang, Jiaxi and Zhou, Jingren and Lin, Junyang and Dang, Kai and Lu, Keming and Bao, Keqin and Yang, Kexin and Yu, Le and Li, Mei and Xue, Mingfeng and Zhang, Pei and Zhu, Qin and Men, Rui and Lin, Runji and Li, Tianhao and Tang, Tianyi and Xia, Tingyu and Ren, Xingzhang and Ren, Xuancheng and Fan, Yang and Su, Yang and Zhang, Yichang and Wan, Yu and Liu, Yuqiong and Cui, Zeyu and Zhang, Zhenru and Qiu, Zihan},
  year = 2025,
  month = jan,
  number = {arXiv:2412.15115},
  eprint = {2412.15115},
  primaryclass = {cs},
  publisher = {arXiv},
  doi = {10.48550/arXiv.2412.15115},
  archiveprefix = {arXiv}
}

@misc{singhaniaLokiLowrankKeys2024,
  title={Loki: Low-rank Keys for Efficient Sparse Attention}, 
  author={Prajwal Singhania and Siddharth Singh and Shwai He and Soheil Feizi and Abhinav Bhatele},
  year={2024},
  eprint={2406.02542},
  archivePrefix={arXiv},
  primaryClass={cs.LG},
  url={https://arxiv.org/abs/2406.02542}, 
}

@misc{suRoFormerEnhancedTransformer2023,
        title={RoFormer: Enhanced Transformer with Rotary Position Embedding}, 
      author={Jianlin Su and Yu Lu and Shengfeng Pan and Ahmed Murtadha and Bo Wen and Yunfeng Liu},
      year={2023},
      eprint={2104.09864},
      archivePrefix={arXiv},
      primaryClass={cs.CL},
      url={https://arxiv.org/abs/2104.09864}, 
}

@misc{teamGemma3Technical2025,
        title={Gemma 3 Technical Report}, 
      author={Gemma Team and Aishwarya Kamath and Johan Ferret and Shreya Pathak and Nino Vieillard and Ramona Merhej and Sarah Perrin and Tatiana Matejovicova and Alexandre Ramé and Morgane Rivière and Louis Rouillard and Thomas Mesnard and Geoffrey Cideron and Jean-bastien Grill and Sabela Ramos and Edouard Yvinec and Michelle Casbon and Etienne Pot and Ivo Penchev and Gaël Liu and Francesco Visin and Kathleen Kenealy and Lucas Beyer and Xiaohai Zhai and Anton Tsitsulin and Robert Busa-Fekete and Alex Feng and Noveen Sachdeva and Benjamin Coleman and Yi Gao and Basil Mustafa and Iain Barr and Emilio Parisotto and David Tian and Matan Eyal and Colin Cherry and Jan-Thorsten Peter and Danila Sinopalnikov and Surya Bhupatiraju and Rishabh Agarwal and Mehran Kazemi and Dan Malkin and Ravin Kumar and David Vilar and Idan Brusilovsky and Jiaming Luo and Andreas Steiner and Abe Friesen and Abhanshu Sharma and Abheesht Sharma and Adi Mayrav Gilady and Adrian Goedeckemeyer and Alaa Saade and Alex Feng and Alexander Kolesnikov and Alexei Bendebury and Alvin Abdagic and Amit Vadi and András György and André Susano Pinto and Anil Das and Ankur Bapna and Antoine Miech and Antoine Yang and Antonia Paterson and Ashish Shenoy and Ayan Chakrabarti and Bilal Piot and Bo Wu and Bobak Shahriari and Bryce Petrini and Charlie Chen and Charline Le Lan and Christopher A. Choquette-Choo and CJ Carey and Cormac Brick and Daniel Deutsch and Danielle Eisenbud and Dee Cattle and Derek Cheng and Dimitris Paparas and Divyashree Shivakumar Sreepathihalli and Doug Reid and Dustin Tran and Dustin Zelle and Eric Noland and Erwin Huizenga and Eugene Kharitonov and Frederick Liu and Gagik Amirkhanyan and Glenn Cameron and Hadi Hashemi and Hanna Klimczak-Plucińska and Harman Singh and Harsh Mehta and Harshal Tushar Lehri and Hussein Hazimeh and Ian Ballantyne and Idan Szpektor and Ivan Nardini and Jean Pouget-Abadie and Jetha Chan and Joe Stanton and John Wieting and Jonathan Lai and Jordi Orbay and Joseph Fernandez and Josh Newlan and Ju-yeong Ji and Jyotinder Singh and Kat Black and Kathy Yu and Kevin Hui and Kiran Vodrahalli and Klaus Greff and Linhai Qiu and Marcella Valentine and Marina Coelho and Marvin Ritter and Matt Hoffman and Matthew Watson and Mayank Chaturvedi and Michael Moynihan and Min Ma and Nabila Babar and Natasha Noy and Nathan Byrd and Nick Roy and Nikola Momchev and Nilay Chauhan and Noveen Sachdeva and Oskar Bunyan and Pankil Botarda and Paul Caron and Paul Kishan Rubenstein and Phil Culliton and Philipp Schmid and Pier Giuseppe Sessa and Pingmei Xu and Piotr Stanczyk and Pouya Tafti and Rakesh Shivanna and Renjie Wu and Renke Pan and Reza Rokni and Rob Willoughby and Rohith Vallu and Ryan Mullins and Sammy Jerome and Sara Smoot and Sertan Girgin and Shariq Iqbal and Shashir Reddy and Shruti Sheth and Siim Põder and Sijal Bhatnagar and Sindhu Raghuram Panyam and Sivan Eiger and Susan Zhang and Tianqi Liu and Trevor Yacovone and Tyler Liechty and Uday Kalra and Utku Evci and Vedant Misra and Vincent Roseberry and Vlad Feinberg and Vlad Kolesnikov and Woohyun Han and Woosuk Kwon and Xi Chen and Yinlam Chow and Yuvein Zhu and Zichuan Wei and Zoltan Egyed and Victor Cotruta and Minh Giang and Phoebe Kirk and Anand Rao and Kat Black and Nabila Babar and Jessica Lo and Erica Moreira and Luiz Gustavo Martins and Omar Sanseviero and Lucas Gonzalez and Zach Gleicher and Tris Warkentin and Vahab Mirrokni and Evan Senter and Eli Collins and Joelle Barral and Zoubin Ghahramani and Raia Hadsell and Yossi Matias and D. Sculley and Slav Petrov and Noah Fiedel and Noam Shazeer and Oriol Vinyals and Jeff Dean and Demis Hassabis and Koray Kavukcuoglu and Clement Farabet and Elena Buchatskaya and Jean-Baptiste Alayrac and Rohan Anil and Dmitry and Lepikhin and Sebastian Borgeaud and Olivier Bachem and Armand Joulin and Alek Andreev and Cassidy Hardin and Robert Dadashi and Léonard Hussenot},
      year={2025},
      eprint={2503.19786},
      archivePrefix={arXiv},
      primaryClass={cs.CL},
      url={https://arxiv.org/abs/2503.19786}, 
}

@misc{vaswaniAttentionAllYou2017,
      title={Attention Is All You Need}, 
      author={Ashish Vaswani and Noam Shazeer and Niki Parmar and Jakob Uszkoreit and Llion Jones and Aidan N. Gomez and Lukasz Kaiser and Illia Polosukhin},
      year={2023},
      eprint={1706.03762},
      archivePrefix={arXiv},
      primaryClass={cs.CL},
      url={https://arxiv.org/abs/1706.03762}, 
}

@misc{wangLinformerSelfAttentionLinear2020,
      title={Linformer: Self-Attention with Linear Complexity}, 
      author={Sinong Wang and Belinda Z. Li and Madian Khabsa and Han Fang and Hao Ma},
      year={2020},
      eprint={2006.04768},
      archivePrefix={arXiv},
      primaryClass={cs.LG},
      url={https://arxiv.org/abs/2006.04768}, 
}

@misc{xiaoEfficientStreamingLanguage2024,
    title={Efficient Streaming Language Models with Attention Sinks}, 
  author={Guangxuan Xiao and Yuandong Tian and Beidi Chen and Song Han and Mike Lewis},
  year={2024},
  eprint={2309.17453},
  archivePrefix={arXiv},
  primaryClass={cs.CL},
  url={https://arxiv.org/abs/2309.17453}, 
}

@misc{yangPyramidInferPyramidKV2024,
    title={PyramidInfer: Pyramid KV Cache Compression for High-throughput LLM Inference}, 
      author={Dongjie Yang and XiaoDong Han and Yan Gao and Yao Hu and Shilin Zhang and Hai Zhao},
      year={2024},
      eprint={2405.12532},
      archivePrefix={arXiv},
      primaryClass={cs.CL},
      url={https://arxiv.org/abs/2405.12532}, 
}

@misc{zandiehKDEformerAcceleratingTransformers2023,
  title={KDEformer: Accelerating Transformers via Kernel Density Estimation}, 
      author={Amir Zandieh and Insu Han and Majid Daliri and Amin Karbasi},
      year={2023},
      eprint={2302.02451},
      archivePrefix={arXiv},
      primaryClass={cs.LG},
      url={https://arxiv.org/abs/2302.02451}, 
}

@misc{schroderWildCatNearLinearAttention2026,
    title={WildCat: Near-Linear Attention in Theory and Practice}, 
  author={Tobias Schröder and Lester Mackey},
  year={2026},
  eprint={2602.10056},
  archivePrefix={arXiv},
  primaryClass={cs.LG},
  url={https://arxiv.org/abs/2602.10056}, 
}

@inproceedings{baiLongBenchBilingualMultitask2024,
  title = {{LongBench}: A Bilingual, Multitask Benchmark for Long Context Understanding},
  author = {Bai, Yushi and Lv, Xin and Zhang, Jiajie and Lyu, Hongchang and Tang, Jiankai and Huang, Zhidian and Du, Zhengxiao and Liu, Xiao and Zeng, Aohan and Hou, Lei and Dong, Yuxiao and Tang, Jie and Li, Juanzi},
  booktitle = {Proceedings of the 62nd Annual Meeting of the Association for Computational Linguistics (Volume 1: Long Papers)},
  year = {2024},
  pages = {3119--3137},
  publisher = {Association for Computational Linguistics},
  address = {Bangkok, Thailand},
  doi = {10.18653/v1/2024.acl-long.172}
}

@article{busingerLinearLeast1965,
  title = {Linear Least Squares Solutions by {{Householder}} Transformations},
  author = {Businger, Peter and Golub, Gene H.},
  year = 1965,
  journal = {Numerische Mathematik},
  volume = {7},
  number = {3},
  pages = {269--276},
  doi = {10.1007/BF01436084}
}

@incollection{highamAnalysisCholesky1990,
    author = {Higham, Nicholas J},
    editor = {Cox, M G and Hammarling, S},
    isbn = {9780198535645},
    title = {Analysis of the Cholesky decomposition of a semi-definite matrix},
    booktitle = {Reliable Numerical Commputation},
    publisher = {Oxford University Press},
    year = {1990},
    month = {09},
    doi = {10.1093/oso/9780198535645.003.0010},
    url = {https://doi.org/10.1093/oso/9780198535645.003.0010},
    eprint = {https://academic.oup.com/book/0/chapter/422058562/chapter-pdf/52393608/isbn-9780198535645-book-part-10.pdf},
}

@misc{gongExpressLanguageModeling2026,
  title = {Express {{Language Modeling}}},
  author = {Gong, Albert and Carrell, Annabelle Michael and Dwivedi, Raaz and Mackey, Lester},
  year = 2026,
  month = jun,
  number = {arXiv:2606.10944},
  eprint = {2606.10944},
  primaryclass = {cs},
  publisher = {arXiv},
  doi = {10.48550/arXiv.2606.10944},
  archiveprefix = {arXiv}
}

@misc{zhangH2OHeavyHitterOracle2023,
    title={H$_2$O: Heavy-Hitter Oracle for Efficient Generative Inference of Large Language Models}, 
      author={Zhenyu Zhang and Ying Sheng and Tianyi Zhou and Tianlong Chen and Lianmin Zheng and Ruisi Cai and Zhao Song and Yuandong Tian and Christopher Ré and Clark Barrett and Zhangyang Wang and Beidi Chen},
      year={2023},
      eprint={2306.14048},
      archivePrefix={arXiv},
      primaryClass={cs.LG},
      url={https://arxiv.org/abs/2306.14048}, 
}

@misc{liSnapKVLLMKnows2024,
  title = {{{SnapKV}}: {{LLM Knows What You}} Are {{Looking}} for {{Before Generation}}},
  shorttitle = {{{SnapKV}}},
  author = {Li, Yuhong and Huang, Yingbing and Yang, Bowen and Venkitesh, Bharat and Locatelli, Acyr and Ye, Hanchen and Cai, Tianle and Lewis, Patrick and Chen, Deming},
  year = 2024,
  month = apr,
  number = {arXiv:2404.14469},
  eprint = {2404.14469},
  primaryclass = {cs},
  publisher = {arXiv},
  doi = {10.48550/arXiv.2404.14469},
  archiveprefix = {arXiv}
}

@misc{serretAttentionMechanismsThrough2026,
        title={Attention Mechanisms Through the Lens of Numerical Methods: Approximation Methods and Alternative Formulations}, 
      author={Michel Fabrice Serret and Alice Cortinovis and Yijun Dong and Diana Halikias and Anna Ma and Fabio Matti and Deanna Needell and Katherine J. Pearce and Elizaveta Rebrova and Disha Shur and Rudi Smith and Hai-Xiao Wang and Laura Grigori},
      year={2026},
      eprint={2604.01757},
      archivePrefix={arXiv},
      primaryClass={math.NA},
      url={https://arxiv.org/abs/2604.01757}, 
}

@inproceedings{rahimiRandomFeaturesLargescale2007,
 author = {Rahimi, Ali and Recht, Benjamin},
 booktitle = {Advances in Neural Information Processing Systems},
 editor = {J. Platt and D. Koller and Y. Singer and S. Roweis},
 pages = {},
 publisher = {Curran Associates, Inc.},
 title = {Random Features for Large-Scale Kernel Machines},
 url = {https://proceedings.neurips.cc/paper_files/paper/2007/file/013a006f03dbc5392effeb8f18fda755-Paper.pdf},
 volume = {20},
 year = {2007}
}

@article{demarchiNearoptimalDataindependentPoint2005,
  title = {Near-Optimal Data-Independent Point Locations for Radial Basis Function Interpolation},
  author = {De Marchi, Stefano and Schaback, Robert and Wendland, Holger},
  year = 2005,
  journal = {Advances in Computational Mathematics},
  volume = {23},
  number = {3},
  pages = {317--330},
  doi = {10.1007/s10444-004-1829-1}
}

@article{santinConvergenceRateDataindependent2017,
  title = {Convergence Rate of the Data-Independent {{P}}-Greedy Algorithm in Kernel-Based Approximation},
  author = {Santin, Gabriele and Haasdonk, Bernard},
  year = 2017,
  journal = {Dolomites Research Notes on Approximation},
  volume = {10},
  number = {Special Issue},
  pages = {68--78},
  doi = {10.14658/PUPJ-DRNA-2017-Special_Issue-9},
  eprint = {1612.02672},
  primaryclass = {math}
}

@article{wenzelNovelClassStabilized2021,
title = {A novel class of stabilized greedy kernel approximation algorithms: Convergence, stability and uniform point distribution},
journal = {Journal of Approximation Theory},
volume = {262},
pages = {105508},
year = {2021},
issn = {0021-9045},
doi = {https://doi.org/10.1016/j.jat.2020.105508},
url = {https://www.sciencedirect.com/science/article/pii/S0021904520301441},
author = {Tizian Wenzel and Gabriele Santin and Bernard Haasdonk}
}

@article{wenzelAnalysisTargetDatadependent2023,
  title = {Analysis of Target Data-Dependent Greedy Kernel Algorithms: Convergence Rates for $f$-, $f \cdot P$- and $f/P$-Greedy},
  shorttitle = {Analysis of Target Data-Dependent Greedy Kernel Algorithms},
  author = {Wenzel, Tizian and Santin, Gabriele and Haasdonk, Bernard},
  year = 2023,
  journal = {Constructive Approximation},
  volume = {57},
  number = {1},
  pages = {45--74},
  doi = {10.1007/s00365-022-09592-3}
}

@article{drmacNewSelectionOperator2016,
author = {Drma{\v c}, Zlatko and Gugercin, Serkan},
title = {A New Selection Operator for the Discrete Empirical Interpolation Method---Improved A Priori Error Bound and Extensions},
journal = {SIAM Journal on Scientific Computing},
volume = {38},
number = {2},
pages = {A631-A648},
year = {2016},
doi = {10.1137/15M1019271},

URL = { 
    
        https://doi.org/10.1137/15M1019271
    
    

},
eprint = { 
    
        https://doi.org/10.1137/15M1019271
    
    

}

}

@article{cortinovisAdaptiveRandomizedPivoting2026,
author = {Cortinovis, Alice and Kressner, Daniel},
title = {Adaptive Randomized Pivoting for Column Subset Selection, DEIM, and Low-Rank Approximation},
journal = {SIAM Journal on Matrix Analysis and Applications},
volume = {47},
number = {1},
pages = {25-47},
year = {2026},
doi = {10.1137/24M1719189},

URL = { 
    
        https://doi.org/10.1137/24M1719189
    
    

},
eprint = { 
    
        https://doi.org/10.1137/24M1719189
    
    

}
}

@misc{fakihEfficientQRbasedColumn2026,
      title={Efficient QR-based Column Subset Selection through Randomized Sparse Embeddings}, 
      author={Israa Fakih and Laura Grigori},
      year={2026},
      eprint={2509.03198},
      archivePrefix={arXiv},
      primaryClass={math.NA},
      url={https://arxiv.org/abs/2509.03198}, 
}

@misc{chenTightBoundsStreaming2026,
      title={Towards Tight Bounds for Streaming Attention}, 
      author={Justin Y. Chen and Ying Feng and Piotr Indyk and Michael Kapralov and Ekaterina Kochetkova and Boris Prokhorov},
      year={2026},
      eprint={2606.07205},
      archivePrefix={arXiv},
      primaryClass={cs.DS},
      url={https://arxiv.org/abs/2606.07205}, 
}

@inproceedings{guo-etal-2020-wiki,
    title = "{W}iki-40{B}: Multilingual Language Model Dataset",
    author = "Guo, Mandy  and
      Dai, Zihang  and
      Vrande{\v{c}}i{\'c}, Denny  and
      Al-Rfou, Rami",
    editor = "Calzolari, Nicoletta  and
      B{\'e}chet, Fr{\'e}d{\'e}ric  and
      Blache, Philippe  and
      Choukri, Khalid  and
      Cieri, Christopher  and
      Declerck, Thierry  and
      Goggi, Sara  and
      Isahara, Hitoshi  and
      Maegaard, Bente  and
      Mariani, Joseph  and
      Mazo, H{\'e}l{\`e}ne  and
      Moreno, Asuncion  and
      Odijk, Jan  and
      Piperidis, Stelios",
    booktitle = "Proceedings of the Twelfth Language Resources and Evaluation Conference",
    month = may,
    year = "2020",
    address = "Marseille, France",
    publisher = "European Language Resources Association",
    url = "https://aclanthology.org/2020.lrec-1.297/",
    pages = "2440--2452",
    language = "eng",
    ISBN = "979-10-95546-34-4"
}

@inproceedings{wolf-etal-2020-transformers,
        title = "Transformers: State-of-the-Art Natural Language Processing",
    author = "Wolf, Thomas  and
      Debut, Lysandre  and
      Sanh, Victor  and
      Chaumond, Julien  and
      Delangue, Clement  and
      Moi, Anthony  and
      Cistac, Pierric  and
      Rault, Tim  and
      Louf, Remi  and
      Funtowicz, Morgan  and
      Davison, Joe  and
      Shleifer, Sam  and
      von Platen, Patrick  and
      Ma, Clara  and
      Jernite, Yacine  and
      Plu, Julien  and
      Xu, Canwen  and
      Le Scao, Teven  and
      Gugger, Sylvain  and
      Drame, Mariama  and
      Lhoest, Quentin  and
      Rush, Alexander",
    editor = "Liu, Qun  and
      Schlangen, David",
    booktitle = "Proceedings of the 2020 Conference on Empirical Methods in Natural Language Processing: System Demonstrations",
    month = oct,
    year = "2020",
    address = "Online",
    publisher = "Association for Computational Linguistics",
    url = "https://aclanthology.org/2020.emnlp-demos.6/",
    doi = "10.18653/v1/2020.emnlp-demos.6",
    pages = "38--45"
}

@article{Kessy02102018,
author = {Agnan Kessy and Alex Lewin and Korbinian Strimmer},
title = {Optimal Whitening and Decorrelation},
journal = {The American Statistician},
volume = {72},
number = {4},
pages = {309--314},
year = {2018},
publisher = {Taylor \& Francis},
doi = {10.1080/00031305.2016.1277159},


URL = { 
    
        https://doi.org/10.1080/00031305.2016.1277159
    
    

},
eprint = { 
    
        https://doi.org/10.1080/00031305.2016.1277159
    
    

}

}

@misc{han2025polarquantquantizingkvcaches,
      title={PolarQuant: Quantizing KV Caches with Polar Transformation}, 
      author={Insu Han and Praneeth Kacham and Amin Karbasi and Vahab Mirrokni and Amir Zandieh},
      year={2025},
      eprint={2502.02617},
      archivePrefix={arXiv},
      primaryClass={cs.LG},
      url={https://arxiv.org/abs/2502.02617}, 
}

@misc{zandieh2025turboquantonlinevectorquantization,
      title={TurboQuant: Online Vector Quantization with Near-optimal Distortion Rate}, 
      author={Amir Zandieh and Majid Daliri and Majid Hadian and Vahab Mirrokni},
      year={2025},
      eprint={2504.19874},
      archivePrefix={arXiv},
      primaryClass={cs.LG},
      url={https://arxiv.org/abs/2504.19874}, 
}

@inproceedings{pati1993orthogonal,
    author={Pati, Y.C. and Rezaiifar, R. and Krishnaprasad, P.S.},
  booktitle={Proceedings of 27th Asilomar Conference on Signals, Systems and Computers}, 
  title={Orthogonal matching pursuit: recursive function approximation with applications to wavelet decomposition}, 
  year={1993},
  volume={},
  number={},
  pages={40-44 vol.1},
  doi={10.1109/ACSSC.1993.342465}}

@article{davis1997greedy,
  title={Greedy adaptive approximation},
  author={Davis, Geoffrey},
  journal={Journal of Constructive Approximation},
  volume={13},
  number={1},
  pages={57--98},
  year={1997}
}

@article{tropp2006algorithms,
title = {Algorithms for simultaneous sparse approximation. Part I: Greedy pursuit},
journal = {Signal Processing},
volume = {86},
number = {3},
pages = {572-588},
year = {2006},
note = {Sparse Approximations in Signal and Image Processing},
issn = {0165-1684},
doi = {https://doi.org/10.1016/j.sigpro.2005.05.030},
url = {https://www.sciencedirect.com/science/article/pii/S0165168405002227},
author = {Joel A. Tropp and Anna C. Gilbert and Martin J. Strauss},
}

@misc{serret2026understandingtransformersattentionmechanisms,
      title={Understanding Transformers and Attention Mechanisms: An Introduction for Applied Mathematicians}, 
      author={Michel Fabrice Serret},
      year={2026},
      eprint={2604.00965},
      archivePrefix={arXiv},
      primaryClass={math.NA},
      url={https://arxiv.org/abs/2604.00965}, 
}

@misc{dosovitskiy2021imageworth16x16words,
      title={An Image is Worth 16x16 Words: Transformers for Image Recognition at Scale}, 
      author={Alexey Dosovitskiy and Lucas Beyer and Alexander Kolesnikov and Dirk Weissenborn and Xiaohua Zhai and Thomas Unterthiner and Mostafa Dehghani and Matthias Minderer and Georg Heigold and Sylvain Gelly and Jakob Uszkoreit and Neil Houlsby},
      year={2021},
      eprint={2010.11929},
      archivePrefix={arXiv},
      primaryClass={cs.CV},
      url={https://arxiv.org/abs/2010.11929}, 
}

@misc{simeoni2025dinov3,
      title={DINOv3}, 
      author={Oriane Siméoni and Huy V. Vo and Maximilian Seitzer and Federico Baldassarre and Maxime Oquab and Cijo Jose and Vasil Khalidov and Marc Szafraniec and Seungeun Yi and Michaël Ramamonjisoa and Francisco Massa and Daniel Haziza and Luca Wehrstedt and Jianyuan Wang and Timothée Darcet and Théo Moutakanni and Leonel Sentana and Claire Roberts and Andrea Vedaldi and Jamie Tolan and John Brandt and Camille Couprie and Julien Mairal and Hervé Jégou and Patrick Labatut and Piotr Bojanowski},
      year={2025},
      eprint={2508.10104},
      archivePrefix={arXiv},
      primaryClass={cs.CV},
      url={https://arxiv.org/abs/2508.10104}, 
}

@article{Sun_2023,
   title={Attention-based graph neural networks: a survey},
   volume={56},
   ISSN={1573-7462},
   url={http://dx.doi.org/10.1007/s10462-023-10577-2},
   DOI={10.1007/s10462-023-10577-2},
   number={S2},
   journal={Artificial Intelligence Review},
   publisher={Springer Science and Business Media LLC},
   author={Sun, Chengcheng and Li, Chenhao and Lin, Xiang and Zheng, Tianji and Meng, Fanrong and Rui, Xiaobin and Wang, Zhixiao},
   year={2023},
   month=Aug, pages={2263–2310} }

@article{stathopoulosBlockOrthogonalizationProcedure2002,
author = {Stathopoulos, Andreas and Wu, Kesheng},
title = {A Block Orthogonalization Procedure with Constant Synchronization Requirements},
journal = {SIAM Journal on Scientific Computing},
volume = {23},
number = {6},
pages = {2165-2182},
year = {2002},
doi = {10.1137/S1064827500370883},
URL = { 
        https://doi.org/10.1137/S1064827500370883
},
eprint = {
        https://doi.org/10.1137/S1064827500370883
}

}
